%% file: main.tex
\documentclass{article}

\usepackage{microtype}
\usepackage{graphicx}
\usepackage{subcaption}
\usepackage{booktabs} 

\usepackage{listings}
\usepackage{xcolor}
\usepackage{float}
\usepackage{hyperref}
\usepackage{amsfonts}       
\usepackage{nicefrac}       
\usepackage{microtype}      
\usepackage{xcolor}         
\usepackage{enumitem}
\usepackage{adjustbox}
\usepackage{framed}
\usepackage{amsmath}
\usepackage{amssymb}
\usepackage{bm}
\usepackage{graphicx}
\usepackage{multirow}
\usepackage[most]{tcolorbox}

\usepackage[table]{xcolor}
\usepackage{calc}
\usepackage[T1]{fontenc}

\usepackage[preprint]{icml2026}

\usepackage{amsmath}
\usepackage{amssymb}
\usepackage{mathtools}
\usepackage{amsthm}

\usepackage[capitalize,noabbrev]{cleveref}

\theoremstyle{plain}

\theoremstyle{definition}

\theoremstyle{remark}

\newcommand{\ours}{{AIR}} 

\usepackage[textsize=tiny]{todonotes}

\definecolor{bgred}{RGB}{255, 220, 220}
\newcommand{\myhl}[1]{\begingroup\setlength{\fboxsep}{0.5pt}\colorbox{bgred}{#1}\endgroup}

\icmltitlerunning{Manuscript}

\begin{document}

\fancypagestyle{firststyle}{
    \fancyhead[R]{%
        \vspace*{2mm}%
        \raisebox{-0.3\height}{\includegraphics[width=53.3mm]{map.pdf}}%
    }
}

\twocolumn[

  \icmltitle{AIR:  Post-training Data Selection for Reasoning via Attention Head Influence}




  \icmlsetsymbol{equal}{*}

  \begin{icmlauthorlist}
    \icmlauthor{Jinrui Liu}{equal,yyy}
    \icmlauthor{Jeff Wu}{equal,comp}
    \icmlauthor{Xuanguang Pan}{yyy}
    \icmlauthor{Gavin Cheung}{comp}
    \icmlauthor{Shuai Ma}{yyy}
    \icmlauthor{Chongyang Tao}{yyy}
  \end{icmlauthorlist}

  \icmlaffiliation{yyy}{Beihang University}
  \icmlaffiliation{comp}{Independent Researcher}


  \icmlkeywords{Machine Learning, ICML}

  \vskip 0.3in
]



\printAffiliationsAndNotice{\icmlEqualContribution}

\begin{abstract}
LLMs achieve remarkable multi-step reasoning capabilities, yet effectively transferring these skills via post-training distillation remains challenging. Existing data selection methods, ranging from manual curation to heuristics based on length, entropy, or overall loss, fail to capture the causal importance of individual reasoning steps, limiting distillation efficiency. To address this, we propose Attention Influence for Reasoning (AIR), a principled, {unsupervised and training-free} framework that leverages mechanistic insights of the retrieval head to select high-value post-training data. AIR first identifies reasoning-critical attention heads of an off-the-shelf model, then constructs a weakened reference model with disabled head influence, and finally quantifies the resulting loss divergence as the Attention Influence Score. This score enables fine-grained assessment at both the step and sample levels, supporting step-level weighted fine-tuning and global sample selection. Experiments across multiple reasoning benchmarks show that AIR consistently improves reasoning accuracy, surpassing heuristic baselines and effectively isolating the most critical steps and samples. Our work establishes a mechanism-driven, data-efficient approach for reasoning distillation in LLMs.

\end{abstract}

\input{1-intro}

\input{3-method}
\input{4-exp}

\input{2-related}

\section{Conclusion}
In this paper, we propose Attention Influence for Reasoning (AIR), an unsupervised and training-free framework for selecting high-quality and reasoning-intensive post-training data. Specifically, the method first identifies specific "retrieval heads" responsible for token-level copying. It then constructs a weakened reference model by disabling their influence and quantifies the loss divergence relative to the strong base model to enable two approaches: step-level weighted supervised fine-tuning and global sample selection. Experiments across multiple reasoning and comprehensive benchmarks demonstrate that \ours{} effectively identifies critical reasoning steps and samples and consistently improves reasoning performance, establishing a mechanistically interpretable and data-efficient approach for reasoning post-training for LLMs.

\section*{Acknowledgments}
We sincerely thank Kai Hua for invaluable discussions and insightful feedback, which substantially strengthened this work.

\nocite{langley00}

\bibliography{icml2026}
\bibliographystyle{icml2026}


\input{5-appendix}

\end{document}

%% file: 1-intro.tex
\definecolor{rliableolive}{HTML}{BBCC33}
\definecolor{rliableblue}{HTML}{77AADD}
\definecolor{rliablered}{HTML}{EE8866}
\definecolor{SDEblue}{RGB}{28 58 88}
\definecolor{cc1}{rgb}{1.0, 0.44, 0.37}
\definecolor{cc2}{rgb}{0.0, 0.2, 0.6}
\definecolor{cc3}{RGB}{255, 191, 0}
\definecolor{cc4}{RGB}{0, 128, 128}

\section{Introduction}
\label{sec:intro}

Large Language Models (LLMs) have demonstrated revolutionary capabilities in solving complex problems through multi-step chain-of-thought (CoT) reasoning~\cite{guo2025deepseek,yang2025qwen3}. The ability to generate high-quality reasoning traces is critical, making research on improving and transferring these skills a primary focus in advanced generative AI.  Seminal work in this field has primarily explored  post-training techniques, including reasoning distillation via Supervised Fine-Tuning (SFT)~\citep{muennighoff2025s1, openr1, guha2025openthoughts, hu2025distillation, ye2025limo} and approaches based on Reinforcement Learning (RL)~\cite{schulman2017proximal,ziegler2019fine,ouyang2022training,guo2025deepseek}. While both paradigms have achieved notable progress in this area, RL is notoriously resource-consuming and difficult to tune, often rendering it prohibitively expensive. Fortunately, recent findings \cite{muennighoff2025s1,hu2025distillation} indicate that applying SFT to strategically curated, high-quality samples can match or even surpass the performance of more costly RL-based methods. This establishes a strong foundation where data-efficient and high-quality distillation is the key to democratizing advanced reasoning intelligence.

Given the demonstrated efficacy of distillation on limited examples, the question of data selection—\emph{identifying the most effective and high-value reasoning traces}—has become paramount for maximizing knowledge transfer and incentive reasoning capability. The community has explored various approaches: while manual curation, such as the hand-crafted sample selection performed by s1K~\cite{muennighoff2025s1} and LIMO~\cite{ye2025limo}, shows strong performance, it is inherently labor-intensive and non-scalable. Automated solutions, including heuristic methods based on length~\cite{olsson2022context} or complexity~\citep{ye2025limo, li2025naturalthoughts,Wang2024} and influence-based scoring techniques~\citep{lin2024rho,humane2025influence, jiang2025importance,qin2025sstoken}, offer scalability. However, a significant gap remains: these existing methods rely on coarse proxy metrics which fail to isolate and measure the causal criticality of {individual steps within a complex reasoning trace}. 
This limitation becomes especially pronounced when the reasoning relies on precise internal factual retrieval, which is performed by specialized mechanisms within Transformer architectures.
Moreover, these existing approaches rarely incorporate a step-level mechanism for data selection capable of discerning the quality or importance of each discrete reasoning step. 
Such fine-grained selection is essential because the distillation objective must prioritize core reasoning patterns such as planning, summarization, and reflection, which are more valuable for the student model to learn and generalize, as highlighted by our case study.

\begin{figure*}[ht] 
    \centering
     \includegraphics[width=\textwidth]{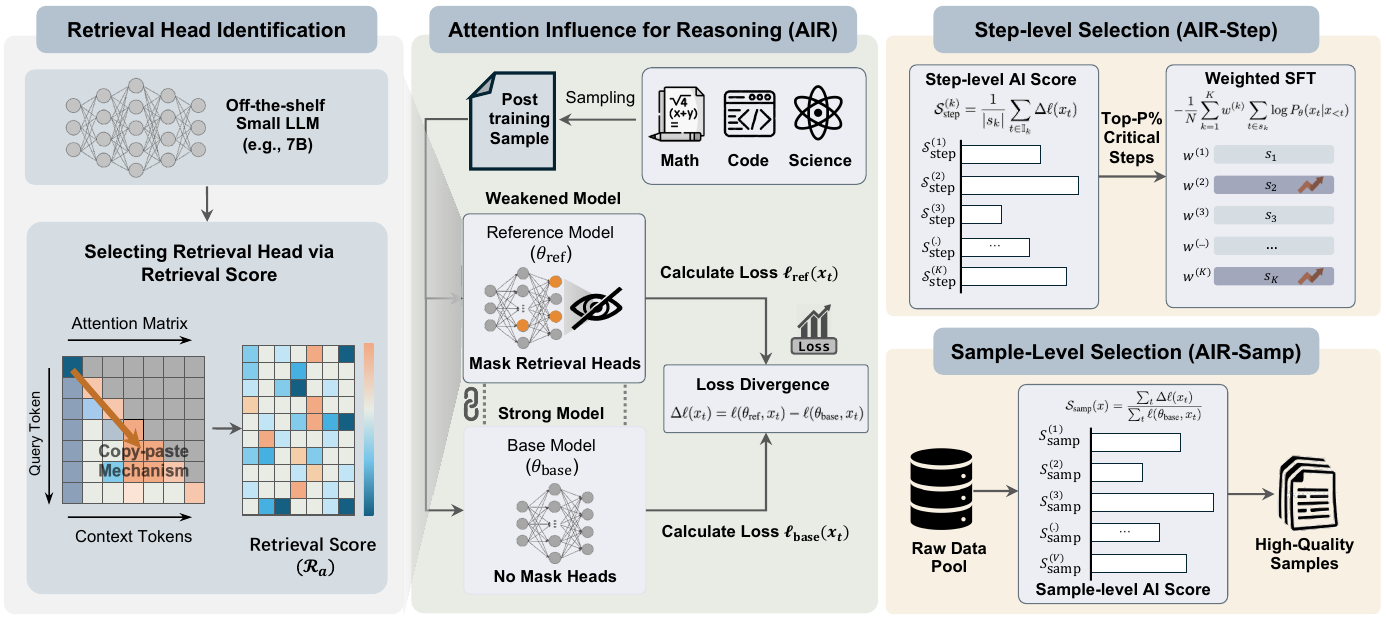} 
    \caption{The illustration of the \ours{} method. Note that the base model and reference model are derived from the same underlying model. For ease of visualization, orange circles depict the masking of retrieval heads.}
    \label{fig:air}
\end{figure*}

\begin{figure*}[t]
\begin{tcolorbox}[colback=rliableolive!10!white,
                  colframe=black,
                  boxrule=1pt,
                  boxsep=2pt,
                  top=3pt,
                  bottom=3pt,
                  left=2pt,
                  right=2pt]
\begin{center}
\textbf{How can we identify high-value reasoning traces in a principled and interpretable way?}
\end{center}
\end{tcolorbox}
\vspace{-4mm}
\end{figure*}

To bridge this gap, we advocate for principled data selection driven by insights from mechanistic interpretability~\cite{olsson2022context,bricken2023towards}. We specifically leverage the finding that certain ``retrieval heads" are functionally responsible for ensuring token-level copying and fidelity—the backbone of multi-step reasoning~\cite{wu2024retrieval,hua2025attentioninfluence}. If a reasoning step is correctly executed, the corresponding retrieval mechanism must be functionally engaged. 
Building on this observation, we propose Attention Influence for Reasoning (\ours), an unsupervised and training-free framework for selecting post-training data based on the causal influence of reasoning-related attention heads.
\ours~ first identifies reasoning-
critical attention heads of an off-the-shelf base model ($\theta_{\text{base}}$).
It then constructs a weakened reference model ($\theta_{\text{ref}}$) by disabling their influence, and measures the resulting loss divergence against the strong base model ($\theta_{\text{base}}$). This divergence directly yields the \ours{} score, a fine-grained metric reflecting the token's reliance on reasoning. 
We further compute the \ours{} Score at both the reasoning step-level and the sample-level, which are then used to drive two distinct data selection strategies: step-level weighted SFT and global sample selection.

We conducted comprehensive data selection experiments across both step-level and sample-level applications using several widely used reasoning datasets. The empirical results demonstrate that models distilled using our influence-based selection achieve notable improvements in accuracy, consistently outperforming conventional prioritization baselines. Furthermore, the effectiveness of our approach is supported by statistical analysis and visualization of the selected data distributions, which provide clear evidence that \ours{} successfully isolates and prioritizes the most critical steps and samples. The main contributions of this work are summarized as follows:
\begin{itemize}
    \item  We propose \ours, a principled, unsupervised, and training-free framework that quantifies the causal influence of reasoning-critical attention heads, enabling a mechanism-driven approach to data selection for reasoning distillation.
    
    \item We formally define the \ours{} score based on the loss divergence between a base model and its deliberately weakened counterpart guided by retrieval heads, enabling accurate step- and sample-level assessments of reasoning criticality.
    
    \item We comprehensively demonstrate that applying \ours{} leads to superior performance in transferring various reasoning skills, outperforming existing heuristic data selection methods and achieving performance comparable to manually curated data.
\end{itemize}

%% file: 3-method.tex
\section{Method: \ours{}}
\label{sec:model_overview}

We introduce \ours{}, a principled framework that employs model introspection based on attention influence to provide precise guidance for high-quality data selection for reasoning. 
Our core insight is derived from recent mechanistic interpretability findings: the existence of specialized {retrieval attention heads} is critical to LLMs' in-context learning, retrieval, and multi-step reasoning. As shown in \autoref{fig:air},
we first construct a {mechanically weakened reference model} ($\theta_\text{ref}$) by specifically masking the identified reasoning-critical retrieval heads in the base model ($\theta_\text{base}$). Subsequently, we quantify data importance using the {loss divergence} between $\theta_\text{ref}$ and $\theta_\text{base}$---a metric that directly measures a sample's {dependence} on the model's fundamental retrieval mechanism. This approach yields the {attention influence score ($\mathcal{S}$)}, which is applied at both the {sample level} (for prioritized data selection) and the {reasoning step level} (for fine-grained weighted supervision). 
Unlike current post-training data selection methods, which primarily rely on heuristic filtering or comparing losses against a strong reference model, \ours{} provides a training-free, cost-effective, precise, and {mechanism-driven} approach for curating impactful data essential for efficient reasoning post-training.

\subsection{Identifying Important Attention Heads}
\label{sec:detect}

\noindent Leveraging insights from mechanistic interpretability, we focus on isolating the specific components within the Transformer architecture that underpin its capacity for information retrieval and complex reasoning. Motivated by the findings that {retrieval heads} play a critical role in maintaining LLMs' factual integrity and reasoning performance~\citep{wu2024retrieval,hua2025attentioninfluence}, we {operationalize the formal identification criterion} to identify these crucial heads with a small LLM (e.g., Qwen2.5-7B-instruct) for our framework.

Formally, a retrieval head $h$ serves to preserve contextual fidelity by enabling accurate, token-level transfer of information from the source text to the generated output. To quantify this behavior, a token-level recall rate based on attention head dynamics is defined. Specifically, consider the LLM at decoding step $t$. Let the initial input context (the prompt) be $\mathbf{x}_{1:n}$, and the sequence of previously generated tokens be $\mathbf{w}_{1:t-1}$. The full input sequence visible to the current query is $\mathbf{x}_{1:n+t-1}$:
\begin{equation}
    \mathbf{x}_{1:n+t-1} = [\underbrace{x_1, \dots, x_n}_{\text{context}}, \underbrace{w_1, \dots, w_{t-1}}_{\text{generated tokens}}]
\end{equation}
The attention score vector produced by head $h$ for the query $w_{t-1}$ (targeting the generation of $w_t$) is $\mathbf{a}_t \in \mathbb{R}^{\,n+t-1}$.  An attention head $h$ is defined to successfully perform a context retrieval, or \emph{``copy-paste"} operation, from a target content set $\mathbf{k} \subseteq \mathbf{x}_{1:n}$ to generate $w_t$ if and only if the following two conditions are satisfied:
\begin{itemize}
    \item \emph{\textbf{ Token Appearance (C1):}} The token generated at step $t$, $w_t$, exists within the target context subset $\mathbf{k}$: $w_t \in \mathbf{k}$.
    \item \emph{\textbf{Maximal Attention (C2):} } The position $j^\ast$ corresponding to the generated token $w_t$ must receive the maximal attention weight across all visible positions in head $h$:
    $$
    j^\ast = \arg\max_{j \in \{1,\dots,n+t-1\}} \mathbf{a}_t[j], \quad \text{such that } x_{j^\ast} = w_t.
    $$
\end{itemize}

\textbf{Retrieval Score ($\mathcal{R}_h$).} Let $\mathbf{g_h}$ denote the set containing all tokens copied and pasted by a given head $h$ (i.e., tokens satisfying both C1 and C2). $\mathcal{R}_h$ quantifies the head's efficacy in retrieving the corresponding content $\mathbf{k}$ from the context, modeling this capability as a recall rate:
\begin{equation}
    \mathcal{R}_h = \frac{|\mathbf{g_h} \cap \mathbf{k} |}{|\mathbf{k}|}
\end{equation}
Attention heads exhibiting a high $\mathcal{R}_h$ are definitively classified as \emph{reasoning-critical retrieval heads} and are subsequently targeted for construction of the weakened reference model.

\subsection{Calculating Attention Influence Score}
\label{sec:weak_model_loss_gap}

After identifying the reasoning-critical retrieval heads using the criterion above, we continue by quantifying their functional impact on model behavior.
To this end, we construct a weakened reference model by selectively masking these heads, thereby noising the contribution of their information pathways. We then measure the resulting loss divergence between the weakened model and the strong base model ($\theta_{\text{base}}$). This divergence, computed in terms of cross-entropy loss, serves as the foundation for defining the Attention Influence Score ($\mathcal{S}$), which captures how strongly each token, reasoning step, or sample is affected by the masked reasoning mechanisms.

\textbf{Weakened Reference Model Construction ($\theta_{ref}$).} 
Let $\mathcal{H}$ be the set of all attention heads in the base model $\theta_{base}$. The set of reasoning-critical retrieval heads, $\mathcal{H}_{\text{critical}}$, is defined by selecting all attention heads whose Retrieval Score ($\mathcal{R}_h$) falls within the top $\delta$ percentile of $\{\mathcal{R}_h \mid h \in \mathcal{H}\}$, formally defined as:
\begin{equation}
    \mathcal{H}_{\text{critical}} = \text{argmax}_{\mathcal{H}_{\text{critical}} \subseteq \mathcal{H}, |\mathcal{H}_{\text{critical}}|  = \delta |\mathcal{H}| }  \{\mathcal{R}_h | {h\in\mathcal{H}}  \}
\end{equation}
Where $\delta \in (0, 1)$ is a hyperparameter used to define the proportion of heads considered reasoning-critical (e.g., $\delta=0.05$ is commonly used in~\citet{wu2024retrieval}).

We then generate the weakened reference model $\theta_{ref}$ by systematically disabling the computational function of the identified retrieval heads $\mathcal{H}_{\text{critical}}$.  
This is achieved through a \emph{masking operation} during the forward pass. For any layer $l$ and identified head $h \in \mathcal{H}_{\text{critical}}$, the attention weights are set to a uniform distribution.  Specifically, if the attention sequence length is $N$, the attention weight $a_{ij}$ for all tokens $j$ is set such that:
\begin{equation}
    a_{ij} = \frac{1}{L}, \quad \text{for all } j \in \{1, \dots, N\}
\end{equation}
This masking operation effectively nullifies the head's specialized retrieval capability, degrading the model's performance specifically in fact retrieval and reasoning while keeping all underlying weight parameters identical to $\theta_{base}$.

\textbf{Quantifying Attention Influence via Loss Divergence.} The weakened model $\theta_{ref}$ serves as a critical counterpart to the strong base model $\theta_{base}$. We leverage the difference in their cross-entropy losses to calculate the \textbf{Attention Influence Score ($\mathcal{S}$)}. Let $\ell(\theta, x_t)$ denote the token-level cross-entropy loss for a given model $\theta$ predicting the token $x_t$. The fundamental measure of influence for a single token is the loss gap between the two models:
\begin{equation}
    \Delta \ell (x_t) = \ell(\theta_\text{ref}, x_t) - \ell(\theta_\text{base}, x_t)
\end{equation}
A positive loss gap ($\Delta \ell (x_t) > 0$) signifies that the base model's performance on token $x_t$ heavily relied on the functional retrieval mechanism. This loss gap is then aggregated to derive the Step-Level and Sample-Level Attention Influence Scores, as detailed in the following sections. 

{Although attention influence is computed at the token level, we do not directly perform token-level selection. Reasoning is inherently structured and step-wise, and individual tokens do not constitute meaningful training units. Moreover, token-level divergence is often noisy and fails to reliably capture underlying reasoning dependencies. More importantly, token-level selection would fragment the reasoning chain and disrupt the semantic coherence required for learning multi-step logical patterns.}
By aggregating token-level influence into step-level and sample-level scores, we can obtain stable, semantically aligned indicators of reasoning importance that better match the sequence-level nature of SFT objectives. Therefore, step-level weighting and sample-level data selection can provide a more faithful way.

\subsection{Sample-Level Attention Influence Score}
\label{sec:sample_level_ai}

\noindent Based on the token-level loss divergence $\Delta \ell (x_t)$ derived from the $\langle \theta_{\text{ref}}, \theta_{\text{base}} \rangle$ model pair, we first quantify the influence at a global reasoning sample level. We define the total sequence loss for a sample $x$ as $\mathcal{L}(\theta, x) = \sum_{t=1}^{N} \ell(\theta, x_t)$. Since intrinsic sample difficulty varies, using absolute loss differences can be sensitive to the sample's scale. To mitigate this, we define the Sample-Level Score as the \textbf{relative loss divergence} (relative loss increase):
\begin{equation}
    \mathcal{S}_{\text{samp}}(x) = \frac{\sum_t \Delta\ell (x_t)}{\sum_{t} \ell(\theta_{\text{base}}, x_t)}
\end{equation}
Samples with higher $\mathcal{S}_{\text{sample}}$ reflect a stronger reliance on the retrieval mechanism for accurate generation, and are therefore prioritized during post-training data selection. 
In our experiments, we select the top $V$ samples with the highest $\mathcal{S}_{\text{samp}}(x)$ scores for fine-tuning. This score provides an important signal for global data curation. Comparisons of $\mathcal{S}_{\text{sample}}$ are typically performed only within individual domains or subsets (for instance, with or without CoT), since loss values cannot be meaningfully compared across domains or subsets.

\subsection{Step-Level Attention Influence Score}
\label{sec:step_level_ai}

\noindent While the sample-level metric curates training data from a global perspective, the step-level metric focuses on fine-grained local importance. We segment the reasoning trajectory into $K$ discrete reasoning steps, $\mathcal{S} = \{s_1, s_2, \dots, s_K\}$, typically delimited by structural separators (e.g., line breaks). Let $\mathbb{I}_k$ denote the set of token indices belonging to step $s_k$. The Step-Level Attention Influence Score for the $k$-th step is defined as the average loss divergence across all tokens within that step:
\begin{equation}
\mathcal{S}_{\text{step}}^{(k)} = \frac{1}{|s_k|} \sum_{t \in \mathbb{I}_k} \Delta \ell (x_t)
\end{equation}
where $|s_k|$ denotes the number of tokens (i.e., the size of the index set $\mathbb{I}_k$) in step $s_k$.

Based on $\mathcal{S}_{\text{step}}^{(k)}$, we adopt a threshold-based weighting strategy to prioritize steps highly dependent on the retrieval heads. We define the set of critical steps $\mathcal{K}_P$ as the top $P$ percent of steps exhibiting the highest scores:
\begin{equation}
\label{eq:step_p}
\mathcal{K}_P = \{ k \mid \mathcal{S}_{\text{step}}^{(k)} \in \text{Top } P\% \text{ of } \mathcal{S}_{\text{step}} \}
\end{equation}
The raw weight $\tilde{w}^{(k)}$ for step $k$ is then assigned based on membership in the critical step set $\mathcal{K}_P$:
\begin{equation}
\label{eq:step_a}
\tilde{w}^{(k)} = 1 + (\alpha-1) \cdot \mathbb{I}_{k \in \mathcal{K}_P}
\end{equation}
where $\alpha >= 1$ is the weight amplification factor that determines the boost applied to critical steps. The term $\mathbb{I}_{k \in \mathcal{K}_P}$ is the indicator function which evaluates to $1$ if the step index $k$ belongs to the set of critical steps $\mathcal{K}_P$ and $0$ otherwise.

All tokens $t$ belonging to step $k$ (i.e., $t \in I_k$) are assigned the uniform raw weight $\tilde{w}_t = \tilde{w}^{(k)}$. We apply sequence-level normalization to prevent weight magnitude from biasing the global learning rate. 
{For each post-training sample, which consists of a clearly defined model input and model output, we denote the total token count across all steps in the sample as $N$, namely the total number of tokens in the output.} The final normalized weight $w^{(k)}$ is scaled such that the weighted sum equals the total token count $N$:
\begin{equation}
w^{(k)} = \tilde{w}^{(k)} \times \frac{N}{\sum_{k=1}^{K} |s_k| \cdot \tilde{w}^{(k)}}
\end{equation}
Finally, the weighted SFT objective function is applied as follows:
\begin{equation}
\mathcal{L}_{\text{SFT}} = - \frac{1}{N} \sum_{k=1}^{K} w^{(k)} \sum_{t \in s_k} \log P_{\theta}(x_t | x_{<t})
\end{equation}

%% file: 4-exp.tex
\section{Experiments}
\label{sec:experiments}

In this section, we present empirical analysis to validate the effectiveness of reasoning-intensive data selected by \ours.
\subsection{Datasets and Metrics}
\label{sec:setup}
Following the experimental protocol of s1~\cite{muennighoff2025s1}, our experiments utilize datasets derived from the s1 project\footnote{\url{https://github.com/simplescaling/s1}}. For \textbf{\emph{the sample-level experiments}}, we apply data selection via \ours{} directly to the raw 59K-full dataset to identify high-value samples for post-training.
We then compare the performance of models trained on our selected subset against those trained on subsets selected by other methods.
Since the 59K-full dataset contains only Gemini-distilled trajectories, we apply the \ours{} filtering mechanism to this raw dataset, which is constructed from Gemini Flash Thinking~\cite{team2023gemini}, to identify high-value samples. For each selected example, we then create a corresponding version distilled from DeepSeek-R1~\cite{guo2025deepseek}\footnote{Due to cost constraints, we do not create the complete 59K R1 reasoning dataset.}. 
Finally, we report the performance of models trained on our \ours{}-selected subsets across both reasoning trajectory sources.
For \textbf{\emph{the step-level experiments}}, we use the s1K-1.1 dataset, a 1K subset curated from the 59K-full dataset and augmented with reasoning trajectories generated by DeepSeek-R1~\cite{guo2025deepseek} for post-training. The curation follows four criteria: \emph{Quality}, \emph{Difficulty}, \emph{Diversity}, and manual selection. In addition, as reported in~\cite{muennighoff2025s1}, the s1K-1.1 dataset achieves better performance than s1K. Further details on the datasets are provided in \autoref{appd:data_details}.

For evaluation, we selected four mainstream benchmarks categorized into mathematical reasoning and general scientific capabilities. In the mathematical domain, we employ {AIME 2024 \& 2025} \cite{aime2024part1,aime2025part1} and {MATH500} \cite{hendrycks2021measuring}. The former consists of challenging problems from the American Invitational Mathematics Examination, assessing competition-level proficiency in arithmetic, algebra, number theory, and geometry. The latter comprises 500 examples selected by OpenAI from the MATH dataset, serving as a comprehensive standard for mathematical problem-solving. 
For scientific and general knowledge, we include {GPQA Diamond} \cite{rein2023gpqa},  a challenging subset of GPQA consisting of 198 PhD-level questions across biology, chemistry, and physics.
All evaluations were conducted using the official code base provided by s1~\cite{muennighoff2025s1}, utilizing greedy decoding (temperature=0), and we report Pass@1 accuracy. 

\subsection{Baselines}
For the sample-level data selection experiments, our primary comparison is against the s1 model, which is trained on a 1K subset curated from a 59K dataset with the assistance of manual selection. This curated subset is available in two versions: s1K, which contains reasoning trajectories distilled from Gemini Thinking Experimental~\cite{geminithinking}, and s1K-1.1, which contains trajectories distilled from DeepSeek-R1~\cite{guo2025deepseek}.
In addition to these manually curated baselines, we also include several heuristic data-selection baselines provided in the s1 paper. These include Random, Random-by-type, Length, and Diverse, each representing a simple heuristic strategy for selecting a representative or informative subset from the full dataset.

For the step-level data selection experiments, we mainly compare against the s1.1 model, as our step-level \ours{} experiments are conducted on the s1K-1.1 dataset. This dataset is a 1K subset curated from the 59K-full dataset and augmented with reasoning trajectories generated by DeepSeek-R1~\cite{guo2025deepseek} for post-training, and it achieves better performance than s1K.
In addition to this baseline, we consider two alternative step-level selection strategies in the discussions. The first is a \emph{Random} strategy, which removes semantic guidance and simply assigns high weights to 20\% of steps randomly selected from each training instance. The second is an \emph{Entropy} strategy, where we use the same base model ($\theta_\text{base}$) employed in \ours{} to compute uncertainty scores. Specifically, we calculate the Shannon entropy for each token from the base model’s output logits and compute the mean token entropy over all tokens within a reasoning step. Following the same protocol used in our main method, the top 20\% of steps with the highest average entropy are selected for weighting.

\begin{table*}[h]
\centering
\small
\setlength{\tabcolsep}{10pt}
\caption{Performance of \colorbox{blue!10}{sample-level \ours{}} and \colorbox{yellow!30}{step-level \ours{}}. }
\resizebox{0.95\linewidth}{!}{%
\begin{tabular}{lcccccc}
\toprule
Model & \# Examples &  AIME 2024 & AIME 2025 & MATH500 & GPQA Diamond  & Average \\
\midrule
R1~\cite{guo2025deepseek} & $\gg$800K & 79.80 & 70.00 & 97.30 & 71.50 & 79.65 \\
\midrule
R1-distill-Qwen-14B~\cite{guo2025deepseek} & 800K & 61.70 & 48.00 & 93.90 & 59.10 & 65.68 \\
R1-distill-Qwen-32B~\cite{guo2025deepseek} & 800K & 58.30 & 49.60 & 94.30 & 62.10 &  66.08 \\
R1-distill-Llama-70B~\cite{guo2025deepseek} & 800K & 57.10 & 56.30 & 94.50 & 65.20 & 68.28 \\ \midrule[1pt]
\multicolumn{7}{c}{ \bf{s1K Setting: Distilling from  \underline{Gemini}}
}\\ \midrule[0.2pt]
Random  & 1K & 30.00 & 20.00 & 90.40 & 51.01  & 47.85 \\
s1K~\cite{muennighoff2025s1} & 1K & 50.00  & 26.70  & 92.60 & 56.60  & 56.48 \\
\rowcolor{blue!10}
\ours{}-Sample & 1K & 50.00 & 23.33 & 90.80 & 55.00  & 54.78 \\ \midrule[1pt]
\multicolumn{7}{c}{ \bf{s1K-1.1 Setting: Distilling from Deepseek \underline{R1}}
}\\  \midrule[0.2pt]
Random & 1K & {50.00} &  36.67 & 94.80& 58.59 &  60.02\\
s1K-1.1~\cite{muennighoff2025s1} & 1K & {56.70} & 50.00 & 94.40 & 60.60 &  65.43 \\
\rowcolor{blue!10}
\ours{}-Sample & 1K & {56.70}& {50.00} & {95.20} & {66.67} & {67.14} \\ \midrule[1pt]
\rowcolor{yellow!30}
s1K-1.1 + \ours{}-Step & 1K & \textbf{66.67} & \textbf{53.33} & \textbf{95.60} & \textbf{65.66}  &  \textbf{70.32} \\
\bottomrule
\end{tabular}%
}
\vspace{-3mm}
\label{tab:evaluation_2}
\end{table*}

\subsection{Implementation Details}
\label{sec:experiment_details}
In our experiments, we select {Qwen2.5-7B-Instruct} \cite{yang2024qwen2} as the base model for identifying attention heads critical to reasoning and constructing the reference model. Specifically, we identified and masked the attention heads ranking in the top 5\% by {retrieval scores} to build this weakened reference model ($\theta_\text{ref}$), which is consistent with~\citet{hua2025attentioninfluence}. We then calculate the {Attention Influence Score} by comparing loss divergence between the base model and the reference model.
Following the experimental protocol of s1~\cite{muennighoff2025s1}, we perform post-training on Qwen2.5-32B-Instruct~\cite{yang2024qwen2} model.

For \emph{the step-level selection setting}, we analyze the ratio of critical steps ($P\%$ in Equation~(\ref{eq:step_p})) and the weight amplification factor ($\alpha$ in Equation~(\ref{eq:step_a})). We then select the top 20\% reasoning steps according to their Attention Influence Score as critical steps and assign a higher loss weight of \(\alpha = 2\) to enhance learning of key reasoning logic.
For \emph{the sample-level selection setting}, to maintain a category distribution (e.g., Math, Science, and Crossword) consistent with s1K/s1K-1.1 datasets ($V=1000$), we do not perform global filtering on the original 59K data pool. Instead, we determine the sampling quota based on the category proportions in the standard dataset. Within each category, we select the top samples according to their $\mathcal{S}_{\text{samp}}(x)$ scores, ultimately constructing a training set that matches the scale of s1K/s1K-1.1 datasets (1,000 samples).

Following~\citet{muennighoff2025s1}, our model is fine-tuned for 5 epochs with a global batch size of 16 using the AdamW optimizer \cite{loshchilov2017decoupled} with parameters $\beta_1=0.9$, $\beta_2=0.95$, and a weight decay of $10^{-4}$.
The learning rate is initialized at $10^{-5}$, with a linear warmup over the first 5\% of training steps, followed by a cosine decay schedule. Furthermore, cross-entropy loss is calculated exclusively on the reasoning paths and final responses, excluding the user prompt. Additional experimental details are provided in \autoref{appd:training_details}.

\subsection{Evaluation Results}

\paragraph{Sample-Level \ours~ Efficacy.}
\autoref{tab:evaluation_2} reports the performance of both step-level and sample-level AIR evaluations. The results clearly demonstrate that replacing random selection with our \ours{} strategy substantially improves performance, raising the average accuracy from 47.85\% to 54.78\% under the s1K setting and from 60.02\% to 67.14\% under the s1K-1.1 setting. Compared with the manually curated s1K/s1K-1.1 baseline, our AIR-Sample approach achieves slightly lower or comparable performance with s1K, while consistently outperforming s1K-1.1 across all four datasets. Notably, our method relies solely on automatic \ours{} scores, eliminating the need for manual filtering, which demonstrates the efficiency and scalability of our approach. In addition, the superior performance of \ours{} on s1K-1.1, which replaces responses with stronger reasoning trajectories from DeepSeek-R1, further confirms that high-quality problems selected by our strategy, when combined with stronger reasoning paths, can more fully unlock the model’s potential. Remarkably, under the s1K-1.1 setting, models trained on only 1K examples selected by \ours{} even outperform R1-distill-Qwen-32B, which is trained on 800K examples distilled from DeepSeek-R1, highlighting the effectiveness of our approach in leveraging small, high-value training subsets.
For a detailed introduction to data processing efficiency,
please refer to \autoref{appd:data_efficiency}.

\paragraph{Step-Level Efficiency.}
The last row of \autoref{tab:evaluation_2} reports the performance of introducing the step-level weighting mechanism (\ours{}-Step). Compared to the s1K-1.1 baseline model, which achieves an average accuracy of 65.43\%, our \ours{}-Step method substantially improves the average accuracy to 70.32\%. Notably, on benchmarks requiring rigorous logical reasoning, such as AIME 2024 and GPQA Diamond, this approach achieves significant gains of +9.97\% and +5.06\%, respectively. These results indicate that applying fine-grained loss weighting to reasoning steps enables the model to more effectively capture critical reasoning logic, thereby enhancing both logical rigor and solution proficiency in complex mathematical and scientific problems. Furthermore, cross-model comparisons indicate that, using only 1,000 training samples, our method achieves higher performance than the R1-distill-Qwen-32B model, which is trained on 800K examples, and surpasses the R1-distill-Qwen-70B model on most benchmarks, demonstrating its strong data efficiency.

\section{Discussion}
\label{sec:discussion}

\begin{figure*}[t!]  
    \centering
    \includegraphics[width=\textwidth]{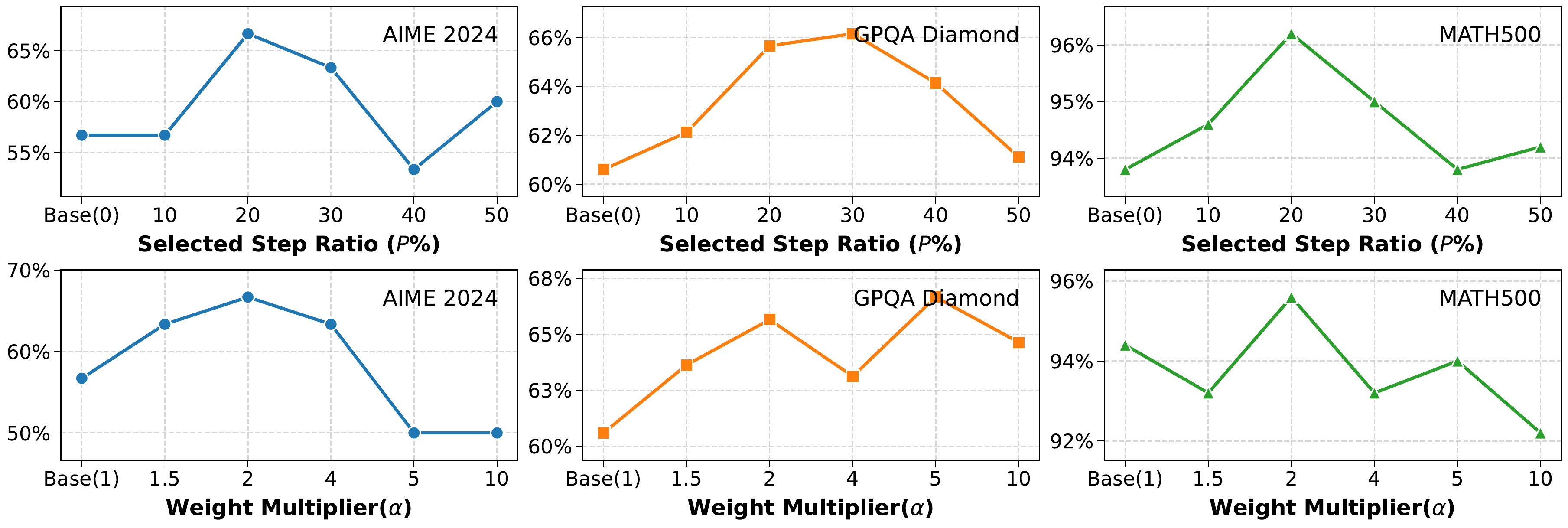} 
    \vspace{-2mm}
    \caption{ Hyperparameter Analysis in the step-level \ours{} SFT.}
    \label{fig:sensitivity}
    \vspace{-3mm}
\end{figure*}

\subsection{Hyperparameter Analysis}
To comprehensively evaluate the robustness of \ours{} and its dependence on hyperparameter settings, we conducted a sensitivity analysis on two key hyperparameters in the step-level \ours{} SFT process: the ratio of selected critical steps ($P\%$ in Equation~(\ref{eq:step_p})) and the weight amplification factor ($\alpha$ in Equation~(\ref{eq:step_a})). \autoref{fig:sensitivity} presents the evaluation results.
We can observe a general pattern: as the ratio of selected critical steps increases, model performance first improves and then declines. When the ratio of critical steps is set to 20\%, the model achieves peak performance across all three major benchmarks (AIME 2024, GPQA, and MATH500) with accuracies of 66.67\%, 65.66\%, and 95.60\%, respectively. However, when the ratio is further increased to 40\% or 50\%, a substantial number of non-critical or redundant steps are included in the high-weight category, leading to diminished performance.

Similarly, the weight amplification factor generally follows the same trend as the ratio of selected critical steps. A normalized weight multiplier of $2\times$ yields the optimal overall gain. In contrast, aggressively increasing the weight multiplier to $5\times$ or $10\times$ results in significant performance degradation, particularly on AIME 2024 and MATH500. This deterioration is likely due to excessive local gradient updates distorting the loss landscape, causing the model to overfit specific reasoning fragments at the expense of the overall coherence of the generative distribution. Therefore, choosing a moderate weight multiplier of $2\times$ is proper for maintaining the model’s general reasoning capability.

\begin{table}[t!]
\centering
\caption{Comparison of sample-level selection strategies. Except for the Full 59K dataset, all other datasets contain 1K samples.}
\setlength{\tabcolsep}{2pt}
\resizebox{\linewidth}{!}{
\begin{tabular}{lccccc}  
\toprule
Method & AIME 2024 & MATH500  & GPQA Diamond& AVG \\
\midrule
Full 59K & 53.30 & 92.80  & 58.10  & 68.07 \\  
\midrule
Random 1K& {36.70}  & 90.60  & 52.00& 59.77 \\ 
Random-type 1K& 30.00 & 90.40 & 51.01  & 57.14 \\ 
Length 1K& 33.30 & 90.40 & {59.60}  & 61.10 \\ 
Diverse 1K& 26.70  & {91.20}  & 54.60& 57.50 \\ 
\midrule
s1K& 50.00 & {92.60}  & 56.60 & 66.40 \\ 
\midrule
\textbf{\ours{} (Ours)} & {50.00} & {90.80} & {55.00} & {65.27} \\ 
\bottomrule
\end{tabular}
}
\vspace{-3mm}
\label{tab:sample_ablation}
\end{table}

\subsection{Comparison of Data Selection Strategies}
\autoref{tab:sample_ablation}  presents a comparison of different sample-level data selection strategies under the s1K setting\footnote{Since the heuristic baseline dataset only provides Gemini-generated outputs consistent with the standard s1k, obtaining the corresponding DeepSeek-R1 inference trajectories for these large-scale datasets is quite difficult and expensive. Therefore, although our method shows some performance differences compared to the base model on certain benchmarks under the s1k setting, we still retain this setting.}, where all datasets except the Full 59K dataset contain only 1,000 samples. Among the heuristic baselines, Random 1K and Random-type 1K generally underperform, indicating that unguided selection is insufficient for capturing high-value training samples. Simple heuristic strategies, such as Length and Diverse, can improve performance on individual benchmarks but show inconsistent results across datasets and are significantly worse than our \ours{}. Among these heuristics, Length is the most competitive, achieving the highest score on GPQA Diamond (59.60\%), which is reasonable because longer samples tend to contain richer reasoning steps and more complex contexts, providing more informative training signals. Notably, the s1K baseline, which relies on manual curation, achieves competitive performance, particularly on MATH500 (92.60\%). While \ours{} is slightly below the manually curated s1K baseline (66.40\%), it achieves these results fully automatically, demonstrating its ability to effectively identify high-value samples without human intervention and providing a practical, scalable alternative to manual selection.

\begin{table}[t!]
\centering
\caption{Comparison of step-level weighting strategies. With the exception of s1k-1.1 baseline, all methods select a subset of 20\% of the steps and emphasize their contribution via \ours{} SFT.}
\setlength{\tabcolsep}{2pt}
\resizebox{\linewidth}{!}{
\begin{tabular}{lcccc} 
\toprule
Method & AIME 2024 & MATH500 & GPQA Diamond  & AVG \\
\midrule
s1K-1.1 & 56.70 & 94.40  & 60.60 & 70.57 \\ 
\midrule
Random & 56.67  & 94.00 & 61.62 & 70.76 \\
Entropy & 63.33 & 93.80 & 64.14 & 73.76 \\
\textbf{\ours-step (Ours)} & \textbf{66.67} & \textbf{95.60} & \textbf{65.66}  & \textbf{75.98} \\
\bottomrule
\end{tabular}
}
\label{tab:step_ablation}
\end{table}

\autoref{tab:step_ablation} compares different step-level weighting strategies under the s1K-1.1 setting, where all methods except s1K-1.1 select 20\% of the steps and apply weighted SFT. We choose s1K-1.1 as the base for step-level experiments because, as shown in \autoref{tab:evaluation_2}, it consistently outperforms s1K, providing a stronger starting point for evaluating step selection methods.
Random achieves performance comparable to s1K-1.1, indicating that unguided step selection does not lead to substantial improvements. The Entropy-based strategy provides moderate gains, suggesting that emphasizing high-uncertainty steps can help the model focus on informative reasoning signals. Our AIR-step method consistently outperforms all baselines, improving the average score from 73.76\% (Entropy) to 75.98\%. This demonstrates that AIR-step effectively identifies and amplifies critical reasoning steps, resulting in stronger step-level supervision and improved overall model performance.

\begin{table}[t!]
\centering
\caption{Comparison of sample characteristics between s1K and \ours{} selected dataset.}
\resizebox{0.95\linewidth}{!}{%
\begin{tabular}{lcc}
\toprule
\textbf{Metric} &  \textbf{s1K-1.1} &\textbf{\ours{} (Ours)}  \\
\midrule
Average Reasoning Steps & 265.15  & 274.05  \\
Prompt Constraint Density & 1.11\%  & 1.45\% \\
Numeric Answer Ratio & 54.48\%  & 42.24\% \\
Symbolic Answer Ratio & 45.52\% & 57.76\%  \\ 
\bottomrule
\end{tabular}%
}
\label{tab:dataset_characteristics}
\end{table}

\begin{figure}[t!]  
    \centering
    \includegraphics[width=0.8\linewidth]{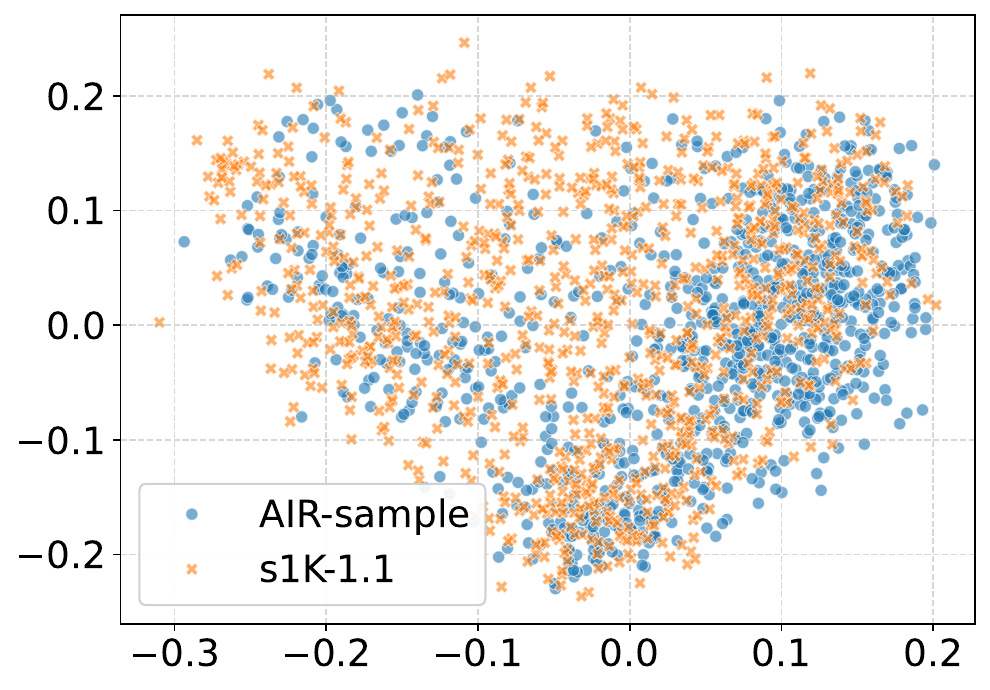} 
    \caption{The PCA visualization of the data distribution of s1K-1.1 and \ours{} selected samples.}
    \label{fig:pca_dataset}
\end{figure}

\begin{table}[t!]
\centering
\caption{Comparison of reasoning characteristics between s1K-1.1 and s1K-1.1 (+\ours{}-Step) models on AIME 2024.}
\resizebox{\linewidth}{!}{%
\begin{tabular}{lcc}
\toprule
Metric & \textbf{s1K-1.1} & \textbf{s1K-1.1 (+\ours{}-Step)} \\
\midrule
\multicolumn{3}{l}{\textit{Macro Statistics}} \\
Avg. Response Length & 10,652.87 & 10,537.00 \\
Transition Density (\%) & 8.846 & 8.862 \\
\midrule
\multicolumn{3}{l}{\textit{Reasoning Connectives (\%)}} \\
Causal & 2.82 & 2.80 \\
Contrast & 1.05 & 1.13 \\
Correction & 1.06 & 1.16 \\
Exploratory & 1.54 & 1.54 \\
Planning  & 0.97 & 1.04 \\
Verification  & 1.02 & 1.08 \\
\midrule
\multicolumn{3}{l}{\textit{Symbolic \& Structure(\%)}} \\
Symbolic Ratio & 2.37 & 2.52 \\
Header Line  & 0.03 & 0.28 \\
List Line& 9.97 & 9.64 \\
\midrule
\multicolumn{3}{l}{\textit{Reasoning Rhythm}} \\
Sentence Volatility (CV \%) & 94.65 & 96.14 \\
\bottomrule
\end{tabular}%
}
\label{tab:linguistic_analysis}
\end{table}

\begin{table*}[t!]
\centering
\small
\caption{Step-by-step comparison of reasoning trajectories on an AIME 2024 example between the base model and step-level \ours{} model. For clarity, only key reasoning steps are shown. Gray-shaded areas indicate explanatory analysis of the models’ reasoning steps.}
\label{tab:case_study_roots_analysis}
\resizebox{\textwidth}{!}{%
\begin{tabular}{p{0.48\textwidth} p{0.48\textwidth}}
\toprule
\multicolumn{2}{p{0.96\textwidth}}{
\textbf{Problem:} Let $\omega\neq 1$ be a 13th root of unity. Find the remainder when $\prod_{k=0}^{12}(2-2\omega^k+\omega^{2k})$ is divided by 1000.
} \\
\midrule
\textbf{s1K-1.1 (Baseline)} & \textbf{s1K-1.1 + \ours{} (Ours)} \\
\cmidrule(r){1-1} \cmidrule(l){2-2}

\textit{...Let me pair terms $y$ and $1/y$. Let $t = y + 1/y$. ... I need the resultant of the minimal polynomial $f(t)$ and $g(t) = 2t^2 - 6t + 5$...} \par
\vspace{2pt}
\noindent\colorbox{gray!30}{%
    \parbox{\dimexpr\linewidth-2\fboxsep}{%
        \small{(\textbf{Methodological Fixation}: Failing to see the complex structure, it retreats to a brute-force substitution $t$, increasing complexity.)}%
    }%
}
\vspace{8pt}

& 
\textit{...Let $x = \omega^k$. The term is $x^2 - 2x + 2$. The roots are $1 \pm i$. So, $2 - 2\omega^k + \omega^{2k} = (\omega^k - (1 + i))(\omega^k - (1 - i))$...} \par
\vspace{2pt}
\noindent\colorbox{gray!30}{%
    \parbox{\dimexpr\linewidth-2\fboxsep}{%
        \small{(\textbf{Critical Path Awareness}: The model identifies the high-value strategy of factoring the quadratic over $\mathbb{C}$ instead of $\mathbb{R}$.)}%
    }%
}
\vspace{8pt}
\\ 

\textit{...Divide $f(t)$ by $g(t)$... First term... Second term... Remainder is $-25t + 187/8$...} \par
\vspace{2pt}
\noindent\colorbox{gray!30}{%
    \parbox{\dimexpr\linewidth-2\fboxsep}{%
        \small{(\textbf{Cascading Error}: The model enters a "Computation Quagmire" of long division. High complexity leads to fractional artifacts.)}%
    }%
}
\vspace{8pt}

& 
\textit{...We know that $\prod_{k=0}^{12} (z - \omega^k) = z^{13} - 1$. Thus, the product collapses to $P = ((1+i)^{13} - 1)((1-i)^{13} - 1)$...} \par
\vspace{2pt}
\noindent\colorbox{gray!30}{%
    \parbox{\dimexpr\linewidth-2\fboxsep}{%
        \small{(\textbf{High-Influence Locking}: Applies the identity to perform Dimensionality Reduction, bypassing iterative calculation.)}%
    }%
}
\vspace{8pt}
\\

\textit{...Wait, the remainder has a fractional coefficient... $S \approx 11.112$... This contradicts the fact that S is an integer...} \par
\vspace{2pt}
\noindent\colorbox{gray!30}{%
    \parbox{\dimexpr\linewidth-2\fboxsep}{%
        \small{(\textbf{Logical Fracture}: Arithmetic errors accumulate, causing the model to lose track of integer constraints and crash.)}%
    }%
}

& 
\textit{...Using Polar form: $(1+i)^{13} = -64(1+i)$. Calculating product... $8321 \equiv 321 \pmod{1000}$.} \par
\vspace{2pt}
\noindent\colorbox{gray!30}{%
    \parbox{\dimexpr\linewidth-2\fboxsep}{%
        \small{(\textbf{Computational Resilience}: Maintains logical consistency through complex arithmetic to reach the correct integer result.)}%
    }%
}
\\ \midrule

\textbf{Final Answer: 7 (Incorrect)} & \textbf{Final Answer: 321 (Correct)} \\
\bottomrule
\end{tabular}%
}
\end{table*}

\subsection{Analysis of \ours{} Selected Samples}
To comprehensively elucidate how different selection strategies shape the underlying data distribution and enhance post-hoc interpretability, we consider several evaluation metrics to systematically compare the dataset selected by \ours{} with the baseline s1K-1.1 dataset. Specifically, the framework assesses logical complexity through the \textit{Average Reasoning Steps} (ARS), which quantifies the granularity and depth of the reasoning path. In addition, we examine \textit{Prompt Constraint Density} (PCD) to capture the richness and specificity of restrictive conditions within input instructions. Finally, we evaluate answer complexity using the \textit{Numeric Answer Ratio} (NAR) and \textit{Symbolic Answer Ratio} (SAR)—defined as the proportions of purely numeric answers and answers containing symbolic expressions, respectively—to characterize the extent to which solutions rely on formal logical deduction.
For detailed metric definitions and computation procedures, please refer to \autoref{appd:dataset_eval_metrics}.

The statistical results presented in Table~\ref{tab:dataset_characteristics} reveal the advantages of \ours{} dataset regarding cognitive load and logical density. {In terms of reasoning characteristics, the \textit{Average Reasoning Steps} of \ours{} (274.05) are higher than those of s1K-1.1 (265.15). This indicates that samples selected via attention influence tend to require more granular problem decomposition and more complex logic}. Regarding cognitive constraints, \ours{} exhibits a higher Prompt Constraint Density (1.45\% vs. 1.11\%), implying that the selected problems often involve more specific or stringent boundary conditions, thereby requiring the model to perform highly consistent deductions within a given logical framework. Crucially, in terms of answer complexity, the proportion of answers containing symbolic and formulaic derivations in \ours{} reaches 57.76\%, substantially surpassing the 45.52\% observed in s1K-1.1. This structural disparity suggests that \ours{} prioritizes the retention of samples necessitating symbolic manipulation, abstract generalization, and structured deduction.

Furthermore, to visually illustrate the topological distribution of the selected datasets within the latent semantic space, we employed Principal Component Analysis (PCA) to perform dimensionality reduction on the high-dimensional embedding representations of the selected data items. As shown in \autoref{fig:pca_dataset}, unlike the s1K-1.1 dataset, which exhibits a relatively dispersed distribution due to its pursuit of diversity metrics, \ours{} selected samples present a distinctly more compact cluster structure on the two-dimensional plane. This is attributed to our method's exclusive reliance on the model's internal attention influence mechanism, without incorporating diversity constraints. However, these clustered regions highly overlap with the core semantic areas of the s1K-1.1 dataset. This indicates that while the data selected via the \ours{} strategy is more structurally compact converged, it maintains strong semantic consistency with the core semantics of the s1K-1.1 dataset.

\subsection{Impact of Step-level \ours{} on Reasoning Behavior}
To investigate changes in reasoning behavior at a granular level, we conducted a quantitative semantic analysis on the responses generated by the s1.1 baseline and our \ours{} model on the AIME 2024 benchmark.\footnote{For a detailed description of the metrics, please refer to \autoref{appd:reasoning_analysis_methord}.} Table \ref{tab:linguistic_analysis} illustrates the disparities in reasoning dynamics between the two models. At the macro level, results based on dependency syntactic analysis reveal that, despite a slight reduction in average response length, the Transition Density of the \ours{} model increased from $8.846\%$ to $8.862\%$, indicating enhanced logical coherence. At the micro level, an analysis of logical connective distribution shows a marked rise in the density of "Correction," "Planning," and "Verification" tokens. This reflects more frequent reflection and self-correction during the reasoning process, demonstrating a more rigorous reasoning chain. Regarding structural organization, in contrast to the baseline’s preference for linear lists (List Lines), the \ours{} model exhibits a significant increase in the usage of Header Lines and mathematical symbols. This shift indicates that the \ours{} strategy successfully encourages the model to decompose complex thought processes into modular logical units. Furthermore, the increased volatility in sentence length mirrors a human-like reasoning rhythm—employing long sentences for complex deduction and short ones for qualitative confirmation—thereby facilitating the construction of a more robust logical framework.

\subsection{Case Study: Step-level \ours{} Model Output}

To empirically evaluate the effectiveness of the \ours{} framework in improving reasoning capabilities at a microscopic level, we conducted a qualitative analysis of the model's generated responses on the AIME24 dataset. As shown in Table~\ref{tab:case_study_roots_analysis}, {s1K-1.1+\ours{}} model, trained with step-level \ours{} weighting, is superior to the baseline {s1K-1.1} model in terms of solution accuracy and logical rigor, with the two exhibiting distinct reasoning styles.
Notably, {s1K-1.1+\ours{}} model demonstrates clear critical path awareness, capable of identifying and locking onto high-value strategies for problem-solving, such as complex number transformations in algebra or coordinate system construction in geometry. In contrast, the baseline model struggles to maintain the coherence of the reasoning chain when dealing with complex tasks involving long-range calculations or multi-branch case discussions. It frequently resorts to random guessing after retrieval failures or tends to abandon rigorous deduction in favor of intuition-based, erroneous generalizations. While this tendency might grant the model some exploratory nature in local steps, it easily triggers a cascading accumulation of errors in long-chain reasoning.
Conversely, {s1K-1.1+\ours{}} model exhibits improved computational resilience and logical consistency. It persists in performing tedious calculations and exhaustive case verifications based on fundamental mathematical principles, thereby effectively avoiding logical fractures. This phenomenon indicates that our method successfully helps the model precisely locate ``High-Influence'' steps in the reasoning trajectory that are decisive for the final result. By reinforcing the execution fidelity of these key logical nodes through weighted training, the model maintains a stable reasoning state when facing non-trivial problems, thereby acquiring deeper mathematical insight.

\begin{figure}[t!]
    \centering
    \refstepcounter{figure} 
    \begin{tcolorbox}[
        enhanced,
        colback=white, 
        colframe=gray!60,       
        fonttitle=\bfseries,
        arc=2mm, 
        boxrule=0.5mm,
        left=2mm, right=2mm, top=2mm, bottom=2mm
    ]
    
    \noindent
    \begin{minipage}{\linewidth}
        \tiny 
        \linespread{0.9}\selectfont 
        \setlength{\parindent}{0pt}
        
        \textbf{Problem}: Given a rational number, write it as a fraction in lowest terms and calculate the product of the resulting numerator and denominator. For how many rational numbers between 0 and 1 will $20!$ be the resulting product?
        
        \vspace{0.8em}
        \hrule  
        \vspace{0.8em}
        
        \textbf{Reasoning Snippet} \\
        \centerline{\textbf{...}}  
        \vspace{0.2em}
        
        Therefore, the number of such coprime pairs is $2^k$, where $k$ is the number of distinct prime factors of $20!$. But wait, $20!$ has prime factors: 2, 3, 5, 7, 11, 13, 17, 19. That's 8 primes.
        
        \vspace{0.2em}
        \myhl{Since all exponents except for 2, 3, 5, 7 have exponent 1, and exponents for 2, 3, 5, 7 are} \myhl{even and odd.}
        
        \vspace{0.2em}
        Therefore, $20!$ is not a perfect square. Thus, there are no solutions where $a = b$. Hence, total coprime pairs are $2^8$. But since each assignment corresponds to a unique pair, total number of unordered pairs would be $2^{8 - 1} = 128$.
        
        \vspace{0.2em}
        \myhl{Wait, perhaps not quite. Since not all assignments will satisfy $a < b$, even for coprime} \myhl{divisors. Wait, confusion arises here.}
        
        \vspace{0.2em}
        Wait, actually, we need to be precise. The assignments consist of selecting a subset $S$ of the set of prime factors. The number of pairs with $a < b$ is equal to the number of possible $a$ such that $a < \sqrt{20!}$. Since $20!$ is not a square, the total number should be $256 / 2 = 128$.
        
        \vspace{0.2em}
        \myhl{But wait, can it be? Let me check with a smaller factorial.}
        
        \vspace{0.2em}
        Suppose $n = 6$. $6! = 720$. Prime factorization: $2^4 \times 3^2 \times 5^1$. Primes are 2, 3, 5 (3 primes). Total assignments $2^3 = 8$. We expect $8/2 = 4$ pairs where $a < b$.
        
        \vspace{0.2em}
        \myhl{So for 720, with prime factors 2, 3, 5: The 8 assignments:} \myhl{1. Assign none to a => a=1...} \myhl{2. Assign 2 to a...} \myhl{...} \myhl{5. Assign 2 and 3 to a => a=144, b=5 (a > b: no)}
        
        \vspace{0.2em}
        Therefore, out of 8 assignments, 4 result in $a < b$. So in general, the answer seems to be half of the number of coprime ordered pairs. For $20!$, it's $2^8 / 2 = 128$.
        
        \vspace{0.2em}
         {\textbf{...}} 
        \vspace{0.8em}
        \vspace{0.8em}
        
        \textbf{The final answer: 128} 
    \end{minipage}
    \end{tcolorbox}
    
    \caption{Visualization of critical reasoning steps selected by our \ours{}. The example illustrates a self-correction process: the model first derives an answer, then questions its logic with respect to the $a<b$ constraint, performs a verification on a smaller case ($6!$), and ultimately confirms the result. For ease of illustration, only key reasoning steps are shown. }
    \label{fig:reasoning_snippet} 
\end{figure}

\subsection{Case Study: Steps Selected by the \ours{} Model}
\autoref{fig:reasoning_snippet} illustrates how the AIR model selects critical reasoning steps. Steps marked in red indicate those chosen by AIR, while some intermediate reasoning steps have been omitted for clarity. The AIR scoring mechanism emphasizes \emph{logical pivots} that guide the reasoning direction and key \emph{self-correction} moments where the model revises incorrect assumptions. Routine declarations or low-information intermediate calculations are assigned lower weights. By focusing optimization on the most information-dense steps, AIR effectively distills the causal structures essential for maintaining coherent reasoning chains, leading to improved performance on complex multi-step tasks. More empirical cases are provided in \autoref{appd:visualization_step}.

%% file: 2-related.tex
\section{Related Work}

\subsection{Data Selection for Pretraining}
Many training-free methods use heuristic filtering rules ~\citep{rae2021scaling, xie2023data} or perplexity of existing LLMs ~\citep{ankner2024perplexed} to assess the quality of pretraining data. For instance, 
Scaling Filter ~\citep{li2024scalingfilter} evaluates text quality by measuring the perplexity difference between a small and a large language model trained on the same dataset. 
Some methods leverage weak supervision from Wikipedia-style text to identify high-quality documents (e.g., Llama 2), while others such as DCLM fit user preferences from behavioral signals.
In contrast, methods that train models using human-labeled or LLM-generated labels—such as Llama 3, FineWeb-Edu, and ProX—have gained more attention due to their higher accuracy and broader applicability. AttentionInfluence \cite{hua2025attentioninfluence}  is the first method to leverage the attention mechanisms of transformers for pretraining data selection, while remaining completely unsupervised and training-free. 
Another line of work ~\citep{wettig2024qurating,zhao2024decoratelm,peng2025dataman} focuses on training multi-task or multi-class classifiers using data labeled by proprietary commercial LLMs such as the GPT series. There are also efforts to train and combine several domain-specific classifiers ~\citep{wettig2025organize,FineFineWeb} for practical usage.

\subsection{Data Selection for Post-training}
Post-training data selection aims to support two key types of tuning: general alignment tuning, which curates high-quality and diverse instruction–response pairs~\cite{ding2023enhancing,zhou2023lima}, and reasoning tuning, which focuses on samples with informative chains of thought~\cite{kumar2025llm}.
Compared to pretraining data selection~\cite{yu2024mates}, post-training data selection operates on a much smaller set of examples,  which allows for more computationally expensive and fine-grained selection strategies.
For instance, manual curation~\cite{muennighoff2025s1, ye2025limo} or assessment by large models such as DeepSeek R1 or GPT-4o has been used to identify high-quality reasoning samples. Post-training samples can be assessed with richer posterior information, for example by measuring the loss difference between the answer alone and the answer conditioned on the question \cite{li2024quantity}, or—when chain-of-thought (CoT) signals are available—by comparing the loss between answers given only the question and answers conditioned on both the question and its CoT \cite{wang2025reverse}. Additional signals, such as question difficulty \cite{li2025naturalthoughts} (e.g., pass rate), sample-level influence \cite{jiang2025importance} or token-level mirror influence \cite{lin2024rho, qin2025sstoken}, can further guide the selection of informative samples or tokens for reasoning distillation.

There is also another line of work \cite{wang2025beyond, wang2025stabilizing, yu2025gpo} that focuses on identifying the critical tokens {or steps} in reinforcement learning settings, which provides insights into how token-level importance can influence model learning and downstream reasoning performance.

\subsection{Mechanistic Interpretability}
Understanding the inner workings of LLMs is crucial for advancing artificial general intelligence in a reliable and safe manner.
\citet{olsson2022context} and \citet{wu2024retrieval} reveal certain heads are responsible for in-context learning and retrieval, respectively. 
\citet{lv2024interpreting} further explores how attention heads and MLPs collaborate for factual recall. 
Sparse autoencoders \citep{anthropic_sparse_autoencoder} and head importance estimation \citep{fu2024not} are also used to analyze or optimize head behaviors. \ours{} adopts a proxy task, proposed by \citet{wu2024retrieval,qiu2024clongeval}, to detect specific important heads, namely the retrieval heads in this paper. \ours{} naturally extends the insights from \citet{wu2024retrieval}, broadening their application beyond model analysis and inference acceleration to include effective and efficient data selection.

\subsection{Influence Measure}
\citet{ruis2024procedural} uses influence
functions to recognize pretraining documents important for learning factual knowledge and mathematical reasoning separately. Mirror Influence ~\citep{ko2024mirrored} realizes an efficient data influence estimation to select high-quality data. MATES ~\citep{yu2024mates} continuously adapts a data influence model to the evolving data preferences of the pretraining model and then selects the most effective data for the current pretraining progress. Our work is similar to Mirror Influence in that we use data influence estimation to select high-quality data. However, while Mirror Influence requires a high-quality dataset to train a strong reference model and create a model pair with significant differences in capabilities to compute the data-loss gap, our approach uses the attention mechanism to derive a weaker reference model from the base model. This enables us to obtain two models with a significant capability gap and compute the data-loss gap to evaluate data quality.

%% file: 5-appendix.tex
\newpage
\appendix
\onecolumn

\section{Dataset Details}
\label{appd:data_details}

In this work, we leverage the datasets introduced by the s1 project~\cite{muennighoff2025s1}. The details of these datasets are described below:

\textbf{1. The 59K-full Dataset.} 
This corpus comprises approximately 59,000 questions compiled from 16 distinct sources. Furthermore, the pool incorporates novel, manually curated quantitative reasoning datasets, such as PhD-level statistics problems (s1-prob) and brain teasers (s1-teasers). A manual inspection was conducted during the collection phase to ensure data quality, resulting in the direct exclusion of datasets characterised by poor formatting or insufficient reasoning depth. We utilize this raw data pool for our sample-level filtering experiments.

\textbf{2. The s1K Dataset.}
The s1K dataset is a high-quality subset of 1,000 samples derived from the 59K-full dataset. To construct this dataset, \citet{muennighoff2025s1} employed a pipeline that integrates automated heuristics with manual curation. The extraction procedure consists of the following stages:

\begin{enumerate}[leftmargin=5em]
    \item[\textbf{Stage 1:}] \textbf{Quality Filtering.}
    The process initiates with an automated filter designed to remove samples that trigger API errors, as well as those containing specific string patterns indicative of formatting issues (e.g., ASCII art, malformed citations). This step reduced the data pool to approximately 51.5K samples.

    \item[\textbf{Stage 2:}] \textbf{Difficulty Screening.}
    To guarantee sufficient problem complexity, a model-based difficulty filter was applied. Each question was evaluated using two models (Qwen2.5-7B-Instruct and Qwen2.5-32B-Instruct). Questions correctly solved by either model were deemed ``too trivial'' and subsequently discarded. This filtering step significantly reduced the pool to approximately 24.5K samples.

    \item[\textbf{Stage 3:}] \textbf{Diversity Awareness.} This stage aims to ensure broad coverage and balanced complexity in the selected data by combining manual curation with domain-stratified sampling.
    \begin{itemize}[leftmargin=0em]
        \item \textbf{Manual Pre-selection (384 samples):} Based on manual quality assessments, 384 samples were directly selected from high-reliability datasets.
        \item \textbf{Diversity Sampling (616 samples):} The remaining quota was filled using a stratified sampling algorithm. Questions were categorized into 50 scientific domains (e.g., number theory, quantum mechanics) via the MSC system. The sampling algorithm ensured uniform domain coverage while prioritizing questions with longer reasoning trajectories, which served as a proxy for complexity.
    \end{itemize}
\end{enumerate}

\textbf{3. The s1K-1.1 Dataset}
The s1K-1.1 dataset retains the identical set of 1,000 questions selected through the aforementioned pipeline but regenerates the reasoning trajectories using \textbf{DeepSeek-R1}. Compared to the Gemini-distilled trajectories utilized in the original s1K, the trajectories in s1K-1.1 are generally more extensive and demonstrate superior reasoning capabilities, thereby providing higher-quality supervision signals for our step-level experiments.

\section{Training Details}
\label{appd:training_details}
All training experiments are run on a platform with {8} NVIDIA A800 GPUs, using DeepSpeed for distributed training. We adopt the context configuration from s1.1 and set the maximum sequence length of DeepSeek-R1–generated reasoning traces to 20,000 tokens. {To optimize memory efficiency, we employ bfloat16 precision throughout the training process,}

\section{Efficiency in Data Selection with AIR}
\label{appd:data_efficiency}
AIR demonstrates high efficiency in selecting high-value training samples for reasoning distillation. Specifically, processing the 59K-sample data pool using Qwen2.5-7B-Instruct required only approximately 6.9 hours on 8 NVIDIA A800 GPUs, corresponding to an average processing time of 0.42 seconds per sample. Importantly, this time reflects offline, one-time data selection and is significantly faster than manual curation of high-quality reasoning traces. Importantly, the selected data can be used to train different models, consistently improving their multi-step reasoning performance.

\section{Evaluation Metrics for Comparing the Selected Dataset}
\label{appd:dataset_eval_metrics}
To quantify the linguistic characteristics and reasoning difficulty of the datasets, we adopted four key metrics in our analysis. The specific calculation logic and definitions are detailed below:

\begin{enumerate}
    \item \textbf{Average Reasoning Steps (ARS)}:    We employ ARS as a proxy metric for the granularity of the logical derivation. The calculation is performed by segmenting the model's response using newline characters as delimiters; each resulting line is treated as a distinct reasoning step. Consequently, the metric is derived by averaging the total count of these steps across all samples in the dataset. A higher number of reasoning steps typically indicates a more detailed problem decomposition, where the model explicitly articulates intermediate sub-goals, calculations, and verification procedures rather than relying on implicit leaps or shallow heuristics.

    \item \textbf{Prompt Constraint Density (PCD)}: This metric aims to measure the strictness and closedness of the problem definition. We first extract the user query from the multi-turn dialogue and perform matching based on a predefined vocabulary of mathematical constraints (including keywords such as "given", "assume", "such that", "satisfy", "where", etc.). Constraint density is defined as the total frequency of these constraint keywords divided by the total number of words in the question. High density implies that the problem is a well-defined, closed-ended problem with clear boundaries, requiring the model to strictly adhere to the given logical framework during deduction.

    \item \textbf{Answer Complexity}: This metric reflects whether the problem-solving process tends towards "numerical convergence" or "symbolic deduction." We utilize regular expressions to classify the content marked by \texttt{\textbackslash boxed\{\}} in the model's output:
    \begin{itemize}
        \item \textbf{Numeric Answer Ratio (NAR):} If the answer contains only digits, decimal points, or negative signs (matching the regex pattern \texttt{\textasciicircum-?\textbackslash d+(\textbackslash .\textbackslash d+)?\$}), it is categorized as numeric. Such problems typically emphasize the robustness of long-chain calculations.
        \item \textbf{Symbolic Answer Ratio (SAR):} If the answer contains any non-numeric characters (such as variables $x$, constants $\pi$, radicals $\sqrt{}$, or function expressions), it is categorized as symbolic. Such problems generally involve structured reasoning, abstract generalization, and symbolic manipulation, thereby placing higher demands on the model's mathematical reasoning abilities.
    \end{itemize}
\end{enumerate}

\section{Evaluation Metrics for Comparing Model Reasoning Outputs} 
\label{appd:reasoning_analysis_methord}

To comprehensively quantify the micro-level reasoning characteristics of the model's Chain-of-Thought (CoT), we establish an automated analysis pipeline based on the Natural Language Processing tool \texttt{spaCy}~\footnote{\url{https://spacy.io/}}. The specific metric definitions and calculation logic are as follows:

\subsection{Macro Statistics}
We load the \texttt{en\_core\_web\_sm} pre-trained model to perform tokenization and dependency parsing on the \textit{complete response text} generated by the model (encompassing both the reasoning process within \texttt{<think>} tags and the final response body). To capture the density of logical flow during the reasoning process, we iterate through every token in the document to extract vocabulary where the dependency tag belonged to the following specific sets as \emph{logic words}:

\begin{itemize}
    \item \textbf{Marker (\texttt{mark} in spaCy)}: Markers introducing clauses or adverbials, typically used to construct causal or conditional relationships (e.g., \textit{because, since, that, if}).
    \item \textbf{Coordinating Conjunction (\texttt{cc} in spaCy)}: Conjunctions used to connect logical branches or indicate transitions/juxtapositions (e.g., \textit{and, but, or}).
    \item \textbf{Adverbial Modifier (\texttt{advmod} in spaCy)}: Modifiers often used to indicate inference conclusions, emphasis, or contextual reversals (e.g., \textit{however, therefore, thus, instead}).
\end{itemize}

Based on the extraction results above, we calculate the following core metrics:
\begin{itemize}
    \item \textbf{Avg. Response Length}: The average total number of tokens as statistically determined by the spaCy tokenizer.
        
    \item \textbf{Transition Density}: Defined as the proportion of logical words relative to the total number of tokens, used to measure the density of logical connectives within the reasoning process.

\end{itemize}

\subsection{Reasoning Connectives}
Reasoning connectives refer to logical and discourse markers that explicitly signal transitions within a reasoning process. We categorize these connectives into six dimensions and identify them using regular-expression-based matching. Unlike generic word frequency measures, this design focuses on connectives that are informative for mathematical and logical reasoning. We report the proportion of each connective type relative to the total output length. The specific dimensions are defined as follows:
\begin{itemize}
    \item \textbf{Causal}: Measures reasoning coherence. Includes causal words like \textit{therefore, thus, hence, consequently, because, since}.
    \item \textbf{Contrast}: Measures adjustments in the thought path. Keywords include \textit{however, but, yet, nevertheless, nonetheless, on the other hand, conversely}.
    \item \textbf{Correction}: Measures the model's ``metacognitive monitoring'' capability, i.e., the frequency of self-negation or pausing. Core keywords include \textit{wait, no, hold on, hang on, actually, mistake, error, incorrect, wrong, let me check}.
    \item \textbf{Exploratory}: Captures divergent reasoning by identifying markers of hypothesis exploration (e.g., \textit{alternatively, maybe, perhaps, possibly, another way, approach, try}).
    \item \textbf{Planning}: Captures anticipatory reasoning by identifying markers of pre-action structuring, such as \textit{let me, let us, let's, I need to, we need to, goal, target}.
    \item \textbf{Verification}: Measures the confirmation behavior regarding intermediate results, such as \textit{ check, verify, ensure, confirm, valid, satisfies, correctly}.
\end{itemize}

\subsection{Symbolic \& Structure}

\textbf{Symbolic Ratio}: Measures the model's reliance on formalized language. Using regular expressions, we extract all formula blocks—including inline formulas (\texttt{\$...\$}) and display formulas (\texttt{\textbackslash[...\textbackslash]})—and calculate the ratio of characters within formulas to the total text length.

\textbf{Structural Metrics}: Quantifies the organizational layout of the text to capture formality and readability:
\begin{itemize}
    \item \textbf{Header Line}: Lines starting with Markdown header symbols (\texttt{\#}), reflecting the degree of modularity in reasoning.
    \item \textbf{List Line}: Lines starting with bullet points (\texttt{-}, \texttt{*}) or numbering (\texttt{1.}), reflecting enumerative features of the thought process.
\end{itemize}

\subsection{Reasoning Rhythm}
We analyze the distributional properties of sentence lengths to characterize the rhythm of reasoning. Specifically, we flatten the generated text and segment it using punctuation, discarding short noise fragments (fewer than five characters). Based on the resulting sentence lengths, we compute Sentence Volatility, which is quantified using the coefficient of variation (CV):
\begin{equation}
    \text{Sentence Volatility (CV \%)} = \frac{\sigma_{L}}{\mu_{L}} \times 100\%
\end{equation}

Where $\sigma_{L}$ is the standard deviation of sentence character length, and $\mu_{L}$ is the average sentence length.

\section{Visualization of Critical Reasoning Steps Selected by AIR}
\label{appd:visualization_step}
We provide visualization of critical reasoning steps selected by our \ours{}. Steps highlighted in red correspond to those chosen by AIR, with color intensity reflecting the magnitude of the score. Some intermediate steps have been omitted for clarity.
\newcommand{\heatbox}[2]{%
  \begingroup%
  \setlength{\fboxsep}{0pt}
  \definecolor{tempcolor}{rgb}{#1}%
  \colorbox{tempcolor}{\strut #2}
  \endgroup%
}
\subsubsection{Case 1}
\begingroup
\tiny
\setlength{\parindent}{0pt}
\noindent
\textless{}|im\_start|\textgreater{}\hspace{0pt}system\hspace{0pt}\textless{}br\textgreater{}\hspace{0pt}%
You\hspace{0pt} are\hspace{0pt} Q\hspace{0pt}wen\hspace{0pt},\hspace{0pt} created\hspace{0pt} by\hspace{0pt}%
 Alibaba\hspace{0pt} Cloud\hspace{0pt}.\hspace{0pt} You\hspace{0pt} are\hspace{0pt} a\hspace{0pt} helpful\hspace{0pt}%
 assistant\hspace{0pt}.\hspace{0pt}\textless{}|im\_end|\textgreater{}\hspace{0pt}\textless{}br\textgreater{}\hspace{0pt}%
\textless{}|im\_start|\textgreater{}\hspace{0pt}user\hspace{0pt}\textless{}br\textgreater{}\hspace{0pt}%
e\hspace{0pt}.\hspace{0pt} Calculate\hspace{0pt} the\hspace{0pt} projected\hspace{0pt} physical\hspace{0pt}%
 separation\hspace{0pt},\hspace{0pt} \$\hspace{0pt}r\hspace{0pt}\_\{\hspace{0pt}p\hspace{0pt}\}\$\hspace{0pt}%
,\hspace{0pt} between\hspace{0pt} the\hspace{0pt} galaxy\hspace{0pt} and\hspace{0pt} the\hspace{0pt} Voor\hspace{0pt}%
werp\hspace{0pt}.g\hspace{0pt}.\hspace{0pt} High\hspace{0pt} precision\hspace{0pt} measurements\hspace{0pt}%
 showed\hspace{0pt} that\hspace{0pt} the\hspace{0pt} Voor\hspace{0pt}werp\hspace{0pt} is\hspace{0pt} slightly\hspace{0pt}%
 further\hspace{0pt} away\hspace{0pt} than\hspace{0pt} the\hspace{0pt} galaxy\hspace{0pt},\hspace{0pt}%
 and\hspace{0pt} so\hspace{0pt} \$\textbackslash{}\hspace{0pt}theta\hspace{0pt}=\hspace{0pt}1\hspace{0pt}%
2\hspace{0pt}5\hspace{0pt}\^{}\hspace{0pt}\{\textbackslash{}\hspace{0pt}circ\hspace{0pt}\}\hspace{0pt}%
\$.\hspace{0pt} Use\hspace{0pt} this\hspace{0pt} with\hspace{0pt} your\hspace{0pt} expression\hspace{0pt}%
 from\hspace{0pt} the\hspace{0pt} previous\hspace{0pt} part\hspace{0pt} of\hspace{0pt} the\hspace{0pt}%
 question\hspace{0pt} to\hspace{0pt} estimate\hspace{0pt} an\hspace{0pt} upper\hspace{0pt} limit\hspace{0pt}%
 for\hspace{0pt} the\hspace{0pt} number\hspace{0pt} of\hspace{0pt} years\hspace{0pt} that\hspace{0pt}%
 have\hspace{0pt} passed\hspace{0pt} since\hspace{0pt} the\hspace{0pt} qu\hspace{0pt}asar\hspace{0pt}%
 was\hspace{0pt} last\hspace{0pt} active\hspace{0pt}.\hspace{0pt}\textless{}|im\_end|\textgreater{}\hspace{0pt}%
\textless{}br\textgreater{}\hspace{0pt}\textless{}|im\_start|\textgreater{}\hspace{0pt}assistant\hspace{0pt}%
\textless{}br\textgreater{}\hspace{0pt}\heatbox{0.965,0.335,0.335}{\textless{}|im\_start|\textgreater{}}\hspace{0pt}%
\heatbox{0.965,0.335,0.335}{think}\hspace{0pt}\textless{}br\textgreater{}\hspace{0pt}Alright\hspace{0pt}%
,\hspace{0pt} let\hspace{0pt} me\hspace{0pt} try\hspace{0pt} to\hspace{0pt} tackle\hspace{0pt} this\hspace{0pt}%
 problem\hspace{0pt} step\hspace{0pt} by\hspace{0pt} step\hspace{0pt}.\hspace{0pt} So\hspace{0pt},\hspace{0pt}%
 we\hspace{0pt} need\hspace{0pt} to\hspace{0pt} calculate\hspace{0pt} the\hspace{0pt} projected\hspace{0pt}%
 physical\hspace{0pt} separation\hspace{0pt},\hspace{0pt} r\hspace{0pt}\_p\hspace{0pt},\hspace{0pt} between\hspace{0pt}%
 the\hspace{0pt} galaxy\hspace{0pt} and\hspace{0pt} the\hspace{0pt} Voor\hspace{0pt}werp\hspace{0pt}.\hspace{0pt}%
 Then\hspace{0pt},\hspace{0pt} given\hspace{0pt} that\hspace{0pt} Î\hspace{0pt}¸\hspace{0pt} is\hspace{0pt}%
 \hspace{0pt}1\hspace{0pt}2\hspace{0pt}5\hspace{0pt} degrees\hspace{0pt},\hspace{0pt} use\hspace{0pt}%
 that\hspace{0pt} to\hspace{0pt} estimate\hspace{0pt} an\hspace{0pt} upper\hspace{0pt} limit\hspace{0pt}%
 for\hspace{0pt} the\hspace{0pt} time\hspace{0pt} since\hspace{0pt} the\hspace{0pt} qu\hspace{0pt}asar\hspace{0pt}%
 was\hspace{0pt} last\hspace{0pt} active\hspace{0pt}.\hspace{0pt} Hmm\hspace{0pt},\hspace{0pt} okay\hspace{0pt}%
.\textless{}br\textgreater{}\textless{}br\textgreater{}\hspace{0pt}First\hspace{0pt},\hspace{0pt} let\hspace{0pt}%
 me\hspace{0pt} recall\hspace{0pt} what\hspace{0pt} the\hspace{0pt} projected\hspace{0pt} physical\hspace{0pt}%
 separation\hspace{0pt} means\hspace{0pt}.\hspace{0pt} In\hspace{0pt} astronomy\hspace{0pt},\hspace{0pt}%
 when\hspace{0pt} we\hspace{0pt} observe\hspace{0pt} objects\hspace{0pt} in\hspace{0pt} the\hspace{0pt}%
 sky\hspace{0pt},\hspace{0pt} we\hspace{0pt} often\hspace{0pt} measure\hspace{0pt} their\hspace{0pt} angular\hspace{0pt}%
 separation\hspace{0pt},\hspace{0pt} which\hspace{0pt} is\hspace{0pt} the\hspace{0pt} angle\hspace{0pt}%
 between\hspace{0pt} them\hspace{0pt} as\hspace{0pt} seen\hspace{0pt} from\hspace{0pt} Earth\hspace{0pt}%
.\hspace{0pt} But\hspace{0pt} to\hspace{0pt} find\hspace{0pt} the\hspace{0pt} actual\hspace{0pt} physical\hspace{0pt}%
 distance\hspace{0pt} between\hspace{0pt} them\hspace{0pt} (\hspace{0pt}the\hspace{0pt} projected\hspace{0pt}%
 separation\hspace{0pt}),\hspace{0pt} we\hspace{0pt} need\hspace{0pt} to\hspace{0pt} know\hspace{0pt}%
 how\hspace{0pt} far\hspace{0pt} away\hspace{0pt} they\hspace{0pt} are\hspace{0pt} from\hspace{0pt} us\hspace{0pt}%
.\hspace{0pt} The\hspace{0pt} projected\hspace{0pt} separation\hspace{0pt} can\hspace{0pt} be\hspace{0pt}%
 calculated\hspace{0pt} if\hspace{0pt} we\hspace{0pt} know\hspace{0pt} the\hspace{0pt} distance\hspace{0pt}%
 to\hspace{0pt} the\hspace{0pt} objects\hspace{0pt},\hspace{0pt} usually\hspace{0pt} using\hspace{0pt}%
 the\hspace{0pt} formula\hspace{0pt}:\textless{}br\textgreater{}\textless{}br\textgreater{}\hspace{0pt}%
\heatbox{0.969,0.420,0.420}{r}\hspace{0pt}\heatbox{0.969,0.420,0.420}{\_p}\hspace{0pt}\heatbox{0.969,0.420,0.420}{ =}\hspace{0pt}%
\heatbox{0.969,0.420,0.420}{ d}\hspace{0pt}\heatbox{0.969,0.420,0.420}{ *}\hspace{0pt}\heatbox{0.969,0.420,0.420}{ tan}\hspace{0pt}%
\heatbox{0.969,0.420,0.420}{(}\hspace{0pt}\heatbox{0.969,0.420,0.420}{Î¸}\hspace{0pt})\textless{}br\textgreater{}\textless{}br\textgreater{}\hspace{0pt}%
Where\hspace{0pt} d\hspace{0pt} is\hspace{0pt} the\hspace{0pt} distance\hspace{0pt} to\hspace{0pt} the\hspace{0pt}%
 objects\hspace{0pt} (\hspace{0pt}assuming\hspace{0pt} they\hspace{0pt} are\hspace{0pt} at\hspace{0pt}%
 the\hspace{0pt} same\hspace{0pt} distance\hspace{0pt}),\hspace{0pt} and\hspace{0pt} Î\hspace{0pt}¸\hspace{0pt}%
 is\hspace{0pt} the\hspace{0pt} angular\hspace{0pt} separation\hspace{0pt}.\hspace{0pt} But\hspace{0pt}%
 wait\hspace{0pt},\hspace{0pt} in\hspace{0pt} this\hspace{0pt} case\hspace{0pt},\hspace{0pt} the\hspace{0pt}%
 problem\hspace{0pt} states\hspace{0pt} that\hspace{0pt} high\hspace{0pt} precision\hspace{0pt} measurements\hspace{0pt}%
 showed\hspace{0pt} that\hspace{0pt} the\hspace{0pt} Voor\hspace{0pt}werp\hspace{0pt} is\hspace{0pt} slightly\hspace{0pt}%
 further\hspace{0pt} away\hspace{0pt} than\hspace{0pt} the\hspace{0pt} galaxy\hspace{0pt}.\hspace{0pt}%
 Does\hspace{0pt} this\hspace{0pt} affect\hspace{0pt} the\hspace{0pt} calculation\hspace{0pt}?\hspace{0pt}%
 Maybe\hspace{0pt} because\hspace{0pt} the\hspace{0pt} actual\hspace{0pt} separation\hspace{0pt} might\hspace{0pt}%
 involve\hspace{0pt} some\hspace{0pt} line\hspace{0pt}-of\hspace{0pt}-s\hspace{0pt}ight\hspace{0pt} component\hspace{0pt}%
.\hspace{0pt} But\hspace{0pt} since\hspace{0pt} it\hspace{0pt}'s\hspace{0pt} asking\hspace{0pt} for\hspace{0pt}%
 the\hspace{0pt} projected\hspace{0pt} physical\hspace{0pt} separation\hspace{0pt},\hspace{0pt} maybe\hspace{0pt}%
 we\hspace{0pt} can\hspace{0pt} still\hspace{0pt} use\hspace{0pt} the\hspace{0pt} angular\hspace{0pt}%
 separation\hspace{0pt} and\hspace{0pt} the\hspace{0pt} distance\hspace{0pt} to\hspace{0pt} one\hspace{0pt}%
 of\hspace{0pt} them\hspace{0pt}?\textless{}br\textgreater{}\textless{}br\textgreater{}\hspace{0pt}Wait\hspace{0pt}%
,\hspace{0pt} but\hspace{0pt} perhaps\hspace{0pt} they\hspace{0pt} are\hspace{0pt} using\hspace{0pt} the\hspace{0pt}%
 fact\hspace{0pt} that\hspace{0pt} even\hspace{0pt} though\hspace{0pt} there\hspace{0pt}'s\hspace{0pt}%
 a\hspace{0pt} slight\hspace{0pt} difference\hspace{0pt} in\hspace{0pt} distance\hspace{0pt},\hspace{0pt}%
 the\hspace{0pt} projected\hspace{0pt} separation\hspace{0pt} r\hspace{0pt}\_p\hspace{0pt} is\hspace{0pt}%
 calculated\hspace{0pt} as\hspace{0pt} the\hspace{0pt} trans\hspace{0pt}verse\hspace{0pt} component\hspace{0pt}%
 of\hspace{0pt} their\hspace{0pt} separation\hspace{0pt}.\hspace{0pt} But\hspace{0pt} I\hspace{0pt} need\hspace{0pt}%
 to\hspace{0pt} check\hspace{0pt} how\hspace{0pt} exactly\hspace{0pt} this\hspace{0pt} is\hspace{0pt}%
 defined\hspace{0pt}.\textless{}br\textgreater{}\textless{}br\textgreater{}\hspace{0pt}Alternatively\hspace{0pt}%
,\hspace{0pt} maybe\hspace{0pt} we\hspace{0pt} use\hspace{0pt} the\hspace{0pt} small\hspace{0pt} angle\hspace{0pt}%
 approximation\hspace{0pt}.\hspace{0pt} Though\hspace{0pt} since\hspace{0pt} Î\hspace{0pt}¸\hspace{0pt}%
 here\hspace{0pt} is\hspace{0pt} given\hspace{0pt} as\hspace{0pt} \hspace{0pt}1\hspace{0pt}2\hspace{0pt}%
5\hspace{0pt} degrees\hspace{0pt},\hspace{0pt} which\hspace{0pt} is\hspace{0pt} a\hspace{0pt} large\hspace{0pt}%
 angle\hspace{0pt}.\hspace{0pt} Wait\hspace{0pt},\hspace{0pt} actually\hspace{0pt} Î\hspace{0pt}¸\hspace{0pt}%
 in\hspace{0pt} part\hspace{0pt} e\hspace{0pt} might\hspace{0pt} be\hspace{0pt} the\hspace{0pt} angular\hspace{0pt}%
 separation\hspace{0pt},\hspace{0pt} but\hspace{0pt} in\hspace{0pt} part\hspace{0pt} g\hspace{0pt},\hspace{0pt}%
 the\hspace{0pt} angle\hspace{0pt} Î\hspace{0pt}¸\hspace{0pt}=\hspace{0pt}1\hspace{0pt}2\hspace{0pt}5\hspace{0pt}%
 degrees\hspace{0pt} refers\hspace{0pt} to\hspace{0pt} the\hspace{0pt} angle\hspace{0pt} between\hspace{0pt}%
 the\hspace{0pt} line\hspace{0pt} connecting\hspace{0pt} the\hspace{0pt} galaxy\hspace{0pt} and\hspace{0pt}%
 the\hspace{0pt} Voor\hspace{0pt}werp\hspace{0pt} and\hspace{0pt} our\hspace{0pt} line\hspace{0pt} of\hspace{0pt}%
 sight\hspace{0pt}?\hspace{0pt} Perhaps\hspace{0pt}.\hspace{0pt} That\hspace{0pt} might\hspace{0pt} make\hspace{0pt}%
 sense\hspace{0pt}.\hspace{0pt} So\hspace{0pt},\hspace{0pt} in\hspace{0pt} part\hspace{0pt} e\hspace{0pt}%
,\hspace{0pt} maybe\hspace{0pt} we\hspace{0pt} just\hspace{0pt} calculate\hspace{0pt} the\hspace{0pt}%
 projected\hspace{0pt} separation\hspace{0pt} using\hspace{0pt} some\hspace{0pt} angular\hspace{0pt} separation\hspace{0pt}%
 (\hspace{0pt}not\hspace{0pt} \hspace{0pt}1\hspace{0pt}2\hspace{0pt}5\hspace{0pt} degrees\hspace{0pt}%
),\hspace{0pt} but\hspace{0pt} then\hspace{0pt} in\hspace{0pt} part\hspace{0pt} g\hspace{0pt},\hspace{0pt}%
 with\hspace{0pt} Î\hspace{0pt}¸\hspace{0pt}=\hspace{0pt}1\hspace{0pt}2\hspace{0pt}5\hspace{0pt} degrees\hspace{0pt}%
,\hspace{0pt} we\hspace{0pt} use\hspace{0pt} trig\hspace{0pt}on\hspace{0pt}ometry\hspace{0pt} to\hspace{0pt}%
 get\hspace{0pt} the\hspace{0pt} actual\hspace{0pt} separation\hspace{0pt} and\hspace{0pt} then\hspace{0pt}%
 compute\hspace{0pt} the\hspace{0pt} light\hspace{0pt} travel\hspace{0pt} time\hspace{0pt}?\textless{}br\textgreater{}\textless{}br\textgreater{}\hspace{0pt}%
\heatbox{0.966,0.349,0.349}{Wait}\hspace{0pt}\heatbox{0.966,0.349,0.349}{,}\hspace{0pt}\heatbox{0.966,0.349,0.349}{ let}\hspace{0pt}%
\heatbox{0.966,0.349,0.349}{ me}\hspace{0pt}\heatbox{0.966,0.349,0.349}{ try}\hspace{0pt}\heatbox{0.966,0.349,0.349}{ to}\hspace{0pt}%
\heatbox{0.966,0.349,0.349}{ parse}\hspace{0pt}\heatbox{0.966,0.349,0.349}{ the}\hspace{0pt}\heatbox{0.966,0.349,0.349}{ problem}\hspace{0pt}%
\heatbox{0.966,0.349,0.349}{ again}\hspace{0pt}\heatbox{0.966,0.349,0.349}{.}\hspace{0pt}\heatbox{0.966,0.349,0.349}{ Part}\hspace{0pt}%
\heatbox{0.966,0.349,0.349}{ e}\hspace{0pt}\heatbox{0.966,0.349,0.349}{ says}\hspace{0pt}\heatbox{0.966,0.349,0.349}{,}\hspace{0pt}%
\heatbox{0.966,0.349,0.349}{ "}\hspace{0pt}\heatbox{0.966,0.349,0.349}{Calculate}\hspace{0pt}\heatbox{0.966,0.349,0.349}{ the}\hspace{0pt}%
\heatbox{0.966,0.349,0.349}{ projected}\hspace{0pt}\heatbox{0.966,0.349,0.349}{ physical}\hspace{0pt}%
\heatbox{0.966,0.349,0.349}{ separation}\hspace{0pt}\heatbox{0.966,0.349,0.349}{,}\hspace{0pt}\heatbox{0.966,0.349,0.349}{ r}\hspace{0pt}%
\heatbox{0.966,0.349,0.349}{\_p}\hspace{0pt}\heatbox{0.966,0.349,0.349}{,}\hspace{0pt}\heatbox{0.966,0.349,0.349}{ between}\hspace{0pt}%
\heatbox{0.966,0.349,0.349}{ the}\hspace{0pt}\heatbox{0.966,0.349,0.349}{ galaxy}\hspace{0pt}\heatbox{0.966,0.349,0.349}{ and}\hspace{0pt}%
\heatbox{0.966,0.349,0.349}{ the}\hspace{0pt}\heatbox{0.966,0.349,0.349}{ Voor}\hspace{0pt}\heatbox{0.966,0.349,0.349}{werp}\hspace{0pt}%
\heatbox{0.966,0.349,0.349}{."}\hspace{0pt}\heatbox{0.966,0.349,0.349}{ Then}\hspace{0pt}\heatbox{0.966,0.349,0.349}{ part}\hspace{0pt}%
\heatbox{0.966,0.349,0.349}{ g}\hspace{0pt}\heatbox{0.966,0.349,0.349}{ states}\hspace{0pt}\heatbox{0.966,0.349,0.349}{,}\hspace{0pt}%
\heatbox{0.966,0.349,0.349}{ "}\hspace{0pt}\heatbox{0.966,0.349,0.349}{High}\hspace{0pt}\heatbox{0.966,0.349,0.349}{ precision}\hspace{0pt}%
\heatbox{0.966,0.349,0.349}{ measurements}\hspace{0pt}\heatbox{0.966,0.349,0.349}{ showed}\hspace{0pt}%
\heatbox{0.966,0.349,0.349}{ that}\hspace{0pt}\heatbox{0.966,0.349,0.349}{ the}\hspace{0pt}\heatbox{0.966,0.349,0.349}{ Voor}\hspace{0pt}%
\heatbox{0.966,0.349,0.349}{werp}\hspace{0pt}\heatbox{0.966,0.349,0.349}{ is}\hspace{0pt}\heatbox{0.966,0.349,0.349}{ slightly}\hspace{0pt}%
\heatbox{0.966,0.349,0.349}{ further}\hspace{0pt}\heatbox{0.966,0.349,0.349}{ away}\hspace{0pt}\heatbox{0.966,0.349,0.349}{ than}\hspace{0pt}%
\heatbox{0.966,0.349,0.349}{ the}\hspace{0pt}\heatbox{0.966,0.349,0.349}{ galaxy}\hspace{0pt}\heatbox{0.966,0.349,0.349}{,}\hspace{0pt}%
\heatbox{0.966,0.349,0.349}{ and}\hspace{0pt}\heatbox{0.966,0.349,0.349}{ so}\hspace{0pt}\heatbox{0.966,0.349,0.349}{ Î}\hspace{0pt}%
\heatbox{0.966,0.349,0.349}{¸}\hspace{0pt}\heatbox{0.966,0.349,0.349}{=}\hspace{0pt}\heatbox{0.966,0.349,0.349}{1}\hspace{0pt}%
\heatbox{0.966,0.349,0.349}{2}\hspace{0pt}\heatbox{0.966,0.349,0.349}{5}\hspace{0pt}\heatbox{0.966,0.349,0.349}{Â°}\hspace{0pt}%
\heatbox{0.966,0.349,0.349}{.}\hspace{0pt}\heatbox{0.966,0.349,0.349}{ Use}\hspace{0pt}\heatbox{0.966,0.349,0.349}{ this}\hspace{0pt}%
\heatbox{0.966,0.349,0.349}{ with}\hspace{0pt}\heatbox{0.966,0.349,0.349}{ your}\hspace{0pt}\heatbox{0.966,0.349,0.349}{ expression}\hspace{0pt}%
\heatbox{0.966,0.349,0.349}{ from}\hspace{0pt}\heatbox{0.966,0.349,0.349}{ the}\hspace{0pt}\heatbox{0.966,0.349,0.349}{ previous}\hspace{0pt}%
\heatbox{0.966,0.349,0.349}{ part}\hspace{0pt}\heatbox{0.966,0.349,0.349}{ of}\hspace{0pt}\heatbox{0.966,0.349,0.349}{ the}\hspace{0pt}%
\heatbox{0.966,0.349,0.349}{ question}\hspace{0pt}\heatbox{0.966,0.349,0.349}{ to}\hspace{0pt}\heatbox{0.966,0.349,0.349}{ estimate}\hspace{0pt}%
\heatbox{0.966,0.349,0.349}{ an}\hspace{0pt}\heatbox{0.966,0.349,0.349}{ upper}\hspace{0pt}\heatbox{0.966,0.349,0.349}{ limit}\hspace{0pt}%
\heatbox{0.966,0.349,0.349}{ for}\hspace{0pt}\heatbox{0.966,0.349,0.349}{ the}\hspace{0pt}\heatbox{0.966,0.349,0.349}{ number}\hspace{0pt}%
\heatbox{0.966,0.349,0.349}{ of}\hspace{0pt}\heatbox{0.966,0.349,0.349}{ years}\hspace{0pt}\heatbox{0.966,0.349,0.349}{ that}\hspace{0pt}%
\heatbox{0.966,0.349,0.349}{ have}\hspace{0pt}\heatbox{0.966,0.349,0.349}{ passed}\hspace{0pt}\heatbox{0.966,0.349,0.349}{ since}\hspace{0pt}%
\heatbox{0.966,0.349,0.349}{ the}\hspace{0pt}\heatbox{0.966,0.349,0.349}{ qu}\hspace{0pt}\heatbox{0.966,0.349,0.349}{asar}\hspace{0pt}%
\heatbox{0.966,0.349,0.349}{ was}\hspace{0pt}\heatbox{0.966,0.349,0.349}{ last}\hspace{0pt}\heatbox{0.966,0.349,0.349}{ active}\hspace{0pt}%
."\textless{}br\textgreater{}\textless{}br\textgreater{}\hspace{0pt}Hmm\hspace{0pt},\hspace{0pt} so\hspace{0pt}%
 maybe\hspace{0pt} in\hspace{0pt} part\hspace{0pt} e\hspace{0pt},\hspace{0pt} they\hspace{0pt} expect\hspace{0pt}%
 an\hspace{0pt} expression\hspace{0pt} for\hspace{0pt} r\hspace{0pt}\_p\hspace{0pt} in\hspace{0pt} terms\hspace{0pt}%
 of\hspace{0pt} distance\hspace{0pt} or\hspace{0pt} something\hspace{0pt},\hspace{0pt} and\hspace{0pt}%
 part\hspace{0pt} g\hspace{0pt} uses\hspace{0pt} the\hspace{0pt} angle\hspace{0pt} Î\hspace{0pt}¸\hspace{0pt}%
=\hspace{0pt}1\hspace{0pt}2\hspace{0pt}5\hspace{0pt} degrees\hspace{0pt} to\hspace{0pt} get\hspace{0pt}%
 the\hspace{0pt} three\hspace{0pt}-dimensional\hspace{0pt} separation\hspace{0pt} from\hspace{0pt} the\hspace{0pt}%
 projected\hspace{0pt} separation\hspace{0pt}.\hspace{0pt} So\hspace{0pt} the\hspace{0pt} angle\hspace{0pt}%
 Î\hspace{0pt}¸\hspace{0pt}=\hspace{0pt}1\hspace{0pt}2\hspace{0pt}5\hspace{0pt} degrees\hspace{0pt} is\hspace{0pt}%
 the\hspace{0pt} angle\hspace{0pt} between\hspace{0pt} the\hspace{0pt} line\hspace{0pt} connecting\hspace{0pt}%
 the\hspace{0pt} galaxy\hspace{0pt} and\hspace{0pt} Voor\hspace{0pt}werp\hspace{0pt} and\hspace{0pt} our\hspace{0pt}%
 line\hspace{0pt} of\hspace{0pt} sight\hspace{0pt},\hspace{0pt} such\hspace{0pt} that\hspace{0pt} the\hspace{0pt}%
 actual\hspace{0pt} separation\hspace{0pt} is\hspace{0pt} r\hspace{0pt} =\hspace{0pt} r\hspace{0pt}\_p\hspace{0pt}%
 /\hspace{0pt} sin\hspace{0pt}(\hspace{0pt}Î¸\hspace{0pt}),\hspace{0pt} assuming\hspace{0pt} Î\hspace{0pt}%
¸\hspace{0pt} is\hspace{0pt} measured\hspace{0pt} from\hspace{0pt} the\hspace{0pt} line\hspace{0pt} of\hspace{0pt}%
 sight\hspace{0pt}.\hspace{0pt} Wait\hspace{0pt},\hspace{0pt} or\hspace{0pt} maybe\hspace{0pt} it\hspace{0pt}%
's\hspace{0pt} cosine\hspace{0pt}?\hspace{0pt} 

...(Omit)...
\hspace{0pt} Wait\hspace{0pt},\hspace{0pt} perhaps\hspace{0pt}%
 in\hspace{0pt} part\hspace{0pt} e\hspace{0pt},\hspace{0pt} when\hspace{0pt} calculating\hspace{0pt} the\hspace{0pt}%
 projected\hspace{0pt} separation\hspace{0pt} r\hspace{0pt}\_p\hspace{0pt},\hspace{0pt} they\hspace{0pt}%
 are\hspace{0pt} considering\hspace{0pt} some\hspace{0pt} other\hspace{0pt} angular\hspace{0pt} separation\hspace{0pt}%
 (\hspace{0pt}like\hspace{0pt} on\hspace{0pt} the\hspace{0pt} sky\hspace{0pt}),\hspace{0pt} so\hspace{0pt}%
 when\hspace{0pt} we\hspace{0pt} compute\hspace{0pt} the\hspace{0pt} physical\hspace{0pt} separation\hspace{0pt}%
,\hspace{0pt} r\hspace{0pt}\_p\hspace{0pt} is\hspace{0pt} already\hspace{0pt} in\hspace{0pt} parse\hspace{0pt}%
cs\hspace{0pt} or\hspace{0pt} light\hspace{0pt}-years\hspace{0pt}.\textless{}br\textgreater{}\textless{}br\textgreater{}\hspace{0pt}%
\heatbox{0.966,0.349,0.349}{Wait}\hspace{0pt}\heatbox{0.966,0.349,0.349}{,}\hspace{0pt}\heatbox{0.966,0.349,0.349}{ maybe}\hspace{0pt}%
\heatbox{0.966,0.349,0.349}{ I}\hspace{0pt}\heatbox{0.966,0.349,0.349}{ should}\hspace{0pt}\heatbox{0.966,0.349,0.349}{ look}\hspace{0pt}%
\heatbox{0.966,0.349,0.349}{ at}\hspace{0pt}\heatbox{0.966,0.349,0.349}{ part}\hspace{0pt}\heatbox{0.966,0.349,0.349}{ e}\hspace{0pt}%
\heatbox{0.966,0.349,0.349}{ first}\hspace{0pt}\heatbox{0.966,0.349,0.349}{.}\hspace{0pt}\heatbox{0.966,0.349,0.349}{ Because}\hspace{0pt}%
\heatbox{0.966,0.349,0.349}{ part}\hspace{0pt}\heatbox{0.966,0.349,0.349}{ g}\hspace{0pt}\heatbox{0.966,0.349,0.349}{ depends}\hspace{0pt}%
\heatbox{0.966,0.349,0.349}{ on}\hspace{0pt}\heatbox{0.966,0.349,0.349}{ part}\hspace{0pt}\heatbox{0.966,0.349,0.349}{ e}\hspace{0pt}%
\heatbox{0.966,0.349,0.349}{.}\hspace{0pt}\heatbox{0.966,0.349,0.349}{ So}\hspace{0pt}\heatbox{0.966,0.349,0.349}{ part}\hspace{0pt}%
\heatbox{0.966,0.349,0.349}{ e}\hspace{0pt}\heatbox{0.966,0.349,0.349}{ is}\hspace{0pt}\heatbox{0.966,0.349,0.349}{ to}\hspace{0pt}%
\heatbox{0.966,0.349,0.349}{ "}\hspace{0pt}\heatbox{0.966,0.349,0.349}{calculate}\hspace{0pt}\heatbox{0.966,0.349,0.349}{ the}\hspace{0pt}%
\heatbox{0.966,0.349,0.349}{ projected}\hspace{0pt}\heatbox{0.966,0.349,0.349}{ physical}\hspace{0pt}%
\heatbox{0.966,0.349,0.349}{ separation}\hspace{0pt}\heatbox{0.966,0.349,0.349}{,}\hspace{0pt}\heatbox{0.966,0.349,0.349}{ r}\hspace{0pt}%
\heatbox{0.966,0.349,0.349}{\_p}\hspace{0pt}\heatbox{0.966,0.349,0.349}{,}\hspace{0pt}\heatbox{0.966,0.349,0.349}{ between}\hspace{0pt}%
\heatbox{0.966,0.349,0.349}{ the}\hspace{0pt}\heatbox{0.966,0.349,0.349}{ galaxy}\hspace{0pt}\heatbox{0.966,0.349,0.349}{ and}\hspace{0pt}%
\heatbox{0.966,0.349,0.349}{ the}\hspace{0pt}\heatbox{0.966,0.349,0.349}{ Voor}\hspace{0pt}\heatbox{0.966,0.349,0.349}{werp}\hspace{0pt}%
\heatbox{0.966,0.349,0.349}{."}\hspace{0pt}\heatbox{0.966,0.349,0.349}{ But}\hspace{0pt}\heatbox{0.966,0.349,0.349}{ the}\hspace{0pt}%
\heatbox{0.966,0.349,0.349}{ problem}\hspace{0pt}\heatbox{0.966,0.349,0.349}{ probably}\hspace{0pt}\heatbox{0.966,0.349,0.349}{ gives}\hspace{0pt}%
\heatbox{0.966,0.349,0.349}{ some}\hspace{0pt}\heatbox{0.966,0.349,0.349}{ data}\hspace{0pt}\heatbox{0.966,0.349,0.349}{ in}\hspace{0pt}%
\heatbox{0.966,0.349,0.349}{ previous}\hspace{0pt}\heatbox{0.966,0.349,0.349}{ parts}\hspace{0pt}\heatbox{0.966,0.349,0.349}{,}\hspace{0pt}%
\heatbox{0.966,0.349,0.349}{ which}\hspace{0pt}\heatbox{0.966,0.349,0.349}{ aren}\hspace{0pt}\heatbox{0.966,0.349,0.349}{'t}\hspace{0pt}%
\heatbox{0.966,0.349,0.349}{ here}\hspace{0pt}\heatbox{0.966,0.349,0.349}{.}\hspace{0pt}\heatbox{0.966,0.349,0.349}{ Since}\hspace{0pt}%
\heatbox{0.966,0.349,0.349}{ this}\hspace{0pt}\heatbox{0.966,0.349,0.349}{ is}\hspace{0pt}\heatbox{0.966,0.349,0.349}{ part}\hspace{0pt}%
\heatbox{0.966,0.349,0.349}{ of}\hspace{0pt}\heatbox{0.966,0.349,0.349}{ a}\hspace{0pt}\heatbox{0.966,0.349,0.349}{ series}\hspace{0pt}%
\heatbox{0.966,0.349,0.349}{ of}\hspace{0pt}\heatbox{0.966,0.349,0.349}{ questions}\hspace{0pt}\heatbox{0.966,0.349,0.349}{,}\hspace{0pt}%
\heatbox{0.966,0.349,0.349}{ starting}\hspace{0pt}\heatbox{0.966,0.349,0.349}{ with}\hspace{0pt}\heatbox{0.966,0.349,0.349}{ part}\hspace{0pt}%
\heatbox{0.966,0.349,0.349}{ e}\hspace{0pt}\heatbox{0.966,0.349,0.349}{,}\hspace{0pt}\heatbox{0.966,0.349,0.349}{ then}\hspace{0pt}%
\heatbox{0.966,0.349,0.349}{ part}\hspace{0pt}\heatbox{0.966,0.349,0.349}{ g}\hspace{0pt}.\textless{}br\textgreater{}\textless{}br\textgreater{}\hspace{0pt}%
However\hspace{0pt},\hspace{0pt} the\hspace{0pt} user\hspace{0pt} hasn\hspace{0pt}'t\hspace{0pt} provided\hspace{0pt}%
 the\hspace{0pt} previous\hspace{0pt} parts\hspace{0pt} a\hspace{0pt} through\hspace{0pt} d\hspace{0pt}%
,\hspace{0pt} so\hspace{0pt} I\hspace{0pt} need\hspace{0pt} to\hspace{0pt} make\hspace{0pt} assumptions\hspace{0pt}%
.\hspace{0pt} But\hspace{0pt} perhaps\hspace{0pt} part\hspace{0pt} e\hspace{0pt} can\hspace{0pt} be\hspace{0pt}%
 answered\hspace{0pt} using\hspace{0pt} common\hspace{0pt} knowledge\hspace{0pt} or\hspace{0pt} formulas\hspace{0pt}%
.\textless{}br\textgreater{}\textless{}br\textgreater{}\hspace{0pt}Let\hspace{0pt} me\hspace{0pt} check\hspace{0pt}%
.\hspace{0pt} To\hspace{0pt} calculate\hspace{0pt} the\hspace{0pt} projected\hspace{0pt} physical\hspace{0pt}%
 separation\hspace{0pt},\hspace{0pt} you\hspace{0pt} need\hspace{0pt} the\hspace{0pt} angular\hspace{0pt}%
 separation\hspace{0pt} in\hspace{0pt} the\hspace{0pt} sky\hspace{0pt} between\hspace{0pt} the\hspace{0pt}%
 galaxy\hspace{0pt} and\hspace{0pt} the\hspace{0pt} Voor\hspace{0pt}werp\hspace{0pt},\hspace{0pt} and\hspace{0pt}%
 the\hspace{0pt} distance\hspace{0pt} to\hspace{0pt} the\hspace{0pt} system\hspace{0pt}.\hspace{0pt} The\hspace{0pt}%
 formula\hspace{0pt} is\hspace{0pt} r\hspace{0pt}\_p\hspace{0pt} =\hspace{0pt} d\hspace{0pt} *\hspace{0pt}%
 tan\hspace{0pt}(\hspace{0pt}Î´\hspace{0pt}Î¸\hspace{0pt}),\hspace{0pt} where\hspace{0pt} Î´\hspace{0pt}%
Î¸\hspace{0pt} is\hspace{0pt} the\hspace{0pt} angular\hspace{0pt} separation\hspace{0pt}.\hspace{0pt}%
 But\hspace{0pt} if\hspace{0pt} Î´\hspace{0pt}Î¸\hspace{0pt} is\hspace{0pt} small\hspace{0pt},\hspace{0pt}%
 you\hspace{0pt} can\hspace{0pt} approximate\hspace{0pt} tan\hspace{0pt}(\hspace{0pt}Î´\hspace{0pt}Î¸\hspace{0pt}%
)\hspace{0pt} âī\hspace{0pt}Ī\hspace{0pt} Î´\hspace{0pt}Î¸\hspace{0pt} in\hspace{0pt} radians\hspace{0pt}%
.\hspace{0pt} However\hspace{0pt},\hspace{0pt} without\hspace{0pt} knowing\hspace{0pt} the\hspace{0pt}%
 actual\hspace{0pt} angular\hspace{0pt} separation\hspace{0pt} or\hspace{0pt} distance\hspace{0pt},\hspace{0pt}%
 maybe\hspace{0pt} some\hspace{0pt} other\hspace{0pt} information\hspace{0pt} was\hspace{0pt} given\hspace{0pt}%
 in\hspace{0pt} previous\hspace{0pt} parts\hspace{0pt}.\hspace{0pt} Since\hspace{0pt} the\hspace{0pt}%
 user\hspace{0pt} hasn\hspace{0pt}'t\hspace{0pt} provided\hspace{0pt} that\hspace{0pt},\hspace{0pt} I\hspace{0pt}%
 need\hspace{0pt} to\hspace{0pt} assume\hspace{0pt}.\hspace{0pt} Alternatively\hspace{0pt},\hspace{0pt}%
 maybe\hspace{0pt} in\hspace{0pt} part\hspace{0pt} g\hspace{0pt},\hspace{0pt} they\hspace{0pt} just\hspace{0pt}%
 need\hspace{0pt} to\hspace{0pt} relate\hspace{0pt} t\hspace{0pt} =\hspace{0pt} r\hspace{0pt}\_p\hspace{0pt}%
 /\hspace{0pt} (\hspace{0pt}c\hspace{0pt} *\hspace{0pt} sin\hspace{0pt} Î\hspace{0pt}¸\hspace{0pt}),\hspace{0pt}%
 where\hspace{0pt} Î\hspace{0pt}¸\hspace{0pt} is\hspace{0pt} \hspace{0pt}1\hspace{0pt}2\hspace{0pt}5\hspace{0pt}%
 degrees\hspace{0pt}.\hspace{0pt} But\hspace{0pt} for\hspace{0pt} r\hspace{0pt}\_p\hspace{0pt},\hspace{0pt}%
 perhaps\hspace{0pt} it\hspace{0pt} was\hspace{0pt} calculated\hspace{0pt} in\hspace{0pt} part\hspace{0pt}%
 e\hspace{0pt}.\textless{}br\textgreater{}\textless{}br\textgreater{}\hspace{0pt}But\hspace{0pt} since\hspace{0pt}%
 the\hspace{0pt} user\hspace{0pt} hasn\hspace{0pt}'t\hspace{0pt} provided\hspace{0pt} part\hspace{0pt}%
 e\hspace{0pt}'s\hspace{0pt} specifics\hspace{0pt},\hspace{0pt} it\hspace{0pt}'s\hspace{0pt} challenging\hspace{0pt}%
.\hspace{0pt} Maybe\hspace{0pt} in\hspace{0pt} part\hspace{0pt} e\hspace{0pt},\hspace{0pt} if\hspace{0pt}%
 you\hspace{0pt} use\hspace{0pt} H\hspace{0pt}ubble\hspace{0pt}'s\hspace{0pt} Law\hspace{0pt},\hspace{0pt}%
 since\hspace{0pt} galaxies\hspace{0pt}'\hspace{0pt} distances\hspace{0pt} are\hspace{0pt} often\hspace{0pt}%
 estimated\hspace{0pt} using\hspace{0pt} red\hspace{0pt}shift\hspace{0pt}.\hspace{0pt} Suppose\hspace{0pt}%
 the\hspace{0pt} galaxy\hspace{0pt} has\hspace{0pt} a\hspace{0pt} red\hspace{0pt}shift\hspace{0pt} z\hspace{0pt}%
,\hspace{0pt} then\hspace{0pt} the\hspace{0pt} distance\hspace{0pt} d\hspace{0pt} is\hspace{0pt} cz\hspace{0pt}%
 /\hspace{0pt} H\hspace{0pt}0\hspace{0pt}.\hspace{0pt} But\hspace{0pt} without\hspace{0pt} specific\hspace{0pt}%
 numbers\hspace{0pt},\hspace{0pt} not\hspace{0pt} sure\hspace{0pt}.\hspace{0pt} Alternatively\hspace{0pt}%
,\hspace{0pt} maybe\hspace{0pt} the\hspace{0pt} Voor\hspace{0pt}werp\hspace{0pt} is\hspace{0pt} at\hspace{0pt}%
 the\hspace{0pt} same\hspace{0pt} distance\hspace{0pt} as\hspace{0pt} the\hspace{0pt} galaxy\hspace{0pt}%
,\hspace{0pt} so\hspace{0pt} the\hspace{0pt} projected\hspace{0pt} separation\hspace{0pt} is\hspace{0pt}%
 just\hspace{0pt} d\hspace{0pt} *\hspace{0pt} Î\hspace{0pt}¸\hspace{0pt} (\hspace{0pt}in\hspace{0pt} radians\hspace{0pt}%
).\hspace{0pt} Wait\hspace{0pt},\hspace{0pt} but\hspace{0pt} the\hspace{0pt} given\hspace{0pt} angle\hspace{0pt}%
 in\hspace{0pt} part\hspace{0pt} g\hspace{0pt} is\hspace{0pt} \hspace{0pt}1\hspace{0pt}2\hspace{0pt}5\hspace{0pt}%
 degrees\hspace{0pt},\hspace{0pt} which\hspace{0pt} might\hspace{0pt} not\hspace{0pt} be\hspace{0pt} the\hspace{0pt}%
 angular\hspace{0pt} separation\hspace{0pt},\hspace{0pt} but\hspace{0pt} the\hspace{0pt} angle\hspace{0pt}%
 of\hspace{0pt} the\hspace{0pt} separation\hspace{0pt} vector\hspace{0pt} with\hspace{0pt} respect\hspace{0pt}%
 to\hspace{0pt} the\hspace{0pt} line\hspace{0pt} of\hspace{0pt} sight\hspace{0pt}.\textless{}br\textgreater{}\textless{}br\textgreater{}\hspace{0pt}%
\heatbox{0.974,0.511,0.511}{Given}\hspace{0pt}\heatbox{0.974,0.511,0.511}{ the}\hspace{0pt}\heatbox{0.974,0.511,0.511}{ confusion}\hspace{0pt}%
\heatbox{0.974,0.511,0.511}{,}\hspace{0pt}\heatbox{0.974,0.511,0.511}{ perhaps}\hspace{0pt}\heatbox{0.974,0.511,0.511}{ best}\hspace{0pt}%
\heatbox{0.974,0.511,0.511}{ to}\hspace{0pt}\heatbox{0.974,0.511,0.511}{ consider}\hspace{0pt}\heatbox{0.974,0.511,0.511}{ hypothetical}\hspace{0pt}%
\heatbox{0.974,0.511,0.511}{ values}\hspace{0pt}\heatbox{0.974,0.511,0.511}{.}\hspace{0pt}\heatbox{0.974,0.511,0.511}{ Let}\hspace{0pt}%
\heatbox{0.974,0.511,0.511}{ me}\hspace{0pt}\heatbox{0.974,0.511,0.511}{ imagine}\hspace{0pt}\heatbox{0.974,0.511,0.511}{ that}\hspace{0pt}%
\heatbox{0.974,0.511,0.511}{ in}\hspace{0pt}\heatbox{0.974,0.511,0.511}{ part}\hspace{0pt}\heatbox{0.974,0.511,0.511}{ e}\hspace{0pt}%
\heatbox{0.974,0.511,0.511}{,}\hspace{0pt}\heatbox{0.974,0.511,0.511}{ the}\hspace{0pt}\heatbox{0.974,0.511,0.511}{ projected}\hspace{0pt}%
\heatbox{0.974,0.511,0.511}{ separation}\hspace{0pt}\heatbox{0.974,0.511,0.511}{ r}\hspace{0pt}\heatbox{0.974,0.511,0.511}{\_p}\hspace{0pt}%
\heatbox{0.974,0.511,0.511}{ is}\hspace{0pt}\heatbox{0.974,0.511,0.511}{ calculated}\hspace{0pt}\heatbox{0.974,0.511,0.511}{ using}\hspace{0pt}%
\heatbox{0.974,0.511,0.511}{ some}\hspace{0pt}\heatbox{0.974,0.511,0.511}{ angular}\hspace{0pt}\heatbox{0.974,0.511,0.511}{ separation}\hspace{0pt}%
\heatbox{0.974,0.511,0.511}{ and}\hspace{0pt}\heatbox{0.974,0.511,0.511}{ distance}\hspace{0pt}\heatbox{0.974,0.511,0.511}{.}\hspace{0pt}%
\heatbox{0.974,0.511,0.511}{ For}\hspace{0pt}\heatbox{0.974,0.511,0.511}{ example}\hspace{0pt}\heatbox{0.974,0.511,0.511}{,}\hspace{0pt}%
\heatbox{0.974,0.511,0.511}{ if}\hspace{0pt}\heatbox{0.974,0.511,0.511}{ the}\hspace{0pt}\heatbox{0.974,0.511,0.511}{ angular}\hspace{0pt}%
\heatbox{0.974,0.511,0.511}{ separation}\hspace{0pt}\heatbox{0.974,0.511,0.511}{ is}\hspace{0pt}\heatbox{0.974,0.511,0.511}{ Î´}\hspace{0pt}%
\heatbox{0.974,0.511,0.511}{Î¸}\hspace{0pt}\heatbox{0.974,0.511,0.511}{,}\hspace{0pt}\heatbox{0.974,0.511,0.511}{ and}\hspace{0pt}%
\heatbox{0.974,0.511,0.511}{ the}\hspace{0pt}\heatbox{0.974,0.511,0.511}{ distance}\hspace{0pt}\heatbox{0.974,0.511,0.511}{ to}\hspace{0pt}%
\heatbox{0.974,0.511,0.511}{ the}\hspace{0pt}\heatbox{0.974,0.511,0.511}{ galaxy}\hspace{0pt}\heatbox{0.974,0.511,0.511}{ is}\hspace{0pt}%
\heatbox{0.974,0.511,0.511}{ d}\hspace{0pt}\heatbox{0.974,0.511,0.511}{,}\hspace{0pt}\heatbox{0.974,0.511,0.511}{ then}\hspace{0pt}%
\heatbox{0.974,0.511,0.511}{ r}\hspace{0pt}\heatbox{0.974,0.511,0.511}{\_p}\hspace{0pt}\heatbox{0.974,0.511,0.511}{ âī}\hspace{0pt}%
\heatbox{0.974,0.511,0.511}{Ī}\hspace{0pt}\heatbox{0.974,0.511,0.511}{ d}\hspace{0pt}\heatbox{0.974,0.511,0.511}{ *}\hspace{0pt}%
\heatbox{0.974,0.511,0.511}{ Î´}\hspace{0pt}\heatbox{0.974,0.511,0.511}{Î¸}\hspace{0pt}\heatbox{0.974,0.511,0.511}{,}\hspace{0pt}%
\heatbox{0.974,0.511,0.511}{ where}\hspace{0pt}\heatbox{0.974,0.511,0.511}{ Î´}\hspace{0pt}\heatbox{0.974,0.511,0.511}{Î¸}\hspace{0pt}%
\heatbox{0.974,0.511,0.511}{ is}\hspace{0pt}\heatbox{0.974,0.511,0.511}{ in}\hspace{0pt}\heatbox{0.974,0.511,0.511}{ radians}\hspace{0pt}%
\heatbox{0.974,0.511,0.511}{.}\hspace{0pt}\heatbox{0.974,0.511,0.511}{ Then}\hspace{0pt}\heatbox{0.974,0.511,0.511}{,}\hspace{0pt}%
\heatbox{0.974,0.511,0.511}{ in}\hspace{0pt}\heatbox{0.974,0.511,0.511}{ part}\hspace{0pt}\heatbox{0.974,0.511,0.511}{ g}\hspace{0pt}%
\heatbox{0.974,0.511,0.511}{,}\hspace{0pt}\heatbox{0.974,0.511,0.511}{ given}\hspace{0pt}\heatbox{0.974,0.511,0.511}{ Î}\hspace{0pt}%
\heatbox{0.974,0.511,0.511}{¸}\hspace{0pt}\heatbox{0.974,0.511,0.511}{ =}\hspace{0pt}\heatbox{0.974,0.511,0.511}{ }\hspace{0pt}%
\heatbox{0.974,0.511,0.511}{1}\hspace{0pt}\heatbox{0.974,0.511,0.511}{2}\hspace{0pt}\heatbox{0.974,0.511,0.511}{5}\hspace{0pt}%
\heatbox{0.974,0.511,0.511}{Â°}\hspace{0pt}\heatbox{0.974,0.511,0.511}{,}\hspace{0pt}\heatbox{0.974,0.511,0.511}{ the}\hspace{0pt}%
\heatbox{0.974,0.511,0.511}{ angle}\hspace{0pt}\heatbox{0.974,0.511,0.511}{ between}\hspace{0pt}\heatbox{0.974,0.511,0.511}{ the}\hspace{0pt}%
\heatbox{0.974,0.511,0.511}{ separation}\hspace{0pt}\heatbox{0.974,0.511,0.511}{ vector}\hspace{0pt}\heatbox{0.974,0.511,0.511}{ and}\hspace{0pt}%
\heatbox{0.974,0.511,0.511}{ the}\hspace{0pt}\heatbox{0.974,0.511,0.511}{ line}\hspace{0pt}\heatbox{0.974,0.511,0.511}{ of}\hspace{0pt}%
\heatbox{0.974,0.511,0.511}{ sight}\hspace{0pt}\heatbox{0.974,0.511,0.511}{,}\hspace{0pt}\heatbox{0.974,0.511,0.511}{ then}\hspace{0pt}%
\heatbox{0.974,0.511,0.511}{ the}\hspace{0pt}\heatbox{0.974,0.511,0.511}{ actual}\hspace{0pt}\heatbox{0.974,0.511,0.511}{ separation}\hspace{0pt}%
\heatbox{0.974,0.511,0.511}{ is}\hspace{0pt}\heatbox{0.974,0.511,0.511}{ r}\hspace{0pt}\heatbox{0.974,0.511,0.511}{ =}\hspace{0pt}%
\heatbox{0.974,0.511,0.511}{ r}\hspace{0pt}\heatbox{0.974,0.511,0.511}{\_p}\hspace{0pt}\heatbox{0.974,0.511,0.511}{ /}\hspace{0pt}%
\heatbox{0.974,0.511,0.511}{ sin}\hspace{0pt}\heatbox{0.974,0.511,0.511}{Î¸}\hspace{0pt}\heatbox{0.974,0.511,0.511}{.}\hspace{0pt}%
\heatbox{0.974,0.511,0.511}{ Therefore}\hspace{0pt}\heatbox{0.974,0.511,0.511}{,}\hspace{0pt}\heatbox{0.974,0.511,0.511}{ the}\hspace{0pt}%
\heatbox{0.974,0.511,0.511}{ light}\hspace{0pt}\heatbox{0.974,0.511,0.511}{ travel}\hspace{0pt}\heatbox{0.974,0.511,0.511}{ time}\hspace{0pt}%
\heatbox{0.974,0.511,0.511}{ would}\hspace{0pt}\heatbox{0.974,0.511,0.511}{ be}\hspace{0pt}\heatbox{0.974,0.511,0.511}{ t}\hspace{0pt}%
\heatbox{0.974,0.511,0.511}{ =}\hspace{0pt}\heatbox{0.974,0.511,0.511}{ r}\hspace{0pt}\heatbox{0.974,0.511,0.511}{ /}\hspace{0pt}%
\heatbox{0.974,0.511,0.511}{ c}\hspace{0pt}\heatbox{0.974,0.511,0.511}{ =}\hspace{0pt}\heatbox{0.974,0.511,0.511}{ (}\hspace{0pt}%
\heatbox{0.974,0.511,0.511}{r}\hspace{0pt}\heatbox{0.974,0.511,0.511}{\_p}\hspace{0pt}\heatbox{0.974,0.511,0.511}{ /}\hspace{0pt}%
\heatbox{0.974,0.511,0.511}{ sin}\hspace{0pt}\heatbox{0.974,0.511,0.511}{Î¸}\hspace{0pt}\heatbox{0.974,0.511,0.511}{)}\hspace{0pt}%
\heatbox{0.974,0.511,0.511}{ /}\hspace{0pt}\heatbox{0.974,0.511,0.511}{ c}\hspace{0pt}\heatbox{0.974,0.511,0.511}{.}\hspace{0pt}%
\heatbox{0.974,0.511,0.511}{ So}\hspace{0pt}\heatbox{0.974,0.511,0.511}{ the}\hspace{0pt}\heatbox{0.974,0.511,0.511}{ upper}\hspace{0pt}%
\heatbox{0.974,0.511,0.511}{ limit}\hspace{0pt}\heatbox{0.974,0.511,0.511}{ on}\hspace{0pt}\heatbox{0.974,0.511,0.511}{ the}\hspace{0pt}%
\heatbox{0.974,0.511,0.511}{ time}\hspace{0pt}\heatbox{0.974,0.511,0.511}{ would}\hspace{0pt}\heatbox{0.974,0.511,0.511}{ be}\hspace{0pt}%
\heatbox{0.974,0.511,0.511}{ if}\hspace{0pt}\heatbox{0.974,0.511,0.511}{ the}\hspace{0pt}\heatbox{0.974,0.511,0.511}{ separation}\hspace{0pt}%
\heatbox{0.974,0.511,0.511}{ is}\hspace{0pt}\heatbox{0.974,0.511,0.511}{ maxim}\hspace{0pt}\heatbox{0.974,0.511,0.511}{ized}\hspace{0pt}%
\heatbox{0.974,0.511,0.511}{,}\hspace{0pt}\heatbox{0.974,0.511,0.511}{ which}\hspace{0pt}\heatbox{0.974,0.511,0.511}{ would}\hspace{0pt}%
\heatbox{0.974,0.511,0.511}{ correspond}\hspace{0pt}\heatbox{0.974,0.511,0.511}{ to}\hspace{0pt}\heatbox{0.974,0.511,0.511}{ sin}\hspace{0pt}%
\heatbox{0.974,0.511,0.511}{Î¸}\hspace{0pt}\heatbox{0.974,0.511,0.511}{ being}\hspace{0pt}\heatbox{0.974,0.511,0.511}{ minimized}\hspace{0pt}%
\heatbox{0.974,0.511,0.511}{.}\hspace{0pt}\heatbox{0.974,0.511,0.511}{ However}\hspace{0pt}\heatbox{0.974,0.511,0.511}{,}\hspace{0pt}%
\heatbox{0.974,0.511,0.511}{ given}\hspace{0pt}\heatbox{0.974,0.511,0.511}{ that}\hspace{0pt}\heatbox{0.974,0.511,0.511}{ Î}\hspace{0pt}%
\heatbox{0.974,0.511,0.511}{¸}\hspace{0pt}\heatbox{0.974,0.511,0.511}{ is}\hspace{0pt}\heatbox{0.974,0.511,0.511}{ fixed}\hspace{0pt}%
\heatbox{0.974,0.511,0.511}{ at}\hspace{0pt}\heatbox{0.974,0.511,0.511}{ }\hspace{0pt}\heatbox{0.974,0.511,0.511}{1}\hspace{0pt}%
\heatbox{0.974,0.511,0.511}{2}\hspace{0pt}\heatbox{0.974,0.511,0.511}{5}\hspace{0pt}\heatbox{0.974,0.511,0.511}{Â°}\hspace{0pt}%
\heatbox{0.974,0.511,0.511}{,}\hspace{0pt}\heatbox{0.974,0.511,0.511}{ sin}\hspace{0pt}\heatbox{0.974,0.511,0.511}{Î¸}\hspace{0pt}%
\heatbox{0.974,0.511,0.511}{ is}\hspace{0pt}\heatbox{0.974,0.511,0.511}{ a}\hspace{0pt}\heatbox{0.974,0.511,0.511}{ known}\hspace{0pt}%
\heatbox{0.974,0.511,0.511}{ value}\hspace{0pt}\heatbox{0.974,0.511,0.511}{.}\hspace{0pt}\heatbox{0.974,0.511,0.511}{ Therefore}\hspace{0pt}%
\heatbox{0.974,0.511,0.511}{,}\hspace{0pt}\heatbox{0.974,0.511,0.511}{ the}\hspace{0pt}\heatbox{0.974,0.511,0.511}{ upper}\hspace{0pt}%
\heatbox{0.974,0.511,0.511}{ limit}\hspace{0pt}\heatbox{0.974,0.511,0.511}{ comes}\hspace{0pt}\heatbox{0.974,0.511,0.511}{ from}\hspace{0pt}%
\heatbox{0.974,0.511,0.511}{ using}\hspace{0pt}\heatbox{0.974,0.511,0.511}{ Î}\hspace{0pt}\heatbox{0.974,0.511,0.511}{¸}\hspace{0pt}%
\heatbox{0.974,0.511,0.511}{=}\hspace{0pt}\heatbox{0.974,0.511,0.511}{1}\hspace{0pt}\heatbox{0.974,0.511,0.511}{2}\hspace{0pt}%
\heatbox{0.974,0.511,0.511}{5}\hspace{0pt}\heatbox{0.974,0.511,0.511}{Â°}\hspace{0pt}\heatbox{0.974,0.511,0.511}{,}\hspace{0pt}%
\heatbox{0.974,0.511,0.511}{ which}\hspace{0pt}\heatbox{0.974,0.511,0.511}{ gives}\hspace{0pt}\heatbox{0.974,0.511,0.511}{ a}\hspace{0pt}%
\heatbox{0.974,0.511,0.511}{ certain}\hspace{0pt}\heatbox{0.974,0.511,0.511}{ actual}\hspace{0pt}\heatbox{0.974,0.511,0.511}{ separation}\hspace{0pt}%
\heatbox{0.974,0.511,0.511}{,}\hspace{0pt}\heatbox{0.974,0.511,0.511}{ which}\hspace{0pt}\heatbox{0.974,0.511,0.511}{ when}\hspace{0pt}%
\heatbox{0.974,0.511,0.511}{ divided}\hspace{0pt}\heatbox{0.974,0.511,0.511}{ by}\hspace{0pt}\heatbox{0.974,0.511,0.511}{ the}\hspace{0pt}%
\heatbox{0.974,0.511,0.511}{ speed}\hspace{0pt}\heatbox{0.974,0.511,0.511}{ of}\hspace{0pt}\heatbox{0.974,0.511,0.511}{ light}\hspace{0pt}%
\heatbox{0.974,0.511,0.511}{ gives}\hspace{0pt}\heatbox{0.974,0.511,0.511}{ the}\hspace{0pt}\heatbox{0.974,0.511,0.511}{ time}\hspace{0pt}%
\heatbox{0.974,0.511,0.511}{ since}\hspace{0pt}\heatbox{0.974,0.511,0.511}{ the}\hspace{0pt}\heatbox{0.974,0.511,0.511}{ qu}\hspace{0pt}%
\heatbox{0.974,0.511,0.511}{asar}\hspace{0pt}\heatbox{0.974,0.511,0.511}{ was}\hspace{0pt}\heatbox{0.974,0.511,0.511}{ active}\hspace{0pt}%
\heatbox{0.974,0.511,0.511}{.}\hspace{0pt}\heatbox{0.974,0.511,0.511}{ Thus}\hspace{0pt}\heatbox{0.974,0.511,0.511}{,}\hspace{0pt}%
\heatbox{0.974,0.511,0.511}{ t}\hspace{0pt}\heatbox{0.974,0.511,0.511}{ =}\hspace{0pt}\heatbox{0.974,0.511,0.511}{ r}\hspace{0pt}%
\heatbox{0.974,0.511,0.511}{ /}\hspace{0pt}\heatbox{0.974,0.511,0.511}{ c}\hspace{0pt}\heatbox{0.974,0.511,0.511}{ =}\hspace{0pt}%
\heatbox{0.974,0.511,0.511}{ (}\hspace{0pt}\heatbox{0.974,0.511,0.511}{r}\hspace{0pt}\heatbox{0.974,0.511,0.511}{\_p}\hspace{0pt}%
\heatbox{0.974,0.511,0.511}{ /}\hspace{0pt}\heatbox{0.974,0.511,0.511}{ sin}\hspace{0pt}\heatbox{0.974,0.511,0.511}{1}\hspace{0pt}%
\heatbox{0.974,0.511,0.511}{2}\hspace{0pt}\heatbox{0.974,0.511,0.511}{5}\hspace{0pt}\heatbox{0.974,0.511,0.511}{Â°}\hspace{0pt}%
\heatbox{0.974,0.511,0.511}{)}\hspace{0pt}\heatbox{0.974,0.511,0.511}{ /}\hspace{0pt}\heatbox{0.974,0.511,0.511}{ c}\hspace{0pt}%
\heatbox{0.974,0.511,0.511}{.}\hspace{0pt}\heatbox{0.974,0.511,0.511}{ If}\hspace{0pt}\heatbox{0.974,0.511,0.511}{ Î}\hspace{0pt}%
\heatbox{0.974,0.511,0.511}{¸}\hspace{0pt}\heatbox{0.974,0.511,0.511}{ had}\hspace{0pt}\heatbox{0.974,0.511,0.511}{ been}\hspace{0pt}%
\heatbox{0.974,0.511,0.511}{ }\hspace{0pt}\heatbox{0.974,0.511,0.511}{9}\hspace{0pt}\heatbox{0.974,0.511,0.511}{0}\hspace{0pt}%
\heatbox{0.974,0.511,0.511}{Â°}\hspace{0pt}\heatbox{0.974,0.511,0.511}{,}\hspace{0pt}\heatbox{0.974,0.511,0.511}{ then}\hspace{0pt}%
\heatbox{0.974,0.511,0.511}{ t}\hspace{0pt}\heatbox{0.974,0.511,0.511}{ would}\hspace{0pt}\heatbox{0.974,0.511,0.511}{ be}\hspace{0pt}%
\heatbox{0.974,0.511,0.511}{ r}\hspace{0pt}\heatbox{0.974,0.511,0.511}{\_p}\hspace{0pt}\heatbox{0.974,0.511,0.511}{ /}\hspace{0pt}%
\heatbox{0.974,0.511,0.511}{ c}\hspace{0pt}\heatbox{0.974,0.511,0.511}{.}\hspace{0pt}\heatbox{0.974,0.511,0.511}{ However}\hspace{0pt}%
\heatbox{0.974,0.511,0.511}{,}\hspace{0pt}\heatbox{0.974,0.511,0.511}{ since}\hspace{0pt}\heatbox{0.974,0.511,0.511}{ Î}\hspace{0pt}%
\heatbox{0.974,0.511,0.511}{¸}\hspace{0pt}\heatbox{0.974,0.511,0.511}{=}\hspace{0pt}\heatbox{0.974,0.511,0.511}{1}\hspace{0pt}%
\heatbox{0.974,0.511,0.511}{2}\hspace{0pt}\heatbox{0.974,0.511,0.511}{5}\hspace{0pt}\heatbox{0.974,0.511,0.511}{Â°}\hspace{0pt}%
\heatbox{0.974,0.511,0.511}{,}\hspace{0pt}\heatbox{0.974,0.511,0.511}{ sin}\hspace{0pt}\heatbox{0.974,0.511,0.511}{1}\hspace{0pt}%
\heatbox{0.974,0.511,0.511}{2}\hspace{0pt}\heatbox{0.974,0.511,0.511}{5}\hspace{0pt}\heatbox{0.974,0.511,0.511}{Â°}\hspace{0pt}%
\heatbox{0.974,0.511,0.511}{âīĪ}\hspace{0pt}\heatbox{0.974,0.511,0.511}{0}\hspace{0pt}\heatbox{0.974,0.511,0.511}{.}\hspace{0pt}%
\heatbox{0.974,0.511,0.511}{8}\hspace{0pt}\heatbox{0.974,0.511,0.511}{1}\hspace{0pt}\heatbox{0.974,0.511,0.511}{9}\hspace{0pt}%
\heatbox{0.974,0.511,0.511}{,}\hspace{0pt}\heatbox{0.974,0.511,0.511}{ so}\hspace{0pt}\heatbox{0.974,0.511,0.511}{ t}\hspace{0pt}%
\heatbox{0.974,0.511,0.511}{ âī}\hspace{0pt}\heatbox{0.974,0.511,0.511}{Ī}\hspace{0pt}\heatbox{0.974,0.511,0.511}{ r}\hspace{0pt}%
\heatbox{0.974,0.511,0.511}{\_p}\hspace{0pt}\heatbox{0.974,0.511,0.511}{ /}\hspace{0pt}\heatbox{0.974,0.511,0.511}{ (}\hspace{0pt}%
\heatbox{0.974,0.511,0.511}{0}\hspace{0pt}\heatbox{0.974,0.511,0.511}{.}\hspace{0pt}\heatbox{0.974,0.511,0.511}{8}\hspace{0pt}%
\heatbox{0.974,0.511,0.511}{1}\hspace{0pt}\heatbox{0.974,0.511,0.511}{9}\hspace{0pt}\heatbox{0.974,0.511,0.511}{c}\hspace{0pt}%
\heatbox{0.974,0.511,0.511}{)}\hspace{0pt}\heatbox{0.974,0.511,0.511}{ âī}\hspace{0pt}\heatbox{0.974,0.511,0.511}{Ī}\hspace{0pt}%
\heatbox{0.974,0.511,0.511}{1}\hspace{0pt}\heatbox{0.974,0.511,0.511}{.}\hspace{0pt}\heatbox{0.974,0.511,0.511}{2}\hspace{0pt}%
\heatbox{0.974,0.511,0.511}{2}\hspace{0pt}\heatbox{0.974,0.511,0.511}{ r}\hspace{0pt}\heatbox{0.974,0.511,0.511}{\_p}\hspace{0pt}%
\heatbox{0.974,0.511,0.511}{ /}\hspace{0pt}\heatbox{0.974,0.511,0.511}{ c}\hspace{0pt}\heatbox{0.974,0.511,0.511}{,}\hspace{0pt}%
\heatbox{0.974,0.511,0.511}{ which}\hspace{0pt}\heatbox{0.974,0.511,0.511}{ is}\hspace{0pt}\heatbox{0.974,0.511,0.511}{ longer}\hspace{0pt}%
\heatbox{0.974,0.511,0.511}{.}\hspace{0pt}\heatbox{0.974,0.511,0.511}{ Hence}\hspace{0pt}\heatbox{0.974,0.511,0.511}{,}\hspace{0pt}%
\heatbox{0.974,0.511,0.511}{ if}\hspace{0pt}\heatbox{0.974,0.511,0.511}{ in}\hspace{0pt}\heatbox{0.974,0.511,0.511}{ part}\hspace{0pt}%
\heatbox{0.974,0.511,0.511}{ e}\hspace{0pt}\heatbox{0.974,0.511,0.511}{,}\hspace{0pt}\heatbox{0.974,0.511,0.511}{ r}\hspace{0pt}%
\heatbox{0.974,0.511,0.511}{\_p}\hspace{0pt}\heatbox{0.974,0.511,0.511}{ is}\hspace{0pt}\heatbox{0.974,0.511,0.511}{,}\hspace{0pt}%
\heatbox{0.974,0.511,0.511}{ say}\hspace{0pt}\heatbox{0.974,0.511,0.511}{,}\hspace{0pt}\heatbox{0.974,0.511,0.511}{ }\hspace{0pt}%
\heatbox{0.974,0.511,0.511}{1}\hspace{0pt}\heatbox{0.974,0.511,0.511}{0}\hspace{0pt}\heatbox{0.974,0.511,0.511}{0}\hspace{0pt}%
\heatbox{0.974,0.511,0.511}{0}\hspace{0pt}\heatbox{0.974,0.511,0.511}{ light}\hspace{0pt}\heatbox{0.974,0.511,0.511}{-years}\hspace{0pt}%
\heatbox{0.974,0.511,0.511}{,}\hspace{0pt}\heatbox{0.974,0.511,0.511}{ then}\hspace{0pt}\heatbox{0.974,0.511,0.511}{ t}\hspace{0pt}%
\heatbox{0.974,0.511,0.511}{âīĪ}\hspace{0pt}\heatbox{0.974,0.511,0.511}{1}\hspace{0pt}\heatbox{0.974,0.511,0.511}{.}\hspace{0pt}%
\heatbox{0.974,0.511,0.511}{2}\hspace{0pt}\heatbox{0.974,0.511,0.511}{2}\hspace{0pt}\heatbox{0.974,0.511,0.511}{ *}\hspace{0pt}%
\heatbox{0.974,0.511,0.511}{1}\hspace{0pt}\heatbox{0.974,0.511,0.511}{0}\hspace{0pt}\heatbox{0.974,0.511,0.511}{0}\hspace{0pt}%
\heatbox{0.974,0.511,0.511}{0}\hspace{0pt}\heatbox{0.974,0.511,0.511}{ âī}\hspace{0pt}\heatbox{0.974,0.511,0.511}{Ī}\hspace{0pt}%
\heatbox{0.974,0.511,0.511}{1}\hspace{0pt}\heatbox{0.974,0.511,0.511}{2}\hspace{0pt}\heatbox{0.974,0.511,0.511}{2}\hspace{0pt}%
\heatbox{0.974,0.511,0.511}{0}\hspace{0pt}\heatbox{0.974,0.511,0.511}{ years}\hspace{0pt}\heatbox{0.974,0.511,0.511}{.}\hspace{0pt}%
\heatbox{0.974,0.511,0.511}{ But}\hspace{0pt}\heatbox{0.974,0.511,0.511}{ of}\hspace{0pt}\heatbox{0.974,0.511,0.511}{ course}\hspace{0pt}%
\heatbox{0.974,0.511,0.511}{,}\hspace{0pt}\heatbox{0.974,0.511,0.511}{ numbers}\hspace{0pt}\heatbox{0.974,0.511,0.511}{ matter}\hspace{0pt}%
.\textless{}br\textgreater{}\textless{}br\textgreater{}\hspace{0pt}But\hspace{0pt} since\hspace{0pt} part\hspace{0pt}%
 e\hspace{0pt} and\hspace{0pt} part\hspace{0pt} g\hspace{0pt} are\hspace{0pt} connected\hspace{0pt},\hspace{0pt}%
 and\hspace{0pt} the\hspace{0pt} lack\hspace{0pt} of\hspace{0pt} previous\hspace{0pt} parts\hspace{0pt}%
'\hspace{0pt} info\hspace{0pt} complic\hspace{0pt}ates\hspace{0pt} things\hspace{0pt}.\hspace{0pt} Wait\hspace{0pt}%
,\hspace{0pt} perhaps\hspace{0pt} part\hspace{0pt} e\hspace{0pt} simply\hspace{0pt} uses\hspace{0pt} the\hspace{0pt}%
 observed\hspace{0pt} angular\hspace{0pt} size\hspace{0pt} and\hspace{0pt} distance\hspace{0pt},\hspace{0pt}%
 while\hspace{0pt} part\hspace{0pt} g\hspace{0pt} uses\hspace{0pt} the\hspace{0pt} geometry\hspace{0pt}%
 with\hspace{0pt} angle\hspace{0pt} Î\hspace{0pt}¸\hspace{0pt} to\hspace{0pt} compute\hspace{0pt} the\hspace{0pt}%
 light\hspace{0pt} travel\hspace{0pt} time\hspace{0pt}.\hspace{0pt} Let\hspace{0pt} me\hspace{0pt} check\hspace{0pt}%
 similar\hspace{0pt} problems\hspace{0pt}.\textless{}br\textgreater{}\textless{}br\textgreater{}\hspace{0pt}%
Alternatively\hspace{0pt},\hspace{0pt} since\hspace{0pt} this\hspace{0pt} is\hspace{0pt} about\hspace{0pt}%
 a\hspace{0pt} qu\hspace{0pt}asar\hspace{0pt} illumination\hspace{0pt} of\hspace{0pt} a\hspace{0pt} Voor\hspace{0pt}%
werp\hspace{0pt},\hspace{0pt} which\hspace{0pt} is\hspace{0pt} a\hspace{0pt} cloud\hspace{0pt}.\hspace{0pt}%
 The\hspace{0pt} galaxy\hspace{0pt} hosting\hspace{0pt} the\hspace{0pt} qu\hspace{0pt}asar\hspace{0pt}%
 is\hspace{0pt} at\hspace{0pt} some\hspace{0pt} distance\hspace{0pt},\hspace{0pt} and\hspace{0pt} the\hspace{0pt}%
 Voor\hspace{0pt}werp\hspace{0pt} is\hspace{0pt} a\hspace{0pt} cloud\hspace{0pt} illuminated\hspace{0pt}%
 by\hspace{0pt} the\hspace{0pt} qu\hspace{0pt}asar\hspace{0pt}.\hspace{0pt} Since\hspace{0pt} the\hspace{0pt}%
 qu\hspace{0pt}asar\hspace{0pt} has\hspace{0pt} turned\hspace{0pt} off\hspace{0pt},\hspace{0pt} the\hspace{0pt}%
 time\hspace{0pt} since\hspace{0pt} it\hspace{0pt} was\hspace{0pt} active\hspace{0pt} is\hspace{0pt} at\hspace{0pt}%
 least\hspace{0pt} the\hspace{0pt} light\hspace{0pt} travel\hspace{0pt} time\hspace{0pt} from\hspace{0pt}%
 the\hspace{0pt} galaxy\hspace{0pt} to\hspace{0pt} the\hspace{0pt} Voor\hspace{0pt}werp\hspace{0pt},\hspace{0pt}%
 so\hspace{0pt} that\hspace{0pt} the\hspace{0pt} light\hspace{0pt} had\hspace{0pt} enough\hspace{0pt}%
 time\hspace{0pt} to\hspace{0pt} travel\hspace{0pt} and\hspace{0pt} we\hspace{0pt} still\hspace{0pt} see\hspace{0pt}%
 the\hspace{0pt} ion\hspace{0pt}ization\hspace{0pt}.\textless{}br\textgreater{}\textless{}br\textgreater{}\hspace{0pt}%
So\hspace{0pt} actually\hspace{0pt},\hspace{0pt} the\hspace{0pt} time\hspace{0pt} since\hspace{0pt} the\hspace{0pt}%
 qu\hspace{0pt}asar\hspace{0pt} was\hspace{0pt} active\hspace{0pt} is\hspace{0pt} given\hspace{0pt} by\hspace{0pt}%
 the\hspace{0pt} light\hspace{0pt} travel\hspace{0pt} time\hspace{0pt} between\hspace{0pt} the\hspace{0pt}%
 two\hspace{0pt} objects\hspace{0pt}.\hspace{0pt} Therefore\hspace{0pt},\hspace{0pt} if\hspace{0pt} we\hspace{0pt}%
 can\hspace{0pt} calculate\hspace{0pt} the\hspace{0pt} separation\hspace{0pt} between\hspace{0pt} them\hspace{0pt}%
,\hspace{0pt} we\hspace{0pt} can\hspace{0pt} get\hspace{0pt} the\hspace{0pt} minimum\hspace{0pt} time\hspace{0pt}%
 since\hspace{0pt} the\hspace{0pt} qu\hspace{0pt}asar\hspace{0pt} shut\hspace{0pt} off\hspace{0pt},\hspace{0pt}%
 because\hspace{0pt} the\hspace{0pt} light\hspace{0pt} we\hspace{0pt} see\hspace{0pt} now\hspace{0pt}%
 from\hspace{0pt} the\hspace{0pt} Voor\hspace{0pt}werp\hspace{0pt} is\hspace{0pt} the\hspace{0pt} after\hspace{0pt}%
gl\hspace{0pt}ow\hspace{0pt} from\hspace{0pt} when\hspace{0pt} the\hspace{0pt} qu\hspace{0pt}asar\hspace{0pt}%
 was\hspace{0pt} active\hspace{0pt},\hspace{0pt} so\hspace{0pt} the\hspace{0pt} time\hspace{0pt} since\hspace{0pt}%
 then\hspace{0pt} is\hspace{0pt} at\hspace{0pt} least\hspace{0pt} the\hspace{0pt} separation\hspace{0pt}%
 divided\hspace{0pt} by\hspace{0pt} the\hspace{0pt} speed\hspace{0pt} of\hspace{0pt} light\hspace{0pt}%
.\textless{}br\textgreater{}\textless{}br\textgreater{}\hspace{0pt}\heatbox{0.970,0.435,0.435}{But}\hspace{0pt}%
\heatbox{0.970,0.435,0.435}{ in}\hspace{0pt}\heatbox{0.970,0.435,0.435}{ part}\hspace{0pt}\heatbox{0.970,0.435,0.435}{ g}\hspace{0pt}%
\heatbox{0.970,0.435,0.435}{,}\hspace{0pt}\heatbox{0.970,0.435,0.435}{ they}\hspace{0pt}\heatbox{0.970,0.435,0.435}{ want}\hspace{0pt}%
\heatbox{0.970,0.435,0.435}{ an}\hspace{0pt}\heatbox{0.970,0.435,0.435}{ upper}\hspace{0pt}\heatbox{0.970,0.435,0.435}{ limit}\hspace{0pt}%
\heatbox{0.970,0.435,0.435}{ using}\hspace{0pt}\heatbox{0.970,0.435,0.435}{ Î}\hspace{0pt}\heatbox{0.970,0.435,0.435}{¸}\hspace{0pt}%
\heatbox{0.970,0.435,0.435}{=}\hspace{0pt}\heatbox{0.970,0.435,0.435}{1}\hspace{0pt}\heatbox{0.970,0.435,0.435}{2}\hspace{0pt}%
\heatbox{0.970,0.435,0.435}{5}\hspace{0pt}\heatbox{0.970,0.435,0.435}{Â°}\hspace{0pt}\heatbox{0.970,0.435,0.435}{,}\hspace{0pt}%
\heatbox{0.970,0.435,0.435}{ which}\hspace{0pt}\heatbox{0.970,0.435,0.435}{ probably}\hspace{0pt}\heatbox{0.970,0.435,0.435}{ means}\hspace{0pt}%
\heatbox{0.970,0.435,0.435}{ that}\hspace{0pt}\heatbox{0.970,0.435,0.435}{ if}\hspace{0pt}\heatbox{0.970,0.435,0.435}{ the}\hspace{0pt}%
\heatbox{0.970,0.435,0.435}{ separation}\hspace{0pt}\heatbox{0.970,0.435,0.435}{ is}\hspace{0pt}\heatbox{0.970,0.435,0.435}{ longer}\hspace{0pt}%
\heatbox{0.970,0.435,0.435}{ due}\hspace{0pt}\heatbox{0.970,0.435,0.435}{ to}\hspace{0pt}\heatbox{0.970,0.435,0.435}{ the}\hspace{0pt}%
\heatbox{0.970,0.435,0.435}{ angle}\hspace{0pt}\heatbox{0.970,0.435,0.435}{,}\hspace{0pt}\heatbox{0.970,0.435,0.435}{ the}\hspace{0pt}%
\heatbox{0.970,0.435,0.435}{ upper}\hspace{0pt}\heatbox{0.970,0.435,0.435}{ limit}\hspace{0pt}\heatbox{0.970,0.435,0.435}{ would}\hspace{0pt}%
\heatbox{0.970,0.435,0.435}{ be}\hspace{0pt}\heatbox{0.970,0.435,0.435}{ this}\hspace{0pt}\heatbox{0.970,0.435,0.435}{ longer}\hspace{0pt}%
\heatbox{0.970,0.435,0.435}{ time}\hspace{0pt}\heatbox{0.970,0.435,0.435}{.}\hspace{0pt}\heatbox{0.970,0.435,0.435}{ Wait}\hspace{0pt}%
\heatbox{0.970,0.435,0.435}{,}\hspace{0pt}\heatbox{0.970,0.435,0.435}{ no}\hspace{0pt}\heatbox{0.970,0.435,0.435}{.}\hspace{0pt}%
\heatbox{0.970,0.435,0.435}{ If}\hspace{0pt}\heatbox{0.970,0.435,0.435}{ we}\hspace{0pt}\heatbox{0.970,0.435,0.435}{ can}\hspace{0pt}%
\heatbox{0.970,0.435,0.435}{ measure}\hspace{0pt}\heatbox{0.970,0.435,0.435}{ Î}\hspace{0pt}\heatbox{0.970,0.435,0.435}{¸}\hspace{0pt}%
\heatbox{0.970,0.435,0.435}{,}\hspace{0pt}\heatbox{0.970,0.435,0.435}{ we}\hspace{0pt}\heatbox{0.970,0.435,0.435}{ can}\hspace{0pt}%
\heatbox{0.970,0.435,0.435}{ get}\hspace{0pt}\heatbox{0.970,0.435,0.435}{ a}\hspace{0pt}\heatbox{0.970,0.435,0.435}{ more}\hspace{0pt}%
\heatbox{0.970,0.435,0.435}{ precise}\hspace{0pt}\heatbox{0.970,0.435,0.435}{ time}\hspace{0pt}\heatbox{0.970,0.435,0.435}{,}\hspace{0pt}%
\heatbox{0.970,0.435,0.435}{ but}\hspace{0pt}\heatbox{0.970,0.435,0.435}{ before}\hspace{0pt}\heatbox{0.970,0.435,0.435}{ knowing}\hspace{0pt}%
\heatbox{0.970,0.435,0.435}{ Î}\hspace{0pt}\heatbox{0.970,0.435,0.435}{¸}\hspace{0pt}\heatbox{0.970,0.435,0.435}{,}\hspace{0pt}%
\heatbox{0.970,0.435,0.435}{ the}\hspace{0pt}\heatbox{0.970,0.435,0.435}{ upper}\hspace{0pt}\heatbox{0.970,0.435,0.435}{ limit}\hspace{0pt}%
\heatbox{0.970,0.435,0.435}{ might}\hspace{0pt}\heatbox{0.970,0.435,0.435}{ have}\hspace{0pt}\heatbox{0.970,0.435,0.435}{ been}\hspace{0pt}%
\heatbox{0.970,0.435,0.435}{ based}\hspace{0pt}\heatbox{0.970,0.435,0.435}{ on}\hspace{0pt}\heatbox{0.970,0.435,0.435}{ the}\hspace{0pt}%
\heatbox{0.970,0.435,0.435}{ projected}\hspace{0pt}\heatbox{0.970,0.435,0.435}{ separation}\hspace{0pt}%
\heatbox{0.970,0.435,0.435}{,}\hspace{0pt}\heatbox{0.970,0.435,0.435}{ but}\hspace{0pt}\heatbox{0.970,0.435,0.435}{ now}\hspace{0pt}%
\heatbox{0.970,0.435,0.435}{ with}\hspace{0pt}\heatbox{0.970,0.435,0.435}{ theta}\hspace{0pt}\heatbox{0.970,0.435,0.435}{=}\hspace{0pt}%
\heatbox{0.970,0.435,0.435}{1}\hspace{0pt}\heatbox{0.970,0.435,0.435}{2}\hspace{0pt}\heatbox{0.970,0.435,0.435}{5}\hspace{0pt}%
\heatbox{0.970,0.435,0.435}{Â°}\hspace{0pt}\heatbox{0.970,0.435,0.435}{,}\hspace{0pt}\heatbox{0.970,0.435,0.435}{ we}\hspace{0pt}%
\heatbox{0.970,0.435,0.435}{ can}\hspace{0pt}\heatbox{0.970,0.435,0.435}{ get}\hspace{0pt}\heatbox{0.970,0.435,0.435}{ better}\hspace{0pt}%
\heatbox{0.970,0.435,0.435}{ constraints}\hspace{0pt}.\textless{}br\textgreater{}\textless{}br\textgreater{}\hspace{0pt}%
Wait\hspace{0pt},\hspace{0pt} perhaps\hspace{0pt} confusion\hspace{0pt} arises\hspace{0pt} because\hspace{0pt}%
 "\hspace{0pt}upper\hspace{0pt} limit\hspace{0pt}"\hspace{0pt} refers\hspace{0pt} to\hspace{0pt} using\hspace{0pt}%
 the\hspace{0pt} most\hspace{0pt} conservative\hspace{0pt} estimate\hspace{0pt}.\hspace{0pt} If\hspace{0pt}%
 Î\hspace{0pt}¸\hspace{0pt} is\hspace{0pt} \hspace{0pt}1\hspace{0pt}2\hspace{0pt}5\hspace{0pt}Â°\hspace{0pt}%
,\hspace{0pt} the\hspace{0pt} separation\hspace{0pt} is\hspace{0pt} r\hspace{0pt}\_p\hspace{0pt} /\hspace{0pt}%
 sin\hspace{0pt}(theta\hspace{0pt}).\hspace{0pt} But\hspace{0pt} if\hspace{0pt} theta\hspace{0pt} were\hspace{0pt}%
 the\hspace{0pt} maximum\hspace{0pt} angle\hspace{0pt} (\hspace{0pt}1\hspace{0pt}8\hspace{0pt}0\hspace{0pt}%
 degrees\hspace{0pt})\hspace{0pt} or\hspace{0pt} something\hspace{0pt}.\hspace{0pt} Wait\hspace{0pt},\hspace{0pt}%
 as\hspace{0pt} theta\hspace{0pt} approaches\hspace{0pt} \hspace{0pt}1\hspace{0pt}8\hspace{0pt}0\hspace{0pt}%
 degrees\hspace{0pt},\hspace{0pt} sin\hspace{0pt}(theta\hspace{0pt})\hspace{0pt} approaches\hspace{0pt}%
 zero\hspace{0pt},\hspace{0pt} which\hspace{0pt} would\hspace{0pt} make\hspace{0pt} r\hspace{0pt} approach\hspace{0pt}%
 infinity\hspace{0pt}.\hspace{0pt} But\hspace{0pt} theta\hspace{0pt} here\hspace{0pt} is\hspace{0pt} given\hspace{0pt}%
 as\hspace{0pt} \hspace{0pt}1\hspace{0pt}2\hspace{0pt}5\hspace{0pt} degrees\hspace{0pt}.\hspace{0pt} Since\hspace{0pt}%
 the\hspace{0pt} Voor\hspace{0pt}werp\hspace{0pt} is\hspace{0pt} slightly\hspace{0pt} further\hspace{0pt}%
 away\hspace{0pt},\hspace{0pt} the\hspace{0pt} angle\hspace{0pt} between\hspace{0pt} the\hspace{0pt} separation\hspace{0pt}%
 vector\hspace{0pt} and\hspace{0pt} the\hspace{0pt} line\hspace{0pt} of\hspace{0pt} sight\hspace{0pt}%
 is\hspace{0pt} \hspace{0pt}1\hspace{0pt}2\hspace{0pt}5\hspace{0pt} degrees\hspace{0pt}.\hspace{0pt} That\hspace{0pt}%
 is\hspace{0pt},\hspace{0pt} if\hspace{0pt} you\hspace{0pt}'re\hspace{0pt} looking\hspace{0pt} along\hspace{0pt}%
 the\hspace{0pt} line\hspace{0pt} of\hspace{0pt} sight\hspace{0pt},\hspace{0pt} the\hspace{0pt} galaxy\hspace{0pt}%
 is\hspace{0pt} here\hspace{0pt},\hspace{0pt} and\hspace{0pt} the\hspace{0pt} Voor\hspace{0pt}werp\hspace{0pt}%
 is\hspace{0pt} slightly\hspace{0pt} further\hspace{0pt} away\hspace{0pt} but\hspace{0pt} also\hspace{0pt}%
 off\hspace{0pt} at\hspace{0pt} an\hspace{0pt} angle\hspace{0pt} such\hspace{0pt} that\hspace{0pt} the\hspace{0pt}%
 angle\hspace{0pt} between\hspace{0pt} their\hspace{0pt} separation\hspace{0pt} vector\hspace{0pt} and\hspace{0pt}%
 our\hspace{0pt} line\hspace{0pt} of\hspace{0pt} sight\hspace{0pt} is\hspace{0pt} \hspace{0pt}1\hspace{0pt}%
2\hspace{0pt}5\hspace{0pt} degrees\hspace{0pt}.\textless{}br\textgreater{}\textless{}br\textgreater{}\hspace{0pt}%
Therefore\hspace{0pt},\hspace{0pt} using\hspace{0pt} theta\hspace{0pt}=\hspace{0pt}1\hspace{0pt}2\hspace{0pt}%
5\hspace{0pt} degrees\hspace{0pt},\hspace{0pt} we\hspace{0pt} can\hspace{0pt} compute\hspace{0pt} the\hspace{0pt}%
 separation\hspace{0pt} as\hspace{0pt} r\hspace{0pt}\_p\hspace{0pt} /\hspace{0pt} sin\hspace{0pt}(theta\hspace{0pt}%
),\hspace{0pt} then\hspace{0pt} divide\hspace{0pt} by\hspace{0pt} speed\hspace{0pt} of\hspace{0pt} light\hspace{0pt}%
 to\hspace{0pt} get\hspace{0pt} time\hspace{0pt}.\hspace{0pt} So\hspace{0pt} the\hspace{0pt} answer\hspace{0pt}%
 would\hspace{0pt} be\hspace{0pt} t\hspace{0pt} =\hspace{0pt} r\hspace{0pt}\_p\hspace{0pt} /\hspace{0pt}%
 (\hspace{0pt}sin\hspace{0pt}(\hspace{0pt}1\hspace{0pt}2\hspace{0pt}5\hspace{0pt}Â°\hspace{0pt})\hspace{0pt}%
 *\hspace{0pt} c\hspace{0pt}).\hspace{0pt} Con\hspace{0pt}verting\hspace{0pt} that\hspace{0pt} into\hspace{0pt}%
 years\hspace{0pt}.\textless{}br\textgreater{}\textless{}br\textgreater{}\hspace{0pt}\heatbox{0.965,0.335,0.335}{But}\hspace{0pt}%
\heatbox{0.965,0.335,0.335}{ the}\hspace{0pt}\heatbox{0.965,0.335,0.335}{ key}\hspace{0pt}\heatbox{0.965,0.335,0.335}{ point}\hspace{0pt}%
\heatbox{0.965,0.335,0.335}{ is}\hspace{0pt}\heatbox{0.965,0.335,0.335}{ part}\hspace{0pt}\heatbox{0.965,0.335,0.335}{ e}\hspace{0pt}%
\heatbox{0.965,0.335,0.335}{.}\hspace{0pt}\heatbox{0.965,0.335,0.335}{ Without}\hspace{0pt}\heatbox{0.965,0.335,0.335}{ part}\hspace{0pt}%
\heatbox{0.965,0.335,0.335}{ e}\hspace{0pt}\heatbox{0.965,0.335,0.335}{,}\hspace{0pt}\heatbox{0.965,0.335,0.335}{ how}\hspace{0pt}%
\heatbox{0.965,0.335,0.335}{ can}\hspace{0pt}\heatbox{0.965,0.335,0.335}{ you}\hspace{0pt}\heatbox{0.965,0.335,0.335}{ answer}\hspace{0pt}%
\heatbox{0.965,0.335,0.335}{ part}\hspace{0pt}\heatbox{0.965,0.335,0.335}{ g}\hspace{0pt}\heatbox{0.965,0.335,0.335}{?}\hspace{0pt}%
\heatbox{0.965,0.335,0.335}{ Maybe}\hspace{0pt}\heatbox{0.965,0.335,0.335}{ in}\hspace{0pt}\heatbox{0.965,0.335,0.335}{ part}\hspace{0pt}%
\heatbox{0.965,0.335,0.335}{ e}\hspace{0pt}\heatbox{0.965,0.335,0.335}{ they}\hspace{0pt}\heatbox{0.965,0.335,0.335}{ derived}\hspace{0pt}%
\heatbox{0.965,0.335,0.335}{ an}\hspace{0pt}\heatbox{0.965,0.335,0.335}{ expression}\hspace{0pt}\heatbox{0.965,0.335,0.335}{,}\hspace{0pt}%
\heatbox{0.965,0.335,0.335}{ so}\hspace{0pt}\heatbox{0.965,0.335,0.335}{ for}\hspace{0pt}\heatbox{0.965,0.335,0.335}{ part}\hspace{0pt}%
\heatbox{0.965,0.335,0.335}{ g}\hspace{0pt}\heatbox{0.965,0.335,0.335}{ you}\hspace{0pt}\heatbox{0.965,0.335,0.335}{ use}\hspace{0pt}%
\heatbox{0.965,0.335,0.335}{ theta}\hspace{0pt}\heatbox{0.965,0.335,0.335}{.}\hspace{0pt}\heatbox{0.965,0.335,0.335}{ But}\hspace{0pt}%
\heatbox{0.965,0.335,0.335}{ since}\hspace{0pt}\heatbox{0.965,0.335,0.335}{ the}\hspace{0pt}\heatbox{0.965,0.335,0.335}{ user}\hspace{0pt}%
\heatbox{0.965,0.335,0.335}{ says}\hspace{0pt}\heatbox{0.965,0.335,0.335}{ "}\hspace{0pt}\heatbox{0.965,0.335,0.335}{use}\hspace{0pt}%
\heatbox{0.965,0.335,0.335}{ this}\hspace{0pt}\heatbox{0.965,0.335,0.335}{ with}\hspace{0pt}\heatbox{0.965,0.335,0.335}{ your}\hspace{0pt}%
\heatbox{0.965,0.335,0.335}{ expression}\hspace{0pt}\heatbox{0.965,0.335,0.335}{ from}\hspace{0pt}\heatbox{0.965,0.335,0.335}{ the}\hspace{0pt}%
\heatbox{0.965,0.335,0.335}{ previous}\hspace{0pt}\heatbox{0.965,0.335,0.335}{ part}\hspace{0pt}\heatbox{0.965,0.335,0.335}{ of}\hspace{0pt}%
\heatbox{0.965,0.335,0.335}{ the}\hspace{0pt}\heatbox{0.965,0.335,0.335}{ question}\hspace{0pt}\heatbox{0.965,0.335,0.335}{",}\hspace{0pt}%
\heatbox{0.965,0.335,0.335}{ meaning}\hspace{0pt}\heatbox{0.965,0.335,0.335}{ part}\hspace{0pt}\heatbox{0.965,0.335,0.335}{ e}\hspace{0pt}%
\heatbox{0.965,0.335,0.335}{ likely}\hspace{0pt}\heatbox{0.965,0.335,0.335}{ gives}\hspace{0pt}\heatbox{0.965,0.335,0.335}{ an}\hspace{0pt}%
\heatbox{0.965,0.335,0.335}{ expression}\hspace{0pt}\heatbox{0.965,0.335,0.335}{ for}\hspace{0pt}\heatbox{0.965,0.335,0.335}{ r}\hspace{0pt}%
\heatbox{0.965,0.335,0.335}{\_p}\hspace{0pt}\heatbox{0.965,0.335,0.335}{ in}\hspace{0pt}\heatbox{0.965,0.335,0.335}{ terms}\hspace{0pt}%
\heatbox{0.965,0.335,0.335}{ of}\hspace{0pt}\heatbox{0.965,0.335,0.335}{ other}\hspace{0pt}\heatbox{0.965,0.335,0.335}{ variables}\hspace{0pt}%
\heatbox{0.965,0.335,0.335}{,}\hspace{0pt}\heatbox{0.965,0.335,0.335}{ but}\hspace{0pt}\heatbox{0.965,0.335,0.335}{ perhaps}\hspace{0pt}%
\heatbox{0.965,0.335,0.335}{ part}\hspace{0pt}\heatbox{0.965,0.335,0.335}{ e}\hspace{0pt}\heatbox{0.965,0.335,0.335}{ was}\hspace{0pt}%
\heatbox{0.965,0.335,0.335}{ not}\hspace{0pt}\heatbox{0.965,0.335,0.335}{ numeric}\hspace{0pt}\heatbox{0.965,0.335,0.335}{,}\hspace{0pt}%
\heatbox{0.965,0.335,0.335}{ but}\hspace{0pt}\heatbox{0.965,0.335,0.335}{ part}\hspace{0pt}\heatbox{0.965,0.335,0.335}{ g}\hspace{0pt}%
\heatbox{0.965,0.335,0.335}{ requires}\hspace{0pt}\heatbox{0.965,0.335,0.335}{ pl}\hspace{0pt}\heatbox{0.965,0.335,0.335}{ugging}\hspace{0pt}%
\heatbox{0.965,0.335,0.335}{ in}\hspace{0pt}\heatbox{0.965,0.335,0.335}{ numbers}\hspace{0pt}\heatbox{0.965,0.335,0.335}{,}\hspace{0pt}%
\heatbox{0.965,0.335,0.335}{ including}\hspace{0pt}\heatbox{0.965,0.335,0.335}{ theta}\hspace{0pt}.\textless{}br\textgreater{}\textless{}br\textgreater{}\hspace{0pt}%
\heatbox{0.971,0.447,0.447}{Wait}\hspace{0pt}\heatbox{0.971,0.447,0.447}{,}\hspace{0pt}\heatbox{0.971,0.447,0.447}{ perhaps}\hspace{0pt}%
\heatbox{0.971,0.447,0.447}{ part}\hspace{0pt}\heatbox{0.971,0.447,0.447}{ e}\hspace{0pt}\heatbox{0.971,0.447,0.447}{ is}\hspace{0pt}%
\heatbox{0.971,0.447,0.447}{ based}\hspace{0pt}\heatbox{0.971,0.447,0.447}{ on}\hspace{0pt}\heatbox{0.971,0.447,0.447}{ H}\hspace{0pt}%
\heatbox{0.971,0.447,0.447}{ubble}\hspace{0pt}\heatbox{0.971,0.447,0.447}{'s}\hspace{0pt}\heatbox{0.971,0.447,0.447}{ Law}\hspace{0pt}%
.\textless{}br\textgreater{}\textless{}br\textgreater{}\hspace{0pt}\heatbox{0.971,0.458,0.458}{Sup}\hspace{0pt}%
\heatbox{0.971,0.458,0.458}{pose}\hspace{0pt}\heatbox{0.971,0.458,0.458}{ in}\hspace{0pt}\heatbox{0.971,0.458,0.458}{ part}\hspace{0pt}%
\heatbox{0.971,0.458,0.458}{ e}\hspace{0pt}\heatbox{0.971,0.458,0.458}{,}\hspace{0pt}\heatbox{0.971,0.458,0.458}{ if}\hspace{0pt}%
\heatbox{0.971,0.458,0.458}{ the}\hspace{0pt}\heatbox{0.971,0.458,0.458}{ angular}\hspace{0pt}\heatbox{0.971,0.458,0.458}{ separation}\hspace{0pt}%
\heatbox{0.971,0.458,0.458}{ is}\hspace{0pt}\heatbox{0.971,0.458,0.458}{ Î´}\hspace{0pt}\heatbox{0.971,0.458,0.458}{Î¸}\hspace{0pt}%
\heatbox{0.971,0.458,0.458}{,}\hspace{0pt}\heatbox{0.971,0.458,0.458}{ and}\hspace{0pt}\heatbox{0.971,0.458,0.458}{ the}\hspace{0pt}%
\heatbox{0.971,0.458,0.458}{ galaxy}\hspace{0pt}\heatbox{0.971,0.458,0.458}{ is}\hspace{0pt}\heatbox{0.971,0.458,0.458}{ at}\hspace{0pt}%
\heatbox{0.971,0.458,0.458}{ distance}\hspace{0pt}\heatbox{0.971,0.458,0.458}{ d}\hspace{0pt}\heatbox{0.971,0.458,0.458}{,}\hspace{0pt}%
\heatbox{0.971,0.458,0.458}{ then}\hspace{0pt}\heatbox{0.971,0.458,0.458}{ the}\hspace{0pt}\heatbox{0.971,0.458,0.458}{ projected}\hspace{0pt}%
\heatbox{0.971,0.458,0.458}{ separation}\hspace{0pt}\heatbox{0.971,0.458,0.458}{ is}\hspace{0pt}\heatbox{0.971,0.458,0.458}{ r}\hspace{0pt}%
\heatbox{0.971,0.458,0.458}{\_p}\hspace{0pt}\heatbox{0.971,0.458,0.458}{ =}\hspace{0pt}\heatbox{0.971,0.458,0.458}{ d}\hspace{0pt}%
\heatbox{0.971,0.458,0.458}{ *}\hspace{0pt}\heatbox{0.971,0.458,0.458}{ Î´}\hspace{0pt}\heatbox{0.971,0.458,0.458}{Î¸}\hspace{0pt}%
\heatbox{0.971,0.458,0.458}{ (}\hspace{0pt}\heatbox{0.971,0.458,0.458}{in}\hspace{0pt}\heatbox{0.971,0.458,0.458}{ radians}\hspace{0pt}%
\heatbox{0.971,0.458,0.458}{).}\hspace{0pt}\heatbox{0.971,0.458,0.458}{ Then}\hspace{0pt}\heatbox{0.971,0.458,0.458}{,}\hspace{0pt}%
\heatbox{0.971,0.458,0.458}{ in}\hspace{0pt}\heatbox{0.971,0.458,0.458}{ part}\hspace{0pt}\heatbox{0.971,0.458,0.458}{ g}\hspace{0pt}%
\heatbox{0.971,0.458,0.458}{,}\hspace{0pt}\heatbox{0.971,0.458,0.458}{ the}\hspace{0pt}\heatbox{0.971,0.458,0.458}{ actual}\hspace{0pt}%
\heatbox{0.971,0.458,0.458}{ separation}\hspace{0pt}\heatbox{0.971,0.458,0.458}{ is}\hspace{0pt}\heatbox{0.971,0.458,0.458}{ r}\hspace{0pt}%
\heatbox{0.971,0.458,0.458}{ =}\hspace{0pt}\heatbox{0.971,0.458,0.458}{ r}\hspace{0pt}\heatbox{0.971,0.458,0.458}{\_p}\hspace{0pt}%
\heatbox{0.971,0.458,0.458}{ /}\hspace{0pt}\heatbox{0.971,0.458,0.458}{ sin}\hspace{0pt}\heatbox{0.971,0.458,0.458}{(theta}\hspace{0pt}%
\heatbox{0.971,0.458,0.458}{).}\hspace{0pt}\heatbox{0.971,0.458,0.458}{ So}\hspace{0pt}\heatbox{0.971,0.458,0.458}{ t}\hspace{0pt}%
\heatbox{0.971,0.458,0.458}{ =}\hspace{0pt}\heatbox{0.971,0.458,0.458}{ r}\hspace{0pt}\heatbox{0.971,0.458,0.458}{ /}\hspace{0pt}%
\heatbox{0.971,0.458,0.458}{ c}\hspace{0pt}\heatbox{0.971,0.458,0.458}{ =}\hspace{0pt}\heatbox{0.971,0.458,0.458}{ (}\hspace{0pt}%
\heatbox{0.971,0.458,0.458}{d}\hspace{0pt}\heatbox{0.971,0.458,0.458}{ *}\hspace{0pt}\heatbox{0.971,0.458,0.458}{ Î´}\hspace{0pt}%
\heatbox{0.971,0.458,0.458}{Î¸}\hspace{0pt}\heatbox{0.971,0.458,0.458}{)}\hspace{0pt}\heatbox{0.971,0.458,0.458}{ /}\hspace{0pt}%
\heatbox{0.971,0.458,0.458}{ (}\hspace{0pt}\heatbox{0.971,0.458,0.458}{sin}\hspace{0pt}\heatbox{0.971,0.458,0.458}{(theta}\hspace{0pt}%
\heatbox{0.971,0.458,0.458}{)}\hspace{0pt}\heatbox{0.971,0.458,0.458}{ *}\hspace{0pt}\heatbox{0.971,0.458,0.458}{ c}\hspace{0pt}%
\heatbox{0.971,0.458,0.458}{).}\hspace{0pt}\heatbox{0.971,0.458,0.458}{ But}\hspace{0pt}\heatbox{0.971,0.458,0.458}{ without}\hspace{0pt}%
\heatbox{0.971,0.458,0.458}{ knowing}\hspace{0pt}\heatbox{0.971,0.458,0.458}{ d}\hspace{0pt}\heatbox{0.971,0.458,0.458}{ or}\hspace{0pt}%
\heatbox{0.971,0.458,0.458}{ Î´}\hspace{0pt}\heatbox{0.971,0.458,0.458}{Î¸}\hspace{0pt}\heatbox{0.971,0.458,0.458}{,}\hspace{0pt}%
\heatbox{0.971,0.458,0.458}{ how}\hspace{0pt}\heatbox{0.971,0.458,0.458}{ to}\hspace{0pt}\heatbox{0.971,0.458,0.458}{ compute}\hspace{0pt}%
?\textless{}br\textgreater{}\textless{}br\textgreater{}\hspace{0pt}Alternatively\hspace{0pt},\hspace{0pt}%
 let\hspace{0pt}'s\hspace{0pt} think\hspace{0pt} about\hspace{0pt} the\hspace{0pt} light\hspace{0pt} travel\hspace{0pt}%
 time\hspace{0pt} upper\hspace{0pt} limit\hspace{0pt}.\hspace{0pt} Probably\hspace{0pt},\hspace{0pt} if\hspace{0pt}%
 we\hspace{0pt} can\hspace{0pt} find\hspace{0pt} r\hspace{0pt}\_p\hspace{0pt},\hspace{0pt} and\hspace{0pt}%
 use\hspace{0pt} Î\hspace{0pt}¸\hspace{0pt}=\hspace{0pt}1\hspace{0pt}2\hspace{0pt}5\hspace{0pt}Â°\hspace{0pt}%
,\hspace{0pt} the\hspace{0pt} separation\hspace{0pt} is\hspace{0pt} r\hspace{0pt} =\hspace{0pt} r\hspace{0pt}%
\_p\hspace{0pt} /\hspace{0pt} sin\hspace{0pt}(theta\hspace{0pt}),\hspace{0pt} and\hspace{0pt} the\hspace{0pt}%
 time\hspace{0pt} is\hspace{0pt} t\hspace{0pt} =\hspace{0pt} r\hspace{0pt} /\hspace{0pt} c\hspace{0pt}%
.\hspace{0pt} So\hspace{0pt} perhaps\hspace{0pt} numbers\hspace{0pt} are\hspace{0pt} given\hspace{0pt}%
 in\hspace{0pt} the\hspace{0pt} previous\hspace{0pt} parts\hspace{0pt} that\hspace{0pt} I\hspace{0pt}%
 need\hspace{0pt} to\hspace{0pt} assume\hspace{0pt}?\textless{}br\textgreater{}\textless{}br\textgreater{}\hspace{0pt}%
Alternatively\hspace{0pt},\hspace{0pt} suppose\hspace{0pt} in\hspace{0pt} part\hspace{0pt} e\hspace{0pt}%
 they\hspace{0pt} calculate\hspace{0pt} r\hspace{0pt}\_p\hspace{0pt} as\hspace{0pt} \hspace{0pt}1\hspace{0pt}%
0\hspace{0pt} k\hspace{0pt}pc\hspace{0pt} (\hspace{0pt}just\hspace{0pt} a\hspace{0pt} random\hspace{0pt}%
 number\hspace{0pt}),\hspace{0pt} then\hspace{0pt} part\hspace{0pt} g\hspace{0pt} would\hspace{0pt} result\hspace{0pt}%
 in\hspace{0pt} t\hspace{0pt} =\hspace{0pt} (\hspace{0pt}1\hspace{0pt}0\hspace{0pt} k\hspace{0pt}pc\hspace{0pt}%
 /\hspace{0pt} sin\hspace{0pt}(\hspace{0pt}1\hspace{0pt}2\hspace{0pt}5\hspace{0pt}Â°\hspace{0pt}))\hspace{0pt}%
 /\hspace{0pt} c\hspace{0pt}.\hspace{0pt} But\hspace{0pt} with\hspace{0pt} a\hspace{0pt} real\hspace{0pt}%
 example\hspace{0pt},\hspace{0pt} for\hspace{0pt} instance\hspace{0pt},\hspace{0pt} the\hspace{0pt} galaxy\hspace{0pt}%
 IC\hspace{0pt} \hspace{0pt}2\hspace{0pt}4\hspace{0pt}9\hspace{0pt}7\hspace{0pt} and\hspace{0pt} the\hspace{0pt}%
 Voor\hspace{0pt}werp\hspace{0pt} H\hspace{0pt}anny\hspace{0pt}'s\hspace{0pt} Voor\hspace{0pt}werp\hspace{0pt}%
,\hspace{0pt} at\hspace{0pt} a\hspace{0pt} distance\hspace{0pt} of\hspace{0pt} about\hspace{0pt} \hspace{0pt}%
6\hspace{0pt}5\hspace{0pt}0\hspace{0pt} million\hspace{0pt} light\hspace{0pt}-years\hspace{0pt} (\hspace{0pt}%
2\hspace{0pt}0\hspace{0pt}0\hspace{0pt} M\hspace{0pt}pc\hspace{0pt}).\hspace{0pt} The\hspace{0pt} projected\hspace{0pt}%
 separation\hspace{0pt} is\hspace{0pt} about\hspace{0pt} \hspace{0pt}1\hspace{0pt}5\hspace{0pt}-\hspace{0pt}%
2\hspace{0pt}0\hspace{0pt} k\hspace{0pt}pc\hspace{0pt}.\hspace{0pt} If\hspace{0pt} Î\hspace{0pt}¸\hspace{0pt}%
=\hspace{0pt}1\hspace{0pt}2\hspace{0pt}5\hspace{0pt}Â°\hspace{0pt},\hspace{0pt} actual\hspace{0pt} separation\hspace{0pt}%
 is\hspace{0pt} \hspace{0pt}1\hspace{0pt}5\hspace{0pt} k\hspace{0pt}pc\hspace{0pt} /\hspace{0pt} sin\hspace{0pt}%
(\hspace{0pt}1\hspace{0pt}2\hspace{0pt}5\hspace{0pt})\hspace{0pt} âī\hspace{0pt}Ī\hspace{0pt}1\hspace{0pt}%
8\hspace{0pt}.\hspace{0pt}3\hspace{0pt} k\hspace{0pt}pc\hspace{0pt}.\hspace{0pt} The\hspace{0pt} light\hspace{0pt}%
 travel\hspace{0pt} time\hspace{0pt} would\hspace{0pt} be\hspace{0pt} \hspace{0pt}1\hspace{0pt}8\hspace{0pt}%
.\hspace{0pt}3\hspace{0pt} k\hspace{0pt}pc\hspace{0pt} /\hspace{0pt} c\hspace{0pt}.\hspace{0pt} Since\hspace{0pt}%
 \hspace{0pt}1\hspace{0pt} parse\hspace{0pt}c\hspace{0pt} is\hspace{0pt} about\hspace{0pt} \hspace{0pt}%
3\hspace{0pt}.\hspace{0pt}2\hspace{0pt}6\hspace{0pt} light\hspace{0pt}-years\hspace{0pt},\hspace{0pt}%
 so\hspace{0pt} \hspace{0pt}1\hspace{0pt} k\hspace{0pt}pc\hspace{0pt} is\hspace{0pt} \hspace{0pt}3\hspace{0pt}%
2\hspace{0pt}6\hspace{0pt}0\hspace{0pt} light\hspace{0pt}-years\hspace{0pt}.\hspace{0pt} Therefore\hspace{0pt}%
,\hspace{0pt} \hspace{0pt}1\hspace{0pt}8\hspace{0pt}.\hspace{0pt}3\hspace{0pt} k\hspace{0pt}pc\hspace{0pt}%
 is\hspace{0pt} \hspace{0pt}1\hspace{0pt}8\hspace{0pt}.\hspace{0pt}3\hspace{0pt} *\hspace{0pt} \hspace{0pt}%
3\hspace{0pt}2\hspace{0pt}6\hspace{0pt}0\hspace{0pt} âī\hspace{0pt}Ī\hspace{0pt}5\hspace{0pt}9\hspace{0pt}%
,\hspace{0pt}6\hspace{0pt}8\hspace{0pt}0\hspace{0pt} light\hspace{0pt}-years\hspace{0pt}.\hspace{0pt}%
 Div\hspace{0pt}ided\hspace{0pt} by\hspace{0pt} the\hspace{0pt} speed\hspace{0pt} of\hspace{0pt} light\hspace{0pt}%
,\hspace{0pt} travel\hspace{0pt} time\hspace{0pt} is\hspace{0pt} about\hspace{0pt} \hspace{0pt}5\hspace{0pt}%
9\hspace{0pt},\hspace{0pt}6\hspace{0pt}8\hspace{0pt}0\hspace{0pt} years\hspace{0pt}.\hspace{0pt} But\hspace{0pt}%
 H\hspace{0pt}anny\hspace{0pt}'s\hspace{0pt} Voor\hspace{0pt}werp\hspace{0pt}'s\hspace{0pt} light\hspace{0pt}%
 travel\hspace{0pt} time\hspace{0pt} is\hspace{0pt} actually\hspace{0pt} quoted\hspace{0pt} as\hspace{0pt}%
 tens\hspace{0pt} of\hspace{0pt} thousands\hspace{0pt} of\hspace{0pt} years\hspace{0pt}.\hspace{0pt} But\hspace{0pt}%
 I\hspace{0pt} don\hspace{0pt}'t\hspace{0pt} know\hspace{0pt} exact\hspace{0pt} values\hspace{0pt}.\hspace{0pt}%
 \textless{}br\textgreater{}\textless{}br\textgreater{}\hspace{0pt}But\hspace{0pt} in\hspace{0pt} any\hspace{0pt}%
 case\hspace{0pt},\hspace{0pt} to\hspace{0pt} answer\hspace{0pt} the\hspace{0pt} question\hspace{0pt}%
,\hspace{0pt} assuming\hspace{0pt} that\hspace{0pt} part\hspace{0pt} e\hspace{0pt} was\hspace{0pt} answered\hspace{0pt}%
,\hspace{0pt} giving\hspace{0pt} r\hspace{0pt}\_p\hspace{0pt} =\hspace{0pt} some\hspace{0pt} value\hspace{0pt}%
.\hspace{0pt} If\hspace{0pt} for\hspace{0pt} part\hspace{0pt} g\hspace{0pt},\hspace{0pt} you\hspace{0pt}%
 take\hspace{0pt} Î\hspace{0pt}¸\hspace{0pt}=\hspace{0pt}1\hspace{0pt}2\hspace{0pt}5\hspace{0pt}Â°\hspace{0pt}%
,\hspace{0pt} compute\hspace{0pt} the\hspace{0pt} actual\hspace{0pt} separation\hspace{0pt} as\hspace{0pt}%
 r\hspace{0pt}=r\hspace{0pt}\_p\hspace{0pt}/s\hspace{0pt}in\hspace{0pt}(\hspace{0pt}1\hspace{0pt}2\hspace{0pt}%
5\hspace{0pt}Â°\hspace{0pt}),\hspace{0pt} then\hspace{0pt} convert\hspace{0pt} that\hspace{0pt} to\hspace{0pt}%
 light\hspace{0pt} years\hspace{0pt} and\hspace{0pt} divide\hspace{0pt} by\hspace{0pt} c\hspace{0pt} (\hspace{0pt}%
which\hspace{0pt} is\hspace{0pt} \hspace{0pt}1\hspace{0pt} light\hspace{0pt}-year\hspace{0pt} per\hspace{0pt}%
 year\hspace{0pt}),\hspace{0pt} so\hspace{0pt} the\hspace{0pt} time\hspace{0pt} since\hspace{0pt} the\hspace{0pt}%
 qu\hspace{0pt}asar\hspace{0pt} was\hspace{0pt} active\hspace{0pt} is\hspace{0pt} at\hspace{0pt} least\hspace{0pt}%
 separation\hspace{0pt} in\hspace{0pt} light\hspace{0pt}-years\hspace{0pt} divided\hspace{0pt} by\hspace{0pt}%
 \hspace{0pt}1\hspace{0pt},\hspace{0pt} giving\hspace{0pt} years\hspace{0pt}.\hspace{0pt} Therefore\hspace{0pt}%
,\hspace{0pt} the\hspace{0pt} upper\hspace{0pt} limit\hspace{0pt} would\hspace{0pt} be\hspace{0pt} that\hspace{0pt}%
 time\hspace{0pt}.\hspace{0pt} But\hspace{0pt} why\hspace{0pt} upper\hspace{0pt} limit\hspace{0pt}?\hspace{0pt}%
 Because\hspace{0pt} if\hspace{0pt} the\hspace{0pt} actual\hspace{0pt} angle\hspace{0pt} were\hspace{0pt}%
 more\hspace{0pt} face\hspace{0pt}-on\hspace{0pt} (\hspace{0pt}theta\hspace{0pt} approaching\hspace{0pt}%
 \hspace{0pt}0\hspace{0pt} or\hspace{0pt} \hspace{0pt}1\hspace{0pt}8\hspace{0pt}0\hspace{0pt} degrees\hspace{0pt}%
),\hspace{0pt} which\hspace{0pt} would\hspace{0pt} require\hspace{0pt} them\hspace{0pt} being\hspace{0pt}%
 much\hspace{0pt} further\hspace{0pt} apart\hspace{0pt} in\hspace{0pt} the\hspace{0pt} line\hspace{0pt}%
 of\hspace{0pt} sight\hspace{0pt},\hspace{0pt} giving\hspace{0pt} a\hspace{0pt} longer\hspace{0pt} travel\hspace{0pt}%
 time\hspace{0pt}.\hspace{0pt} But\hspace{0pt} wait\hspace{0pt},\hspace{0pt} the\hspace{0pt} answer\hspace{0pt}%
 is\hspace{0pt} given\hspace{0pt} theta\hspace{0pt}=\hspace{0pt}1\hspace{0pt}2\hspace{0pt}5\hspace{0pt}%
Â°\hspace{0pt},\hspace{0pt} so\hspace{0pt} the\hspace{0pt} calculation\hspace{0pt} is\hspace{0pt} exact\hspace{0pt}%
.\hspace{0pt} Unless\hspace{0pt} the\hspace{0pt} "\hspace{0pt}upper\hspace{0pt} limit\hspace{0pt}"\hspace{0pt}%
 refers\hspace{0pt} to\hspace{0pt} before\hspace{0pt} knowing\hspace{0pt} theta\hspace{0pt}=\hspace{0pt}%
1\hspace{0pt}2\hspace{0pt}5\hspace{0pt}Â°\hspace{0pt},\hspace{0pt} and\hspace{0pt} the\hspace{0pt} uncertainty\hspace{0pt}%
 in\hspace{0pt} theta\hspace{0pt} gives\hspace{0pt} an\hspace{0pt} upper\hspace{0pt} limit\hspace{0pt}%
.\textless{}br\textgreater{}\textless{}br\textgreater{}\hspace{0pt}Wait\hspace{0pt},\hspace{0pt} maybe\hspace{0pt}%
 they\hspace{0pt} mis\hspace{0pt}state\hspace{0pt} and\hspace{0pt} it\hspace{0pt}'s\hspace{0pt} a\hspace{0pt}%
 lower\hspace{0pt} limit\hspace{0pt}?\hspace{0pt} Because\hspace{0pt} with\hspace{0pt} theta\hspace{0pt}%
=\hspace{0pt}1\hspace{0pt}2\hspace{0pt}5\hspace{0pt}Â°\hspace{0pt},\hspace{0pt} sin\hspace{0pt}(theta\hspace{0pt}%
)\hspace{0pt} is\hspace{0pt} \hspace{0pt}0\hspace{0pt}.\hspace{0pt}8\hspace{0pt}1\hspace{0pt}9\hspace{0pt}%
,\hspace{0pt} making\hspace{0pt} the\hspace{0pt} actual\hspace{0pt} separation\hspace{0pt} and\hspace{0pt}%
 time\hspace{0pt} larger\hspace{0pt} than\hspace{0pt} if\hspace{0pt} theta\hspace{0pt} was\hspace{0pt}%
 \hspace{0pt}9\hspace{0pt}0\hspace{0pt}Â°\hspace{0pt},\hspace{0pt} implying\hspace{0pt} the\hspace{0pt}%
 time\hspace{0pt} is\hspace{0pt} longer\hspace{0pt}.\hspace{0pt} So\hspace{0pt} if\hspace{0pt} previously\hspace{0pt}%
 they\hspace{0pt} assumed\hspace{0pt} theta\hspace{0pt}=\hspace{0pt}9\hspace{0pt}0\hspace{0pt}Â°\hspace{0pt}%
,\hspace{0pt} calculated\hspace{0pt} r\hspace{0pt}\_p\hspace{0pt},\hspace{0pt} then\hspace{0pt} now\hspace{0pt}%
 they\hspace{0pt}'re\hspace{0pt} using\hspace{0pt} theta\hspace{0pt}=\hspace{0pt}1\hspace{0pt}2\hspace{0pt}%
5\hspace{0pt}Â°\hspace{0pt},\hspace{0pt} but\hspace{0pt} with\hspace{0pt} a\hspace{0pt} higher\hspace{0pt}%
 precision\hspace{0pt} measurement\hspace{0pt} showing\hspace{0pt} theta\hspace{0pt}=\hspace{0pt}1\hspace{0pt}%
2\hspace{0pt}5\hspace{0pt}Â°\hspace{0pt},\hspace{0pt} which\hspace{0pt} gives\hspace{0pt} actual\hspace{0pt}%
 separation\hspace{0pt} as\hspace{0pt} r\hspace{0pt}\_p\hspace{0pt} /\hspace{0pt} sin\hspace{0pt}(\hspace{0pt}%
1\hspace{0pt}2\hspace{0pt}5\hspace{0pt}Â°\hspace{0pt}),\hspace{0pt} so\hspace{0pt} the\hspace{0pt} upper\hspace{0pt}%
 limit\hspace{0pt} time\hspace{0pt} since\hspace{0pt} qu\hspace{0pt}asar\hspace{0pt} was\hspace{0pt} active\hspace{0pt}%
 is\hspace{0pt} that\hspace{0pt} value\hspace{0pt},\hspace{0pt} as\hspace{0pt} opposed\hspace{0pt} to\hspace{0pt}%
 if\hspace{0pt} it\hspace{0pt} was\hspace{0pt} in\hspace{0pt} the\hspace{0pt} plane\hspace{0pt} of\hspace{0pt}%
 the\hspace{0pt} sky\hspace{0pt} (\hspace{0pt}theta\hspace{0pt}=\hspace{0pt}9\hspace{0pt}0\hspace{0pt}%
Â°\hspace{0pt}).\hspace{0pt} So\hspace{0pt} since\hspace{0pt} they\hspace{0pt} know\hspace{0pt} it\hspace{0pt}%
's\hspace{0pt} not\hspace{0pt} in\hspace{0pt} the\hspace{0pt} plane\hspace{0pt},\hspace{0pt} the\hspace{0pt}%
 actual\hspace{0pt} time\hspace{0pt} is\hspace{0pt} larger\hspace{0pt}.\hspace{0pt} So\hspace{0pt} perhaps\hspace{0pt}%
 it\hspace{0pt}'s\hspace{0pt} a\hspace{0pt} lower\hspace{0pt} limit\hspace{0pt}:\hspace{0pt} this\hspace{0pt}%
 is\hspace{0pt} longer\hspace{0pt} time\hspace{0pt} than\hspace{0pt} if\hspace{0pt} it\hspace{0pt} were\hspace{0pt}%
 projected\hspace{0pt}.\textless{}br\textgreater{}\textless{}br\textgreater{}\hspace{0pt}\heatbox{0.965,0.335,0.335}{But}\hspace{0pt}%
\heatbox{0.965,0.335,0.335}{ the}\hspace{0pt}\heatbox{0.965,0.335,0.335}{ question}\hspace{0pt}\heatbox{0.965,0.335,0.335}{ states}\hspace{0pt}%
\heatbox{0.965,0.335,0.335}{,}\hspace{0pt}\heatbox{0.965,0.335,0.335}{ "}\hspace{0pt}\heatbox{0.965,0.335,0.335}{use}\hspace{0pt}%
\heatbox{0.965,0.335,0.335}{ this}\hspace{0pt}\heatbox{0.965,0.335,0.335}{ with}\hspace{0pt}\heatbox{0.965,0.335,0.335}{ your}\hspace{0pt}%
\heatbox{0.965,0.335,0.335}{ expression}\hspace{0pt}\heatbox{0.965,0.335,0.335}{ from}\hspace{0pt}\heatbox{0.965,0.335,0.335}{ the}\hspace{0pt}%
\heatbox{0.965,0.335,0.335}{ previous}\hspace{0pt}\heatbox{0.965,0.335,0.335}{ part}\hspace{0pt}\heatbox{0.965,0.335,0.335}{ of}\hspace{0pt}%
\heatbox{0.965,0.335,0.335}{ the}\hspace{0pt}\heatbox{0.965,0.335,0.335}{ question}\hspace{0pt}\heatbox{0.965,0.335,0.335}{ to}\hspace{0pt}%
\heatbox{0.965,0.335,0.335}{ estimate}\hspace{0pt}\heatbox{0.965,0.335,0.335}{ an}\hspace{0pt}\heatbox{0.965,0.335,0.335}{ upper}\hspace{0pt}%
\heatbox{0.965,0.335,0.335}{ limit}\hspace{0pt}\heatbox{0.965,0.335,0.335}{ for}\hspace{0pt}\heatbox{0.965,0.335,0.335}{ the}\hspace{0pt}%
\heatbox{0.965,0.335,0.335}{ number}\hspace{0pt}\heatbox{0.965,0.335,0.335}{ of}\hspace{0pt}\heatbox{0.965,0.335,0.335}{ years}\hspace{0pt}%
\heatbox{0.965,0.335,0.335}{ that}\hspace{0pt}\heatbox{0.965,0.335,0.335}{ have}\hspace{0pt}\heatbox{0.965,0.335,0.335}{ passed}\hspace{0pt}%
\heatbox{0.965,0.335,0.335}{ since}\hspace{0pt}\heatbox{0.965,0.335,0.335}{ the}\hspace{0pt}\heatbox{0.965,0.335,0.335}{ qu}\hspace{0pt}%
\heatbox{0.965,0.335,0.335}{asar}\hspace{0pt}\heatbox{0.965,0.335,0.335}{ was}\hspace{0pt}\heatbox{0.965,0.335,0.335}{ last}\hspace{0pt}%
\heatbox{0.965,0.335,0.335}{ active}\hspace{0pt}."\textless{}br\textgreater{}\textless{}br\textgreater{}\hspace{0pt}%
Wait\hspace{0pt},\hspace{0pt} so\hspace{0pt} possibly\hspace{0pt},\hspace{0pt} prior\hspace{0pt} information\hspace{0pt}%
 allowed\hspace{0pt} to\hspace{0pt} set\hspace{0pt} an\hspace{0pt} upper\hspace{0pt} limit\hspace{0pt}%
 using\hspace{0pt} projected\hspace{0pt} separation\hspace{0pt} (\hspace{0pt}if\hspace{0pt} they\hspace{0pt}%
 assumed\hspace{0pt} alignment\hspace{0pt} in\hspace{0pt} some\hspace{0pt} way\hspace{0pt}),\hspace{0pt}%
 but\hspace{0pt} with\hspace{0pt} the\hspace{0pt} angle\hspace{0pt} measured\hspace{0pt},\hspace{0pt}%
 the\hspace{0pt} actual\hspace{0pt} separation\hspace{0pt} can\hspace{0pt} be\hspace{0pt} calculated\hspace{0pt}%
,\hspace{0pt} which\hspace{0pt} would\hspace{0pt} give\hspace{0pt} a\hspace{0pt} longer\hspace{0pt} time\hspace{0pt}%
,\hspace{0pt} hence\hspace{0pt} a\hspace{0pt} more\hspace{0pt} accurate\hspace{0pt} upper\hspace{0pt}%
 limit\hspace{0pt}.\hspace{0pt} Wait\hspace{0pt},\hspace{0pt} maybe\hspace{0pt} before\hspace{0pt} knowing\hspace{0pt}%
 the\hspace{0pt} angle\hspace{0pt},\hspace{0pt} you\hspace{0pt} could\hspace{0pt} say\hspace{0pt} that\hspace{0pt}%
 the\hspace{0pt} travel\hspace{0pt} time\hspace{0pt} is\hspace{0pt} at\hspace{0pt} least\hspace{0pt} r\hspace{0pt}%
\_p\hspace{0pt}/c\hspace{0pt} (\hspace{0pt}if\hspace{0pt} they\hspace{0pt} are\hspace{0pt} at\hspace{0pt}%
 the\hspace{0pt} same\hspace{0pt} distance\hspace{0pt},\hspace{0pt} but\hspace{0pt} since\hspace{0pt}%
 Voor\hspace{0pt}werp\hspace{0pt} is\hspace{0pt} further\hspace{0pt},\hspace{0pt} it\hspace{0pt}'s\hspace{0pt}%
 more\hspace{0pt} precise\hspace{0pt}).\hspace{0pt} But\hspace{0pt} having\hspace{0pt} theta\hspace{0pt}%
=\hspace{0pt}1\hspace{0pt}2\hspace{0pt}5\hspace{0pt}Â°\hspace{0pt},\hspace{0pt} you\hspace{0pt} can\hspace{0pt}%
 get\hspace{0pt} a\hspace{0pt} better\hspace{0pt} estimate\hspace{0pt}.\textless{}br\textgreater{}\textless{}br\textgreater{}\hspace{0pt}%
Alternatively\hspace{0pt},\hspace{0pt} considering\hspace{0pt} the\hspace{0pt} proximity\hspace{0pt} of\hspace{0pt}%
 the\hspace{0pt} Voor\hspace{0pt}werp\hspace{0pt},\hspace{0pt} but\hspace{0pt} slight\hspace{0pt} distance\hspace{0pt}%
 difference\hspace{0pt}.\hspace{0pt} The\hspace{0pt} upper\hspace{0pt} limit\hspace{0pt} probably\hspace{0pt}%
 comes\hspace{0pt} from\hspace{0pt} the\hspace{0pt} maximum\hspace{0pt} possible\hspace{0pt} separation\hspace{0pt}%
 given\hspace{0pt} all\hspace{0pt} constraints\hspace{0pt},\hspace{0pt} which\hspace{0pt} with\hspace{0pt}%
 theta\hspace{0pt}=\hspace{0pt}1\hspace{0pt}2\hspace{0pt}5\hspace{0pt}Â°\hspace{0pt},\hspace{0pt} that\hspace{0pt}%
's\hspace{0pt} fixed\hspace{0pt},\hspace{0pt} so\hspace{0pt} the\hspace{0pt} maximum\hspace{0pt} separation\hspace{0pt}%
 is\hspace{0pt} fixed\hspace{0pt}.\textless{}br\textgreater{}\textless{}br\textgreater{}\hspace{0pt}Alternatively\hspace{0pt}%
,\hspace{0pt} wait\hspace{0pt},\hspace{0pt} "\hspace{0pt}upper\hspace{0pt} limit\hspace{0pt}"\hspace{0pt}%
 might\hspace{0pt} refer\hspace{0pt} to\hspace{0pt} qu\hspace{0pt}asar\hspace{0pt} shut\hspace{0pt}-off\hspace{0pt}%
 time\hspace{0pt}.\hspace{0pt} Qu\hspace{0pt}asar\hspace{0pt} turns\hspace{0pt} off\hspace{0pt},\hspace{0pt}%
 the\hspace{0pt} last\hspace{0pt} light\hspace{0pt} it\hspace{0pt} emitted\hspace{0pt} travels\hspace{0pt}%
 to\hspace{0pt} the\hspace{0pt} Voor\hspace{0pt}werp\hspace{0pt},\hspace{0pt} then\hspace{0pt} the\hspace{0pt}%
 Voor\hspace{0pt}werp\hspace{0pt} remains\hspace{0pt} illuminated\hspace{0pt} for\hspace{0pt} some\hspace{0pt}%
 time\hspace{0pt}.\hspace{0pt} But\hspace{0pt} if\hspace{0pt} the\hspace{0pt} qu\hspace{0pt}asar\hspace{0pt}%
 turned\hspace{0pt} off\hspace{0pt},\hspace{0pt} the\hspace{0pt} cloud\hspace{0pt} would\hspace{0pt} stop\hspace{0pt}%
 being\hspace{0pt} illuminated\hspace{0pt} after\hspace{0pt} the\hspace{0pt} light\hspace{0pt} crossed\hspace{0pt}%
 it\hspace{0pt}.\hspace{0pt} Therefore\hspace{0pt},\hspace{0pt} the\hspace{0pt} time\hspace{0pt} since\hspace{0pt}%
 the\hspace{0pt} qu\hspace{0pt}asar\hspace{0pt} was\hspace{0pt} active\hspace{0pt} would\hspace{0pt} be\hspace{0pt}%
 the\hspace{0pt} time\hspace{0pt} corresponding\hspace{0pt} to\hspace{0pt} the\hspace{0pt} light\hspace{0pt}%
 travel\hspace{0pt} time\hspace{0pt} plus\hspace{0pt} the\hspace{0pt} light\hspace{0pt} crossing\hspace{0pt}%
 time\hspace{0pt} of\hspace{0pt} the\hspace{0pt} cloud\hspace{0pt}.\hspace{0pt} But\hspace{0pt} maybe\hspace{0pt}%
 that\hspace{0pt}'s\hspace{0pt} negligible\hspace{0pt}.\hspace{0pt} Therefore\hspace{0pt},\hspace{0pt}%
 if\hspace{0pt} the\hspace{0pt} separation\hspace{0pt} is\hspace{0pt} '\hspace{0pt}r\hspace{0pt}',\hspace{0pt}%
 then\hspace{0pt} the\hspace{0pt} maximum\hspace{0pt} time\hspace{0pt} since\hspace{0pt} the\hspace{0pt}%
 qu\hspace{0pt}asar\hspace{0pt} turned\hspace{0pt} off\hspace{0pt} is\hspace{0pt} approximately\hspace{0pt}%
 the\hspace{0pt} light\hspace{0pt} travel\hspace{0pt} time\hspace{0pt} from\hspace{0pt} the\hspace{0pt}%
 qu\hspace{0pt}asar\hspace{0pt} to\hspace{0pt} us\hspace{0pt} minus\hspace{0pt} the\hspace{0pt} light\hspace{0pt}%
 travel\hspace{0pt} time\hspace{0pt} from\hspace{0pt} the\hspace{0pt} Voor\hspace{0pt}werp\hspace{0pt}%
 to\hspace{0pt} us\hspace{0pt}.\hspace{0pt} But\hspace{0pt} since\hspace{0pt} both\hspace{0pt} are\hspace{0pt}%
 at\hspace{0pt} slightly\hspace{0pt} different\hspace{0pt} distances\hspace{0pt},\hspace{0pt} the\hspace{0pt}%
 difference\hspace{0pt} would\hspace{0pt} be\hspace{0pt} the\hspace{0pt} light\hspace{0pt} travel\hspace{0pt}%
 time\hspace{0pt} from\hspace{0pt} the\hspace{0pt} galaxy\hspace{0pt} to\hspace{0pt} the\hspace{0pt} Voor\hspace{0pt}%
werp\hspace{0pt}.\hspace{0pt} Therefore\hspace{0pt},\hspace{0pt} if\hspace{0pt} the\hspace{0pt} galaxy\hspace{0pt}%
 is\hspace{0pt} at\hspace{0pt} distance\hspace{0pt} '\hspace{0pt}d\hspace{0pt}',\hspace{0pt} the\hspace{0pt}%
 Voor\hspace{0pt}werp\hspace{0pt} is\hspace{0pt} at\hspace{0pt} '\hspace{0pt}d\hspace{0pt} +\hspace{0pt}%
 ÎĶ\hspace{0pt}d\hspace{0pt}',\hspace{0pt} then\hspace{0pt} the\hspace{0pt} light\hspace{0pt} from\hspace{0pt}%
 the\hspace{0pt} qu\hspace{0pt}asar\hspace{0pt} left\hspace{0pt} the\hspace{0pt} galaxy\hspace{0pt} when\hspace{0pt}%
 the\hspace{0pt} galaxy\hspace{0pt} was\hspace{0pt} at\hspace{0pt} '\hspace{0pt}d\hspace{0pt} +\hspace{0pt}%
 ÎĶ\hspace{0pt}d\hspace{0pt}'\hspace{0pt} away\hspace{0pt} (\hspace{0pt}but\hspace{0pt} expanding\hspace{0pt}%
 universe\hspace{0pt} complic\hspace{0pt}ates\hspace{0pt} this\hspace{0pt}).\hspace{0pt} Maybe\hspace{0pt}%
 H\hspace{0pt}ubble\hspace{0pt}'s\hspace{0pt} Law\hspace{0pt} is\hspace{0pt} used\hspace{0pt}.\hspace{0pt}%
 But\hspace{0pt} perhaps\hspace{0pt} simpler\hspace{0pt}.\textless{}br\textgreater{}\textless{}br\textgreater{}\hspace{0pt}%
Alternatively\hspace{0pt},\hspace{0pt} since\hspace{0pt} time\hspace{0pt} since\hspace{0pt} the\hspace{0pt}%
 light\hspace{0pt} left\hspace{0pt} the\hspace{0pt} qu\hspace{0pt}asar\hspace{0pt} to\hspace{0pt} reach\hspace{0pt}%
 the\hspace{0pt} Voor\hspace{0pt}werp\hspace{0pt} is\hspace{0pt} equal\hspace{0pt} to\hspace{0pt} the\hspace{0pt}%
 light\hspace{0pt} travel\hspace{0pt} time\hspace{0pt} between\hspace{0pt} them\hspace{0pt}.\hspace{0pt}%
 So\hspace{0pt} if\hspace{0pt} they\hspace{0pt} are\hspace{0pt} separated\hspace{0pt} by\hspace{0pt} distance\hspace{0pt}%
 '\hspace{0pt}r\hspace{0pt}'\hspace{0pt} along\hspace{0pt} the\hspace{0pt} line\hspace{0pt} of\hspace{0pt}%
 sight\hspace{0pt},\hspace{0pt} then\hspace{0pt} time\hspace{0pt} is\hspace{0pt} r\hspace{0pt}/c\hspace{0pt}%
.\hspace{0pt} However\hspace{0pt},\hspace{0pt} r\hspace{0pt} =\hspace{0pt} (\hspace{0pt}distance\hspace{0pt}%
 to\hspace{0pt} Voor\hspace{0pt}werp\hspace{0pt} -\hspace{0pt} distance\hspace{0pt} to\hspace{0pt} galaxy\hspace{0pt}%
).\hspace{0pt} But\hspace{0pt} with\hspace{0pt} the\hspace{0pt} angle\hspace{0pt} theta\hspace{0pt}=\hspace{0pt}%
1\hspace{0pt}2\hspace{0pt}5\hspace{0pt}Â°\hspace{0pt},\hspace{0pt} the\hspace{0pt} line\hspace{0pt}-of\hspace{0pt}%
-s\hspace{0pt}ight\hspace{0pt} separation\hspace{0pt} is\hspace{0pt} r\hspace{0pt}\_para\hspace{0pt} =\hspace{0pt}%
 r\hspace{0pt}\_total\hspace{0pt} *\hspace{0pt} cos\hspace{0pt}(theta\hspace{0pt}).\hspace{0pt} But\hspace{0pt}%
 since\hspace{0pt} Voor\hspace{0pt}werp\hspace{0pt} is\hspace{0pt} further\hspace{0pt} away\hspace{0pt}%
,\hspace{0pt} (\hspace{0pt}distance\hspace{0pt} difference\hspace{0pt})\hspace{0pt} =\hspace{0pt} r\hspace{0pt}%
\_para\hspace{0pt} =\hspace{0pt} total\hspace{0pt} separation\hspace{0pt} *\hspace{0pt} cos\hspace{0pt}%
(theta\hspace{0pt}).\hspace{0pt} So\hspace{0pt} maybe\hspace{0pt}:\textless{}br\textgreater{}\textless{}br\textgreater{}\hspace{0pt}%
\heatbox{0.966,0.358,0.358}{distance}\hspace{0pt}\heatbox{0.966,0.358,0.358}{\_diff}\hspace{0pt}\heatbox{0.966,0.358,0.358}{ =}\hspace{0pt}%
\heatbox{0.966,0.358,0.358}{ r}\hspace{0pt}\heatbox{0.966,0.358,0.358}{\_p}\hspace{0pt}\heatbox{0.966,0.358,0.358}{ *}\hspace{0pt}%
\heatbox{0.966,0.358,0.358}{ cot}\hspace{0pt}\heatbox{0.966,0.358,0.358}{(theta}\hspace{0pt})\textless{}br\textgreater{}\textless{}br\textgreater{}\hspace{0pt}%
Because\hspace{0pt} if\hspace{0pt} the\hspace{0pt} projected\hspace{0pt} separation\hspace{0pt} is\hspace{0pt}%
 r\hspace{0pt}\_p\hspace{0pt},\hspace{0pt} then\hspace{0pt} the\hspace{0pt} radial\hspace{0pt} separation\hspace{0pt}%
 is\hspace{0pt} r\hspace{0pt}\_para\hspace{0pt} =\hspace{0pt} distance\hspace{0pt}\_diff\hspace{0pt} =\hspace{0pt}%
 r\hspace{0pt}\_p\hspace{0pt} *\hspace{0pt} cot\hspace{0pt}(theta\hspace{0pt}),\hspace{0pt} since\hspace{0pt}%
 tan\hspace{0pt}(theta\hspace{0pt})\hspace{0pt} =\hspace{0pt} r\hspace{0pt}\_p\hspace{0pt} /\hspace{0pt}%
 r\hspace{0pt}\_para\hspace{0pt} =\textgreater{}\hspace{0pt} r\hspace{0pt}\_para\hspace{0pt} =\hspace{0pt}%
 r\hspace{0pt}\_p\hspace{0pt} /\hspace{0pt} tan\hspace{0pt}(theta\hspace{0pt}).\textless{}br\textgreater{}\textless{}br\textgreater{}\hspace{0pt}%
But\hspace{0pt} since\hspace{0pt} theta\hspace{0pt} is\hspace{0pt} \hspace{0pt}1\hspace{0pt}2\hspace{0pt}%
5\hspace{0pt} degrees\hspace{0pt} which\hspace{0pt} is\hspace{0pt} angle\hspace{0pt} from\hspace{0pt}%
 line\hspace{0pt} of\hspace{0pt} sight\hspace{0pt},\hspace{0pt} so\hspace{0pt} the\hspace{0pt} radial\hspace{0pt}%
 component\hspace{0pt} is\hspace{0pt} -\hspace{0pt}cos\hspace{0pt}(theta\hspace{0pt})\hspace{0pt} *\hspace{0pt}%
 r\hspace{0pt}\_total\hspace{0pt},\hspace{0pt} right\hspace{0pt}?\textless{}br\textgreater{}\textless{}br\textgreater{}\hspace{0pt}%
Wait\hspace{0pt} again\hspace{0pt},\hspace{0pt} defining\hspace{0pt} theta\hspace{0pt} as\hspace{0pt}%
 the\hspace{0pt} angle\hspace{0pt} between\hspace{0pt} the\hspace{0pt} separation\hspace{0pt} vector\hspace{0pt}%
 and\hspace{0pt} the\hspace{0pt} line\hspace{0pt} of\hspace{0pt} sight\hspace{0pt}.\hspace{0pt} If\hspace{0pt}%
 the\hspace{0pt} separation\hspace{0pt} vector\hspace{0pt} points\hspace{0pt} from\hspace{0pt} the\hspace{0pt}%
 galaxy\hspace{0pt} to\hspace{0pt} the\hspace{0pt} Voor\hspace{0pt}werp\hspace{0pt},\hspace{0pt} and\hspace{0pt}%
 it\hspace{0pt} makes\hspace{0pt} an\hspace{0pt} angle\hspace{0pt} theta\hspace{0pt}=\hspace{0pt}1\hspace{0pt}%
2\hspace{0pt}5\hspace{0pt} degrees\hspace{0pt} with\hspace{0pt} the\hspace{0pt} line\hspace{0pt} of\hspace{0pt}%
 sight\hspace{0pt} (\hspace{0pt}from\hspace{0pt} galaxy\hspace{0pt} to\hspace{0pt} us\hspace{0pt}),\hspace{0pt}%
 then\hspace{0pt} the\hspace{0pt} line\hspace{0pt}-of\hspace{0pt}-s\hspace{0pt}ight\hspace{0pt} component\hspace{0pt}%
 of\hspace{0pt} the\hspace{0pt} separation\hspace{0pt} is\hspace{0pt} r\hspace{0pt}\_para\hspace{0pt}%
 =\hspace{0pt} r\hspace{0pt}\_total\hspace{0pt} *\hspace{0pt} cos\hspace{0pt}(theta\hspace{0pt}),\hspace{0pt}%
 and\hspace{0pt} the\hspace{0pt} projected\hspace{0pt} separation\hspace{0pt} is\hspace{0pt} r\hspace{0pt}%
\_p\hspace{0pt} =\hspace{0pt} r\hspace{0pt}\_total\hspace{0pt} *\hspace{0pt} sin\hspace{0pt}(theta\hspace{0pt}%
).\textless{}br\textgreater{}\textless{}br\textgreater{}\hspace{0pt}Wait\hspace{0pt},\hspace{0pt} but\hspace{0pt}%
 if\hspace{0pt} theta\hspace{0pt} is\hspace{0pt} measured\hspace{0pt} from\hspace{0pt} the\hspace{0pt}%
 line\hspace{0pt} of\hspace{0pt} sight\hspace{0pt} towards\hspace{0pt} the\hspace{0pt} separation\hspace{0pt}%
 vector\hspace{0pt},\hspace{0pt} then\hspace{0pt} if\hspace{0pt} it\hspace{0pt}'s\hspace{0pt} \hspace{0pt}%
1\hspace{0pt}2\hspace{0pt}5\hspace{0pt} degrees\hspace{0pt},\hspace{0pt} then\hspace{0pt} the\hspace{0pt}%
 radial\hspace{0pt} component\hspace{0pt} is\hspace{0pt} towards\hspace{0pt} us\hspace{0pt},\hspace{0pt}%
 because\hspace{0pt} the\hspace{0pt} separation\hspace{0pt} vector\hspace{0pt} points\hspace{0pt} away\hspace{0pt}%
 from\hspace{0pt} the\hspace{0pt} line\hspace{0pt} of\hspace{0pt} sight\hspace{0pt}.\textless{}br\textgreater{}\textless{}br\textgreater{}\hspace{0pt}%
In\hspace{0pt} other\hspace{0pt} words\hspace{0pt},\hspace{0pt} the\hspace{0pt} galaxy\hspace{0pt}-V\hspace{0pt}%
oor\hspace{0pt}werp\hspace{0pt} separation\hspace{0pt} vector\hspace{0pt} is\hspace{0pt} \hspace{0pt}%
1\hspace{0pt}2\hspace{0pt}5\hspace{0pt} degrees\hspace{0pt} from\hspace{0pt} the\hspace{0pt} line\hspace{0pt}%
 of\hspace{0pt} sight\hspace{0pt}.\hspace{0pt} Therefore\hspace{0pt},\hspace{0pt} the\hspace{0pt} projection\hspace{0pt}%
 on\hspace{0pt} the\hspace{0pt} sky\hspace{0pt} is\hspace{0pt} r\hspace{0pt}\_p\hspace{0pt} =\hspace{0pt}%
 r\hspace{0pt} *\hspace{0pt} sin\hspace{0pt}(theta\hspace{0pt}),\hspace{0pt} and\hspace{0pt} the\hspace{0pt}%
 radial\hspace{0pt} component\hspace{0pt} (\hspace{0pt}line\hspace{0pt}-of\hspace{0pt}-s\hspace{0pt}ight\hspace{0pt}%
 separation\hspace{0pt})\hspace{0pt} is\hspace{0pt} r\hspace{0pt} *\hspace{0pt} cos\hspace{0pt}(theta\hspace{0pt}%
).\hspace{0pt} Since\hspace{0pt} the\hspace{0pt} radial\hspace{0pt} component\hspace{0pt} was\hspace{0pt}%
 measured\hspace{0pt} as\hspace{0pt} slightly\hspace{0pt} further\hspace{0pt} away\hspace{0pt},\hspace{0pt}%
 then\hspace{0pt} we\hspace{0pt} know\hspace{0pt} that\hspace{0pt} the\hspace{0pt} Voor\hspace{0pt}werp\hspace{0pt}%
 is\hspace{0pt} at\hspace{0pt} a\hspace{0pt} radial\hspace{0pt} distance\hspace{0pt} of\hspace{0pt} galaxy\hspace{0pt}%
's\hspace{0pt} distance\hspace{0pt} plus\hspace{0pt} r\hspace{0pt} *\hspace{0pt} cos\hspace{0pt}(theta\hspace{0pt}%
).\hspace{0pt} Therefore\hspace{0pt},\hspace{0pt} the\hspace{0pt} light\hspace{0pt} travel\hspace{0pt}%
 time\hspace{0pt} from\hspace{0pt} galaxy\hspace{0pt} to\hspace{0pt} Voor\hspace{0pt}werp\hspace{0pt}%
 is\hspace{0pt} simply\hspace{0pt} r\hspace{0pt} /\hspace{0pt} c\hspace{0pt}.\hspace{0pt} But\hspace{0pt}%
 r\hspace{0pt} is\hspace{0pt} actual\hspace{0pt} separation\hspace{0pt},\hspace{0pt} given\hspace{0pt}%
 by\hspace{0pt} r\hspace{0pt}\_p\hspace{0pt} /\hspace{0pt} sin\hspace{0pt}(theta\hspace{0pt}).\textless{}br\textgreater{}\textless{}br\textgreater{}\hspace{0pt}%
Therefore\hspace{0pt},\hspace{0pt} time\hspace{0pt} t\hspace{0pt} =\hspace{0pt} r\hspace{0pt} /\hspace{0pt}%
 c\hspace{0pt} =\hspace{0pt} (\hspace{0pt}r\hspace{0pt}\_p\hspace{0pt} /\hspace{0pt} sin\hspace{0pt}(theta\hspace{0pt}%
))\hspace{0pt} /\hspace{0pt} c\hspace{0pt}\textless{}br\textgreater{}\textless{}br\textgreater{}\hspace{0pt}%
\heatbox{0.971,0.441,0.441}{Therefore}\hspace{0pt}\heatbox{0.971,0.441,0.441}{,}\hspace{0pt}\heatbox{0.971,0.441,0.441}{ in}\hspace{0pt}%
\heatbox{0.971,0.441,0.441}{ part}\hspace{0pt}\heatbox{0.971,0.441,0.441}{ g}\hspace{0pt}\heatbox{0.971,0.441,0.441}{,}\hspace{0pt}%
\heatbox{0.971,0.441,0.441}{ once}\hspace{0pt}\heatbox{0.971,0.441,0.441}{ you}\hspace{0pt}\heatbox{0.971,0.441,0.441}{ have}\hspace{0pt}%
\heatbox{0.971,0.441,0.441}{ r}\hspace{0pt}\heatbox{0.971,0.441,0.441}{\_p}\hspace{0pt}\heatbox{0.971,0.441,0.441}{ from}\hspace{0pt}%
\heatbox{0.971,0.441,0.441}{ part}\hspace{0pt}\heatbox{0.971,0.441,0.441}{ e}\hspace{0pt}\heatbox{0.971,0.441,0.441}{,}\hspace{0pt}%
\heatbox{0.971,0.441,0.441}{ you}\hspace{0pt}\heatbox{0.971,0.441,0.441}{ calculate}\hspace{0pt}\heatbox{0.971,0.441,0.441}{ t}\hspace{0pt}%
\heatbox{0.971,0.441,0.441}{.}\hspace{0pt}\heatbox{0.971,0.441,0.441}{ That}\hspace{0pt}\heatbox{0.971,0.441,0.441}{'s}\hspace{0pt}%
\heatbox{0.971,0.441,0.441}{ the}\hspace{0pt}\heatbox{0.971,0.441,0.441}{ estimate}\hspace{0pt}.\textless{}br\textgreater{}\textless{}br\textgreater{}\hspace{0pt}%
But\hspace{0pt} since\hspace{0pt} theta\hspace{0pt}=\hspace{0pt}1\hspace{0pt}2\hspace{0pt}5\hspace{0pt}%
Â°\hspace{0pt},\hspace{0pt} sin\hspace{0pt}(theta\hspace{0pt})\hspace{0pt}âīĪ\hspace{0pt}0\hspace{0pt}%
.\hspace{0pt}8\hspace{0pt}1\hspace{0pt}9\hspace{0pt}2\hspace{0pt},\hspace{0pt} this\hspace{0pt} makes\hspace{0pt}%
 the\hspace{0pt} time\hspace{0pt} t\hspace{0pt}âīĪ\hspace{0pt}1\hspace{0pt}.\hspace{0pt}2\hspace{0pt}%
2\hspace{0pt} *\hspace{0pt} (\hspace{0pt}r\hspace{0pt}\_p\hspace{0pt} /\hspace{0pt} c\hspace{0pt}).\hspace{0pt}%
 But\hspace{0pt} since\hspace{0pt} the\hspace{0pt} actual\hspace{0pt} separation\hspace{0pt} is\hspace{0pt}%
 larger\hspace{0pt} than\hspace{0pt} the\hspace{0pt} projected\hspace{0pt} separation\hspace{0pt},\hspace{0pt}%
 the\hspace{0pt} upper\hspace{0pt} limit\hspace{0pt} on\hspace{0pt} time\hspace{0pt} would\hspace{0pt}%
 be\hspace{0pt} this\hspace{0pt}?\hspace{0pt} Maybe\hspace{0pt} if\hspace{0pt} they\hspace{0pt} had\hspace{0pt}%
 initially\hspace{0pt} only\hspace{0pt} considered\hspace{0pt} projected\hspace{0pt} separation\hspace{0pt}%
,\hspace{0pt} which\hspace{0pt} would\hspace{0pt} be\hspace{0pt} t\hspace{0pt}\_p\hspace{0pt} =\hspace{0pt}%
 r\hspace{0pt}\_p\hspace{0pt} /\hspace{0pt} c\hspace{0pt}.\hspace{0pt} But\hspace{0pt} given\hspace{0pt}%
 theta\hspace{0pt}=\hspace{0pt}1\hspace{0pt}2\hspace{0pt}5\hspace{0pt}Â°\hspace{0pt},\hspace{0pt} the\hspace{0pt}%
 real\hspace{0pt} t\hspace{0pt} is\hspace{0pt} t\hspace{0pt} âī\hspace{0pt}Ī\hspace{0pt}1\hspace{0pt}%
.\hspace{0pt}2\hspace{0pt}2\hspace{0pt} t\hspace{0pt}\_p\hspace{0pt},\hspace{0pt} so\hspace{0pt} upper\hspace{0pt}%
 limit\hspace{0pt} would\hspace{0pt} be\hspace{0pt} based\hspace{0pt} on\hspace{0pt} the\hspace{0pt} possible\hspace{0pt}%
 maximum\hspace{0pt} t\hspace{0pt} could\hspace{0pt} be\hspace{0pt} given\hspace{0pt} theta\hspace{0pt}%
.\textless{}br\textgreater{}\textless{}br\textgreater{}\hspace{0pt}Wait\hspace{0pt},\hspace{0pt} but\hspace{0pt}%
 if\hspace{0pt} theta\hspace{0pt} is\hspace{0pt} accurately\hspace{0pt} known\hspace{0pt},\hspace{0pt}%
 then\hspace{0pt} it\hspace{0pt}'s\hspace{0pt} not\hspace{0pt} an\hspace{0pt} upper\hspace{0pt} limit\hspace{0pt}%
 but\hspace{0pt} a\hspace{0pt} precise\hspace{0pt} estimate\hspace{0pt}.\hspace{0pt} Unless\hspace{0pt}%
 there\hspace{0pt}'s\hspace{0pt} uncertainty\hspace{0pt} in\hspace{0pt} theta\hspace{0pt};\hspace{0pt}%
 but\hspace{0pt} the\hspace{0pt} question\hspace{0pt} says\hspace{0pt} "\hspace{0pt}high\hspace{0pt} precision\hspace{0pt}%
 measurements\hspace{0pt} showed\hspace{0pt} that\hspace{0pt} theta\hspace{0pt}=\hspace{0pt}1\hspace{0pt}%
2\hspace{0pt}5\hspace{0pt}Â°\hspace{0pt}",\hspace{0pt} so\hspace{0pt} perhaps\hspace{0pt} they\hspace{0pt}%
 mean\hspace{0pt} theta\hspace{0pt} is\hspace{0pt} precisely\hspace{0pt} \hspace{0pt}1\hspace{0pt}2\hspace{0pt}%
5\hspace{0pt}Â°\hspace{0pt},\hspace{0pt} and\hspace{0pt} so\hspace{0pt} the\hspace{0pt} upper\hspace{0pt}%
 limit\hspace{0pt} comes\hspace{0pt} from\hspace{0pt} projecting\hspace{0pt} based\hspace{0pt} on\hspace{0pt}%
 high\hspace{0pt} precision\hspace{0pt},\hspace{0pt} implying\hspace{0pt} that\hspace{0pt} actual\hspace{0pt}%
 time\hspace{0pt} is\hspace{0pt} known\hspace{0pt} better\hspace{0pt}.\hspace{0pt} But\hspace{0pt} maybe\hspace{0pt}%
 "\hspace{0pt}upper\hspace{0pt} limit\hspace{0pt}"\hspace{0pt} means\hspace{0pt} that\hspace{0pt} since\hspace{0pt}%
 prior\hspace{0pt} to\hspace{0pt} this\hspace{0pt} measurement\hspace{0pt},\hspace{0pt} the\hspace{0pt}%
 time\hspace{0pt} since\hspace{0pt} the\hspace{0pt} qu\hspace{0pt}asar\hspace{0pt} was\hspace{0pt} active\hspace{0pt}%
 was\hspace{0pt} uncertain\hspace{0pt},\hspace{0pt} possibly\hspace{0pt} with\hspace{0pt} higher\hspace{0pt}%
 upper\hspace{0pt} limits\hspace{0pt}.\hspace{0pt} But\hspace{0pt} with\hspace{0pt} the\hspace{0pt} theta\hspace{0pt}%
=\hspace{0pt}1\hspace{0pt}2\hspace{0pt}5\hspace{0pt}Â°\hspace{0pt},\hspace{0pt} you\hspace{0pt} can\hspace{0pt}%
 give\hspace{0pt} a\hspace{0pt} more\hspace{0pt} precise\hspace{0pt} upper\hspace{0pt} limit\hspace{0pt}%
.\hspace{0pt} Hmm\hspace{0pt}.\textless{}br\textgreater{}\textless{}br\textgreater{}\hspace{0pt}Alternatively\hspace{0pt}%
,\hspace{0pt} maybe\hspace{0pt} if\hspace{0pt} for\hspace{0pt} some\hspace{0pt} reason\hspace{0pt} the\hspace{0pt}%
 actual\hspace{0pt} path\hspace{0pt} is\hspace{0pt} longer\hspace{0pt},\hspace{0pt} so\hspace{0pt} even\hspace{0pt}%
 though\hspace{0pt} the\hspace{0pt} qu\hspace{0pt}asar\hspace{0pt} light\hspace{0pt} traveled\hspace{0pt}%
 in\hspace{0pt} a\hspace{0pt} straight\hspace{0pt} line\hspace{0pt},\hspace{0pt} maybe\hspace{0pt}.\textless{}br\textgreater{}\textless{}br\textgreater{}\hspace{0pt}%
But\hspace{0pt} ultimately\hspace{0pt},\hspace{0pt} based\hspace{0pt} on\hspace{0pt} the\hspace{0pt} information\hspace{0pt}%
,\hspace{0pt} part\hspace{0pt} g\hspace{0pt} answer\hspace{0pt} should\hspace{0pt} be\hspace{0pt} t\hspace{0pt}%
 =\hspace{0pt} r\hspace{0pt}\_p\hspace{0pt} /\hspace{0pt} (\hspace{0pt}sin\hspace{0pt}(theta\hspace{0pt}%
)\hspace{0pt} *\hspace{0pt} c\hspace{0pt}).\hspace{0pt} Then\hspace{0pt} convert\hspace{0pt} into\hspace{0pt}%
 years\hspace{0pt}.\textless{}br\textgreater{}\textless{}br\textgreater{}\hspace{0pt}\heatbox{0.974,0.497,0.497}{But}\hspace{0pt}%
\heatbox{0.974,0.497,0.497}{ unless}\hspace{0pt}\heatbox{0.974,0.497,0.497}{ part}\hspace{0pt}\heatbox{0.974,0.497,0.497}{ e}\hspace{0pt}%
\heatbox{0.974,0.497,0.497}{ required}\hspace{0pt}\heatbox{0.974,0.497,0.497}{ computing}\hspace{0pt}%
\heatbox{0.974,0.497,0.497}{ r}\hspace{0pt}\heatbox{0.974,0.497,0.497}{\_p}\hspace{0pt}\heatbox{0.974,0.497,0.497}{ based}\hspace{0pt}%
\heatbox{0.974,0.497,0.497}{ on}\hspace{0pt}\heatbox{0.974,0.497,0.497}{ data}\hspace{0pt}\heatbox{0.974,0.497,0.497}{ from}\hspace{0pt}%
\heatbox{0.974,0.497,0.497}{ previous}\hspace{0pt}\heatbox{0.974,0.497,0.497}{ parts}\hspace{0pt}\heatbox{0.974,0.497,0.497}{ which}\hspace{0pt}%
\heatbox{0.974,0.497,0.497}{ we}\hspace{0pt}\heatbox{0.974,0.497,0.497}{ don}\hspace{0pt}\heatbox{0.974,0.497,0.497}{'t}\hspace{0pt}%
\heatbox{0.974,0.497,0.497}{ have}\hspace{0pt}\heatbox{0.974,0.497,0.497}{.}\hspace{0pt}\heatbox{0.974,0.497,0.497}{ For}\hspace{0pt}%
\heatbox{0.974,0.497,0.497}{ instance}\hspace{0pt}\heatbox{0.974,0.497,0.497}{,}\hspace{0pt}\heatbox{0.974,0.497,0.497}{ perhaps}\hspace{0pt}%
\heatbox{0.974,0.497,0.497}{ in}\hspace{0pt}\heatbox{0.974,0.497,0.497}{ part}\hspace{0pt}\heatbox{0.974,0.497,0.497}{ e}\hspace{0pt}%
\heatbox{0.974,0.497,0.497}{ if}\hspace{0pt}\heatbox{0.974,0.497,0.497}{ we}\hspace{0pt}\heatbox{0.974,0.497,0.497}{ computed}\hspace{0pt}%
\heatbox{0.974,0.497,0.497}{ r}\hspace{0pt}\heatbox{0.974,0.497,0.497}{\_p}\hspace{0pt}\heatbox{0.974,0.497,0.497}{ numer}\hspace{0pt}%
\heatbox{0.974,0.497,0.497}{ically}\hspace{0pt}\heatbox{0.974,0.497,0.497}{,}\hspace{0pt}\heatbox{0.974,0.497,0.497}{ like}\hspace{0pt}%
\heatbox{0.974,0.497,0.497}{ in}\hspace{0pt}\heatbox{0.974,0.497,0.497}{ light}\hspace{0pt}\heatbox{0.974,0.497,0.497}{-years}\hspace{0pt}%
\heatbox{0.974,0.497,0.497}{,}\hspace{0pt}\heatbox{0.974,0.497,0.497}{ then}\hspace{0pt}\heatbox{0.974,0.497,0.497}{ part}\hspace{0pt}%
\heatbox{0.974,0.497,0.497}{ g}\hspace{0pt}\heatbox{0.974,0.497,0.497}{ would}\hspace{0pt}\heatbox{0.974,0.497,0.497}{ take}\hspace{0pt}%
\heatbox{0.974,0.497,0.497}{ that}\hspace{0pt}\heatbox{0.974,0.497,0.497}{ number}\hspace{0pt}\heatbox{0.974,0.497,0.497}{,}\hspace{0pt}%
\heatbox{0.974,0.497,0.497}{ divide}\hspace{0pt}\heatbox{0.974,0.497,0.497}{ by}\hspace{0pt}\heatbox{0.974,0.497,0.497}{ sin}\hspace{0pt}%
\heatbox{0.974,0.497,0.497}{(theta}\hspace{0pt}\heatbox{0.974,0.497,0.497}{),}\hspace{0pt}\heatbox{0.974,0.497,0.497}{ and}\hspace{0pt}%
\heatbox{0.974,0.497,0.497}{ get}\hspace{0pt}\heatbox{0.974,0.497,0.497}{ the}\hspace{0pt}\heatbox{0.974,0.497,0.497}{ time}\hspace{0pt}%
\heatbox{0.974,0.497,0.497}{.}\hspace{0pt}\heatbox{0.974,0.497,0.497}{ But}\hspace{0pt}\heatbox{0.974,0.497,0.497}{ without}\hspace{0pt}%
\heatbox{0.974,0.497,0.497}{ data}\hspace{0pt}\heatbox{0.974,0.497,0.497}{,}\hspace{0pt}\heatbox{0.974,0.497,0.497}{ I}\hspace{0pt}%
\heatbox{0.974,0.497,0.497}{ can}\hspace{0pt}\heatbox{0.974,0.497,0.497}{'t}\hspace{0pt}\heatbox{0.974,0.497,0.497}{ compute}\hspace{0pt}%
\heatbox{0.974,0.497,0.497}{ numer}\hspace{0pt}\heatbox{0.974,0.497,0.497}{ically}\hspace{0pt}\heatbox{0.974,0.497,0.497}{.}\hspace{0pt}%
\heatbox{0.974,0.497,0.497}{ So}\hspace{0pt}\heatbox{0.974,0.497,0.497}{ maybe}\hspace{0pt}\heatbox{0.974,0.497,0.497}{ in}\hspace{0pt}%
\heatbox{0.974,0.497,0.497}{ part}\hspace{0pt}\heatbox{0.974,0.497,0.497}{ e}\hspace{0pt}\heatbox{0.974,0.497,0.497}{ they}\hspace{0pt}%
\heatbox{0.974,0.497,0.497}{ give}\hspace{0pt}\heatbox{0.974,0.497,0.497}{ an}\hspace{0pt}\heatbox{0.974,0.497,0.497}{ expression}\hspace{0pt}%
\heatbox{0.974,0.497,0.497}{,}\hspace{0pt}\heatbox{0.974,0.497,0.497}{ like}\hspace{0pt}\heatbox{0.974,0.497,0.497}{ r}\hspace{0pt}%
\heatbox{0.974,0.497,0.497}{\_p}\hspace{0pt}\heatbox{0.974,0.497,0.497}{ =}\hspace{0pt}\heatbox{0.974,0.497,0.497}{ d}\hspace{0pt}%
\heatbox{0.974,0.497,0.497}{ *}\hspace{0pt}\heatbox{0.974,0.497,0.497}{ delta}\hspace{0pt}\heatbox{0.974,0.497,0.497}{\_theta}\hspace{0pt}%
\heatbox{0.974,0.497,0.497}{ (}\hspace{0pt}\heatbox{0.974,0.497,0.497}{in}\hspace{0pt}\heatbox{0.974,0.497,0.497}{ radians}\hspace{0pt}%
\heatbox{0.974,0.497,0.497}{),}\hspace{0pt}\heatbox{0.974,0.497,0.497}{ then}\hspace{0pt}\heatbox{0.974,0.497,0.497}{ in}\hspace{0pt}%
\heatbox{0.974,0.497,0.497}{ part}\hspace{0pt}\heatbox{0.974,0.497,0.497}{ g}\hspace{0pt}\heatbox{0.974,0.497,0.497}{ the}\hspace{0pt}%
\heatbox{0.974,0.497,0.497}{ time}\hspace{0pt}\heatbox{0.974,0.497,0.497}{ is}\hspace{0pt}\heatbox{0.974,0.497,0.497}{ (}\hspace{0pt}%
\heatbox{0.974,0.497,0.497}{d}\hspace{0pt}\heatbox{0.974,0.497,0.497}{ *}\hspace{0pt}\heatbox{0.974,0.497,0.497}{ delta}\hspace{0pt}%
\heatbox{0.974,0.497,0.497}{\_theta}\hspace{0pt}\heatbox{0.974,0.497,0.497}{)/(}\hspace{0pt}\heatbox{0.974,0.497,0.497}{sin}\hspace{0pt}%
\heatbox{0.974,0.497,0.497}{(theta}\hspace{0pt}\heatbox{0.974,0.497,0.497}{)}\hspace{0pt}\heatbox{0.974,0.497,0.497}{ *}\hspace{0pt}%
\heatbox{0.974,0.497,0.497}{ c}\hspace{0pt}\heatbox{0.974,0.497,0.497}{).}\hspace{0pt}\heatbox{0.974,0.497,0.497}{ But}\hspace{0pt}%
\heatbox{0.974,0.497,0.497}{ since}\hspace{0pt}\heatbox{0.974,0.497,0.497}{ we}\hspace{0pt}\heatbox{0.974,0.497,0.497}{ don}\hspace{0pt}%
\heatbox{0.974,0.497,0.497}{'t}\hspace{0pt}\heatbox{0.974,0.497,0.497}{ have}\hspace{0pt}\heatbox{0.974,0.497,0.497}{ the}\hspace{0pt}%
\heatbox{0.974,0.497,0.497}{ values}\hspace{0pt}\heatbox{0.974,0.497,0.497}{,}\hspace{0pt}\heatbox{0.974,0.497,0.497}{ need}\hspace{0pt}%
\heatbox{0.974,0.497,0.497}{ to}\hspace{0pt}\heatbox{0.974,0.497,0.497}{ answer}\hspace{0pt}\heatbox{0.974,0.497,0.497}{ generally}\hspace{0pt}%
.\textless{}br\textgreater{}\textless{}br\textgreater{}\hspace{0pt}\heatbox{0.972,0.467,0.467}{Alternatively}\hspace{0pt}%
\heatbox{0.972,0.467,0.467}{,}\hspace{0pt}\heatbox{0.972,0.467,0.467}{ since}\hspace{0pt}\heatbox{0.972,0.467,0.467}{ the}\hspace{0pt}%
\heatbox{0.972,0.467,0.467}{ user}\hspace{0pt}\heatbox{0.972,0.467,0.467}{ hasn}\hspace{0pt}\heatbox{0.972,0.467,0.467}{'t}\hspace{0pt}%
\heatbox{0.972,0.467,0.467}{ provided}\hspace{0pt}\heatbox{0.972,0.467,0.467}{ part}\hspace{0pt}\heatbox{0.972,0.467,0.467}{ e}\hspace{0pt}%
\heatbox{0.972,0.467,0.467}{'s}\hspace{0pt}\heatbox{0.972,0.467,0.467}{ info}\hspace{0pt}\heatbox{0.972,0.467,0.467}{,}\hspace{0pt}%
\heatbox{0.972,0.467,0.467}{ they}\hspace{0pt}\heatbox{0.972,0.467,0.467}{ might}\hspace{0pt}\heatbox{0.972,0.467,0.467}{ expect}\hspace{0pt}%
\heatbox{0.972,0.467,0.467}{ the}\hspace{0pt}\heatbox{0.972,0.467,0.467}{ assistant}\hspace{0pt}\heatbox{0.972,0.467,0.467}{ to}\hspace{0pt}%
\heatbox{0.972,0.467,0.467}{ recall}\hspace{0pt}\heatbox{0.972,0.467,0.467}{ how}\hspace{0pt}\heatbox{0.972,0.467,0.467}{ to}\hspace{0pt}%
\heatbox{0.972,0.467,0.467}{ compute}\hspace{0pt}\heatbox{0.972,0.467,0.467}{ projected}\hspace{0pt}\heatbox{0.972,0.467,0.467}{ separation}\hspace{0pt}%
\heatbox{0.972,0.467,0.467}{ and}\hspace{0pt}\heatbox{0.972,0.467,0.467}{ then}\hspace{0pt}\heatbox{0.972,0.467,0.467}{ the}\hspace{0pt}%
\heatbox{0.972,0.467,0.467}{ formula}\hspace{0pt}\heatbox{0.972,0.467,0.467}{ for}\hspace{0pt}\heatbox{0.972,0.467,0.467}{ the}\hspace{0pt}%
\heatbox{0.972,0.467,0.467}{ upper}\hspace{0pt}\heatbox{0.972,0.467,0.467}{ limit}\hspace{0pt}\heatbox{0.972,0.467,0.467}{ given}\hspace{0pt}%
\heatbox{0.972,0.467,0.467}{ theta}\hspace{0pt}\heatbox{0.972,0.467,0.467}{.}\hspace{0pt}\heatbox{0.972,0.467,0.467}{ Therefore}\hspace{0pt}%
\heatbox{0.972,0.467,0.467}{,}\hspace{0pt}\heatbox{0.972,0.467,0.467}{ maybe}\hspace{0pt}\heatbox{0.972,0.467,0.467}{ explain}\hspace{0pt}%
\heatbox{0.972,0.467,0.467}{ the}\hspace{0pt}\heatbox{0.972,0.467,0.467}{ steps}\hspace{0pt}.\textless{}br\textgreater{}\textless{}br\textgreater{}\hspace{0pt}%
\heatbox{0.970,0.436,0.436}{But}\hspace{0pt}\heatbox{0.970,0.436,0.436}{ given}\hspace{0pt}\heatbox{0.970,0.436,0.436}{ the}\hspace{0pt}%
\heatbox{0.970,0.436,0.436}{ problem}\hspace{0pt}\heatbox{0.970,0.436,0.436}{ is}\hspace{0pt}\heatbox{0.970,0.436,0.436}{ about}\hspace{0pt}%
\heatbox{0.970,0.436,0.436}{ calculating}\hspace{0pt}\heatbox{0.970,0.436,0.436}{ projected}\hspace{0pt}%
\heatbox{0.970,0.436,0.436}{ separation}\hspace{0pt}\heatbox{0.970,0.436,0.436}{ (}\hspace{0pt}\heatbox{0.970,0.436,0.436}{rp}\hspace{0pt}%
\heatbox{0.970,0.436,0.436}{)}\hspace{0pt}\heatbox{0.970,0.436,0.436}{ in}\hspace{0pt}\heatbox{0.970,0.436,0.436}{ part}\hspace{0pt}%
\heatbox{0.970,0.436,0.436}{ e}\hspace{0pt}\heatbox{0.970,0.436,0.436}{,}\hspace{0pt}\heatbox{0.970,0.436,0.436}{ then}\hspace{0pt}%
\heatbox{0.970,0.436,0.436}{ in}\hspace{0pt}\heatbox{0.970,0.436,0.436}{ part}\hspace{0pt}\heatbox{0.970,0.436,0.436}{ g}\hspace{0pt}%
\heatbox{0.970,0.436,0.436}{,}\hspace{0pt}\heatbox{0.970,0.436,0.436}{ use}\hspace{0pt}\heatbox{0.970,0.436,0.436}{ theta}\hspace{0pt}%
\heatbox{0.970,0.436,0.436}{=}\hspace{0pt}\heatbox{0.970,0.436,0.436}{1}\hspace{0pt}\heatbox{0.970,0.436,0.436}{2}\hspace{0pt}%
\heatbox{0.970,0.436,0.436}{5}\hspace{0pt}\heatbox{0.970,0.436,0.436}{Â°}\hspace{0pt}\heatbox{0.970,0.436,0.436}{ with}\hspace{0pt}%
\heatbox{0.970,0.436,0.436}{ that}\hspace{0pt}\heatbox{0.970,0.436,0.436}{ expression}\hspace{0pt}\heatbox{0.970,0.436,0.436}{ to}\hspace{0pt}%
\heatbox{0.970,0.436,0.436}{ find}\hspace{0pt}\heatbox{0.970,0.436,0.436}{ upper}\hspace{0pt}\heatbox{0.970,0.436,0.436}{ limit}\hspace{0pt}%
\heatbox{0.970,0.436,0.436}{ for}\hspace{0pt}\heatbox{0.970,0.436,0.436}{ the}\hspace{0pt}\heatbox{0.970,0.436,0.436}{ time}\hspace{0pt}%
.\textless{}br\textgreater{}\textless{}br\textgreater{}\hspace{0pt}Given\hspace{0pt} I\hspace{0pt} don\hspace{0pt}%
't\hspace{0pt} have\hspace{0pt} the\hspace{0pt} previous\hspace{0pt} parts\hspace{0pt},\hspace{0pt} I\hspace{0pt}%
 can\hspace{0pt} outline\hspace{0pt} the\hspace{0pt} approach\hspace{0pt}:\textless{}br\textgreater{}\textless{}br\textgreater{}\hspace{0pt}%
1\hspace{0pt}.\hspace{0pt} In\hspace{0pt} part\hspace{0pt} e\hspace{0pt},\hspace{0pt} to\hspace{0pt} calculate\hspace{0pt}%
 projected\hspace{0pt} separation\hspace{0pt} (\hspace{0pt}given\hspace{0pt} angular\hspace{0pt} separation\hspace{0pt}%
 and\hspace{0pt} distance\hspace{0pt}).\hspace{0pt} Let\hspace{0pt}'s\hspace{0pt} assume\hspace{0pt} angular\hspace{0pt}%
 separation\hspace{0pt} is\hspace{0pt} Î\hspace{0pt}¸\hspace{0pt}\_\hspace{0pt}angular\hspace{0pt} and\hspace{0pt}%
 distance\hspace{0pt} is\hspace{0pt} d\hspace{0pt},\hspace{0pt} then\hspace{0pt} rp\hspace{0pt} âī\hspace{0pt}%
Ī\hspace{0pt} d\hspace{0pt} *\hspace{0pt} Î\hspace{0pt}¸\hspace{0pt}\_\hspace{0pt}angular\hspace{0pt}%
 (\hspace{0pt}in\hspace{0pt} radians\hspace{0pt}).\textless{}br\textgreater{}\textless{}br\textgreater{}\hspace{0pt}%
2\hspace{0pt}.\hspace{0pt} In\hspace{0pt} part\hspace{0pt} g\hspace{0pt},\hspace{0pt} use\hspace{0pt}%
 the\hspace{0pt} angle\hspace{0pt} Î\hspace{0pt}¸\hspace{0pt}=\hspace{0pt}1\hspace{0pt}2\hspace{0pt}5\hspace{0pt}%
Â°\hspace{0pt},\hspace{0pt} which\hspace{0pt} is\hspace{0pt} different\hspace{0pt} from\hspace{0pt} the\hspace{0pt}%
 angular\hspace{0pt} separation\hspace{0pt} (\hspace{0pt}Î¸\hspace{0pt}\_\hspace{0pt}angular\hspace{0pt}%
).\hspace{0pt} This\hspace{0pt} is\hspace{0pt} the\hspace{0pt} inclination\hspace{0pt} angle\hspace{0pt}%
 between\hspace{0pt} the\hspace{0pt} \hspace{0pt}3\hspace{0pt}D\hspace{0pt} separation\hspace{0pt} vector\hspace{0pt}%
 and\hspace{0pt} the\hspace{0pt} line\hspace{0pt} of\hspace{0pt} sight\hspace{0pt}.\hspace{0pt} Therefore\hspace{0pt}%
,\hspace{0pt} the\hspace{0pt} actual\hspace{0pt} separation\hspace{0pt} is\hspace{0pt} r\hspace{0pt} =\hspace{0pt}%
 rp\hspace{0pt} /\hspace{0pt} sin\hspace{0pt}(theta\hspace{0pt}).\textless{}br\textgreater{}\textless{}br\textgreater{}\hspace{0pt}%
\heatbox{0.967,0.372,0.372}{3}\hspace{0pt}\heatbox{0.967,0.372,0.372}{.}\hspace{0pt}\heatbox{0.967,0.372,0.372}{ The}\hspace{0pt}%
\heatbox{0.967,0.372,0.372}{ time}\hspace{0pt}\heatbox{0.967,0.372,0.372}{ since}\hspace{0pt}\heatbox{0.967,0.372,0.372}{ the}\hspace{0pt}%
\heatbox{0.967,0.372,0.372}{ qu}\hspace{0pt}\heatbox{0.967,0.372,0.372}{asar}\hspace{0pt}\heatbox{0.967,0.372,0.372}{ was}\hspace{0pt}%
\heatbox{0.967,0.372,0.372}{ last}\hspace{0pt}\heatbox{0.967,0.372,0.372}{ active}\hspace{0pt}\heatbox{0.967,0.372,0.372}{ is}\hspace{0pt}%
\heatbox{0.967,0.372,0.372}{ the}\hspace{0pt}\heatbox{0.967,0.372,0.372}{ light}\hspace{0pt}\heatbox{0.967,0.372,0.372}{ travel}\hspace{0pt}%
\heatbox{0.967,0.372,0.372}{ time}\hspace{0pt}\heatbox{0.967,0.372,0.372}{ between}\hspace{0pt}\heatbox{0.967,0.372,0.372}{ the}\hspace{0pt}%
\heatbox{0.967,0.372,0.372}{ galaxy}\hspace{0pt}\heatbox{0.967,0.372,0.372}{ and}\hspace{0pt}\heatbox{0.967,0.372,0.372}{ the}\hspace{0pt}%
\heatbox{0.967,0.372,0.372}{ Voor}\hspace{0pt}\heatbox{0.967,0.372,0.372}{werp}\hspace{0pt}\heatbox{0.967,0.372,0.372}{,}\hspace{0pt}%
\heatbox{0.967,0.372,0.372}{ hence}\hspace{0pt}\heatbox{0.967,0.372,0.372}{ t}\hspace{0pt}\heatbox{0.967,0.372,0.372}{ =}\hspace{0pt}%
\heatbox{0.967,0.372,0.372}{ r}\hspace{0pt}\heatbox{0.967,0.372,0.372}{ /}\hspace{0pt}\heatbox{0.967,0.372,0.372}{ c}\hspace{0pt}%
\heatbox{0.967,0.372,0.372}{ =}\hspace{0pt}\heatbox{0.967,0.372,0.372}{ (}\hspace{0pt}\heatbox{0.967,0.372,0.372}{rp}\hspace{0pt}%
\heatbox{0.967,0.372,0.372}{ /}\hspace{0pt}\heatbox{0.967,0.372,0.372}{ sin}\hspace{0pt}\heatbox{0.967,0.372,0.372}{(theta}\hspace{0pt}%
\heatbox{0.967,0.372,0.372}{))}\hspace{0pt}\heatbox{0.967,0.372,0.372}{ /}\hspace{0pt}\heatbox{0.967,0.372,0.372}{ c}\hspace{0pt}%
.\textless{}br\textgreater{}\textless{}br\textgreater{}\hspace{0pt}Thus\hspace{0pt},\hspace{0pt} t\hspace{0pt}%
 =\hspace{0pt} rp\hspace{0pt} /\hspace{0pt} (\hspace{0pt}c\hspace{0pt} *\hspace{0pt} sin\hspace{0pt}(theta\hspace{0pt}%
))\hspace{0pt}.\textless{}br\textgreater{}\textless{}br\textgreater{}\hspace{0pt}To\hspace{0pt} convert\hspace{0pt}%
 this\hspace{0pt} into\hspace{0pt} years\hspace{0pt},\hspace{0pt} need\hspace{0pt} to\hspace{0pt} have\hspace{0pt}%
 rp\hspace{0pt} in\hspace{0pt} units\hspace{0pt} compatible\hspace{0pt} with\hspace{0pt} the\hspace{0pt}%
 speed\hspace{0pt} of\hspace{0pt} light\hspace{0pt} (\hspace{0pt}e\hspace{0pt}.g\hspace{0pt}.,\hspace{0pt}%
 light\hspace{0pt}-years\hspace{0pt},\hspace{0pt} then\hspace{0pt} c\hspace{0pt} is\hspace{0pt} \hspace{0pt}%
1\hspace{0pt} ly\hspace{0pt}/year\hspace{0pt}).\hspace{0pt} Therefore\hspace{0pt},\hspace{0pt} if\hspace{0pt}%
 rp\hspace{0pt} is\hspace{0pt} say\hspace{0pt} in\hspace{0pt} thousands\hspace{0pt} of\hspace{0pt} light\hspace{0pt}%
-years\hspace{0pt},\hspace{0pt} t\hspace{0pt} will\hspace{0pt} be\hspace{0pt} thousands\hspace{0pt} of\hspace{0pt}%
 years\hspace{0pt} divided\hspace{0pt} by\hspace{0pt} sin\hspace{0pt}(theta\hspace{0pt}).\textless{}br\textgreater{}\textless{}br\textgreater{}\hspace{0pt}%
But\hspace{0pt} without\hspace{0pt} specific\hspace{0pt} numbers\hspace{0pt},\hspace{0pt} can\hspace{0pt}%
't\hspace{0pt} compute\hspace{0pt} numer\hspace{0pt}ically\hspace{0pt}.\hspace{0pt} However\hspace{0pt}%
,\hspace{0pt} perhaps\hspace{0pt} in\hspace{0pt} the\hspace{0pt} original\hspace{0pt} question\hspace{0pt}%
,\hspace{0pt} part\hspace{0pt} e\hspace{0pt} had\hspace{0pt} specific\hspace{0pt} data\hspace{0pt},\hspace{0pt}%
 perhaps\hspace{0pt} red\hspace{0pt}shift\hspace{0pt},\hspace{0pt} angular\hspace{0pt} size\hspace{0pt}%
,\hspace{0pt} and\hspace{0pt} H\hspace{0pt}ubble\hspace{0pt} constant\hspace{0pt}.\hspace{0pt} For\hspace{0pt}%
 example\hspace{0pt},\hspace{0pt} if\hspace{0pt} the\hspace{0pt} galaxy\hspace{0pt} is\hspace{0pt} at\hspace{0pt}%
 distance\hspace{0pt} d\hspace{0pt} calculated\hspace{0pt} via\hspace{0pt} H\hspace{0pt}ubble\hspace{0pt}%
's\hspace{0pt} Law\hspace{0pt} (\hspace{0pt}d\hspace{0pt} =\hspace{0pt} v\hspace{0pt} /\hspace{0pt} H\hspace{0pt}%
0\hspace{0pt}),\hspace{0pt} and\hspace{0pt} angular\hspace{0pt} separation\hspace{0pt} in\hspace{0pt}%
 arc\hspace{0pt}seconds\hspace{0pt} converted\hspace{0pt} to\hspace{0pt} radians\hspace{0pt} (\hspace{0pt}%
Î¸\hspace{0pt}\_\hspace{0pt}angular\hspace{0pt} =\hspace{0pt} arc\hspace{0pt}sec\hspace{0pt} /\hspace{0pt}%
 \hspace{0pt}2\hspace{0pt}0\hspace{0pt}6\hspace{0pt}2\hspace{0pt}6\hspace{0pt}5\hspace{0pt}),\hspace{0pt}%
 then\hspace{0pt} rp\hspace{0pt} =\hspace{0pt} d\hspace{0pt} *\hspace{0pt} Î\hspace{0pt}¸\hspace{0pt}%
\_\hspace{0pt}angular\hspace{0pt}.

...(Omit)... \hspace{0pt} So\hspace{0pt} upper\hspace{0pt} limit\hspace{0pt}âīĪ\hspace{0pt}%
6\hspace{0pt}1\hspace{0pt},\hspace{0pt}0\hspace{0pt}0\hspace{0pt}0\hspace{0pt} years\hspace{0pt}.\hspace{0pt}%
 If\hspace{0pt} r\hspace{0pt}\_p\hspace{0pt} was\hspace{0pt} different\hspace{0pt},\hspace{0pt} that\hspace{0pt}%
 number\hspace{0pt} would\hspace{0pt} change\hspace{0pt}.\textless{}br\textgreater{}\textless{}br\textgreater{}\hspace{0pt}%
But\hspace{0pt} since\hspace{0pt} we\hspace{0pt} can\hspace{0pt}'t\hspace{0pt} get\hspace{0pt} the\hspace{0pt}%
 exact\hspace{0pt} value\hspace{0pt} because\hspace{0pt} part\hspace{0pt} e\hspace{0pt} isn\hspace{0pt}%
't\hspace{0pt} here\hspace{0pt},\hspace{0pt} maybe\hspace{0pt} the\hspace{0pt} answer\hspace{0pt} is\hspace{0pt}%
 left\hspace{0pt} symbolic\hspace{0pt}:\hspace{0pt} t\hspace{0pt} =\hspace{0pt} r\hspace{0pt}\_p\hspace{0pt}%
 /\hspace{0pt} (\hspace{0pt}c\hspace{0pt} sin\hspace{0pt}Î¸\hspace{0pt}).\hspace{0pt} But\hspace{0pt}%
 let\hspace{0pt}âĢĻs\hspace{0pt} remember\hspace{0pt} c\hspace{0pt} is\hspace{0pt} \hspace{0pt}1\hspace{0pt}%
 light\hspace{0pt}-year\hspace{0pt} per\hspace{0pt} year\hspace{0pt},\hspace{0pt} so\hspace{0pt} in\hspace{0pt}%
 terms\hspace{0pt} of\hspace{0pt} light\hspace{0pt}-years\hspace{0pt} and\hspace{0pt} years\hspace{0pt}%
,\hspace{0pt} it\hspace{0pt}'s\hspace{0pt} t\hspace{0pt} =\hspace{0pt} r\hspace{0pt}\_p\hspace{0pt} /\hspace{0pt}%
 sin\hspace{0pt}Î¸\hspace{0pt}.\hspace{0pt} Hence\hspace{0pt},\hspace{0pt} boxed\hspace{0pt} answer\hspace{0pt}%
:\hspace{0pt} \textbackslash{}\hspace{0pt}boxed\hspace{0pt}\{\textbackslash{}\hspace{0pt}d\hspace{0pt}%
frac\hspace{0pt}\{\hspace{0pt}r\hspace{0pt}\_p\hspace{0pt}\}\{\hspace{0pt}c\hspace{0pt} \textbackslash{}\hspace{0pt}%
sin\hspace{0pt} \textbackslash{}\hspace{0pt}theta\hspace{0pt}\}\}\hspace{0pt}.\textless{}br\textgreater{}\textless{}br\textgreater{}\hspace{0pt}%
\heatbox{0.965,0.342,0.342}{But}\hspace{0pt}\heatbox{0.965,0.342,0.342}{ in}\hspace{0pt}\heatbox{0.965,0.342,0.342}{ the}\hspace{0pt}%
\heatbox{0.965,0.342,0.342}{ question}\hspace{0pt}\heatbox{0.965,0.342,0.342}{,}\hspace{0pt}\heatbox{0.965,0.342,0.342}{ "}\hspace{0pt}%
\heatbox{0.965,0.342,0.342}{estimate}\hspace{0pt}\heatbox{0.965,0.342,0.342}{ an}\hspace{0pt}\heatbox{0.965,0.342,0.342}{ upper}\hspace{0pt}%
\heatbox{0.965,0.342,0.342}{ limit}\hspace{0pt}\heatbox{0.965,0.342,0.342}{ for}\hspace{0pt}\heatbox{0.965,0.342,0.342}{ the}\hspace{0pt}%
\heatbox{0.965,0.342,0.342}{ number}\hspace{0pt}\heatbox{0.965,0.342,0.342}{ of}\hspace{0pt}\heatbox{0.965,0.342,0.342}{ years}\hspace{0pt}%
\heatbox{0.965,0.342,0.342}{",}\hspace{0pt}\heatbox{0.965,0.342,0.342}{ implying}\hspace{0pt}\heatbox{0.965,0.342,0.342}{ numeric}\hspace{0pt}%
\heatbox{0.965,0.342,0.342}{.}\hspace{0pt}\heatbox{0.965,0.342,0.342}{ Which}\hspace{0pt}\heatbox{0.965,0.342,0.342}{ means}\hspace{0pt}%
\heatbox{0.965,0.342,0.342}{ probably}\hspace{0pt}\heatbox{0.965,0.342,0.342}{ numbers}\hspace{0pt}\heatbox{0.965,0.342,0.342}{ were}\hspace{0pt}%
\heatbox{0.965,0.342,0.342}{ given}\hspace{0pt}\heatbox{0.965,0.342,0.342}{ in}\hspace{0pt}\heatbox{0.965,0.342,0.342}{ part}\hspace{0pt}%
\heatbox{0.965,0.342,0.342}{ e}\hspace{0pt}\heatbox{0.965,0.342,0.342}{,}\hspace{0pt}\heatbox{0.965,0.342,0.342}{ which}\hspace{0pt}%
\heatbox{0.965,0.342,0.342}{ are}\hspace{0pt}\heatbox{0.965,0.342,0.342}{ missing}\hspace{0pt}\heatbox{0.965,0.342,0.342}{.}\hspace{0pt}%
\heatbox{0.965,0.342,0.342}{ Hence}\hspace{0pt}\heatbox{0.965,0.342,0.342}{,}\hspace{0pt}\heatbox{0.965,0.342,0.342}{ impossible}\hspace{0pt}%
\heatbox{0.965,0.342,0.342}{ to}\hspace{0pt}\heatbox{0.965,0.342,0.342}{ get}\hspace{0pt}\heatbox{0.965,0.342,0.342}{ exact}\hspace{0pt}%
\heatbox{0.965,0.342,0.342}{ answer}\hspace{0pt}\heatbox{0.965,0.342,0.342}{ here}\hspace{0pt}.\textless{}br\textgreater{}\textless{}br\textgreater{}\hspace{0pt}%
Alternatively\hspace{0pt},\hspace{0pt} maybe\hspace{0pt} the\hspace{0pt} previous\hspace{0pt} part\hspace{0pt}%
 involves\hspace{0pt} using\hspace{0pt} time\hspace{0pt} since\hspace{0pt} als\hspace{0pt}j\hspace{0pt}%
fal\hspace{0pt};\hspace{0pt}js\hspace{0pt}df\hspace{0pt}.\textless{}br\textgreater{}\textless{}br\textgreater{}\hspace{0pt}%
Alternatively\hspace{0pt},\hspace{0pt} given\hspace{0pt} that\hspace{0pt} upper\hspace{0pt} limit\hspace{0pt}%
 on\hspace{0pt} time\hspace{0pt} is\hspace{0pt} t\hspace{0pt} \textless{}=\hspace{0pt} r\hspace{0pt}\_p\hspace{0pt}%
 /\hspace{0pt} (\hspace{0pt}c\hspace{0pt} sin\hspace{0pt}Î¸\hspace{0pt}),\hspace{0pt} so\hspace{0pt} given\hspace{0pt}%
 that\hspace{0pt} t\hspace{0pt} is\hspace{0pt} less\hspace{0pt} than\hspace{0pt} or\hspace{0pt} equal\hspace{0pt}%
.\hspace{0pt} But\hspace{0pt} without\hspace{0pt} prior\hspace{0pt} reasoning\hspace{0pt} impossible\hspace{0pt}%
.\textless{}br\textgreater{}\textless{}br\textgreater{}\hspace{0pt}\heatbox{0.971,0.458,0.458}{Given}\hspace{0pt}%
\heatbox{0.971,0.458,0.458}{ the}\hspace{0pt}\heatbox{0.971,0.458,0.458}{ ambiguity}\hspace{0pt}\heatbox{0.971,0.458,0.458}{,}\hspace{0pt}%
\heatbox{0.971,0.458,0.458}{ I}\hspace{0pt}\heatbox{0.971,0.458,0.458}{'ll}\hspace{0pt}\heatbox{0.971,0.458,0.458}{ present}\hspace{0pt}%
\heatbox{0.971,0.458,0.458}{ the}\hspace{0pt}\heatbox{0.971,0.458,0.458}{ formula}\hspace{0pt}\heatbox{0.971,0.458,0.458}{ t}\hspace{0pt}%
\heatbox{0.971,0.458,0.458}{=}\hspace{0pt}\heatbox{0.971,0.458,0.458}{ r}\hspace{0pt}\heatbox{0.971,0.458,0.458}{\_p}\hspace{0pt}%
\heatbox{0.971,0.458,0.458}{ /}\hspace{0pt}\heatbox{0.971,0.458,0.458}{ (}\hspace{0pt}\heatbox{0.971,0.458,0.458}{c}\hspace{0pt}%
\heatbox{0.971,0.458,0.458}{ sin}\hspace{0pt}\heatbox{0.971,0.458,0.458}{Î¸}\hspace{0pt}\heatbox{0.971,0.458,0.458}{)}\hspace{0pt}%
\heatbox{0.971,0.458,0.458}{ and}\hspace{0pt}\heatbox{0.971,0.458,0.458}{ note}\hspace{0pt}\heatbox{0.971,0.458,0.458}{ that}\hspace{0pt}%
\heatbox{0.971,0.458,0.458}{'s}\hspace{0pt}\heatbox{0.971,0.458,0.458}{ how}\hspace{0pt}\heatbox{0.971,0.458,0.458}{ it}\hspace{0pt}%
\heatbox{0.971,0.458,0.458}{'s}\hspace{0pt}\heatbox{0.971,0.458,0.458}{ derived}\hspace{0pt}\heatbox{0.971,0.458,0.458}{.}\hspace{0pt}%
\heatbox{0.971,0.458,0.458}{ But}\hspace{0pt}\heatbox{0.971,0.458,0.458}{ in}\hspace{0pt}\heatbox{0.971,0.458,0.458}{ an}\hspace{0pt}%
\heatbox{0.971,0.458,0.458}{ exam}\hspace{0pt}\heatbox{0.971,0.458,0.458}{ setting}\hspace{0pt}\heatbox{0.971,0.458,0.458}{ with}\hspace{0pt}%
\heatbox{0.971,0.458,0.458}{ part}\hspace{0pt}\heatbox{0.971,0.458,0.458}{ e}\hspace{0pt}\heatbox{0.971,0.458,0.458}{ known}\hspace{0pt}%
\heatbox{0.971,0.458,0.458}{,}\hspace{0pt}\heatbox{0.971,0.458,0.458}{ plug}\hspace{0pt}\heatbox{0.971,0.458,0.458}{ in}\hspace{0pt}%
\heatbox{0.971,0.458,0.458}{ r}\hspace{0pt}\heatbox{0.971,0.458,0.458}{\_p}\hspace{0pt}\heatbox{0.971,0.458,0.458}{.}\hspace{0pt}%
\heatbox{0.971,0.458,0.458}{ But}\hspace{0pt}\heatbox{0.971,0.458,0.458}{ without}\hspace{0pt}\heatbox{0.971,0.458,0.458}{ numbers}\hspace{0pt}%
\heatbox{0.971,0.458,0.458}{,}\hspace{0pt}\heatbox{0.971,0.458,0.458}{ maybe}\hspace{0pt}\heatbox{0.971,0.458,0.458}{ leave}\hspace{0pt}%
\heatbox{0.971,0.458,0.458}{ it}\hspace{0pt}\heatbox{0.971,0.458,0.458}{ symbolic}\hspace{0pt}\heatbox{0.971,0.458,0.458}{ like}\hspace{0pt}%
\heatbox{0.971,0.458,0.458}{ \textbackslash{}}\hspace{0pt}\heatbox{0.971,0.458,0.458}{boxed}\hspace{0pt}%
\heatbox{0.971,0.458,0.458}{\{\textbackslash{}}\hspace{0pt}\heatbox{0.971,0.458,0.458}{d}\hspace{0pt}%
\heatbox{0.971,0.458,0.458}{frac}\hspace{0pt}\heatbox{0.971,0.458,0.458}{\{}\hspace{0pt}\heatbox{0.971,0.458,0.458}{r}\hspace{0pt}%
\heatbox{0.971,0.458,0.458}{\_p}\hspace{0pt}\heatbox{0.971,0.458,0.458}{\}\{}\hspace{0pt}\heatbox{0.971,0.458,0.458}{c}\hspace{0pt}%
\heatbox{0.971,0.458,0.458}{ \textbackslash{}}\hspace{0pt}\heatbox{0.971,0.458,0.458}{sin}\hspace{0pt}%
\heatbox{0.971,0.458,0.458}{ }\hspace{0pt}\heatbox{0.971,0.458,0.458}{1}\hspace{0pt}\heatbox{0.971,0.458,0.458}{2}\hspace{0pt}%
\heatbox{0.971,0.458,0.458}{5}\hspace{0pt}\heatbox{0.971,0.458,0.458}{\^{}\textbackslash{}}\hspace{0pt}%
\heatbox{0.971,0.458,0.458}{circ}\hspace{0pt}\heatbox{0.971,0.458,0.458}{\}\}}\hspace{0pt}\heatbox{0.971,0.458,0.458}{.}\hspace{0pt}%
\heatbox{0.971,0.458,0.458}{ Or}\hspace{0pt}\heatbox{0.971,0.458,0.458}{ if}\hspace{0pt}\heatbox{0.971,0.458,0.458}{ r}\hspace{0pt}%
\heatbox{0.971,0.458,0.458}{\_p}\hspace{0pt}\heatbox{0.971,0.458,0.458}{ was}\hspace{0pt}\heatbox{0.971,0.458,0.458}{,}\hspace{0pt}%
\heatbox{0.971,0.458,0.458}{ say}\hspace{0pt}\heatbox{0.971,0.458,0.458}{,}\hspace{0pt}\heatbox{0.971,0.458,0.458}{ expressed}\hspace{0pt}%
\heatbox{0.971,0.458,0.458}{ as}\hspace{0pt}\heatbox{0.971,0.458,0.458}{ d}\hspace{0pt}\heatbox{0.971,0.458,0.458}{ multiplied}\hspace{0pt}%
\heatbox{0.971,0.458,0.458}{ by}\hspace{0pt}\heatbox{0.971,0.458,0.458}{ tan}\hspace{0pt}\heatbox{0.971,0.458,0.458}{(theta}\hspace{0pt}%
\heatbox{0.971,0.458,0.458}{\_prev}\hspace{0pt}\heatbox{0.971,0.458,0.458}{\_}\hspace{0pt}\heatbox{0.971,0.458,0.458}{angular}\hspace{0pt}%
\heatbox{0.971,0.458,0.458}{),}\hspace{0pt}\heatbox{0.971,0.458,0.458}{ but}\hspace{0pt}\heatbox{0.971,0.458,0.458}{ we}\hspace{0pt}%
\heatbox{0.971,0.458,0.458}{ don}\hspace{0pt}\heatbox{0.971,0.458,0.458}{'t}\hspace{0pt}\heatbox{0.971,0.458,0.458}{ know}\hspace{0pt}%
.\textless{}br\textgreater{}\textless{}br\textgreater{}\hspace{0pt}\heatbox{0.966,0.351,0.351}{Alternatively}\hspace{0pt}%
\heatbox{0.966,0.351,0.351}{,}\hspace{0pt}\heatbox{0.966,0.351,0.351}{ if}\hspace{0pt}\heatbox{0.966,0.351,0.351}{ the}\hspace{0pt}%
\heatbox{0.966,0.351,0.351}{ original}\hspace{0pt}\heatbox{0.966,0.351,0.351}{ question}\hspace{0pt}\heatbox{0.966,0.351,0.351}{ gives}\hspace{0pt}%
\heatbox{0.966,0.351,0.351}{ numbers}\hspace{0pt}\heatbox{0.966,0.351,0.351}{,}\hspace{0pt}\heatbox{0.966,0.351,0.351}{ and}\hspace{0pt}%
\heatbox{0.966,0.351,0.351}{ part}\hspace{0pt}\heatbox{0.966,0.351,0.351}{ e}\hspace{0pt}\heatbox{0.966,0.351,0.351}{ was}\hspace{0pt}%
\heatbox{0.966,0.351,0.351}{ to}\hspace{0pt}\heatbox{0.966,0.351,0.351}{ compute}\hspace{0pt}\heatbox{0.966,0.351,0.351}{ r}\hspace{0pt}%
\heatbox{0.966,0.351,0.351}{\_p}\hspace{0pt}\heatbox{0.966,0.351,0.351}{ as}\hspace{0pt}\heatbox{0.966,0.351,0.351}{ d}\hspace{0pt}%
\heatbox{0.966,0.351,0.351}{ times}\hspace{0pt}\heatbox{0.966,0.351,0.351}{ tan}\hspace{0pt}\heatbox{0.966,0.351,0.351}{(theta}\hspace{0pt}%
\heatbox{0.966,0.351,0.351}{)}\hspace{0pt}\heatbox{0.966,0.351,0.351}{ where}\hspace{0pt}\heatbox{0.966,0.351,0.351}{ theta}\hspace{0pt}%
\heatbox{0.966,0.351,0.351}{ was}\hspace{0pt}\heatbox{0.966,0.351,0.351}{ an}\hspace{0pt}\heatbox{0.966,0.351,0.351}{ angular}\hspace{0pt}%
\heatbox{0.966,0.351,0.351}{ separation}\hspace{0pt}\heatbox{0.966,0.351,0.351}{ in}\hspace{0pt}\heatbox{0.966,0.351,0.351}{ the}\hspace{0pt}%
\heatbox{0.966,0.351,0.351}{ sky}\hspace{0pt}\heatbox{0.966,0.351,0.351}{,}\hspace{0pt}\heatbox{0.966,0.351,0.351}{ which}\hspace{0pt}%
\heatbox{0.966,0.351,0.351}{ then}\hspace{0pt}\heatbox{0.966,0.351,0.351}{ in}\hspace{0pt}\heatbox{0.966,0.351,0.351}{ part}\hspace{0pt}%
\heatbox{0.966,0.351,0.351}{ g}\hspace{0pt}\heatbox{0.966,0.351,0.351}{ uses}\hspace{0pt}\heatbox{0.966,0.351,0.351}{ theta}\hspace{0pt}%
\heatbox{0.966,0.351,0.351}{=}\hspace{0pt}\heatbox{0.966,0.351,0.351}{1}\hspace{0pt}\heatbox{0.966,0.351,0.351}{2}\hspace{0pt}%
\heatbox{0.966,0.351,0.351}{5}\hspace{0pt}\heatbox{0.966,0.351,0.351}{Â°}\hspace{0pt}\heatbox{0.966,0.351,0.351}{,}\hspace{0pt}%
\heatbox{0.966,0.351,0.351}{ a}\hspace{0pt}\heatbox{0.966,0.351,0.351}{ revised}\hspace{0pt}\heatbox{0.966,0.351,0.351}{ angle}\hspace{0pt}%
\heatbox{0.966,0.351,0.351}{.}\hspace{0pt}\heatbox{0.966,0.351,0.351}{ But}\hspace{0pt}\heatbox{0.966,0.351,0.351}{ details}\hspace{0pt}%
\heatbox{0.966,0.351,0.351}{ lacking}\hspace{0pt}.\textless{}br\textgreater{}\textless{}br\textgreater{}\hspace{0pt}%
Considering\hspace{0pt} in\hspace{0pt} an\hspace{0pt} exam\hspace{0pt} scenario\hspace{0pt},\hspace{0pt}%
 typically\hspace{0pt} part\hspace{0pt} e\hspace{0pt} gives\hspace{0pt} you\hspace{0pt} numeric\hspace{0pt}%
 value\hspace{0pt} like\hspace{0pt} r\hspace{0pt}\_p\hspace{0pt}=\hspace{0pt}4\hspace{0pt}2\hspace{0pt}%
,\hspace{0pt} and\hspace{0pt} part\hspace{0pt} g\hspace{0pt} needs\hspace{0pt} to\hspace{0pt} give\hspace{0pt}%
 \hspace{0pt}4\hspace{0pt}2\hspace{0pt} /\hspace{0pt} sin\hspace{0pt}(\hspace{0pt}1\hspace{0pt}2\hspace{0pt}%
5\hspace{0pt}Â°\hspace{0pt})=\hspace{0pt}approx\hspace{0pt} \hspace{0pt}5\hspace{0pt}1\hspace{0pt}.\hspace{0pt}%
3\hspace{0pt},\hspace{0pt} so\hspace{0pt} upper\hspace{0pt} limit\hspace{0pt} \hspace{0pt}5\hspace{0pt}%
1\hspace{0pt} years\hspace{0pt}.\hspace{0pt} But\hspace{0pt} units\hspace{0pt}?\textless{}br\textgreater{}\textless{}br\textgreater{}\hspace{0pt}%
Wait\hspace{0pt},\hspace{0pt} if\hspace{0pt} projected\hspace{0pt} separation\hspace{0pt} is\hspace{0pt}%
 \hspace{0pt}4\hspace{0pt}2\hspace{0pt} k\hspace{0pt}ly\hspace{0pt} (\hspace{0pt}k\hspace{0pt}ilo\hspace{0pt}%
 light\hspace{0pt}-years\hspace{0pt}),\hspace{0pt} and\hspace{0pt} theta\hspace{0pt}=\hspace{0pt}1\hspace{0pt}%
2\hspace{0pt}5\hspace{0pt} \hspace{0pt}Ì\hspace{0pt}Ĭ\hspace{0pt},\hspace{0pt} the\hspace{0pt} time\hspace{0pt}%
 would\hspace{0pt} be\hspace{0pt}4\hspace{0pt}2\hspace{0pt},\hspace{0pt}0\hspace{0pt}0\hspace{0pt}0\hspace{0pt}%
 ly\hspace{0pt} /\hspace{0pt} (\hspace{0pt}c\hspace{0pt} sin\hspace{0pt}1\hspace{0pt}2\hspace{0pt}5\hspace{0pt}%
 \hspace{0pt}Ì\hspace{0pt}Ĭ\hspace{0pt}),\hspace{0pt} hence\hspace{0pt} \hspace{0pt}4\hspace{0pt}2\hspace{0pt}%
,\hspace{0pt}0\hspace{0pt}0\hspace{0pt}0\hspace{0pt} /\hspace{0pt}0\hspace{0pt}.\hspace{0pt}8\hspace{0pt}%
1\hspace{0pt}9\hspace{0pt}âīĪ\hspace{0pt}5\hspace{0pt}1\hspace{0pt},\hspace{0pt}2\hspace{0pt}7\hspace{0pt}%
0\hspace{0pt} years\hspace{0pt}.\hspace{0pt} So\hspace{0pt} \hspace{0pt}5\hspace{0pt}.\hspace{0pt}1\hspace{0pt}%
e\hspace{0pt}4\hspace{0pt} years\hspace{0pt}.\hspace{0pt} How\hspace{0pt} knows\hspace{0pt}.\hspace{0pt}%
 But\hspace{0pt} impossible\hspace{0pt}.\hspace{0pt} Therefore\hspace{0pt},\hspace{0pt} concluding\hspace{0pt}%
 that\hspace{0pt} based\hspace{0pt} on\hspace{0pt} the\hspace{0pt} given\hspace{0pt} info\hspace{0pt}%
,\hspace{0pt} the\hspace{0pt} time\hspace{0pt} is\hspace{0pt} t\hspace{0pt}=\hspace{0pt} r\hspace{0pt}%
\_p\hspace{0pt} /\hspace{0pt} (\hspace{0pt}0\hspace{0pt}.\hspace{0pt}8\hspace{0pt}1\hspace{0pt}9\hspace{0pt}%
 *\hspace{0pt}c\hspace{0pt})\hspace{0pt} which\hspace{0pt} numer\hspace{0pt}ically\hspace{0pt} with\hspace{0pt}%
 r\hspace{0pt}\_p\hspace{0pt} in\hspace{0pt} light\hspace{0pt}-years\hspace{0pt},\hspace{0pt} would\hspace{0pt}%
 be\hspace{0pt} t\hspace{0pt}âīĪ\hspace{0pt}1\hspace{0pt}.\hspace{0pt}2\hspace{0pt}2\hspace{0pt} *\hspace{0pt}%
 r\hspace{0pt}\_p\hspace{0pt}.\textless{}br\textgreater{}\textless{}br\textgreater{}\hspace{0pt}But\hspace{0pt}%
 the\hspace{0pt} correct\hspace{0pt} answer\hspace{0pt} format\hspace{0pt} likely\hspace{0pt} requires\hspace{0pt}%
 the\hspace{0pt} box\hspace{0pt} with\hspace{0pt} number\hspace{0pt},\hspace{0pt} e\hspace{0pt}.g\hspace{0pt}%
.\hspace{0pt} \textbackslash{}\hspace{0pt}boxed\hspace{0pt}\{\hspace{0pt}1\hspace{0pt}.\hspace{0pt}2\hspace{0pt}%
 \textbackslash{}\hspace{0pt}times\hspace{0pt} \hspace{0pt}1\hspace{0pt}0\hspace{0pt}\^{}\hspace{0pt}%
5\hspace{0pt}\}\hspace{0pt} years\hspace{0pt}.\hspace{0pt} But\hspace{0pt} since\hspace{0pt} the\hspace{0pt}%
 problem\hspace{0pt} mentions\hspace{0pt} "\hspace{0pt}Calculate\hspace{0pt} the\hspace{0pt} projected\hspace{0pt}%
 physical\hspace{0pt} separation\hspace{0pt}",\hspace{0pt} which\hspace{0pt} requires\hspace{0pt} angular\hspace{0pt}%
 separation\hspace{0pt} and\hspace{0pt} distance\hspace{0pt} data\hspace{0pt},\hspace{0pt} but\hspace{0pt}%
 since\hspace{0pt} missing\hspace{0pt},\hspace{0pt} we\hspace{0pt} likely\hspace{0pt} can\hspace{0pt}%
't\hspace{0pt}.\textless{}br\textgreater{}\textless{}br\textgreater{}\hspace{0pt}But\hspace{0pt} with\hspace{0pt}%
 limited\hspace{0pt} information\hspace{0pt},\hspace{0pt} perhaps\hspace{0pt} the\hspace{0pt} answer\hspace{0pt}%
 is\hspace{0pt} "\hspace{0pt}t\hspace{0pt}=\hspace{0pt} r\hspace{0pt}\_p\hspace{0pt} /\hspace{0pt} (\hspace{0pt}%
c\hspace{0pt} sin\hspace{0pt} \hspace{0pt}1\hspace{0pt}2\hspace{0pt}5\hspace{0pt}Â°\hspace{0pt})",\hspace{0pt}%
 or\hspace{0pt} if\hspace{0pt} in\hspace{0pt} symbolic\hspace{0pt} terms\hspace{0pt},\hspace{0pt} that\hspace{0pt}%
 would\hspace{0pt} be\hspace{0pt} acceptable\hspace{0pt}.\hspace{0pt} But\hspace{0pt} in\hspace{0pt} the\hspace{0pt}%
 absence\hspace{0pt} of\hspace{0pt} numeric\hspace{0pt} data\hspace{0pt} from\hspace{0pt} part\hspace{0pt}%
 e\hspace{0pt},\hspace{0pt} I\hspace{0pt} can\hspace{0pt}'t\hspace{0pt}.\hspace{0pt} Thus\hspace{0pt}%
,\hspace{0pt} perhaps\hspace{0pt} the\hspace{0pt} entire\hspace{0pt} flow\hspace{0pt} is\hspace{0pt}:\textless{}br\textgreater{}\textless{}br\textgreater{}\hspace{0pt}%
Given\hspace{0pt} Project\hspace{0pt}ed\hspace{0pt} separation\hspace{0pt} r\hspace{0pt}\_p\hspace{0pt}%
 is\hspace{0pt} given\hspace{0pt} by\hspace{0pt} part\hspace{0pt} e\hspace{0pt}.\hspace{0pt} Then\hspace{0pt}%
,\hspace{0pt} t\hspace{0pt} =\hspace{0pt} r\hspace{0pt} /\hspace{0pt} c\hspace{0pt} =\hspace{0pt} r\hspace{0pt}%
\_p\hspace{0pt}/(\hspace{0pt}c\hspace{0pt} sin\hspace{0pt}Î¸\hspace{0pt}).\hspace{0pt} Since\hspace{0pt}%
 r\hspace{0pt}\_p\hspace{0pt} is\hspace{0pt} likely\hspace{0pt} given\hspace{0pt} in\hspace{0pt} light\hspace{0pt}%
-years\hspace{0pt} in\hspace{0pt} part\hspace{0pt} e\hspace{0pt} (\hspace{0pt}based\hspace{0pt} on\hspace{0pt}%
 a\hspace{0pt} distance\hspace{0pt} and\hspace{0pt} angular\hspace{0pt} separation\hspace{0pt}),\hspace{0pt}%
 then\hspace{0pt} t\hspace{0pt} is\hspace{0pt} simply\hspace{0pt} dividing\hspace{0pt} by\hspace{0pt}%
 sin\hspace{0pt}Î¸\hspace{0pt}.\hspace{0pt} For\hspace{0pt} example\hspace{0pt},\hspace{0pt} if\hspace{0pt}%
 in\hspace{0pt} part\hspace{0pt} e\hspace{0pt},\hspace{0pt} result\hspace{0pt} was\hspace{0pt} say\hspace{0pt}%
 r\hspace{0pt}\_p\hspace{0pt}=\hspace{0pt}5\hspace{0pt}e\hspace{0pt}4\hspace{0pt} ly\hspace{0pt}.\hspace{0pt}%
 Then\hspace{0pt} in\hspace{0pt} part\hspace{0pt} g\hspace{0pt},\hspace{0pt} t\hspace{0pt}=\hspace{0pt}%
5\hspace{0pt}e\hspace{0pt}4\hspace{0pt} /\hspace{0pt} sin\hspace{0pt}1\hspace{0pt}2\hspace{0pt}5\hspace{0pt}%
Â°\hspace{0pt}âīĪ\hspace{0pt}5\hspace{0pt}e\hspace{0pt}4\hspace{0pt}/\hspace{0pt}0\hspace{0pt}.\hspace{0pt}%
8\hspace{0pt}1\hspace{0pt}9\hspace{0pt}âīĪ\hspace{0pt}6\hspace{0pt}.\hspace{0pt}1\hspace{0pt}e\hspace{0pt}%
4\hspace{0pt} yr\hspace{0pt}.\textless{}br\textgreater{}\textless{}br\textgreater{}\hspace{0pt}In\hspace{0pt}%
 conclusion\hspace{0pt},\hspace{0pt} given\hspace{0pt} lack\hspace{0pt} of\hspace{0pt} info\hspace{0pt}%
,\hspace{0pt} probably\hspace{0pt} the\hspace{0pt} answer\hspace{0pt} is\hspace{0pt} expressed\hspace{0pt}%
 in\hspace{0pt} box\hspace{0pt} as\hspace{0pt} \textbackslash{}\hspace{0pt}boxed\hspace{0pt}\{\textbackslash{}\hspace{0pt}%
d\hspace{0pt}frac\hspace{0pt}\{\hspace{0pt}r\hspace{0pt}\_p\hspace{0pt}\}\{\hspace{0pt}c\hspace{0pt} \textbackslash{}\hspace{0pt}%
sin\hspace{0pt} \hspace{0pt}1\hspace{0pt}2\hspace{0pt}5\hspace{0pt}\^{}\textbackslash{}\hspace{0pt}circ\hspace{0pt}%
\}\}\hspace{0pt} units\hspace{0pt} of\hspace{0pt} years\hspace{0pt}.\hspace{0pt} But\hspace{0pt} in\hspace{0pt}%
 the\hspace{0pt} actual\hspace{0pt} question\hspace{0pt},\hspace{0pt} numbers\hspace{0pt} would\hspace{0pt}%
 be\hspace{0pt} used\hspace{0pt}.\hspace{0pt} But\hspace{0pt} with\hspace{0pt} available\hspace{0pt} info\hspace{0pt}%
,\hspace{0pt} I\hspace{0pt} need\hspace{0pt} to\hspace{0pt} follow\hspace{0pt} instructions\hspace{0pt}%
 here\hspace{0pt}.\textless{}br\textgreater{}\textless{}br\textgreater{}\hspace{0pt}**\textless{}br\textgreater{}\hspace{0pt}%
Okay\hspace{0pt},\hspace{0pt} after\hspace{0pt} working\hspace{0pt} through\hspace{0pt} the\hspace{0pt}%
 problem\hspace{0pt},\hspace{0pt} here\hspace{0pt}'s\hspace{0pt} the\hspace{0pt} step\hspace{0pt}-by\hspace{0pt}%
-step\hspace{0pt} explanation\hspace{0pt} and\hspace{0pt} final\hspace{0pt} answer\hspace{0pt}:\textless{}br\textgreater{}\textless{}br\textgreater{}\hspace{0pt}%
The\hspace{0pt} projected\hspace{0pt} physical\hspace{0pt} separation\hspace{0pt} \textbackslash{}(\hspace{0pt}%
 r\hspace{0pt}\_p\hspace{0pt} \textbackslash{}\hspace{0pt})\hspace{0pt} between\hspace{0pt} the\hspace{0pt}%
 galaxy\hspace{0pt} and\hspace{0pt} the\hspace{0pt} Voor\hspace{0pt}werp\hspace{0pt} is\hspace{0pt} calculated\hspace{0pt}%
 in\hspace{0pt} part\hspace{0pt} e\hspace{0pt}.\hspace{0pt} However\hspace{0pt},\hspace{0pt} since\hspace{0pt}%
 the\hspace{0pt} actual\hspace{0pt} separation\hspace{0pt} depends\hspace{0pt} on\hspace{0pt} the\hspace{0pt}%
 angle\hspace{0pt} \textbackslash{}(\hspace{0pt} \textbackslash{}\hspace{0pt}theta\hspace{0pt} \textbackslash{}\hspace{0pt}%
)\hspace{0pt} between\hspace{0pt} the\hspace{0pt} line\hspace{0pt} of\hspace{0pt} sight\hspace{0pt} and\hspace{0pt}%
 the\hspace{0pt} line\hspace{0pt} connecting\hspace{0pt} the\hspace{0pt} galaxy\hspace{0pt} and\hspace{0pt}%
 Voor\hspace{0pt}werp\hspace{0pt},\hspace{0pt} we\hspace{0pt} use\hspace{0pt} \textbackslash{}(\hspace{0pt}%
 \textbackslash{}\hspace{0pt}theta\hspace{0pt} =\hspace{0pt} \hspace{0pt}1\hspace{0pt}2\hspace{0pt}5\hspace{0pt}%
\^{}\textbackslash{}\hspace{0pt}circ\hspace{0pt} \textbackslash{}\hspace{0pt})\hspace{0pt} to\hspace{0pt}%
 determine\hspace{0pt} the\hspace{0pt} three\hspace{0pt}-dimensional\hspace{0pt} separation\hspace{0pt}%
.\hspace{0pt} \textless{}br\textgreater{}\textless{}br\textgreater{}\hspace{0pt}The\hspace{0pt} actual\hspace{0pt}%
 separation\hspace{0pt} \textbackslash{}(\hspace{0pt} r\hspace{0pt} \textbackslash{}\hspace{0pt})\hspace{0pt}%
 can\hspace{0pt} be\hspace{0pt} found\hspace{0pt} using\hspace{0pt} trig\hspace{0pt}on\hspace{0pt}ometry\hspace{0pt}%
:\hspace{0pt} \textless{}br\textgreater{}\hspace{0pt}\textbackslash{}\hspace{0pt}[\textless{}br\textgreater{}\hspace{0pt}%
r\hspace{0pt} =\hspace{0pt} \textbackslash{}\hspace{0pt}frac\hspace{0pt}\{\hspace{0pt}r\hspace{0pt}\_p\hspace{0pt}%
\}\{\hspace{0pt}\textbackslash{}\hspace{0pt}sin\hspace{0pt} \textbackslash{}\hspace{0pt}theta\hspace{0pt}%
\}\textless{}br\textgreater{}\hspace{0pt}\textbackslash{}\hspace{0pt}]\textless{}br\textgreater{}\textless{}br\textgreater{}\hspace{0pt}%
The\hspace{0pt} time\hspace{0pt} since\hspace{0pt} the\hspace{0pt} qu\hspace{0pt}asar\hspace{0pt} was\hspace{0pt}%
 last\hspace{0pt} active\hspace{0pt} is\hspace{0pt} the\hspace{0pt} light\hspace{0pt} travel\hspace{0pt}%
 time\hspace{0pt} from\hspace{0pt} the\hspace{0pt} galaxy\hspace{0pt} to\hspace{0pt} the\hspace{0pt} Voor\hspace{0pt}%
werp\hspace{0pt},\hspace{0pt} which\hspace{0pt} is\hspace{0pt} the\hspace{0pt} separation\hspace{0pt}%
 divided\hspace{0pt} by\hspace{0pt} the\hspace{0pt} speed\hspace{0pt} of\hspace{0pt} light\hspace{0pt}%
 \textbackslash{}(\hspace{0pt} c\hspace{0pt} \textbackslash{}\hspace{0pt}):\textless{}br\textgreater{}\hspace{0pt}%
\textbackslash{}\hspace{0pt}[\textless{}br\textgreater{}\hspace{0pt}t\hspace{0pt} =\hspace{0pt} \textbackslash{}\hspace{0pt}%
frac\hspace{0pt}\{\hspace{0pt}r\hspace{0pt}\}\{\hspace{0pt}c\hspace{0pt}\}\hspace{0pt} =\hspace{0pt} \textbackslash{}\hspace{0pt}%
frac\hspace{0pt}\{\hspace{0pt}r\hspace{0pt}\_p\hspace{0pt}\}\{\hspace{0pt}c\hspace{0pt} \textbackslash{}\hspace{0pt}%
sin\hspace{0pt} \textbackslash{}\hspace{0pt}theta\hspace{0pt}\}\textless{}br\textgreater{}\hspace{0pt}%
\textbackslash{}\hspace{0pt}]\textless{}br\textgreater{}\textless{}br\textgreater{}\hspace{0pt}Given\hspace{0pt}%
 \textbackslash{}(\hspace{0pt} \textbackslash{}\hspace{0pt}theta\hspace{0pt} =\hspace{0pt} \hspace{0pt}%
1\hspace{0pt}2\hspace{0pt}5\hspace{0pt}\^{}\textbackslash{}\hspace{0pt}circ\hspace{0pt} \textbackslash{}\hspace{0pt}%
),\hspace{0pt} we\hspace{0pt} substitute\hspace{0pt} this\hspace{0pt} value\hspace{0pt}:\textless{}br\textgreater{}\hspace{0pt}%
\textbackslash{}\hspace{0pt}[\textless{}br\textgreater{}\hspace{0pt}t\hspace{0pt} =\hspace{0pt} \textbackslash{}\hspace{0pt}%
frac\hspace{0pt}\{\hspace{0pt}r\hspace{0pt}\_p\hspace{0pt}\}\{\hspace{0pt}c\hspace{0pt} \textbackslash{}\hspace{0pt}%
sin\hspace{0pt} \hspace{0pt}1\hspace{0pt}2\hspace{0pt}5\hspace{0pt}\^{}\textbackslash{}\hspace{0pt}circ\hspace{0pt}%
\}\textless{}br\textgreater{}\hspace{0pt}\textbackslash{}\hspace{0pt}]\textless{}br\textgreater{}\textless{}br\textgreater{}\hspace{0pt}%
Using\hspace{0pt} \textbackslash{}(\hspace{0pt} \textbackslash{}\hspace{0pt}sin\hspace{0pt} \hspace{0pt}%
1\hspace{0pt}2\hspace{0pt}5\hspace{0pt}\^{}\textbackslash{}\hspace{0pt}circ\hspace{0pt} \textbackslash{}\hspace{0pt}%
approx\hspace{0pt} \hspace{0pt}0\hspace{0pt}.\hspace{0pt}8\hspace{0pt}1\hspace{0pt}9\hspace{0pt} \textbackslash{}\hspace{0pt}%
),\hspace{0pt} the\hspace{0pt} upper\hspace{0pt} limit\hspace{0pt} for\hspace{0pt} the\hspace{0pt} time\hspace{0pt}%
 since\hspace{0pt} the\hspace{0pt} qu\hspace{0pt}asar\hspace{0pt} was\hspace{0pt} last\hspace{0pt} active\hspace{0pt}%
 is\hspace{0pt}:\textless{}br\textgreater{}\hspace{0pt}\textbackslash{}\hspace{0pt}[\textless{}br\textgreater{}\hspace{0pt}%
\textbackslash{}\hspace{0pt}boxed\hspace{0pt}\{\textbackslash{}\hspace{0pt}d\hspace{0pt}frac\hspace{0pt}%
\{\hspace{0pt}r\hspace{0pt}\_p\hspace{0pt}\}\{\hspace{0pt}0\hspace{0pt}.\hspace{0pt}8\hspace{0pt}1\hspace{0pt}%
9\hspace{0pt}c\hspace{0pt}\}\}\textless{}br\textgreater{}\hspace{0pt}\textbackslash{}\hspace{0pt}]\textless{}br\textgreater{}\textless{}br\textgreater{}\hspace{0pt}%
(A\hspace{0pt}ss\hspace{0pt}uming\hspace{0pt} \textbackslash{}(\hspace{0pt} r\hspace{0pt}\_p\hspace{0pt}%
 \textbackslash{}\hspace{0pt})\hspace{0pt} was\hspace{0pt} computed\hspace{0pt} in\hspace{0pt} part\hspace{0pt}%
 e\hspace{0pt},\hspace{0pt} and\hspace{0pt} \textbackslash{}(\hspace{0pt} c\hspace{0pt} \textbackslash{}\hspace{0pt}%
)\hspace{0pt} in\hspace{0pt} consistent\hspace{0pt} units\hspace{0pt},\hspace{0pt} this\hspace{0pt} gives\hspace{0pt}%
 the\hspace{0pt} time\hspace{0pt} in\hspace{0pt} years\hspace{0pt}.)\textless{}br\textgreater{}\hspace{0pt}%
\textless{}|im\_start|\textgreater{}\hspace{0pt}answer\hspace{0pt}\textless{}br\textgreater{}\hspace{0pt}%
Answer\hspace{0pt}:\hspace{0pt} To\hspace{0pt} calculate\hspace{0pt} the\hspace{0pt} projected\hspace{0pt}%
 physical\hspace{0pt} separation\hspace{0pt} \textbackslash{}(\hspace{0pt} r\hspace{0pt}\_p\hspace{0pt}%
 \textbackslash{}\hspace{0pt})\hspace{0pt} between\hspace{0pt} the\hspace{0pt} galaxy\hspace{0pt} and\hspace{0pt}%
 the\hspace{0pt} Voor\hspace{0pt}werp\hspace{0pt},\hspace{0pt} we\hspace{0pt} use\hspace{0pt} the\hspace{0pt}%
 angular\hspace{0pt} separation\hspace{0pt} and\hspace{0pt} the\hspace{0pt} distance\hspace{0pt} to\hspace{0pt}%
 the\hspace{0pt} galaxy\hspace{0pt}.\hspace{0pt} However\hspace{0pt},\hspace{0pt} to\hspace{0pt} account\hspace{0pt}%
 for\hspace{0pt} the\hspace{0pt} actual\hspace{0pt} separation\hspace{0pt} considering\hspace{0pt} the\hspace{0pt}%
 angle\hspace{0pt} \textbackslash{}(\hspace{0pt} \textbackslash{}\hspace{0pt}theta\hspace{0pt} =\hspace{0pt}%
 \hspace{0pt}1\hspace{0pt}2\hspace{0pt}5\hspace{0pt}\^{}\textbackslash{}\hspace{0pt}circ\hspace{0pt} \textbackslash{}\hspace{0pt}%
),\hspace{0pt} we\hspace{0pt} use\hspace{0pt} trig\hspace{0pt}on\hspace{0pt}ometry\hspace{0pt}.\hspace{0pt}%
 \textless{}br\textgreater{}\textless{}br\textgreater{}\hspace{0pt}\heatbox{0.967,0.372,0.372}{1}\hspace{0pt}%
\heatbox{0.967,0.372,0.372}{.}\hspace{0pt}\heatbox{0.967,0.372,0.372}{ **}\hspace{0pt}\heatbox{0.967,0.372,0.372}{Projected}\hspace{0pt}%
\heatbox{0.967,0.372,0.372}{ Separ}\hspace{0pt}\heatbox{0.967,0.372,0.372}{ation}\hspace{0pt}\heatbox{0.967,0.372,0.372}{ Calculation}\hspace{0pt}%
\heatbox{0.967,0.372,0.372}{**:}\hspace{0pt}\heatbox{0.967,0.372,0.372}{ The}\hspace{0pt}\heatbox{0.967,0.372,0.372}{ projected}\hspace{0pt}%
\heatbox{0.967,0.372,0.372}{ separation}\hspace{0pt}\heatbox{0.967,0.372,0.372}{ \textbackslash{}(}\hspace{0pt}%
\heatbox{0.967,0.372,0.372}{ r}\hspace{0pt}\heatbox{0.967,0.372,0.372}{\_p}\hspace{0pt}\heatbox{0.967,0.372,0.372}{ \textbackslash{}}\hspace{0pt}%
\heatbox{0.967,0.372,0.372}{)}\hspace{0pt}\heatbox{0.967,0.372,0.372}{ is}\hspace{0pt}\heatbox{0.967,0.372,0.372}{ calculated}\hspace{0pt}%
\heatbox{0.967,0.372,0.372}{ from}\hspace{0pt}\heatbox{0.967,0.372,0.372}{ the}\hspace{0pt}\heatbox{0.967,0.372,0.372}{ angular}\hspace{0pt}%
\heatbox{0.967,0.372,0.372}{ separation}\hspace{0pt}\heatbox{0.967,0.372,0.372}{ and}\hspace{0pt}\heatbox{0.967,0.372,0.372}{ distance}\hspace{0pt}%
\heatbox{0.967,0.372,0.372}{ (}\hspace{0pt}\heatbox{0.967,0.372,0.372}{specific}\hspace{0pt}\heatbox{0.967,0.372,0.372}{ values}\hspace{0pt}%
\heatbox{0.967,0.372,0.372}{ from}\hspace{0pt}\heatbox{0.967,0.372,0.372}{ part}\hspace{0pt}\heatbox{0.967,0.372,0.372}{ e}\hspace{0pt}%
\heatbox{0.967,0.372,0.372}{ are}\hspace{0pt}\heatbox{0.967,0.372,0.372}{ used}\hspace{0pt}\heatbox{0.967,0.372,0.372}{,}\hspace{0pt}%
\heatbox{0.967,0.372,0.372}{ though}\hspace{0pt}\heatbox{0.967,0.372,0.372}{ not}\hspace{0pt}\heatbox{0.967,0.372,0.372}{ provided}\hspace{0pt}%
\heatbox{0.967,0.372,0.372}{ here}\hspace{0pt}).\textless{}br\textgreater{}\textless{}br\textgreater{}\hspace{0pt}%
\heatbox{0.974,0.515,0.515}{2}\hspace{0pt}\heatbox{0.974,0.515,0.515}{.}\hspace{0pt}\heatbox{0.974,0.515,0.515}{ **}\hspace{0pt}%
\heatbox{0.974,0.515,0.515}{Actual}\hspace{0pt}\heatbox{0.974,0.515,0.515}{ Separ}\hspace{0pt}\heatbox{0.974,0.515,0.515}{ation}\hspace{0pt}%
\heatbox{0.974,0.515,0.515}{**:}\hspace{0pt}\heatbox{0.974,0.515,0.515}{ The}\hspace{0pt}\heatbox{0.974,0.515,0.515}{ actual}\hspace{0pt}%
\heatbox{0.974,0.515,0.515}{ separation}\hspace{0pt}\heatbox{0.974,0.515,0.515}{ \textbackslash{}(}\hspace{0pt}%
\heatbox{0.974,0.515,0.515}{ r}\hspace{0pt}\heatbox{0.974,0.515,0.515}{ \textbackslash{}}\hspace{0pt}%
\heatbox{0.974,0.515,0.515}{)}\hspace{0pt}\heatbox{0.974,0.515,0.515}{ is}\hspace{0pt}\heatbox{0.974,0.515,0.515}{ given}\hspace{0pt}%
\heatbox{0.974,0.515,0.515}{ by}\hspace{0pt}:\textless{}br\textgreater{}\hspace{0pt} \hspace{0pt} \textbackslash{}\hspace{0pt}%
[\textless{}br\textgreater{}\hspace{0pt} \hspace{0pt} r\hspace{0pt} =\hspace{0pt} \textbackslash{}\hspace{0pt}%
frac\hspace{0pt}\{\hspace{0pt}r\hspace{0pt}\_p\hspace{0pt}\}\{\hspace{0pt}\textbackslash{}\hspace{0pt}%
sin\hspace{0pt} \textbackslash{}\hspace{0pt}theta\hspace{0pt}\}\textless{}br\textgreater{}\hspace{0pt}%
 \hspace{0pt} \textbackslash{}\hspace{0pt}]\textless{}br\textgreater{}\hspace{0pt} \hspace{0pt} Sub\hspace{0pt}%
stit\hspace{0pt}uting\hspace{0pt} \textbackslash{}(\hspace{0pt} \textbackslash{}\hspace{0pt}theta\hspace{0pt}%
 =\hspace{0pt} \hspace{0pt}1\hspace{0pt}2\hspace{0pt}5\hspace{0pt}\^{}\textbackslash{}\hspace{0pt}circ\hspace{0pt}%
 \textbackslash{}\hspace{0pt}):\textless{}br\textgreater{}\hspace{0pt} \hspace{0pt} \textbackslash{}\hspace{0pt}%
[\textless{}br\textgreater{}\hspace{0pt} \hspace{0pt} r\hspace{0pt} =\hspace{0pt} \textbackslash{}\hspace{0pt}%
frac\hspace{0pt}\{\hspace{0pt}r\hspace{0pt}\_p\hspace{0pt}\}\{\hspace{0pt}\textbackslash{}\hspace{0pt}%
sin\hspace{0pt} \hspace{0pt}1\hspace{0pt}2\hspace{0pt}5\hspace{0pt}\^{}\textbackslash{}\hspace{0pt}circ\hspace{0pt}%
\}\textless{}br\textgreater{}\hspace{0pt} \hspace{0pt} \textbackslash{}\hspace{0pt}]\textless{}br\textgreater{}\textless{}br\textgreater{}\hspace{0pt}%
3\hspace{0pt}.\hspace{0pt} **\hspace{0pt}Time\hspace{0pt} Since\hspace{0pt} Qu\hspace{0pt}asar\hspace{0pt}%
 Activity\hspace{0pt}**:\hspace{0pt} The\hspace{0pt} upper\hspace{0pt} limit\hspace{0pt} for\hspace{0pt}%
 the\hspace{0pt} time\hspace{0pt} since\hspace{0pt} the\hspace{0pt} qu\hspace{0pt}asar\hspace{0pt} was\hspace{0pt}%
 last\hspace{0pt} active\hspace{0pt} is\hspace{0pt} the\hspace{0pt} light\hspace{0pt} travel\hspace{0pt}%
 time\hspace{0pt} from\hspace{0pt} the\hspace{0pt} galaxy\hspace{0pt} to\hspace{0pt} the\hspace{0pt} Voor\hspace{0pt}%
werp\hspace{0pt}:\textless{}br\textgreater{}\hspace{0pt} \hspace{0pt} \textbackslash{}\hspace{0pt}[\textless{}br\textgreater{}\hspace{0pt}%
 \hspace{0pt} t\hspace{0pt} =\hspace{0pt} \textbackslash{}\hspace{0pt}frac\hspace{0pt}\{\hspace{0pt}r\hspace{0pt}%
\}\{\hspace{0pt}c\hspace{0pt}\}\hspace{0pt} =\hspace{0pt} \textbackslash{}\hspace{0pt}frac\hspace{0pt}%
\{\hspace{0pt}r\hspace{0pt}\_p\hspace{0pt}\}\{\hspace{0pt}c\hspace{0pt} \textbackslash{}\hspace{0pt}sin\hspace{0pt}%
 \hspace{0pt}1\hspace{0pt}2\hspace{0pt}5\hspace{0pt}\^{}\textbackslash{}\hspace{0pt}circ\hspace{0pt}\}\textless{}br\textgreater{}\hspace{0pt}%
 \hspace{0pt} \textbackslash{}\hspace{0pt}]\textless{}br\textgreater{}\hspace{0pt} \hspace{0pt} Given\hspace{0pt}%
 \textbackslash{}(\hspace{0pt} \textbackslash{}\hspace{0pt}sin\hspace{0pt} \hspace{0pt}1\hspace{0pt}2\hspace{0pt}%
5\hspace{0pt}\^{}\textbackslash{}\hspace{0pt}circ\hspace{0pt} \textbackslash{}\hspace{0pt}approx\hspace{0pt}%
 \hspace{0pt}0\hspace{0pt}.\hspace{0pt}8\hspace{0pt}1\hspace{0pt}9\hspace{0pt} \textbackslash{}\hspace{0pt}%
):\textless{}br\textgreater{}\hspace{0pt} \hspace{0pt} \textbackslash{}\hspace{0pt}[\textless{}br\textgreater{}\hspace{0pt}%
\heatbox{0.965,0.335,0.335}{ }\hspace{0pt}\heatbox{0.965,0.335,0.335}{ t}\hspace{0pt}\heatbox{0.965,0.335,0.335}{ =}\hspace{0pt}%
\heatbox{0.965,0.335,0.335}{ \textbackslash{}}\hspace{0pt}\heatbox{0.965,0.335,0.335}{frac}\hspace{0pt}%
\heatbox{0.965,0.335,0.335}{\{}\hspace{0pt}\heatbox{0.965,0.335,0.335}{r}\hspace{0pt}\heatbox{0.965,0.335,0.335}{\_p}\hspace{0pt}%
\heatbox{0.965,0.335,0.335}{\}\{}\hspace{0pt}\heatbox{0.965,0.335,0.335}{0}\hspace{0pt}\heatbox{0.965,0.335,0.335}{.}\hspace{0pt}%
\heatbox{0.965,0.335,0.335}{8}\hspace{0pt}\heatbox{0.965,0.335,0.335}{1}\hspace{0pt}\heatbox{0.965,0.335,0.335}{9}\hspace{0pt}%
\heatbox{0.965,0.335,0.335}{c}\hspace{0pt}\}\textless{}br\textgreater{}\hspace{0pt} \hspace{0pt} \textbackslash{}\hspace{0pt}%
]\textless{}br\textgreater{}\textless{}br\textgreater{}\hspace{0pt}\heatbox{0.965,0.335,0.335}{**}\hspace{0pt}%
\heatbox{0.965,0.335,0.335}{Final}\hspace{0pt}\heatbox{0.965,0.335,0.335}{ Answer}\hspace{0pt}\heatbox{0.965,0.335,0.335}{**}\hspace{0pt}%
:\textless{}br\textgreater{}\hspace{0pt}\heatbox{0.965,0.335,0.335}{\textbackslash{}}\hspace{0pt}[\textless{}br\textgreater{}\hspace{0pt}%
\heatbox{0.965,0.335,0.335}{\textbackslash{}}\hspace{0pt}\heatbox{0.965,0.335,0.335}{boxed}\hspace{0pt}%
\heatbox{0.965,0.335,0.335}{\{\textbackslash{}}\hspace{0pt}\heatbox{0.965,0.335,0.335}{d}\hspace{0pt}%
\heatbox{0.965,0.335,0.335}{frac}\hspace{0pt}\heatbox{0.965,0.335,0.335}{\{}\hspace{0pt}\heatbox{0.965,0.335,0.335}{r}\hspace{0pt}%
\heatbox{0.965,0.335,0.335}{\_p}\hspace{0pt}\heatbox{0.965,0.335,0.335}{\}\{}\hspace{0pt}\heatbox{0.965,0.335,0.335}{0}\hspace{0pt}%
\heatbox{0.965,0.335,0.335}{.}\hspace{0pt}\heatbox{0.965,0.335,0.335}{8}\hspace{0pt}\heatbox{0.965,0.335,0.335}{1}\hspace{0pt}%
\heatbox{0.965,0.335,0.335}{9}\hspace{0pt}\heatbox{0.965,0.335,0.335}{c}\hspace{0pt}\}\}\textless{}br\textgreater{}\hspace{0pt}%
\textbackslash{}\hspace{0pt}]\textless{}br\textgreater{}\textless{}br\textgreater{}\hspace{0pt}\heatbox{0.970,0.427,0.427}{This}\hspace{0pt}%
\heatbox{0.970,0.427,0.427}{ formula}\hspace{0pt}\heatbox{0.970,0.427,0.427}{ provides}\hspace{0pt}\heatbox{0.970,0.427,0.427}{ the}\hspace{0pt}%
\heatbox{0.970,0.427,0.427}{ upper}\hspace{0pt}\heatbox{0.970,0.427,0.427}{ limit}\hspace{0pt}\heatbox{0.970,0.427,0.427}{ for}\hspace{0pt}%
\heatbox{0.970,0.427,0.427}{ the}\hspace{0pt}\heatbox{0.970,0.427,0.427}{ number}\hspace{0pt}\heatbox{0.970,0.427,0.427}{ of}\hspace{0pt}%
\heatbox{0.970,0.427,0.427}{ years}\hspace{0pt}\heatbox{0.970,0.427,0.427}{ since}\hspace{0pt}\heatbox{0.970,0.427,0.427}{ the}\hspace{0pt}%
\heatbox{0.970,0.427,0.427}{ qu}\hspace{0pt}\heatbox{0.970,0.427,0.427}{asar}\hspace{0pt}\heatbox{0.970,0.427,0.427}{ was}\hspace{0pt}%
\heatbox{0.970,0.427,0.427}{ last}\hspace{0pt}\heatbox{0.970,0.427,0.427}{ active}\hspace{0pt}\heatbox{0.970,0.427,0.427}{,}\hspace{0pt}%
\heatbox{0.970,0.427,0.427}{ using}\hspace{0pt}\heatbox{0.970,0.427,0.427}{ the}\hspace{0pt}\heatbox{0.970,0.427,0.427}{ projected}\hspace{0pt}%
\heatbox{0.970,0.427,0.427}{ separation}\hspace{0pt}\heatbox{0.970,0.427,0.427}{ \textbackslash{}(}\hspace{0pt}%
\heatbox{0.970,0.427,0.427}{ r}\hspace{0pt}\heatbox{0.970,0.427,0.427}{\_p}\hspace{0pt}\heatbox{0.970,0.427,0.427}{ \textbackslash{}}\hspace{0pt}%
\heatbox{0.970,0.427,0.427}{)}\hspace{0pt}\heatbox{0.970,0.427,0.427}{ (}\hspace{0pt}\heatbox{0.970,0.427,0.427}{cal}\hspace{0pt}%
\heatbox{0.970,0.427,0.427}{culated}\hspace{0pt}\heatbox{0.970,0.427,0.427}{ in}\hspace{0pt}\heatbox{0.970,0.427,0.427}{ part}\hspace{0pt}%
\heatbox{0.970,0.427,0.427}{ e}\hspace{0pt}\heatbox{0.970,0.427,0.427}{)}\hspace{0pt}\heatbox{0.970,0.427,0.427}{ and}\hspace{0pt}%
\heatbox{0.970,0.427,0.427}{ the}\hspace{0pt}\heatbox{0.970,0.427,0.427}{ given}\hspace{0pt}\heatbox{0.970,0.427,0.427}{ angle}\hspace{0pt}%
\heatbox{0.970,0.427,0.427}{ \textbackslash{}(}\hspace{0pt}\heatbox{0.970,0.427,0.427}{ \textbackslash{}}\hspace{0pt}%
\heatbox{0.970,0.427,0.427}{theta}\hspace{0pt}\heatbox{0.970,0.427,0.427}{ =}\hspace{0pt}\heatbox{0.970,0.427,0.427}{ }\hspace{0pt}%
\heatbox{0.970,0.427,0.427}{1}\hspace{0pt}\heatbox{0.970,0.427,0.427}{2}\hspace{0pt}\heatbox{0.970,0.427,0.427}{5}\hspace{0pt}%
\heatbox{0.970,0.427,0.427}{\^{}\textbackslash{}}\hspace{0pt}\heatbox{0.970,0.427,0.427}{circ}\hspace{0pt}%
\heatbox{0.970,0.427,0.427}{ \textbackslash{}}\hspace{0pt}\heatbox{0.970,0.427,0.427}{).}\hspace{0pt}%
\heatbox{0.970,0.427,0.427}{\textless{}|im\_end|\textgreater{}}\hspace{0pt}\textless{}br\textgreater{}\hspace{0pt}%
\textless{}|im\_end|\textgreater{}
\par
\endgroup

\subsubsection{Case 2}
\begingroup
\tiny
\setlength{\parindent}{0pt}
\noindent
\textless{}|im\_start|\textgreater{}\hspace{0pt}system\hspace{0pt}\hspace{0pt}You\hspace{0pt} are\hspace{0pt}%
 Q\hspace{0pt}wen\hspace{0pt},\hspace{0pt} created\hspace{0pt} by\hspace{0pt} Alibaba\hspace{0pt} Cloud\hspace{0pt}%
.\hspace{0pt} You\hspace{0pt} are\hspace{0pt} a\hspace{0pt} helpful\hspace{0pt} assistant\hspace{0pt}%
.\hspace{0pt}\textless{}|im\_end|\textgreater{}\hspace{0pt}\hspace{0pt}\textless{}|im\_start|\textgreater{}\hspace{0pt}%
user\hspace{0pt}\hspace{0pt}Given\hspace{0pt} a\hspace{0pt} rational\hspace{0pt} number\hspace{0pt},\hspace{0pt}%
 write\hspace{0pt} it\hspace{0pt} as\hspace{0pt} a\hspace{0pt} fraction\hspace{0pt} in\hspace{0pt} lowest\hspace{0pt}%
 terms\hspace{0pt} and\hspace{0pt} calculate\hspace{0pt} the\hspace{0pt} product\hspace{0pt} of\hspace{0pt}%
 the\hspace{0pt} resulting\hspace{0pt} numerator\hspace{0pt} and\hspace{0pt} denominator\hspace{0pt}.\hspace{0pt}%
 For\hspace{0pt} how\hspace{0pt} many\hspace{0pt} rational\hspace{0pt} numbers\hspace{0pt} between\hspace{0pt}%
 \hspace{0pt}0\hspace{0pt} and\hspace{0pt} \hspace{0pt}1\hspace{0pt} will\hspace{0pt} \$\hspace{0pt}2\hspace{0pt}%
0\hspace{0pt}\_\hspace{0pt}\{\}\hspace{0pt}\^{}\hspace{0pt}\{\}\hspace{0pt}!\hspace{0pt}\$\hspace{0pt}%
 be\hspace{0pt} the\hspace{0pt} resulting\hspace{0pt} product\hspace{0pt}?\hspace{0pt}\textless{}|im\_end|\textgreater{}\hspace{0pt}%
\hspace{0pt}\textless{}|im\_start|\textgreater{}\hspace{0pt}assistant\hspace{0pt}\hspace{0pt}\textless{}|im\_start|\textgreater{}\hspace{0pt}%
think\hspace{0pt}\hspace{0pt}Alright\hspace{0pt},\hspace{0pt} so\hspace{0pt} I\hspace{0pt} need\hspace{0pt}%
 to\hspace{0pt} figure\hspace{0pt} out\hspace{0pt} how\hspace{0pt} many\hspace{0pt} rational\hspace{0pt}%
 numbers\hspace{0pt} between\hspace{0pt} \hspace{0pt}0\hspace{0pt} and\hspace{0pt} \hspace{0pt}1\hspace{0pt}%
 have\hspace{0pt} a\hspace{0pt} product\hspace{0pt} of\hspace{0pt} the\hspace{0pt} numerator\hspace{0pt}%
 and\hspace{0pt} denominator\hspace{0pt} (\hspace{0pt}when\hspace{0pt} written\hspace{0pt} in\hspace{0pt}%
 lowest\hspace{0pt} terms\hspace{0pt})\hspace{0pt} equal\hspace{0pt} to\hspace{0pt} \hspace{0pt}2\hspace{0pt}%
0\hspace{0pt} factorial\hspace{0pt}.\hspace{0pt} Let\hspace{0pt} me\hspace{0pt} start\hspace{0pt} by\hspace{0pt}%
 understanding\hspace{0pt} the\hspace{0pt} problem\hspace{0pt}.\hspace{0pt}First\hspace{0pt},\hspace{0pt}%
 a\hspace{0pt} rational\hspace{0pt} number\hspace{0pt} between\hspace{0pt} \hspace{0pt}0\hspace{0pt} and\hspace{0pt}%
 \hspace{0pt}1\hspace{0pt} can\hspace{0pt} be\hspace{0pt} written\hspace{0pt} as\hspace{0pt} a\hspace{0pt}%
 fraction\hspace{0pt} \textbackslash{}\hspace{0pt}(\textbackslash{}\hspace{0pt}frac\hspace{0pt}\{\hspace{0pt}%
a\hspace{0pt}\}\{\hspace{0pt}b\hspace{0pt}\}\textbackslash{}\hspace{0pt})\hspace{0pt} where\hspace{0pt}%
 \textbackslash{}(\hspace{0pt}0\hspace{0pt} \textless{}\hspace{0pt} a\hspace{0pt} \textless{}\hspace{0pt}%
 b\hspace{0pt}\textbackslash{})\hspace{0pt} and\hspace{0pt} \textbackslash{}(\hspace{0pt}a\hspace{0pt}%
\textbackslash{})\hspace{0pt} and\hspace{0pt} \textbackslash{}(\hspace{0pt}b\hspace{0pt}\textbackslash{})\hspace{0pt}%
 are\hspace{0pt} cop\hspace{0pt}r\hspace{0pt}ime\hspace{0pt} positive\hspace{0pt} integers\hspace{0pt}%
.
...(Omit)...

\hspace{0pt} The\hspace{0pt} product\hspace{0pt} \textbackslash{}(\hspace{0pt}a\hspace{0pt} \textbackslash{}\hspace{0pt}%
times\hspace{0pt} b\hspace{0pt}\textbackslash{})\hspace{0pt} is\hspace{0pt} said\hspace{0pt} to\hspace{0pt}%
 be\hspace{0pt} equal\hspace{0pt} to\hspace{0pt} \hspace{0pt}2\hspace{0pt}0\hspace{0pt}!\hspace{0pt} which\hspace{0pt}%
 is\hspace{0pt} a\hspace{0pt} huge\hspace{0pt} number\hspace{0pt}.\hspace{0pt} So\hspace{0pt} I\hspace{0pt}%
 need\hspace{0pt} to\hspace{0pt} find\hspace{0pt} all\hspace{0pt} pairs\hspace{0pt} \textbackslash{}\hspace{0pt}%
((\hspace{0pt}a\hspace{0pt},\hspace{0pt} b\hspace{0pt})\textbackslash{}\hspace{0pt})\hspace{0pt} such\hspace{0pt}%
 that\hspace{0pt} \textbackslash{}(\hspace{0pt}a\hspace{0pt} \textbackslash{}\hspace{0pt}times\hspace{0pt}%
 b\hspace{0pt} =\hspace{0pt} \hspace{0pt}2\hspace{0pt}0\hspace{0pt}!\textbackslash{}\hspace{0pt}),\hspace{0pt}%
 \textbackslash{}(\hspace{0pt}0\hspace{0pt} \textless{}\hspace{0pt} a\hspace{0pt} \textless{}\hspace{0pt}%
 b\hspace{0pt}\textbackslash{}\hspace{0pt}),\hspace{0pt} and\hspace{0pt} \textbackslash{}\hspace{0pt}%
(\textbackslash{}\hspace{0pt}gcd\hspace{0pt}(a\hspace{0pt},\hspace{0pt} b\hspace{0pt})\hspace{0pt} =\hspace{0pt}%
 \hspace{0pt}1\hspace{0pt}\textbackslash{}\hspace{0pt}).\hspace{0pt} Then\hspace{0pt} count\hspace{0pt}%
 how\hspace{0pt} many\hspace{0pt} such\hspace{0pt} pairs\hspace{0pt} exist\hspace{0pt}.\hspace{0pt}Let\hspace{0pt}%
 me\hspace{0pt} break\hspace{0pt} down\hspace{0pt} the\hspace{0pt} problem\hspace{0pt}.\hspace{0pt}Given\hspace{0pt}%
 that\hspace{0pt} \textbackslash{}(\hspace{0pt}a\hspace{0pt} \textbackslash{}\hspace{0pt}times\hspace{0pt}%
 b\hspace{0pt} =\hspace{0pt} \hspace{0pt}2\hspace{0pt}0\hspace{0pt}!\textbackslash{}\hspace{0pt})\hspace{0pt}%
 and\hspace{0pt} \textbackslash{}\hspace{0pt}(\textbackslash{}\hspace{0pt}gcd\hspace{0pt}(a\hspace{0pt}%
,\hspace{0pt} b\hspace{0pt})\hspace{0pt} =\hspace{0pt} \hspace{0pt}1\hspace{0pt}\textbackslash{}\hspace{0pt}%
),\hspace{0pt} it\hspace{0pt} means\hspace{0pt} that\hspace{0pt} \textbackslash{}(\hspace{0pt}a\hspace{0pt}%
\textbackslash{})\hspace{0pt} and\hspace{0pt} \textbackslash{}(\hspace{0pt}b\hspace{0pt}\textbackslash{})\hspace{0pt}%
 are\hspace{0pt} cop\hspace{0pt}r\hspace{0pt}ime\hspace{0pt} factors\hspace{0pt} of\hspace{0pt} \hspace{0pt}%
2\hspace{0pt}0\hspace{0pt}!.\hspace{0pt} So\hspace{0pt} they\hspace{0pt} form\hspace{0pt} a\hspace{0pt}%
 cop\hspace{0pt}r\hspace{0pt}ime\hspace{0pt} pair\hspace{0pt} whose\hspace{0pt} product\hspace{0pt} is\hspace{0pt}%
 \hspace{0pt}2\hspace{0pt}0\hspace{0pt}!.\hspace{0pt} Such\hspace{0pt} pairs\hspace{0pt} are\hspace{0pt}%
 related\hspace{0pt} to\hspace{0pt} the\hspace{0pt} div\hspace{0pt}isors\hspace{0pt} of\hspace{0pt} \hspace{0pt}%
2\hspace{0pt}0\hspace{0pt}!\hspace{0pt} where\hspace{0pt} each\hspace{0pt} pair\hspace{0pt} consists\hspace{0pt}%
 of\hspace{0pt} two\hspace{0pt} cop\hspace{0pt}r\hspace{0pt}ime\hspace{0pt} numbers\hspace{0pt}.\hspace{0pt}%
But\hspace{0pt} since\hspace{0pt} the\hspace{0pt} fraction\hspace{0pt} is\hspace{0pt} between\hspace{0pt}%
 \hspace{0pt}0\hspace{0pt} and\hspace{0pt} \hspace{0pt}1\hspace{0pt},\hspace{0pt} \textbackslash{}(\hspace{0pt}%
a\hspace{0pt} \textless{}\hspace{0pt} b\hspace{0pt}\textbackslash{}\hspace{0pt}).\hspace{0pt} So\hspace{0pt}%
 for\hspace{0pt} each\hspace{0pt} such\hspace{0pt} cop\hspace{0pt}r\hspace{0pt}ime\hspace{0pt} pair\hspace{0pt}%
 where\hspace{0pt} \textbackslash{}(\hspace{0pt}a\hspace{0pt} \textless{}\hspace{0pt} b\hspace{0pt}\textbackslash{}\hspace{0pt}%
),\hspace{0pt} we\hspace{0pt} have\hspace{0pt} a\hspace{0pt} distinct\hspace{0pt} rational\hspace{0pt}%
 number\hspace{0pt},\hspace{0pt} and\hspace{0pt} each\hspace{0pt} of\hspace{0pt} these\hspace{0pt} would\hspace{0pt}%
 satisfy\hspace{0pt} \textbackslash{}(\hspace{0pt}a\hspace{0pt} \textbackslash{}\hspace{0pt}times\hspace{0pt}%
 b\hspace{0pt} =\hspace{0pt} \hspace{0pt}2\hspace{0pt}0\hspace{0pt}!\textbackslash{}\hspace{0pt}).\hspace{0pt}%
 Hence\hspace{0pt},\hspace{0pt} the\hspace{0pt} problem\hspace{0pt} reduces\hspace{0pt} to\hspace{0pt}%
 finding\hspace{0pt} the\hspace{0pt} number\hspace{0pt} of\hspace{0pt} cop\hspace{0pt}r\hspace{0pt}ime\hspace{0pt}%
 pairs\hspace{0pt} \textbackslash{}\hspace{0pt}((\hspace{0pt}a\hspace{0pt},\hspace{0pt} b\hspace{0pt}%
)\textbackslash{}\hspace{0pt})\hspace{0pt} where\hspace{0pt} \textbackslash{}(\hspace{0pt}a\hspace{0pt}%
 \textbackslash{}\hspace{0pt}times\hspace{0pt} b\hspace{0pt} =\hspace{0pt} \hspace{0pt}2\hspace{0pt}0\hspace{0pt}%
!\textbackslash{}\hspace{0pt})\hspace{0pt} and\hspace{0pt} \textbackslash{}(\hspace{0pt}a\hspace{0pt}%
 \textless{}\hspace{0pt} b\hspace{0pt}\textbackslash{}\hspace{0pt}).\hspace{0pt}Now\hspace{0pt},\hspace{0pt}%
 how\hspace{0pt} do\hspace{0pt} I\hspace{0pt} count\hspace{0pt} such\hspace{0pt} pairs\hspace{0pt}?\hspace{0pt}%
Alternatively\hspace{0pt},\hspace{0pt} in\hspace{0pt} number\hspace{0pt} theory\hspace{0pt},\hspace{0pt}%
 this\hspace{0pt} is\hspace{0pt} equivalent\hspace{0pt} to\hspace{0pt} counting\hspace{0pt} the\hspace{0pt}%
 number\hspace{0pt} of\hspace{0pt} ways\hspace{0pt} to\hspace{0pt} write\hspace{0pt} \hspace{0pt}2\hspace{0pt}%
0\hspace{0pt}!\hspace{0pt} as\hspace{0pt} a\hspace{0pt} product\hspace{0pt} of\hspace{0pt} two\hspace{0pt}%
 cop\hspace{0pt}r\hspace{0pt}ime\hspace{0pt} numbers\hspace{0pt},\hspace{0pt} \textbackslash{}(\hspace{0pt}%
a\hspace{0pt}\textbackslash{})\hspace{0pt} and\hspace{0pt} \textbackslash{}(\hspace{0pt}b\hspace{0pt}%
\textbackslash{}\hspace{0pt}),\hspace{0pt} with\hspace{0pt} \textbackslash{}(\hspace{0pt}a\hspace{0pt}%
 \textless{}\hspace{0pt} b\hspace{0pt}\textbackslash{}\hspace{0pt}).\hspace{0pt} Since\hspace{0pt} \textbackslash{}(\hspace{0pt}%
a\hspace{0pt}\textbackslash{})\hspace{0pt} and\hspace{0pt} \textbackslash{}(\hspace{0pt}b\hspace{0pt}%
\textbackslash{})\hspace{0pt} are\hspace{0pt} cop\hspace{0pt}r\hspace{0pt}ime\hspace{0pt},\hspace{0pt}%
 each\hspace{0pt} prime\hspace{0pt} factor\hspace{0pt} in\hspace{0pt} \hspace{0pt}2\hspace{0pt}0\hspace{0pt}%
!\hspace{0pt} must\hspace{0pt} go\hspace{0pt} entirely\hspace{0pt} to\hspace{0pt} \textbackslash{}(\hspace{0pt}%
a\hspace{0pt}\textbackslash{})\hspace{0pt} or\hspace{0pt} entirely\hspace{0pt} to\hspace{0pt} \textbackslash{}(\hspace{0pt}%
b\hspace{0pt}\textbackslash{}\hspace{0pt}).\hspace{0pt} Hence\hspace{0pt},\hspace{0pt} each\hspace{0pt}%
 cop\hspace{0pt}r\hspace{0pt}ime\hspace{0pt} pair\hspace{0pt} corresponds\hspace{0pt} to\hspace{0pt} a\hspace{0pt}%
 subset\hspace{0pt} of\hspace{0pt} the\hspace{0pt} prime\hspace{0pt} factors\hspace{0pt} of\hspace{0pt}%
 \hspace{0pt}2\hspace{0pt}0\hspace{0pt}!,\hspace{0pt} where\hspace{0pt} \textbackslash{}(\hspace{0pt}%
a\hspace{0pt}\textbackslash{})\hspace{0pt} is\hspace{0pt} the\hspace{0pt} product\hspace{0pt} of\hspace{0pt}%
 the\hspace{0pt} primes\hspace{0pt} in\hspace{0pt} the\hspace{0pt} subset\hspace{0pt} and\hspace{0pt}%
 \textbackslash{}(\hspace{0pt}b\hspace{0pt}\textbackslash{})\hspace{0pt} is\hspace{0pt} the\hspace{0pt}%
 product\hspace{0pt} of\hspace{0pt} the\hspace{0pt} remaining\hspace{0pt} primes\hspace{0pt}.\hspace{0pt}%
 However\hspace{0pt},\hspace{0pt} since\hspace{0pt} \hspace{0pt}2\hspace{0pt}0\hspace{0pt}!\hspace{0pt}%
 is\hspace{0pt} not\hspace{0pt} just\hspace{0pt} the\hspace{0pt} product\hspace{0pt} of\hspace{0pt} primes\hspace{0pt}%
,\hspace{0pt} but\hspace{0pt} the\hspace{0pt} product\hspace{0pt} of\hspace{0pt} primes\hspace{0pt} raised\hspace{0pt}%
 to\hspace{0pt} various\hspace{0pt} powers\hspace{0pt}.\hspace{0pt} Wait\hspace{0pt},\hspace{0pt} so\hspace{0pt}%
 maybe\hspace{0pt} I\hspace{0pt} need\hspace{0pt} to\hspace{0pt} rethink\hspace{0pt} that\hspace{0pt}%
.\hspace{0pt}Wait\hspace{0pt},\hspace{0pt} actually\hspace{0pt} \hspace{0pt}2\hspace{0pt}0\hspace{0pt}%
!\hspace{0pt} has\hspace{0pt} prime\hspace{0pt} factors\hspace{0pt} with\hspace{0pt} ex\hspace{0pt}ponents\hspace{0pt}%
.\hspace{0pt} For\hspace{0pt} example\hspace{0pt},\hspace{0pt} \hspace{0pt}2\hspace{0pt}\^{}\hspace{0pt}%
1\hspace{0pt}8\hspace{0pt},\hspace{0pt} \hspace{0pt}3\hspace{0pt}\^{}\hspace{0pt}8\hspace{0pt},\hspace{0pt}%
 \hspace{0pt}5\hspace{0pt}\^{}\hspace{0pt}4\hspace{0pt},\hspace{0pt} etc\hspace{0pt}.,\hspace{0pt} up\hspace{0pt}%
 to\hspace{0pt} the\hspace{0pt} primes\hspace{0pt} less\hspace{0pt} than\hspace{0pt} or\hspace{0pt} equal\hspace{0pt}%
 to\hspace{0pt} \hspace{0pt}2\hspace{0pt}0\hspace{0pt}.\hspace{0pt} But\hspace{0pt} if\hspace{0pt} I\hspace{0pt}%
 need\hspace{0pt} two\hspace{0pt} cop\hspace{0pt}r\hspace{0pt}ime\hspace{0pt} numbers\hspace{0pt} \textbackslash{}(\hspace{0pt}%
a\hspace{0pt}\textbackslash{})\hspace{0pt} and\hspace{0pt} \textbackslash{}(\hspace{0pt}b\hspace{0pt}%
\textbackslash{})\hspace{0pt} such\hspace{0pt} that\hspace{0pt} \textbackslash{}(\hspace{0pt}a\hspace{0pt}%
 \textbackslash{}\hspace{0pt}times\hspace{0pt} b\hspace{0pt} =\hspace{0pt} \hspace{0pt}2\hspace{0pt}0\hspace{0pt}%
!\textbackslash{}\hspace{0pt}),\hspace{0pt} then\hspace{0pt} each\hspace{0pt} of\hspace{0pt} the\hspace{0pt}%
 prime\hspace{0pt} powers\hspace{0pt} in\hspace{0pt} the\hspace{0pt} factor\hspace{0pt}ization\hspace{0pt}%
 of\hspace{0pt} \hspace{0pt}2\hspace{0pt}0\hspace{0pt}!\hspace{0pt} must\hspace{0pt} go\hspace{0pt} entirely\hspace{0pt}%
 to\hspace{0pt} one\hspace{0pt} of\hspace{0pt} the\hspace{0pt} numbers\hspace{0pt} \textbackslash{}(\hspace{0pt}%
a\hspace{0pt}\textbackslash{})\hspace{0pt} or\hspace{0pt} \textbackslash{}(\hspace{0pt}b\hspace{0pt}\textbackslash{}\hspace{0pt}%
).\hspace{0pt} So\hspace{0pt} for\hspace{0pt} each\hspace{0pt} prime\hspace{0pt} factor\hspace{0pt},\hspace{0pt}%
 we\hspace{0pt} assign\hspace{0pt} it\hspace{0pt} to\hspace{0pt} either\hspace{0pt} \textbackslash{}(\hspace{0pt}%
a\hspace{0pt}\textbackslash{})\hspace{0pt} or\hspace{0pt} \textbackslash{}(\hspace{0pt}b\hspace{0pt}\textbackslash{}\hspace{0pt}%
),\hspace{0pt} but\hspace{0pt} not\hspace{0pt} both\hspace{0pt}.\hspace{0pt}Therefore\hspace{0pt},\hspace{0pt}%
 the\hspace{0pt} number\hspace{0pt} of\hspace{0pt} such\hspace{0pt} cop\hspace{0pt}r\hspace{0pt}ime\hspace{0pt}%
 pairs\hspace{0pt} is\hspace{0pt} \textbackslash{}(\hspace{0pt}2\hspace{0pt}\^{}\hspace{0pt}k\hspace{0pt}%
\textbackslash{}\hspace{0pt}),\hspace{0pt} where\hspace{0pt} \textbackslash{}(\hspace{0pt}k\hspace{0pt}%
\textbackslash{})\hspace{0pt} is\hspace{0pt} the\hspace{0pt} number\hspace{0pt} of\hspace{0pt} distinct\hspace{0pt}%
 prime\hspace{0pt} factors\hspace{0pt} of\hspace{0pt} \hspace{0pt}2\hspace{0pt}0\hspace{0pt}!.\hspace{0pt}%
But\hspace{0pt} wait\hspace{0pt},\hspace{0pt} \hspace{0pt}2\hspace{0pt}0\hspace{0pt}!\hspace{0pt} has\hspace{0pt}%
 prime\hspace{0pt} factors\hspace{0pt}:\hspace{0pt} \hspace{0pt}2\hspace{0pt},\hspace{0pt} \hspace{0pt}%
3\hspace{0pt},\hspace{0pt} \hspace{0pt}5\hspace{0pt},\hspace{0pt} \hspace{0pt}7\hspace{0pt},\hspace{0pt}%
 \hspace{0pt}1\hspace{0pt}1\hspace{0pt},\hspace{0pt} \hspace{0pt}1\hspace{0pt}3\hspace{0pt},\hspace{0pt}%
 \hspace{0pt}1\hspace{0pt}7\hspace{0pt},\hspace{0pt} \hspace{0pt}1\hspace{0pt}9\hspace{0pt}.\hspace{0pt}%
 Let\hspace{0pt} me\hspace{0pt} check\hspace{0pt}.\hspace{0pt} The\hspace{0pt} primes\hspace{0pt} less\hspace{0pt}%
 than\hspace{0pt} or\hspace{0pt} equal\hspace{0pt} to\hspace{0pt} \hspace{0pt}2\hspace{0pt}0\hspace{0pt}%
 are\hspace{0pt}:\hspace{0pt} \hspace{0pt}2\hspace{0pt},\hspace{0pt} \hspace{0pt}3\hspace{0pt},\hspace{0pt}%
 \hspace{0pt}5\hspace{0pt},\hspace{0pt} \hspace{0pt}7\hspace{0pt},\hspace{0pt} \hspace{0pt}1\hspace{0pt}%
1\hspace{0pt},\hspace{0pt} \hspace{0pt}1\hspace{0pt}3\hspace{0pt},\hspace{0pt} \hspace{0pt}1\hspace{0pt}%
7\hspace{0pt},\hspace{0pt} \hspace{0pt}1\hspace{0pt}9\hspace{0pt}.\hspace{0pt} So\hspace{0pt} that\hspace{0pt}%
's\hspace{0pt} \hspace{0pt}8\hspace{0pt} primes\hspace{0pt}.\hspace{0pt} Therefore\hspace{0pt},\hspace{0pt}%
 the\hspace{0pt} number\hspace{0pt} of\hspace{0pt} cop\hspace{0pt}r\hspace{0pt}ime\hspace{0pt} pairs\hspace{0pt}%
 would\hspace{0pt} be\hspace{0pt} \textbackslash{}(\hspace{0pt}2\hspace{0pt}\^{}\hspace{0pt}8\hspace{0pt}%
 =\hspace{0pt} \hspace{0pt}2\hspace{0pt}5\hspace{0pt}6\hspace{0pt}\textbackslash{}\hspace{0pt}),\hspace{0pt}%
 but\hspace{0pt} each\hspace{0pt} pair\hspace{0pt} is\hspace{0pt} counted\hspace{0pt} twice\hspace{0pt}%
 except\hspace{0pt} when\hspace{0pt} \textbackslash{}(\hspace{0pt}a\hspace{0pt} =\hspace{0pt} b\hspace{0pt}%
\textbackslash{}\hspace{0pt}).\hspace{0pt} But\hspace{0pt} since\hspace{0pt} \textbackslash{}(\hspace{0pt}%
a\hspace{0pt} \textbackslash{}\hspace{0pt}times\hspace{0pt} b\hspace{0pt} =\hspace{0pt} \hspace{0pt}2\hspace{0pt}%
0\hspace{0pt}!\textbackslash{}\hspace{0pt})\hspace{0pt} and\hspace{0pt} \textbackslash{}(\hspace{0pt}%
a\hspace{0pt}\textbackslash{})\hspace{0pt} and\hspace{0pt} \textbackslash{}(\hspace{0pt}b\hspace{0pt}%
\textbackslash{})\hspace{0pt} are\hspace{0pt} cop\hspace{0pt}r\hspace{0pt}ime\hspace{0pt},\hspace{0pt}%
 \textbackslash{}(\hspace{0pt}a\hspace{0pt} =\hspace{0pt} \hspace{0pt}1\hspace{0pt}\textbackslash{})\hspace{0pt}%
 and\hspace{0pt} \textbackslash{}(\hspace{0pt}b\hspace{0pt} =\hspace{0pt} \hspace{0pt}2\hspace{0pt}0\hspace{0pt}%
!\textbackslash{}\hspace{0pt}),\hspace{0pt} or\hspace{0pt} vice\hspace{0pt} versa\hspace{0pt};\hspace{0pt}%
 however\hspace{0pt},\hspace{0pt} \textbackslash{}(\hspace{0pt}a\hspace{0pt}\textbackslash{})\hspace{0pt}%
 and\hspace{0pt} \textbackslash{}(\hspace{0pt}b\hspace{0pt}\textbackslash{})\hspace{0pt} can\hspace{0pt}%
't\hspace{0pt} be\hspace{0pt} equal\hspace{0pt} unless\hspace{0pt} \hspace{0pt}2\hspace{0pt}0\hspace{0pt}%
!\hspace{0pt} is\hspace{0pt} a\hspace{0pt} square\hspace{0pt},\hspace{0pt} which\hspace{0pt} it\hspace{0pt}%
 is\hspace{0pt} not\hspace{0pt} (\hspace{0pt}since\hspace{0pt} ex\hspace{0pt}ponents\hspace{0pt} in\hspace{0pt}%
 its\hspace{0pt} prime\hspace{0pt} factor\hspace{0pt}ization\hspace{0pt} are\hspace{0pt} all\hspace{0pt}%
 even\hspace{0pt}?\hspace{0pt} Wait\hspace{0pt},\hspace{0pt} \hspace{0pt}2\hspace{0pt}0\hspace{0pt}!\hspace{0pt}%
 has\hspace{0pt} ex\hspace{0pt}ponents\hspace{0pt}:\hspace{0pt}Let\hspace{0pt} me\hspace{0pt} do\hspace{0pt}%
 prime\hspace{0pt} factor\hspace{0pt}ization\hspace{0pt} of\hspace{0pt} \hspace{0pt}2\hspace{0pt}0\hspace{0pt}%
!.\hspace{0pt}To\hspace{0pt} compute\hspace{0pt} the\hspace{0pt} ex\hspace{0pt}ponents\hspace{0pt} of\hspace{0pt}%
 primes\hspace{0pt} in\hspace{0pt} \hspace{0pt}2\hspace{0pt}0\hspace{0pt}!\hspace{0pt}:\hspace{0pt}For\hspace{0pt}%
 prime\hspace{0pt} \hspace{0pt}2\hspace{0pt}:\hspace{0pt} floor\hspace{0pt}(\hspace{0pt}2\hspace{0pt}%
0\hspace{0pt}/\hspace{0pt}2\hspace{0pt})\hspace{0pt} +\hspace{0pt} floor\hspace{0pt}(\hspace{0pt}2\hspace{0pt}%
0\hspace{0pt}/\hspace{0pt}4\hspace{0pt})\hspace{0pt} +\hspace{0pt} floor\hspace{0pt}(\hspace{0pt}2\hspace{0pt}%
0\hspace{0pt}/\hspace{0pt}8\hspace{0pt})\hspace{0pt} +\hspace{0pt} floor\hspace{0pt}(\hspace{0pt}2\hspace{0pt}%
0\hspace{0pt}/\hspace{0pt}1\hspace{0pt}6\hspace{0pt})\hspace{0pt} =\hspace{0pt}1\hspace{0pt}0\hspace{0pt}%
+\hspace{0pt}5\hspace{0pt}+\hspace{0pt}2\hspace{0pt}+\hspace{0pt}1\hspace{0pt}=\hspace{0pt}1\hspace{0pt}%
8\hspace{0pt}\hspace{0pt}For\hspace{0pt} prime\hspace{0pt} \hspace{0pt}3\hspace{0pt}:\hspace{0pt} floor\hspace{0pt}%
(\hspace{0pt}2\hspace{0pt}0\hspace{0pt}/\hspace{0pt}3\hspace{0pt})\hspace{0pt} +\hspace{0pt} floor\hspace{0pt}%
(\hspace{0pt}2\hspace{0pt}0\hspace{0pt}/\hspace{0pt}9\hspace{0pt})\hspace{0pt} +\hspace{0pt} floor\hspace{0pt}%
(\hspace{0pt}2\hspace{0pt}0\hspace{0pt}/\hspace{0pt}2\hspace{0pt}7\hspace{0pt})=\hspace{0pt}6\hspace{0pt}%
 +\hspace{0pt}2\hspace{0pt} +\hspace{0pt}0\hspace{0pt}=\hspace{0pt}8\hspace{0pt}\hspace{0pt}For\hspace{0pt}%
 prime\hspace{0pt} \hspace{0pt}5\hspace{0pt}:\hspace{0pt} floor\hspace{0pt}(\hspace{0pt}2\hspace{0pt}%
0\hspace{0pt}/\hspace{0pt}5\hspace{0pt})\hspace{0pt} +\hspace{0pt} floor\hspace{0pt}(\hspace{0pt}2\hspace{0pt}%
0\hspace{0pt}/\hspace{0pt}2\hspace{0pt}5\hspace{0pt})=\hspace{0pt}4\hspace{0pt} +\hspace{0pt}0\hspace{0pt}%
=\hspace{0pt}4\hspace{0pt}\hspace{0pt}Prime\hspace{0pt} \hspace{0pt}7\hspace{0pt}:\hspace{0pt} floor\hspace{0pt}%
(\hspace{0pt}2\hspace{0pt}0\hspace{0pt}/\hspace{0pt}7\hspace{0pt})\hspace{0pt} +\hspace{0pt} floor\hspace{0pt}%
(\hspace{0pt}2\hspace{0pt}0\hspace{0pt}/\hspace{0pt}4\hspace{0pt}9\hspace{0pt})=\hspace{0pt}2\hspace{0pt}%
 +\hspace{0pt}0\hspace{0pt}=\hspace{0pt}2\hspace{0pt}\hspace{0pt}Pr\hspace{0pt}imes\hspace{0pt} \hspace{0pt}%
1\hspace{0pt}1\hspace{0pt}:\hspace{0pt} floor\hspace{0pt}(\hspace{0pt}2\hspace{0pt}0\hspace{0pt}/\hspace{0pt}%
1\hspace{0pt}1\hspace{0pt})=\hspace{0pt}1\hspace{0pt},\hspace{0pt} similar\hspace{0pt} for\hspace{0pt}%
 \hspace{0pt}1\hspace{0pt}3\hspace{0pt},\hspace{0pt} \hspace{0pt}1\hspace{0pt}7\hspace{0pt},\hspace{0pt}%
 \hspace{0pt}1\hspace{0pt}9\hspace{0pt}.\hspace{0pt} So\hspace{0pt} yes\hspace{0pt},\hspace{0pt} primes\hspace{0pt}%
 \hspace{0pt}1\hspace{0pt}1\hspace{0pt},\hspace{0pt} \hspace{0pt}1\hspace{0pt}3\hspace{0pt},\hspace{0pt}%
 \hspace{0pt}1\hspace{0pt}7\hspace{0pt},\hspace{0pt} \hspace{0pt}1\hspace{0pt}9\hspace{0pt} each\hspace{0pt}%
 have\hspace{0pt} exponent\hspace{0pt} \hspace{0pt}1\hspace{0pt} in\hspace{0pt} \hspace{0pt}2\hspace{0pt}%
0\hspace{0pt}!.\hspace{0pt}So\hspace{0pt} ex\hspace{0pt}ponents\hspace{0pt} are\hspace{0pt}:\hspace{0pt}%
 \hspace{0pt}1\hspace{0pt}8\hspace{0pt} (\hspace{0pt}2\hspace{0pt}),\hspace{0pt} \hspace{0pt}8\hspace{0pt}%
 (\hspace{0pt}3\hspace{0pt}),\hspace{0pt} \hspace{0pt}4\hspace{0pt} (\hspace{0pt}5\hspace{0pt}),\hspace{0pt}%
 \hspace{0pt}2\hspace{0pt} (\hspace{0pt}7\hspace{0pt}),\hspace{0pt} \hspace{0pt}1\hspace{0pt} (\hspace{0pt}%
1\hspace{0pt}1\hspace{0pt}),\hspace{0pt} \hspace{0pt}1\hspace{0pt} (\hspace{0pt}1\hspace{0pt}3\hspace{0pt}%
),\hspace{0pt} \hspace{0pt}1\hspace{0pt} (\hspace{0pt}1\hspace{0pt}7\hspace{0pt}),\hspace{0pt} \hspace{0pt}%
1\hspace{0pt} (\hspace{0pt}1\hspace{0pt}9\hspace{0pt})\hspace{0pt}\heatbox{0.980,0.622,0.622}{Since}\hspace{0pt}%
\heatbox{0.980,0.622,0.622}{ all}\hspace{0pt}\heatbox{0.980,0.622,0.622}{ ex}\hspace{0pt}\heatbox{0.980,0.622,0.622}{ponents}\hspace{0pt}%
\heatbox{0.980,0.622,0.622}{ except}\hspace{0pt}\heatbox{0.980,0.622,0.622}{ for}\hspace{0pt}\heatbox{0.980,0.622,0.622}{ }\hspace{0pt}%
\heatbox{0.980,0.622,0.622}{2}\hspace{0pt}\heatbox{0.980,0.622,0.622}{,}\hspace{0pt}\heatbox{0.980,0.622,0.622}{3}\hspace{0pt}%
\heatbox{0.980,0.622,0.622}{,}\hspace{0pt}\heatbox{0.980,0.622,0.622}{5}\hspace{0pt}\heatbox{0.980,0.622,0.622}{,}\hspace{0pt}%
\heatbox{0.980,0.622,0.622}{7}\hspace{0pt}\heatbox{0.980,0.622,0.622}{ have}\hspace{0pt}\heatbox{0.980,0.622,0.622}{ exponent}\hspace{0pt}%
\heatbox{0.980,0.622,0.622}{ }\hspace{0pt}\heatbox{0.980,0.622,0.622}{1}\hspace{0pt}\heatbox{0.980,0.622,0.622}{,}\hspace{0pt}%
\heatbox{0.980,0.622,0.622}{ and}\hspace{0pt}\heatbox{0.980,0.622,0.622}{ ex}\hspace{0pt}\heatbox{0.980,0.622,0.622}{ponents}\hspace{0pt}%
\heatbox{0.980,0.622,0.622}{ for}\hspace{0pt}\heatbox{0.980,0.622,0.622}{ }\hspace{0pt}\heatbox{0.980,0.622,0.622}{2}\hspace{0pt}%
\heatbox{0.980,0.622,0.622}{,}\hspace{0pt}\heatbox{0.980,0.622,0.622}{3}\hspace{0pt}\heatbox{0.980,0.622,0.622}{,}\hspace{0pt}%
\heatbox{0.980,0.622,0.622}{5}\hspace{0pt}\heatbox{0.980,0.622,0.622}{,}\hspace{0pt}\heatbox{0.980,0.622,0.622}{7}\hspace{0pt}%
\heatbox{0.980,0.622,0.622}{ are}\hspace{0pt}\heatbox{0.980,0.622,0.622}{ even}\hspace{0pt}\heatbox{0.980,0.622,0.622}{ and}\hspace{0pt}%
\heatbox{0.980,0.622,0.622}{ odd}\hspace{0pt}.\hspace{0pt}Therefore\hspace{0pt},\hspace{0pt} \hspace{0pt}%
2\hspace{0pt}0\hspace{0pt}!\hspace{0pt} is\hspace{0pt} not\hspace{0pt} a\hspace{0pt} perfect\hspace{0pt}%
 square\hspace{0pt} because\hspace{0pt},\hspace{0pt} for\hspace{0pt} example\hspace{0pt},\hspace{0pt}%
 exponent\hspace{0pt} of\hspace{0pt} \hspace{0pt}3\hspace{0pt} is\hspace{0pt} \hspace{0pt}8\hspace{0pt}%
 (\hspace{0pt}even\hspace{0pt}),\hspace{0pt} exponent\hspace{0pt} of\hspace{0pt} \hspace{0pt}2\hspace{0pt}%
 is\hspace{0pt} \hspace{0pt}1\hspace{0pt}8\hspace{0pt} (\hspace{0pt}even\hspace{0pt}),\hspace{0pt} but\hspace{0pt}%
 exponent\hspace{0pt} of\hspace{0pt} \hspace{0pt}5\hspace{0pt} is\hspace{0pt} \hspace{0pt}4\hspace{0pt}%
 (\hspace{0pt}even\hspace{0pt}),\hspace{0pt} \hspace{0pt}2\hspace{0pt} (\hspace{0pt}7\hspace{0pt})\hspace{0pt}%
 (\hspace{0pt}even\hspace{0pt}?),\hspace{0pt} no\hspace{0pt} \hspace{0pt}7\hspace{0pt} has\hspace{0pt}%
 exponent\hspace{0pt} \hspace{0pt}2\hspace{0pt}?\hspace{0pt} Wait\hspace{0pt} for\hspace{0pt} \hspace{0pt}%
7\hspace{0pt}:\hspace{0pt} floor\hspace{0pt}(\hspace{0pt}2\hspace{0pt}0\hspace{0pt}/\hspace{0pt}7\hspace{0pt}%
)=\hspace{0pt}2\hspace{0pt} (\hspace{0pt}floor\hspace{0pt}(\hspace{0pt}2\hspace{0pt}0\hspace{0pt}/\hspace{0pt}%
4\hspace{0pt}9\hspace{0pt})\hspace{0pt} =\hspace{0pt} \hspace{0pt}0\hspace{0pt}),\hspace{0pt} so\hspace{0pt}%
 that\hspace{0pt}'s\hspace{0pt} exponent\hspace{0pt} \hspace{0pt}2\hspace{0pt}.\hspace{0pt}Wait\hspace{0pt}%
,\hspace{0pt} \hspace{0pt}7\hspace{0pt}\^{}\hspace{0pt}2\hspace{0pt} divides\hspace{0pt} into\hspace{0pt}%
 \hspace{0pt}2\hspace{0pt}0\hspace{0pt}!\hspace{0pt} but\hspace{0pt} \hspace{0pt}7\hspace{0pt}\^{}\hspace{0pt}%
3\hspace{0pt} does\hspace{0pt} not\hspace{0pt}.\hspace{0pt}But\hspace{0pt} exponent\hspace{0pt} of\hspace{0pt}%
 \hspace{0pt}7\hspace{0pt} is\hspace{0pt} \hspace{0pt}2\hspace{0pt},\hspace{0pt} which\hspace{0pt} is\hspace{0pt}%
 even\hspace{0pt}.\hspace{0pt} Wait\hspace{0pt} a\hspace{0pt} second\hspace{0pt}.\hspace{0pt} Then\hspace{0pt}%
 primes\hspace{0pt} \hspace{0pt}1\hspace{0pt}1\hspace{0pt},\hspace{0pt} \hspace{0pt}1\hspace{0pt}3\hspace{0pt}%
,\hspace{0pt}1\hspace{0pt}7\hspace{0pt},\hspace{0pt}1\hspace{0pt}9\hspace{0pt} have\hspace{0pt} exponent\hspace{0pt}%
 \hspace{0pt}1\hspace{0pt}.\hspace{0pt}So\hspace{0pt} primes\hspace{0pt} \hspace{0pt}2\hspace{0pt}:\hspace{0pt}%
 \hspace{0pt}1\hspace{0pt}8\hspace{0pt},\hspace{0pt} \hspace{0pt}3\hspace{0pt}:\hspace{0pt} \hspace{0pt}%
8\hspace{0pt},\hspace{0pt} \hspace{0pt}5\hspace{0pt}:\hspace{0pt}4\hspace{0pt},\hspace{0pt}7\hspace{0pt}%
:\hspace{0pt}2\hspace{0pt},\hspace{0pt} which\hspace{0pt} are\hspace{0pt} all\hspace{0pt} even\hspace{0pt}%
?\hspace{0pt} Wait\hspace{0pt},\hspace{0pt} \hspace{0pt}1\hspace{0pt}8\hspace{0pt},\hspace{0pt}8\hspace{0pt}%
,\hspace{0pt}4\hspace{0pt},\hspace{0pt}2\hspace{0pt},\hspace{0pt} which\hspace{0pt} are\hspace{0pt} even\hspace{0pt}%
.\hspace{0pt} Wait\hspace{0pt},\hspace{0pt} that\hspace{0pt} would\hspace{0pt} mean\hspace{0pt} the\hspace{0pt}%
 only\hspace{0pt} primes\hspace{0pt} with\hspace{0pt} odd\hspace{0pt} ex\hspace{0pt}ponents\hspace{0pt}%
 are\hspace{0pt} \hspace{0pt}1\hspace{0pt}1\hspace{0pt},\hspace{0pt}1\hspace{0pt}3\hspace{0pt},\hspace{0pt}%
1\hspace{0pt}7\hspace{0pt},\hspace{0pt}1\hspace{0pt}9\hspace{0pt}?\hspace{0pt} Wait\hspace{0pt},\hspace{0pt}%
 but\hspace{0pt} their\hspace{0pt} ex\hspace{0pt}ponents\hspace{0pt} are\hspace{0pt} \hspace{0pt}1\hspace{0pt}%
,\hspace{0pt} which\hspace{0pt} is\hspace{0pt} odd\hspace{0pt}.\hspace{0pt} 
 
 ...(Omit)...
 \hspace{0pt} However\hspace{0pt},\hspace{0pt}%
 because\hspace{0pt} \hspace{0pt}2\hspace{0pt}0\hspace{0pt}!\hspace{0pt} is\hspace{0pt} not\hspace{0pt}%
 a\hspace{0pt} square\hspace{0pt} (\hspace{0pt}due\hspace{0pt} to\hspace{0pt} the\hspace{0pt} primes\hspace{0pt}%
 with\hspace{0pt} ex\hspace{0pt}ponents\hspace{0pt} as\hspace{0pt} discussed\hspace{0pt} earlier\hspace{0pt}%
),\hspace{0pt} there\hspace{0pt} are\hspace{0pt} no\hspace{0pt} pairs\hspace{0pt} where\hspace{0pt} a\hspace{0pt}%
 =\hspace{0pt} b\hspace{0pt}.\hspace{0pt} Hence\hspace{0pt},\hspace{0pt} each\hspace{0pt} unordered\hspace{0pt}%
 pair\hspace{0pt} \{\hspace{0pt}a\hspace{0pt},\hspace{0pt} b\hspace{0pt}\}\hspace{0pt} is\hspace{0pt}%
 counted\hspace{0pt} twice\hspace{0pt} in\hspace{0pt} the\hspace{0pt} \hspace{0pt}2\hspace{0pt}5\hspace{0pt}%
6\hspace{0pt} ordered\hspace{0pt} pairs\hspace{0pt}.\hspace{0pt}But\hspace{0pt} in\hspace{0pt} our\hspace{0pt}%
 problem\hspace{0pt},\hspace{0pt} we\hspace{0pt} need\hspace{0pt} the\hspace{0pt} ordered\hspace{0pt}%
 pairs\hspace{0pt} (\hspace{0pt}a\hspace{0pt},\hspace{0pt} b\hspace{0pt})\hspace{0pt} where\hspace{0pt}%
 a\hspace{0pt} \textless{}\hspace{0pt} b\hspace{0pt},\hspace{0pt} which\hspace{0pt} is\hspace{0pt} exactly\hspace{0pt}%
 half\hspace{0pt} of\hspace{0pt} the\hspace{0pt} total\hspace{0pt} ordered\hspace{0pt} pairs\hspace{0pt}%
 (\hspace{0pt}excluding\hspace{0pt} the\hspace{0pt} a\hspace{0pt} =\hspace{0pt} b\hspace{0pt} case\hspace{0pt}%
,\hspace{0pt} which\hspace{0pt} does\hspace{0pt} not\hspace{0pt} exist\hspace{0pt} here\hspace{0pt}).\hspace{0pt}%
 Therefore\hspace{0pt},\hspace{0pt} the\hspace{0pt} total\hspace{0pt} number\hspace{0pt} should\hspace{0pt}%
 be\hspace{0pt} \hspace{0pt}2\hspace{0pt}5\hspace{0pt}6\hspace{0pt} /\hspace{0pt} \hspace{0pt}2\hspace{0pt}%
 =\hspace{0pt} \hspace{0pt}1\hspace{0pt}2\hspace{0pt}8\hspace{0pt}.\hspace{0pt} But\hspace{0pt} hold\hspace{0pt}%
 on\hspace{0pt},\hspace{0pt} is\hspace{0pt} this\hspace{0pt} correct\hspace{0pt}?\hspace{0pt}\heatbox{0.984,0.698,0.698}{Wait}\hspace{0pt}%
\heatbox{0.984,0.698,0.698}{,}\hspace{0pt}\heatbox{0.984,0.698,0.698}{ perhaps}\hspace{0pt}\heatbox{0.984,0.698,0.698}{ not}\hspace{0pt}%
\heatbox{0.984,0.698,0.698}{ quite}\hspace{0pt}\heatbox{0.984,0.698,0.698}{.}\hspace{0pt}\heatbox{0.984,0.698,0.698}{ Since}\hspace{0pt}%
\heatbox{0.984,0.698,0.698}{ not}\hspace{0pt}\heatbox{0.984,0.698,0.698}{ all}\hspace{0pt}\heatbox{0.984,0.698,0.698}{ assignments}\hspace{0pt}%
\heatbox{0.984,0.698,0.698}{ will}\hspace{0pt}\heatbox{0.984,0.698,0.698}{ satisfy}\hspace{0pt}\heatbox{0.984,0.698,0.698}{ a}\hspace{0pt}%
\heatbox{0.984,0.698,0.698}{ \textless{}}\hspace{0pt}\heatbox{0.984,0.698,0.698}{ b}\hspace{0pt}\heatbox{0.984,0.698,0.698}{,}\hspace{0pt}%
\heatbox{0.984,0.698,0.698}{ even}\hspace{0pt}\heatbox{0.984,0.698,0.698}{ for}\hspace{0pt}\heatbox{0.984,0.698,0.698}{ cop}\hspace{0pt}%
\heatbox{0.984,0.698,0.698}{r}\hspace{0pt}\heatbox{0.984,0.698,0.698}{ime}\hspace{0pt}\heatbox{0.984,0.698,0.698}{ div}\hspace{0pt}%
\heatbox{0.984,0.698,0.698}{isors}\hspace{0pt}\heatbox{0.984,0.698,0.698}{.}\hspace{0pt}\heatbox{0.984,0.698,0.698}{ Wait}\hspace{0pt}%
\heatbox{0.984,0.698,0.698}{,}\hspace{0pt}\heatbox{0.984,0.698,0.698}{ but}\hspace{0pt}\heatbox{0.984,0.698,0.698}{ in}\hspace{0pt}%
\heatbox{0.984,0.698,0.698}{ reality}\hspace{0pt}\heatbox{0.984,0.698,0.698}{,}\hspace{0pt}\heatbox{0.984,0.698,0.698}{ since}\hspace{0pt}%
\heatbox{0.984,0.698,0.698}{ a}\hspace{0pt}\heatbox{0.984,0.698,0.698}{ is}\hspace{0pt}\heatbox{0.984,0.698,0.698}{ a}\hspace{0pt}%
\heatbox{0.984,0.698,0.698}{ number}\hspace{0pt}\heatbox{0.984,0.698,0.698}{ composed}\hspace{0pt}\heatbox{0.984,0.698,0.698}{ by}\hspace{0pt}%
\heatbox{0.984,0.698,0.698}{ assigning}\hspace{0pt}\heatbox{0.984,0.698,0.698}{ primes}\hspace{0pt}\heatbox{0.984,0.698,0.698}{ to}\hspace{0pt}%
\heatbox{0.984,0.698,0.698}{ a}\hspace{0pt}\heatbox{0.984,0.698,0.698}{ (}\hspace{0pt}\heatbox{0.984,0.698,0.698}{possibly}\hspace{0pt}%
\heatbox{0.984,0.698,0.698}{ a}\hspace{0pt}\heatbox{0.984,0.698,0.698}{ =}\hspace{0pt}\heatbox{0.984,0.698,0.698}{ product}\hspace{0pt}%
\heatbox{0.984,0.698,0.698}{ of}\hspace{0pt}\heatbox{0.984,0.698,0.698}{ primes}\hspace{0pt}\heatbox{0.984,0.698,0.698}{ raised}\hspace{0pt}%
\heatbox{0.984,0.698,0.698}{ to}\hspace{0pt}\heatbox{0.984,0.698,0.698}{ their}\hspace{0pt}\heatbox{0.984,0.698,0.698}{ ex}\hspace{0pt}%
\heatbox{0.984,0.698,0.698}{ponents}\hspace{0pt}\heatbox{0.984,0.698,0.698}{)..}\hspace{0pt}\heatbox{0.984,0.698,0.698}{ Wait}\hspace{0pt}%
\heatbox{0.984,0.698,0.698}{,}\hspace{0pt}\heatbox{0.984,0.698,0.698}{ confusion}\hspace{0pt}\heatbox{0.984,0.698,0.698}{ arises}\hspace{0pt}%
\heatbox{0.984,0.698,0.698}{ here}\hspace{0pt}.\hspace{0pt}Wait\hspace{0pt},\hspace{0pt} actually\hspace{0pt}%
,\hspace{0pt} we\hspace{0pt} need\hspace{0pt} to\hspace{0pt} be\hspace{0pt} precise\hspace{0pt}.\hspace{0pt}%
 When\hspace{0pt} we\hspace{0pt} say\hspace{0pt} that\hspace{0pt} a\hspace{0pt} and\hspace{0pt} b\hspace{0pt}%
 are\hspace{0pt} cop\hspace{0pt}r\hspace{0pt}ime\hspace{0pt} with\hspace{0pt} a\hspace{0pt}*b\hspace{0pt}%
=\hspace{0pt}2\hspace{0pt}0\hspace{0pt}!,\hspace{0pt} then\hspace{0pt} a\hspace{0pt} and\hspace{0pt} b\hspace{0pt}%
 must\hspace{0pt} be\hspace{0pt} square\hspace{0pt}-free\hspace{0pt}?\hspace{0pt} No\hspace{0pt},\hspace{0pt}%
 that\hspace{0pt}'s\hspace{0pt} not\hspace{0pt} correct\hspace{0pt}.\hspace{0pt} Each\hspace{0pt} must\hspace{0pt}%
 have\hspace{0pt} all\hspace{0pt} primes\hspace{0pt} ex\hspace{0pt}ponents\hspace{0pt} from\hspace{0pt}%
 \hspace{0pt}2\hspace{0pt}0\hspace{0pt}!\hspace{0pt} For\hspace{0pt} example\hspace{0pt},\hspace{0pt}%
 if\hspace{0pt} I\hspace{0pt} assign\hspace{0pt} prime\hspace{0pt} \hspace{0pt}2\hspace{0pt} to\hspace{0pt}%
 a\hspace{0pt},\hspace{0pt} then\hspace{0pt} a\hspace{0pt} would\hspace{0pt} have\hspace{0pt} \hspace{0pt}%
2\hspace{0pt} to\hspace{0pt} the\hspace{0pt} \hspace{0pt}1\hspace{0pt}8\hspace{0pt}th\hspace{0pt} power\hspace{0pt}%
,\hspace{0pt} whereas\hspace{0pt} \hspace{0pt}3\hspace{0pt} might\hspace{0pt} be\hspace{0pt} assigned\hspace{0pt}%
 to\hspace{0pt} b\hspace{0pt},\hspace{0pt} so\hspace{0pt} the\hspace{0pt} exponent\hspace{0pt} of\hspace{0pt}%
 \hspace{0pt}3\hspace{0pt} would\hspace{0pt} be\hspace{0pt} to\hspace{0pt} the\hspace{0pt} \hspace{0pt}%
8\hspace{0pt}th\hspace{0pt} in\hspace{0pt} b\hspace{0pt}.\hspace{0pt} Wait\hspace{0pt},\hspace{0pt} but\hspace{0pt}%
 a\hspace{0pt} and\hspace{0pt} b\hspace{0pt} have\hspace{0pt} ex\hspace{0pt}ponents\hspace{0pt} for\hspace{0pt}%
 the\hspace{0pt} primes\hspace{0pt}?\hspace{0pt} Wait\hspace{0pt} no\hspace{0pt},\hspace{0pt} in\hspace{0pt}%
 the\hspace{0pt} problem\hspace{0pt} statement\hspace{0pt},\hspace{0pt} the\hspace{0pt} numerator\hspace{0pt}%
 and\hspace{0pt} denominator\hspace{0pt} (\hspace{0pt}a\hspace{0pt} and\hspace{0pt} b\hspace{0pt})\hspace{0pt}%
 when\hspace{0pt} written\hspace{0pt} in\hspace{0pt} lowest\hspace{0pt} terms\hspace{0pt}.\hspace{0pt}%
 If\hspace{0pt} the\hspace{0pt} original\hspace{0pt} rational\hspace{0pt} number\hspace{0pt} is\hspace{0pt}%
 \hspace{0pt} in\hspace{0pt} lowest\hspace{0pt} terms\hspace{0pt},\hspace{0pt} that\hspace{0pt} means\hspace{0pt}%
 that\hspace{0pt} a\hspace{0pt} and\hspace{0pt} b\hspace{0pt} are\hspace{0pt} cop\hspace{0pt}r\hspace{0pt}%
ime\hspace{0pt},\hspace{0pt} hence\hspace{0pt} ex\hspace{0pt}ponents\hspace{0pt} of\hspace{0pt} primes\hspace{0pt}%
 in\hspace{0pt} \hspace{0pt}2\hspace{0pt}0\hspace{0pt}!\hspace{0pt} are\hspace{0pt} distributed\hspace{0pt}%
 completely\hspace{0pt} to\hspace{0pt} a\hspace{0pt} or\hspace{0pt} to\hspace{0pt} b\hspace{0pt}.\hspace{0pt}%
 Thus\hspace{0pt},\hspace{0pt} the\hspace{0pt} product\hspace{0pt} a\hspace{0pt}*b\hspace{0pt}=\hspace{0pt}%
2\hspace{0pt}0\hspace{0pt}!\hspace{0pt} means\hspace{0pt} that\hspace{0pt} each\hspace{0pt} prime\hspace{0pt}%
's\hspace{0pt} exponent\hspace{0pt} in\hspace{0pt} \hspace{0pt}2\hspace{0pt}0\hspace{0pt}!\hspace{0pt}%
 is\hspace{0pt} divided\hspace{0pt} between\hspace{0pt} a\hspace{0pt} and\hspace{0pt} b\hspace{0pt}?\hspace{0pt}%
 But\hspace{0pt} since\hspace{0pt} they\hspace{0pt} must\hspace{0pt} be\hspace{0pt} cop\hspace{0pt}r\hspace{0pt}%
ime\hspace{0pt},\hspace{0pt} it\hspace{0pt}'s\hspace{0pt} either\hspace{0pt} assigned\hspace{0pt} entirely\hspace{0pt}%
 to\hspace{0pt} a\hspace{0pt} or\hspace{0pt} to\hspace{0pt} b\hspace{0pt}.\hspace{0pt}Therefore\hspace{0pt}%
,\hspace{0pt} \hspace{0pt}2\hspace{0pt}0\hspace{0pt}!\hspace{0pt} can\hspace{0pt} be\hspace{0pt} considered\hspace{0pt}%
 as\hspace{0pt} a\hspace{0pt} product\hspace{0pt} of\hspace{0pt} cop\hspace{0pt}r\hspace{0pt}ime\hspace{0pt}%
 numbers\hspace{0pt} a\hspace{0pt} and\hspace{0pt} b\hspace{0pt} for\hspace{0pt} which\hspace{0pt} all\hspace{0pt}%
 primes\hspace{0pt} in\hspace{0pt} a\hspace{0pt} share\hspace{0pt} no\hspace{0pt} primes\hspace{0pt} with\hspace{0pt}%
 b\hspace{0pt}.\hspace{0pt} Therefore\hspace{0pt},\hspace{0pt} the\hspace{0pt} cop\hspace{0pt}r\hspace{0pt}%
ime\hspace{0pt} pairs\hspace{0pt} (\hspace{0pt}a\hspace{0pt},b\hspace{0pt})\hspace{0pt} with\hspace{0pt}%
 a\hspace{0pt}*b\hspace{0pt}=\hspace{0pt}2\hspace{0pt}0\hspace{0pt}!\hspace{0pt} are\hspace{0pt} in\hspace{0pt}%
 correspondence\hspace{0pt} with\hspace{0pt} the\hspace{0pt} subsets\hspace{0pt} of\hspace{0pt} the\hspace{0pt}%
 set\hspace{0pt} of\hspace{0pt} prime\hspace{0pt} factors\hspace{0pt} of\hspace{0pt} \hspace{0pt}2\hspace{0pt}%
0\hspace{0pt}!.\hspace{0pt}Each\hspace{0pt} prime\hspace{0pt} factor\hspace{0pt} must\hspace{0pt} go\hspace{0pt}%
 either\hspace{0pt} to\hspace{0pt} a\hspace{0pt} or\hspace{0pt} to\hspace{0pt} b\hspace{0pt},\hspace{0pt}%
 since\hspace{0pt} a\hspace{0pt} and\hspace{0pt} b\hspace{0pt} are\hspace{0pt} cop\hspace{0pt}r\hspace{0pt}%
ime\hspace{0pt}.\hspace{0pt} Therefore\hspace{0pt},\hspace{0pt} each\hspace{0pt} prime\hspace{0pt} is\hspace{0pt}%
 assigned\hspace{0pt} to\hspace{0pt} a\hspace{0pt} or\hspace{0pt} b\hspace{0pt},\hspace{0pt} so\hspace{0pt}%
 the\hspace{0pt} total\hspace{0pt} number\hspace{0pt} of\hspace{0pt} such\hspace{0pt} assignments\hspace{0pt}%
 is\hspace{0pt} \hspace{0pt}2\hspace{0pt}\^{}\hspace{0pt}8\hspace{0pt} (\hspace{0pt}since\hspace{0pt}%
 there\hspace{0pt} are\hspace{0pt} \hspace{0pt}8\hspace{0pt} distinct\hspace{0pt} primes\hspace{0pt} in\hspace{0pt}%
 \hspace{0pt}2\hspace{0pt}0\hspace{0pt}!'\hspace{0pt}s\hspace{0pt} factor\hspace{0pt}ization\hspace{0pt}%
).\hspace{0pt} Each\hspace{0pt} assignment\hspace{0pt} produces\hspace{0pt} a\hspace{0pt} unique\hspace{0pt}%
 cop\hspace{0pt}r\hspace{0pt}ime\hspace{0pt} pair\hspace{0pt} (\hspace{0pt}a\hspace{0pt},\hspace{0pt}%
 b\hspace{0pt})\hspace{0pt} with\hspace{0pt} a\hspace{0pt}*b\hspace{0pt}=\hspace{0pt}2\hspace{0pt}0\hspace{0pt}%
!.\hspace{0pt}But\hspace{0pt} here\hspace{0pt},\hspace{0pt} (\hspace{0pt}a\hspace{0pt},b\hspace{0pt})\hspace{0pt}%
 is\hspace{0pt} an\hspace{0pt} ordered\hspace{0pt} pair\hspace{0pt},\hspace{0pt} hence\hspace{0pt} assignments\hspace{0pt}%
 where\hspace{0pt} primes\hspace{0pt} are\hspace{0pt} assigned\hspace{0pt} to\hspace{0pt} a\hspace{0pt}%
 vs\hspace{0pt}.\hspace{0pt} assigned\hspace{0pt} to\hspace{0pt} b\hspace{0pt} result\hspace{0pt} in\hspace{0pt}%
 a\hspace{0pt} different\hspace{0pt} ordered\hspace{0pt} pair\hspace{0pt}.\hspace{0pt}However\hspace{0pt}%
,\hspace{0pt} our\hspace{0pt} problem\hspace{0pt} wants\hspace{0pt} rational\hspace{0pt} numbers\hspace{0pt}%
 between\hspace{0pt} \hspace{0pt}0\hspace{0pt} and\hspace{0pt} \hspace{0pt}1\hspace{0pt},\hspace{0pt}%
 which\hspace{0pt} corresponds\hspace{0pt} to\hspace{0pt} a\hspace{0pt} \textless{}\hspace{0pt} b\hspace{0pt}%
 (\hspace{0pt}because\hspace{0pt} \hspace{0pt}0\hspace{0pt} \textless{}\hspace{0pt} a\hspace{0pt}/b\hspace{0pt}%
 \textless{}\hspace{0pt} \hspace{0pt}1\hspace{0pt} implies\hspace{0pt} a\hspace{0pt} \textless{}\hspace{0pt}%
 b\hspace{0pt}).\hspace{0pt} Therefore\hspace{0pt},\hspace{0pt} the\hspace{0pt} number\hspace{0pt} of\hspace{0pt}%
 such\hspace{0pt} fractions\hspace{0pt} is\hspace{0pt} equal\hspace{0pt} to\hspace{0pt} the\hspace{0pt}%
 number\hspace{0pt} of\hspace{0pt} cop\hspace{0pt}r\hspace{0pt}ime\hspace{0pt} divisor\hspace{0pt} pairs\hspace{0pt}%
 (\hspace{0pt}a\hspace{0pt},b\hspace{0pt})\hspace{0pt} where\hspace{0pt} a\hspace{0pt}*b\hspace{0pt}=\hspace{0pt}%
2\hspace{0pt}0\hspace{0pt}!\hspace{0pt} and\hspace{0pt} a\hspace{0pt} \textless{}\hspace{0pt} b\hspace{0pt}%
.\hspace{0pt} Since\hspace{0pt} each\hspace{0pt} cop\hspace{0pt}r\hspace{0pt}ime\hspace{0pt} pair\hspace{0pt}%
 (\hspace{0pt}a\hspace{0pt},b\hspace{0pt})\hspace{0pt} with\hspace{0pt} a\hspace{0pt} \textless{}\hspace{0pt}%
 b\hspace{0pt} corresponds\hspace{0pt} to\hspace{0pt} the\hspace{0pt} exact\hspace{0pt} opposite\hspace{0pt}%
 pair\hspace{0pt} (\hspace{0pt}b\hspace{0pt},a\hspace{0pt})\hspace{0pt} with\hspace{0pt} b\hspace{0pt}%
 \textgreater{}\hspace{0pt} a\hspace{0pt}.\hspace{0pt} Therefore\hspace{0pt},\hspace{0pt} half\hspace{0pt}%
 of\hspace{0pt} the\hspace{0pt} total\hspace{0pt} ordered\hspace{0pt} cop\hspace{0pt}r\hspace{0pt}ime\hspace{0pt}%
 pairs\hspace{0pt} (\hspace{0pt}excluding\hspace{0pt} duplicates\hspace{0pt})\hspace{0pt} would\hspace{0pt}%
 give\hspace{0pt} the\hspace{0pt} count\hspace{0pt}.\hspace{0pt}But\hspace{0pt} since\hspace{0pt} in\hspace{0pt}%
 each\hspace{0pt} such\hspace{0pt} cop\hspace{0pt}r\hspace{0pt}ime\hspace{0pt} pair\hspace{0pt} of\hspace{0pt}%
 div\hspace{0pt}isors\hspace{0pt},\hspace{0pt} (\hspace{0pt}a\hspace{0pt},b\hspace{0pt})\hspace{0pt} and\hspace{0pt}%
 (\hspace{0pt}b\hspace{0pt},a\hspace{0pt})\hspace{0pt} are\hspace{0pt} distinct\hspace{0pt} (\hspace{0pt}%
since\hspace{0pt} we\hspace{0pt}'re\hspace{0pt} considering\hspace{0pt} ordered\hspace{0pt} pairs\hspace{0pt}%
),\hspace{0pt} then\hspace{0pt} the\hspace{0pt} number\hspace{0pt} of\hspace{0pt} cop\hspace{0pt}r\hspace{0pt}%
ime\hspace{0pt} pairs\hspace{0pt} with\hspace{0pt} a\hspace{0pt} \textless{}\hspace{0pt} b\hspace{0pt}%
 is\hspace{0pt} exactly\hspace{0pt} half\hspace{0pt} of\hspace{0pt} the\hspace{0pt} total\hspace{0pt}%
 number\hspace{0pt} of\hspace{0pt} cop\hspace{0pt}r\hspace{0pt}ime\hspace{0pt} pairs\hspace{0pt}.\hspace{0pt}%
 Since\hspace{0pt} the\hspace{0pt} total\hspace{0pt} number\hspace{0pt} is\hspace{0pt} \hspace{0pt}2\hspace{0pt}%
\^{}\hspace{0pt}8\hspace{0pt}=\hspace{0pt}2\hspace{0pt}5\hspace{0pt}6\hspace{0pt} (\hspace{0pt}since\hspace{0pt}%
 each\hspace{0pt} prime\hspace{0pt} is\hspace{0pt} either\hspace{0pt} assigned\hspace{0pt} to\hspace{0pt}%
 a\hspace{0pt} or\hspace{0pt} to\hspace{0pt} b\hspace{0pt},\hspace{0pt} \hspace{0pt}2\hspace{0pt} choices\hspace{0pt}%
 per\hspace{0pt} prime\hspace{0pt}),\hspace{0pt} but\hspace{0pt} the\hspace{0pt} assignment\hspace{0pt}%
 where\hspace{0pt} a\hspace{0pt} =\hspace{0pt} \hspace{0pt}1\hspace{0pt} (\hspace{0pt}all\hspace{0pt}%
 primes\hspace{0pt} assigned\hspace{0pt} to\hspace{0pt} b\hspace{0pt})\hspace{0pt} and\hspace{0pt} b\hspace{0pt}%
 =\hspace{0pt} \hspace{0pt}2\hspace{0pt}0\hspace{0pt}!\hspace{0pt} (\hspace{0pt}there\hspace{0pt}'s\hspace{0pt}%
 exactly\hspace{0pt} one\hspace{0pt} such\hspace{0pt} case\hspace{0pt}),\hspace{0pt} and\hspace{0pt} a\hspace{0pt}%
 can\hspace{0pt} be\hspace{0pt} \hspace{0pt}1\hspace{0pt},\hspace{0pt} then\hspace{0pt} proceed\hspace{0pt}%
.\hspace{0pt} Wait\hspace{0pt},\hspace{0pt} \hspace{0pt}1\hspace{0pt} is\hspace{0pt} the\hspace{0pt} product\hspace{0pt}%
 of\hspace{0pt} assigning\hspace{0pt} no\hspace{0pt} primes\hspace{0pt} to\hspace{0pt} a\hspace{0pt} (\hspace{0pt}%
all\hspace{0pt} primes\hspace{0pt} to\hspace{0pt} b\hspace{0pt}),\hspace{0pt} so\hspace{0pt} that\hspace{0pt}%
's\hspace{0pt} included\hspace{0pt}.\hspace{0pt}But\hspace{0pt} if\hspace{0pt} a\hspace{0pt} and\hspace{0pt}%
 b\hspace{0pt} can\hspace{0pt} be\hspace{0pt} written\hspace{0pt} as\hspace{0pt} a\hspace{0pt} =\hspace{0pt}%
 \hspace{0pt}1\hspace{0pt} and\hspace{0pt} b\hspace{0pt} =\hspace{0pt} \hspace{0pt}2\hspace{0pt}0\hspace{0pt}%
!,\hspace{0pt} yes\hspace{0pt},\hspace{0pt} that\hspace{0pt}'s\hspace{0pt} a\hspace{0pt} cop\hspace{0pt}%
r\hspace{0pt}ime\hspace{0pt} pair\hspace{0pt},\hspace{0pt} likewise\hspace{0pt} a\hspace{0pt}=\hspace{0pt}%
2\hspace{0pt}\^{}\hspace{0pt}1\hspace{0pt}8\hspace{0pt} and\hspace{0pt} b\hspace{0pt}=\hspace{0pt} (\hspace{0pt}%
2\hspace{0pt}0\hspace{0pt}!\hspace{0pt})/\hspace{0pt}2\hspace{0pt}\^{}\hspace{0pt}1\hspace{0pt}8\hspace{0pt}%
 which\hspace{0pt} is\hspace{0pt} still\hspace{0pt} an\hspace{0pt} integer\hspace{0pt}.\hspace{0pt} Similarly\hspace{0pt}%
 for\hspace{0pt} all\hspace{0pt} subsets\hspace{0pt}.\hspace{0pt} Wait\hspace{0pt},\hspace{0pt} but\hspace{0pt}%
 when\hspace{0pt} we\hspace{0pt} write\hspace{0pt} a\hspace{0pt} and\hspace{0pt} b\hspace{0pt} as\hspace{0pt}%
 cop\hspace{0pt}r\hspace{0pt}ime\hspace{0pt} factors\hspace{0pt},\hspace{0pt} the\hspace{0pt} way\hspace{0pt}%
 I\hspace{0pt}'m\hspace{0pt} thinking\hspace{0pt} is\hspace{0pt} that\hspace{0pt} the\hspace{0pt} assignment\hspace{0pt}%
 is\hspace{0pt} of\hspace{0pt} each\hspace{0pt} prime\hspace{0pt} as\hspace{0pt} a\hspace{0pt} binary\hspace{0pt}%
 choice\hspace{0pt} (\hspace{0pt}to\hspace{0pt} a\hspace{0pt} or\hspace{0pt} to\hspace{0pt} b\hspace{0pt}%
).\hspace{0pt} So\hspace{0pt} \hspace{0pt}2\hspace{0pt}\^{}\hspace{0pt}8\hspace{0pt} equals\hspace{0pt}%
 \hspace{0pt}2\hspace{0pt}5\hspace{0pt}6\hspace{0pt} possible\hspace{0pt} assignments\hspace{0pt},\hspace{0pt}%
 and\hspace{0pt} each\hspace{0pt} assignment\hspace{0pt} defines\hspace{0pt} a\hspace{0pt} unique\hspace{0pt}%
 cop\hspace{0pt}r\hspace{0pt}ime\hspace{0pt} pair\hspace{0pt} (\hspace{0pt}a\hspace{0pt},b\hspace{0pt}%
)\hspace{0pt} where\hspace{0pt} a\hspace{0pt} *\hspace{0pt} b\hspace{0pt}=\hspace{0pt}2\hspace{0pt}0\hspace{0pt}%
!.\hspace{0pt} Since\hspace{0pt} each\hspace{0pt} assignment\hspace{0pt} is\hspace{0pt} unique\hspace{0pt}%
,\hspace{0pt} each\hspace{0pt} ordered\hspace{0pt} pair\hspace{0pt} is\hspace{0pt} distinct\hspace{0pt}%
,\hspace{0pt} even\hspace{0pt} if\hspace{0pt} a\hspace{0pt} and\hspace{0pt} b\hspace{0pt} are\hspace{0pt}%
 swapped\hspace{0pt}.\hspace{0pt}Since\hspace{0pt} such\hspace{0pt} ordered\hspace{0pt} pairs\hspace{0pt}%
 can\hspace{0pt} have\hspace{0pt} a\hspace{0pt} \textless{}\hspace{0pt} b\hspace{0pt} or\hspace{0pt} a\hspace{0pt}%
 \textgreater{}\hspace{0pt} b\hspace{0pt},\hspace{0pt} so\hspace{0pt} then\hspace{0pt} to\hspace{0pt}%
 get\hspace{0pt} the\hspace{0pt} (\hspace{0pt}a\hspace{0pt},b\hspace{0pt})\hspace{0pt} with\hspace{0pt}%
 a\hspace{0pt} \textless{}\hspace{0pt} b\hspace{0pt},\hspace{0pt} we\hspace{0pt} divide\hspace{0pt} by\hspace{0pt}%
 \hspace{0pt}2\hspace{0pt},\hspace{0pt} which\hspace{0pt} gives\hspace{0pt} \hspace{0pt}1\hspace{0pt}%
2\hspace{0pt}8\hspace{0pt}.\hspace{0pt}\heatbox{0.980,0.618,0.618}{But}\hspace{0pt}\heatbox{0.980,0.618,0.618}{ wait}\hspace{0pt}%
\heatbox{0.980,0.618,0.618}{,}\hspace{0pt}\heatbox{0.980,0.618,0.618}{ can}\hspace{0pt}\heatbox{0.980,0.618,0.618}{ it}\hspace{0pt}%
\heatbox{0.980,0.618,0.618}{ be}\hspace{0pt}\heatbox{0.980,0.618,0.618}{?}\hspace{0pt}\heatbox{0.980,0.618,0.618}{ Let}\hspace{0pt}%
\heatbox{0.980,0.618,0.618}{ me}\hspace{0pt}\heatbox{0.980,0.618,0.618}{ check}\hspace{0pt}\heatbox{0.980,0.618,0.618}{ with}\hspace{0pt}%
\heatbox{0.980,0.618,0.618}{ a}\hspace{0pt}\heatbox{0.980,0.618,0.618}{ smaller}\hspace{0pt}\heatbox{0.980,0.618,0.618}{ factorial}\hspace{0pt}%
.\hspace{0pt}Sup\hspace{0pt}pose\hspace{0pt} instead\hspace{0pt} of\hspace{0pt} \hspace{0pt}2\hspace{0pt}%
0\hspace{0pt}!,\hspace{0pt} take\hspace{0pt} a\hspace{0pt} simpler\hspace{0pt} case\hspace{0pt}.\hspace{0pt}%
 Let\hspace{0pt} me\hspace{0pt} check\hspace{0pt} with\hspace{0pt} n\hspace{0pt} =\hspace{0pt} \hspace{0pt}%
6\hspace{0pt}.\hspace{0pt} Then\hspace{0pt} \hspace{0pt}6\hspace{0pt}!\hspace{0pt} =\hspace{0pt} \hspace{0pt}%
7\hspace{0pt}2\hspace{0pt}0\hspace{0pt}.\hspace{0pt} Let\hspace{0pt}'s\hspace{0pt} ask\hspace{0pt} for\hspace{0pt}%
 ration\hspace{0pt}als\hspace{0pt} between\hspace{0pt} \hspace{0pt}0\hspace{0pt} and\hspace{0pt} \hspace{0pt}%
1\hspace{0pt} with\hspace{0pt} product\hspace{0pt} a\hspace{0pt}*b\hspace{0pt}=\hspace{0pt}7\hspace{0pt}%
2\hspace{0pt}0\hspace{0pt},\hspace{0pt} gcd\hspace{0pt}(a\hspace{0pt},b\hspace{0pt})=\hspace{0pt}1\hspace{0pt}%
,\hspace{0pt} a\hspace{0pt} \textless{}\hspace{0pt} b\hspace{0pt}.\hspace{0pt} Then\hspace{0pt} \hspace{0pt}%
7\hspace{0pt}2\hspace{0pt}0\hspace{0pt} prime\hspace{0pt} factor\hspace{0pt}ization\hspace{0pt}:\hspace{0pt}%
 \hspace{0pt}2\hspace{0pt}\^{}\hspace{0pt}4\hspace{0pt} *\hspace{0pt}3\hspace{0pt}\^{}\hspace{0pt}2\hspace{0pt}%
 *\hspace{0pt}5\hspace{0pt}\^{}\hspace{0pt}1\hspace{0pt}.\hspace{0pt} So\hspace{0pt} number\hspace{0pt}%
 of\hspace{0pt} primes\hspace{0pt} is\hspace{0pt} \hspace{0pt}3\hspace{0pt} (\hspace{0pt}2\hspace{0pt}%
,\hspace{0pt}3\hspace{0pt},\hspace{0pt}5\hspace{0pt}).\hspace{0pt} Each\hspace{0pt} can\hspace{0pt} be\hspace{0pt}%
 assigned\hspace{0pt} to\hspace{0pt} a\hspace{0pt} or\hspace{0pt} b\hspace{0pt},\hspace{0pt} so\hspace{0pt}%
 \hspace{0pt}2\hspace{0pt}\^{}\hspace{0pt}3\hspace{0pt}=\hspace{0pt}8\hspace{0pt} cop\hspace{0pt}r\hspace{0pt}%
ime\hspace{0pt} pairs\hspace{0pt}.\hspace{0pt} Then\hspace{0pt} half\hspace{0pt} of\hspace{0pt} them\hspace{0pt}%
,\hspace{0pt} \hspace{0pt}4\hspace{0pt} pairs\hspace{0pt} would\hspace{0pt} have\hspace{0pt} a\hspace{0pt}%
 \textless{}\hspace{0pt} b\hspace{0pt}.\hspace{0pt} Let\hspace{0pt}'s\hspace{0pt} check\hspace{0pt}.\hspace{0pt}%
 Possible\hspace{0pt} assignments\hspace{0pt}:\hspace{0pt}1\hspace{0pt}.\hspace{0pt} Assign\hspace{0pt}%
 none\hspace{0pt} to\hspace{0pt} a\hspace{0pt}:\hspace{0pt} a\hspace{0pt} =\hspace{0pt}1\hspace{0pt},\hspace{0pt}%
 b\hspace{0pt}=\hspace{0pt}7\hspace{0pt}2\hspace{0pt}0\hspace{0pt}.\hspace{0pt} Then\hspace{0pt} \hspace{0pt}%
1\hspace{0pt}*\hspace{0pt}7\hspace{0pt}2\hspace{0pt}0\hspace{0pt}=\hspace{0pt}7\hspace{0pt}2\hspace{0pt}%
0\hspace{0pt}.\hspace{0pt} \hspace{0pt} a\hspace{0pt}=\hspace{0pt}1\hspace{0pt} \textless{}\hspace{0pt}%
 b\hspace{0pt}=\hspace{0pt}7\hspace{0pt}2\hspace{0pt}0\hspace{0pt} :\hspace{0pt} valid\hspace{0pt}.\hspace{0pt}%
2\hspace{0pt}.\hspace{0pt} Assign\hspace{0pt} \hspace{0pt}2\hspace{0pt} to\hspace{0pt} a\hspace{0pt},\hspace{0pt}%
 others\hspace{0pt} to\hspace{0pt} b\hspace{0pt}:\hspace{0pt} a\hspace{0pt}=\hspace{0pt}1\hspace{0pt}%
6\hspace{0pt} (\hspace{0pt}2\hspace{0pt}\^{}\hspace{0pt}4\hspace{0pt}),\hspace{0pt} b\hspace{0pt}=\hspace{0pt}%
4\hspace{0pt}5\hspace{0pt} (\hspace{0pt}3\hspace{0pt}\^{}\hspace{0pt}2\hspace{0pt} *\hspace{0pt}5\hspace{0pt}%
).\hspace{0pt} gcd\hspace{0pt}(\hspace{0pt}1\hspace{0pt}6\hspace{0pt},\hspace{0pt}4\hspace{0pt}5\hspace{0pt}%
)=\hspace{0pt}1\hspace{0pt}.\hspace{0pt} Check\hspace{0pt} a\hspace{0pt}=\hspace{0pt}1\hspace{0pt}6\hspace{0pt}%
 \textless{}\hspace{0pt} b\hspace{0pt}=\hspace{0pt}4\hspace{0pt}5\hspace{0pt}?\hspace{0pt} \hspace{0pt}%
1\hspace{0pt}6\hspace{0pt} \textless{}\hspace{0pt}4\hspace{0pt}5\hspace{0pt}:\hspace{0pt} yes\hspace{0pt}%
.\hspace{0pt}3\hspace{0pt}.\hspace{0pt} Assign\hspace{0pt} \hspace{0pt}3\hspace{0pt} to\hspace{0pt} a\hspace{0pt}%
,\hspace{0pt} others\hspace{0pt} to\hspace{0pt} b\hspace{0pt}:\hspace{0pt} a\hspace{0pt}=\hspace{0pt}%
9\hspace{0pt} (\hspace{0pt}3\hspace{0pt}\^{}\hspace{0pt}2\hspace{0pt}),\hspace{0pt} b\hspace{0pt}=\hspace{0pt}%
8\hspace{0pt}0\hspace{0pt} (\hspace{0pt}1\hspace{0pt}6\hspace{0pt} *\hspace{0pt}5\hspace{0pt}).\hspace{0pt}%
 Check\hspace{0pt} \hspace{0pt}9\hspace{0pt} \textless{}\hspace{0pt}8\hspace{0pt}0\hspace{0pt}:\hspace{0pt}%
 yes\hspace{0pt}.\hspace{0pt}4\hspace{0pt}.\hspace{0pt} Assign\hspace{0pt} \hspace{0pt}5\hspace{0pt} to\hspace{0pt}%
 a\hspace{0pt},\hspace{0pt} others\hspace{0pt} to\hspace{0pt} b\hspace{0pt}:\hspace{0pt} a\hspace{0pt}%
=\hspace{0pt}5\hspace{0pt},\hspace{0pt} b\hspace{0pt}=\hspace{0pt} \hspace{0pt}1\hspace{0pt}4\hspace{0pt}%
4\hspace{0pt} (\hspace{0pt}1\hspace{0pt}6\hspace{0pt}*\hspace{0pt}9\hspace{0pt}).\hspace{0pt} \hspace{0pt}%
5\hspace{0pt} \textless{}\hspace{0pt}1\hspace{0pt}4\hspace{0pt}4\hspace{0pt}:\hspace{0pt} yes\hspace{0pt}%
.\hspace{0pt}5\hspace{0pt}.\hspace{0pt} Assign\hspace{0pt} \hspace{0pt}2\hspace{0pt} and\hspace{0pt} \hspace{0pt}%
3\hspace{0pt} to\hspace{0pt} a\hspace{0pt}:\hspace{0pt} a\hspace{0pt}=\hspace{0pt}1\hspace{0pt}6\hspace{0pt}%
*\hspace{0pt}9\hspace{0pt}=\hspace{0pt}1\hspace{0pt}4\hspace{0pt}4\hspace{0pt},\hspace{0pt} b\hspace{0pt}%
=\hspace{0pt}5\hspace{0pt}.\hspace{0pt} Now\hspace{0pt} a\hspace{0pt}=\hspace{0pt}1\hspace{0pt}4\hspace{0pt}%
4\hspace{0pt} and\hspace{0pt} b\hspace{0pt}=\hspace{0pt}5\hspace{0pt}.\hspace{0pt} But\hspace{0pt} \hspace{0pt}%
1\hspace{0pt}4\hspace{0pt}4\hspace{0pt} \textgreater{}\hspace{0pt}5\hspace{0pt},\hspace{0pt} so\hspace{0pt}%
 would\hspace{0pt} be\hspace{0pt} considered\hspace{0pt} as\hspace{0pt} a\hspace{0pt} different\hspace{0pt}%
 pair\hspace{0pt} if\hspace{0pt} ordered\hspace{0pt}.\hspace{0pt} But\hspace{0pt} in\hspace{0pt} assignments\hspace{0pt}%
,\hspace{0pt} assigning\hspace{0pt} to\hspace{0pt} a\hspace{0pt} or\hspace{0pt} b\hspace{0pt}.\hspace{0pt}%
 So\hspace{0pt} here\hspace{0pt},\hspace{0pt} ordering\hspace{0pt} matters\hspace{0pt}.\hspace{0pt} Therefore\hspace{0pt}%
,\hspace{0pt} for\hspace{0pt} the\hspace{0pt} purpose\hspace{0pt} of\hspace{0pt} our\hspace{0pt} count\hspace{0pt}%
,\hspace{0pt} pairing\hspace{0pt} where\hspace{0pt} \hspace{0pt}2\hspace{0pt} and\hspace{0pt} \hspace{0pt}%
3\hspace{0pt} are\hspace{0pt} assigned\hspace{0pt} to\hspace{0pt} a\hspace{0pt} would\hspace{0pt} create\hspace{0pt}%
 a\hspace{0pt}=\hspace{0pt}1\hspace{0pt}4\hspace{0pt}4\hspace{0pt},\hspace{0pt} b\hspace{0pt}=\hspace{0pt}%
5\hspace{0pt},\hspace{0pt} but\hspace{0pt} since\hspace{0pt} a\hspace{0pt} \textless{}\hspace{0pt} b\hspace{0pt}%
 is\hspace{0pt} required\hspace{0pt},\hspace{0pt} this\hspace{0pt} does\hspace{0pt} NOT\hspace{0pt} count\hspace{0pt}%
.\hspace{0pt} So\hspace{0pt} the\hspace{0pt} assignment\hspace{0pt} of\hspace{0pt} primes\hspace{0pt}%
 \{\hspace{0pt}2\hspace{0pt},\hspace{0pt}3\hspace{0pt}\}\hspace{0pt} to\hspace{0pt} a\hspace{0pt} would\hspace{0pt}%
 result\hspace{0pt} in\hspace{0pt} a\hspace{0pt} product\hspace{0pt} a\hspace{0pt}=\hspace{0pt}1\hspace{0pt}%
4\hspace{0pt}4\hspace{0pt},\hspace{0pt} b\hspace{0pt}=\hspace{0pt}5\hspace{0pt}.\hspace{0pt} To\hspace{0pt}%
 reach\hspace{0pt} a\hspace{0pt} \textless{}\hspace{0pt} b\hspace{0pt},\hspace{0pt} the\hspace{0pt} valid\hspace{0pt}%
 pairs\hspace{0pt} where\hspace{0pt} assignment\hspace{0pt} to\hspace{0pt} a\hspace{0pt}.\hspace{0pt}%
 Wait\hspace{0pt},\hspace{0pt} This\hspace{0pt} seems\hspace{0pt} m\hspace{0pt}iscal\hspace{0pt}cul\hspace{0pt}%
ous\hspace{0pt}.\hspace{0pt}\heatbox{0.984,0.697,0.697}{Wait}\hspace{0pt}\heatbox{0.984,0.697,0.697}{ no}\hspace{0pt}%
\heatbox{0.984,0.697,0.697}{,}\hspace{0pt}\heatbox{0.984,0.697,0.697}{ in}\hspace{0pt}\heatbox{0.984,0.697,0.697}{ }\hspace{0pt}%
\heatbox{0.984,0.697,0.697}{7}\hspace{0pt}\heatbox{0.984,0.697,0.697}{2}\hspace{0pt}\heatbox{0.984,0.697,0.697}{0}\hspace{0pt}%
\heatbox{0.984,0.697,0.697}{'s}\hspace{0pt}\heatbox{0.984,0.697,0.697}{ case}\hspace{0pt}\heatbox{0.984,0.697,0.697}{.}\hspace{0pt}%
\heatbox{0.984,0.697,0.697}{ Assign}\hspace{0pt}\heatbox{0.984,0.697,0.697}{ }\hspace{0pt}\heatbox{0.984,0.697,0.697}{2}\hspace{0pt}%
\heatbox{0.984,0.697,0.697}{ and}\hspace{0pt}\heatbox{0.984,0.697,0.697}{ }\hspace{0pt}\heatbox{0.984,0.697,0.697}{3}\hspace{0pt}%
\heatbox{0.984,0.697,0.697}{ to}\hspace{0pt}\heatbox{0.984,0.697,0.697}{ a}\hspace{0pt}\heatbox{0.984,0.697,0.697}{ would}\hspace{0pt}%
\heatbox{0.984,0.697,0.697}{ mean}\hspace{0pt}\heatbox{0.984,0.697,0.697}{ a}\hspace{0pt}\heatbox{0.984,0.697,0.697}{=(}\hspace{0pt}%
\heatbox{0.984,0.697,0.697}{2}\hspace{0pt}\heatbox{0.984,0.697,0.697}{\^{}}\hspace{0pt}\heatbox{0.984,0.697,0.697}{4}\hspace{0pt}%
\heatbox{0.984,0.697,0.697}{)*(}\hspace{0pt}\heatbox{0.984,0.697,0.697}{3}\hspace{0pt}\heatbox{0.984,0.697,0.697}{\^{}}\hspace{0pt}%
\heatbox{0.984,0.697,0.697}{2}\hspace{0pt}\heatbox{0.984,0.697,0.697}{)=}\hspace{0pt}\heatbox{0.984,0.697,0.697}{1}\hspace{0pt}%
\heatbox{0.984,0.697,0.697}{4}\hspace{0pt}\heatbox{0.984,0.697,0.697}{4}\hspace{0pt}\heatbox{0.984,0.697,0.697}{,}\hspace{0pt}%
\heatbox{0.984,0.697,0.697}{ while}\hspace{0pt}\heatbox{0.984,0.697,0.697}{ assigning}\hspace{0pt}\heatbox{0.984,0.697,0.697}{ rest}\hspace{0pt}%
\heatbox{0.984,0.697,0.697}{ (}\hspace{0pt}\heatbox{0.984,0.697,0.697}{5}\hspace{0pt}\heatbox{0.984,0.697,0.697}{)}\hspace{0pt}%
\heatbox{0.984,0.697,0.697}{ to}\hspace{0pt}\heatbox{0.984,0.697,0.697}{ b}\hspace{0pt}\heatbox{0.984,0.697,0.697}{=}\hspace{0pt}%
\heatbox{0.984,0.697,0.697}{5}\hspace{0pt}\heatbox{0.984,0.697,0.697}{.}\hspace{0pt}\heatbox{0.984,0.697,0.697}{ Since}\hspace{0pt}%
\heatbox{0.984,0.697,0.697}{ }\hspace{0pt}\heatbox{0.984,0.697,0.697}{1}\hspace{0pt}\heatbox{0.984,0.697,0.697}{4}\hspace{0pt}%
\heatbox{0.984,0.697,0.697}{4}\hspace{0pt}\heatbox{0.984,0.697,0.697}{ and}\hspace{0pt}\heatbox{0.984,0.697,0.697}{ }\hspace{0pt}%
\heatbox{0.984,0.697,0.697}{5}\hspace{0pt}\heatbox{0.984,0.697,0.697}{:}\hspace{0pt}\heatbox{0.984,0.697,0.697}{ gcd}\hspace{0pt}%
\heatbox{0.984,0.697,0.697}{(}\hspace{0pt}\heatbox{0.984,0.697,0.697}{1}\hspace{0pt}\heatbox{0.984,0.697,0.697}{4}\hspace{0pt}%
\heatbox{0.984,0.697,0.697}{4}\hspace{0pt}\heatbox{0.984,0.697,0.697}{,}\hspace{0pt}\heatbox{0.984,0.697,0.697}{5}\hspace{0pt}%
\heatbox{0.984,0.697,0.697}{)=}\hspace{0pt}\heatbox{0.984,0.697,0.697}{1}\hspace{0pt}\heatbox{0.984,0.697,0.697}{,}\hspace{0pt}%
\heatbox{0.984,0.697,0.697}{ and}\hspace{0pt}\heatbox{0.984,0.697,0.697}{ ab}\hspace{0pt}\heatbox{0.984,0.697,0.697}{=}\hspace{0pt}%
\heatbox{0.984,0.697,0.697}{1}\hspace{0pt}\heatbox{0.984,0.697,0.697}{4}\hspace{0pt}\heatbox{0.984,0.697,0.697}{4}\hspace{0pt}%
\heatbox{0.984,0.697,0.697}{*}\hspace{0pt}\heatbox{0.984,0.697,0.697}{5}\hspace{0pt}\heatbox{0.984,0.697,0.697}{=}\hspace{0pt}%
\heatbox{0.984,0.697,0.697}{7}\hspace{0pt}\heatbox{0.984,0.697,0.697}{2}\hspace{0pt}\heatbox{0.984,0.697,0.697}{0}\hspace{0pt}%
\heatbox{0.984,0.697,0.697}{ indeed}\hspace{0pt}\heatbox{0.984,0.697,0.697}{.}\hspace{0pt}\heatbox{0.984,0.697,0.697}{ But}\hspace{0pt}%
\heatbox{0.984,0.697,0.697}{ a}\hspace{0pt}\heatbox{0.984,0.697,0.697}{=}\hspace{0pt}\heatbox{0.984,0.697,0.697}{1}\hspace{0pt}%
\heatbox{0.984,0.697,0.697}{4}\hspace{0pt}\heatbox{0.984,0.697,0.697}{4}\hspace{0pt}\heatbox{0.984,0.697,0.697}{\textgreater{}}\hspace{0pt}%
\heatbox{0.984,0.697,0.697}{5}\hspace{0pt}\heatbox{0.984,0.697,0.697}{,}\hspace{0pt}\heatbox{0.984,0.697,0.697}{ so}\hspace{0pt}%
\heatbox{0.984,0.697,0.697}{ this}\hspace{0pt}\heatbox{0.984,0.697,0.697}{ would}\hspace{0pt}\heatbox{0.984,0.697,0.697}{ not}\hspace{0pt}%
\heatbox{0.984,0.697,0.697}{ satisfy}\hspace{0pt}\heatbox{0.984,0.697,0.697}{ a}\hspace{0pt}\heatbox{0.984,0.697,0.697}{ \textless{}}\hspace{0pt}%
\heatbox{0.984,0.697,0.697}{ b}\hspace{0pt}\heatbox{0.984,0.697,0.697}{,}\hspace{0pt}\heatbox{0.984,0.697,0.697}{ therefore}\hspace{0pt}%
\heatbox{0.984,0.697,0.697}{ this}\hspace{0pt}\heatbox{0.984,0.697,0.697}{ assignment}\hspace{0pt}\heatbox{0.984,0.697,0.697}{ would}\hspace{0pt}%
\heatbox{0.984,0.697,0.697}{ not}\hspace{0pt}\heatbox{0.984,0.697,0.697}{ be}\hspace{0pt}\heatbox{0.984,0.697,0.697}{ included}\hspace{0pt}%
\heatbox{0.984,0.697,0.697}{ in}\hspace{0pt}\heatbox{0.984,0.697,0.697}{ the}\hspace{0pt}\heatbox{0.984,0.697,0.697}{ count}\hspace{0pt}%
\heatbox{0.984,0.697,0.697}{.}\hspace{0pt}\heatbox{0.984,0.697,0.697}{ So}\hspace{0pt}\heatbox{0.984,0.697,0.697}{ how}\hspace{0pt}%
\heatbox{0.984,0.697,0.697}{ do}\hspace{0pt}\heatbox{0.984,0.697,0.697}{ we}\hspace{0pt}\heatbox{0.984,0.697,0.697}{ reconcile}\hspace{0pt}%
\heatbox{0.984,0.697,0.697}{ the}\hspace{0pt}\heatbox{0.984,0.697,0.697}{ count}\hspace{0pt}.\hspace{0pt}%
If\hspace{0pt} the\hspace{0pt} number\hspace{0pt} of\hspace{0pt} cop\hspace{0pt}r\hspace{0pt}ime\hspace{0pt}%
 assignments\hspace{0pt} (\hspace{0pt}with\hspace{0pt} order\hspace{0pt})\hspace{0pt} is\hspace{0pt} \hspace{0pt}%
2\hspace{0pt}\^{}\hspace{0pt}3\hspace{0pt}=\hspace{0pt}8\hspace{0pt},\hspace{0pt} but\hspace{0pt} how\hspace{0pt}%
 many\hspace{0pt} of\hspace{0pt} those\hspace{0pt} assignments\hspace{0pt} result\hspace{0pt} in\hspace{0pt}%
 a\hspace{0pt} \textless{}\hspace{0pt} b\hspace{0pt}?\hspace{0pt}\heatbox{0.984,0.692,0.692}{So}\hspace{0pt}%
\heatbox{0.984,0.692,0.692}{ for}\hspace{0pt}\heatbox{0.984,0.692,0.692}{ }\hspace{0pt}\heatbox{0.984,0.692,0.692}{7}\hspace{0pt}%
\heatbox{0.984,0.692,0.692}{2}\hspace{0pt}\heatbox{0.984,0.692,0.692}{0}\hspace{0pt}\heatbox{0.984,0.692,0.692}{,}\hspace{0pt}%
\heatbox{0.984,0.692,0.692}{ with}\hspace{0pt}\heatbox{0.984,0.692,0.692}{ prime}\hspace{0pt}\heatbox{0.984,0.692,0.692}{ factors}\hspace{0pt}%
\heatbox{0.984,0.692,0.692}{ }\hspace{0pt}\heatbox{0.984,0.692,0.692}{2}\hspace{0pt}\heatbox{0.984,0.692,0.692}{,}\hspace{0pt}%
\heatbox{0.984,0.692,0.692}{3}\hspace{0pt}\heatbox{0.984,0.692,0.692}{,}\hspace{0pt}\heatbox{0.984,0.692,0.692}{5}\hspace{0pt}%
:\hspace{0pt}\heatbox{0.984,0.697,0.697}{The}\hspace{0pt}\heatbox{0.984,0.697,0.697}{ }\hspace{0pt}\heatbox{0.984,0.697,0.697}{8}\hspace{0pt}%
\heatbox{0.984,0.697,0.697}{ assignments}\hspace{0pt}:\hspace{0pt}\heatbox{0.977,0.567,0.567}{1}\hspace{0pt}%
\heatbox{0.977,0.567,0.567}{.}\hspace{0pt}\heatbox{0.977,0.567,0.567}{ Assign}\hspace{0pt}\heatbox{0.977,0.567,0.567}{ none}\hspace{0pt}%
\heatbox{0.977,0.567,0.567}{ to}\hspace{0pt}\heatbox{0.977,0.567,0.567}{ a}\hspace{0pt}\heatbox{0.977,0.567,0.567}{ =\textgreater{}}\hspace{0pt}%
\heatbox{0.977,0.567,0.567}{ a}\hspace{0pt}\heatbox{0.977,0.567,0.567}{=}\hspace{0pt}\heatbox{0.977,0.567,0.567}{1}\hspace{0pt}%
\heatbox{0.977,0.567,0.567}{,}\hspace{0pt}\heatbox{0.977,0.567,0.567}{ b}\hspace{0pt}\heatbox{0.977,0.567,0.567}{=}\hspace{0pt}%
\heatbox{0.977,0.567,0.567}{7}\hspace{0pt}\heatbox{0.977,0.567,0.567}{2}\hspace{0pt}\heatbox{0.977,0.567,0.567}{0}\hspace{0pt}%
\heatbox{0.977,0.567,0.567}{ (}\hspace{0pt}\heatbox{0.977,0.567,0.567}{1}\hspace{0pt}\heatbox{0.977,0.567,0.567}{ \textless{}}\hspace{0pt}%
\heatbox{0.977,0.567,0.567}{7}\hspace{0pt}\heatbox{0.977,0.567,0.567}{2}\hspace{0pt}\heatbox{0.977,0.567,0.567}{0}\hspace{0pt}%
\heatbox{0.977,0.567,0.567}{:}\hspace{0pt}\heatbox{0.977,0.567,0.567}{ yes}\hspace{0pt})\hspace{0pt}2\hspace{0pt}%
.\hspace{0pt} Assign\hspace{0pt} \hspace{0pt}2\hspace{0pt} to\hspace{0pt} a\hspace{0pt} =\textgreater{}\hspace{0pt}%
 a\hspace{0pt}=\hspace{0pt}1\hspace{0pt}6\hspace{0pt},\hspace{0pt} b\hspace{0pt}=\hspace{0pt}4\hspace{0pt}%
5\hspace{0pt} (\hspace{0pt}1\hspace{0pt}6\hspace{0pt} \textless{}\hspace{0pt}4\hspace{0pt}5\hspace{0pt}%
:\hspace{0pt} yes\hspace{0pt})\hspace{0pt}3\hspace{0pt}.\hspace{0pt} Assign\hspace{0pt} \hspace{0pt}3\hspace{0pt}%
 to\hspace{0pt} a\hspace{0pt} =\textgreater{}\hspace{0pt}a\hspace{0pt}=\hspace{0pt}9\hspace{0pt},\hspace{0pt}%
 b\hspace{0pt}=\hspace{0pt}8\hspace{0pt}0\hspace{0pt} (\hspace{0pt}9\hspace{0pt} \textless{}\hspace{0pt}%
8\hspace{0pt}0\hspace{0pt}:\hspace{0pt} yes\hspace{0pt})\hspace{0pt}4\hspace{0pt}.\hspace{0pt} Assign\hspace{0pt}%
 \hspace{0pt}5\hspace{0pt} to\hspace{0pt} a\hspace{0pt} =\textgreater{}\hspace{0pt}a\hspace{0pt}=\hspace{0pt}%
5\hspace{0pt},\hspace{0pt} b\hspace{0pt}=\hspace{0pt}1\hspace{0pt}4\hspace{0pt}4\hspace{0pt} (\hspace{0pt}%
5\hspace{0pt} \textless{}\hspace{0pt}1\hspace{0pt}4\hspace{0pt}4\hspace{0pt}:\hspace{0pt} yes\hspace{0pt}%
)\hspace{0pt}5\hspace{0pt}.\hspace{0pt} Assign\hspace{0pt} \hspace{0pt}2\hspace{0pt} and\hspace{0pt}3\hspace{0pt}%
 to\hspace{0pt} a\hspace{0pt} =\textgreater{}\hspace{0pt} a\hspace{0pt}=\hspace{0pt}1\hspace{0pt}4\hspace{0pt}%
4\hspace{0pt},\hspace{0pt} b\hspace{0pt}=\hspace{0pt}5\hspace{0pt} (\hspace{0pt}a\hspace{0pt} \textgreater{}\hspace{0pt}%
b\hspace{0pt}:\hspace{0pt} no\hspace{0pt})\hspace{0pt}6\hspace{0pt}.\hspace{0pt} Assign\hspace{0pt} \hspace{0pt}%
2\hspace{0pt} and\hspace{0pt}5\hspace{0pt} to\hspace{0pt} a\hspace{0pt} =\textgreater{}\hspace{0pt} a\hspace{0pt}%
=\hspace{0pt}1\hspace{0pt}6\hspace{0pt}*\hspace{0pt}5\hspace{0pt}=\hspace{0pt}8\hspace{0pt}0\hspace{0pt}%
,\hspace{0pt} b\hspace{0pt}=\hspace{0pt}9\hspace{0pt} (\hspace{0pt}8\hspace{0pt}0\hspace{0pt} \textgreater{}\hspace{0pt}%
9\hspace{0pt}:\hspace{0pt} no\hspace{0pt})\hspace{0pt}7\hspace{0pt}.\hspace{0pt} Assign\hspace{0pt}3\hspace{0pt}%
 and\hspace{0pt}5\hspace{0pt} to\hspace{0pt} a\hspace{0pt} =\textgreater{}\hspace{0pt} a\hspace{0pt}=\hspace{0pt}%
9\hspace{0pt}*\hspace{0pt}5\hspace{0pt}=\hspace{0pt}4\hspace{0pt}5\hspace{0pt},\hspace{0pt} b\hspace{0pt}%
=\hspace{0pt}1\hspace{0pt}6\hspace{0pt} (\hspace{0pt}4\hspace{0pt}5\hspace{0pt}\textgreater{}\hspace{0pt}%
1\hspace{0pt}6\hspace{0pt}:\hspace{0pt} no\hspace{0pt})\hspace{0pt}8\hspace{0pt}.\hspace{0pt} Assign\hspace{0pt}%
 all\hspace{0pt} three\hspace{0pt} primes\hspace{0pt} to\hspace{0pt} a\hspace{0pt} =\textgreater{}\hspace{0pt}%
a\hspace{0pt}=\hspace{0pt}7\hspace{0pt}2\hspace{0pt}0\hspace{0pt},\hspace{0pt} b\hspace{0pt}=\hspace{0pt}%
1\hspace{0pt} (\hspace{0pt}a\hspace{0pt}\textgreater{}\hspace{0pt} b\hspace{0pt}:\hspace{0pt} no\hspace{0pt}%
)\hspace{0pt}Therefore\hspace{0pt},\hspace{0pt} out\hspace{0pt} of\hspace{0pt} \hspace{0pt}8\hspace{0pt}%
 assignments\hspace{0pt},\hspace{0pt} \hspace{0pt}4\hspace{0pt} result\hspace{0pt} in\hspace{0pt} a\hspace{0pt}%
 \textless{}\hspace{0pt} b\hspace{0pt} and\hspace{0pt} \hspace{0pt}4\hspace{0pt} in\hspace{0pt} a\hspace{0pt}%
 \textgreater{}\hspace{0pt}b\hspace{0pt}.\hspace{0pt} So\hspace{0pt} when\hspace{0pt} \hspace{0pt}7\hspace{0pt}%
2\hspace{0pt}0\hspace{0pt} is\hspace{0pt} not\hspace{0pt} a\hspace{0pt} square\hspace{0pt},\hspace{0pt}%
 which\hspace{0pt} it\hspace{0pt} is\hspace{0pt} not\hspace{0pt} (\hspace{0pt}since\hspace{0pt} exponent\hspace{0pt}%
 of\hspace{0pt} \hspace{0pt}5\hspace{0pt} is\hspace{0pt} \hspace{0pt}1\hspace{0pt},\hspace{0pt} odd\hspace{0pt}%
),\hspace{0pt} thus\hspace{0pt} the\hspace{0pt} case\hspace{0pt} a\hspace{0pt}=b\hspace{0pt} is\hspace{0pt}%
 not\hspace{0pt} possible\hspace{0pt},\hspace{0pt} hence\hspace{0pt} the\hspace{0pt} number\hspace{0pt}%
 of\hspace{0pt} ordered\hspace{0pt} cop\hspace{0pt}r\hspace{0pt}ime\hspace{0pt} pairs\hspace{0pt} with\hspace{0pt}%
 a\hspace{0pt} \textless{}\hspace{0pt} b\hspace{0pt} is\hspace{0pt} half\hspace{0pt} the\hspace{0pt} total\hspace{0pt}%
,\hspace{0pt} which\hspace{0pt} is\hspace{0pt} \hspace{0pt}4\hspace{0pt}.\hspace{0pt} So\hspace{0pt} in\hspace{0pt}%
 general\hspace{0pt},\hspace{0pt} the\hspace{0pt} answer\hspace{0pt} seems\hspace{0pt} to\hspace{0pt}%
 be\hspace{0pt} half\hspace{0pt} of\hspace{0pt} the\hspace{0pt} number\hspace{0pt} of\hspace{0pt} cop\hspace{0pt}%
r\hspace{0pt}ime\hspace{0pt} ordered\hspace{0pt} pairs\hspace{0pt}.\hspace{0pt}Therefore\hspace{0pt},\hspace{0pt}%
 again\hspace{0pt},\hspace{0pt} number\hspace{0pt} for\hspace{0pt} \hspace{0pt}2\hspace{0pt}0\hspace{0pt}%
!\hspace{0pt} would\hspace{0pt} be\hspace{0pt} \hspace{0pt}2\hspace{0pt}\^{}\hspace{0pt}8\hspace{0pt}%
 /\hspace{0pt}2\hspace{0pt} =\hspace{0pt}1\hspace{0pt}2\hspace{0pt}8\hspace{0pt}.\hspace{0pt} So\hspace{0pt}%
 \hspace{0pt}1\hspace{0pt}2\hspace{0pt}8\hspace{0pt} cop\hspace{0pt}r\hspace{0pt}ime\hspace{0pt} pairs\hspace{0pt}%
 (\hspace{0pt}a\hspace{0pt},b\hspace{0pt})\hspace{0pt} with\hspace{0pt} a\hspace{0pt} \textless{}\hspace{0pt}%
 b\hspace{0pt}.\hspace{0pt} However\hspace{0pt},\hspace{0pt} perhaps\hspace{0pt} there\hspace{0pt}'s\hspace{0pt}%
 a\hspace{0pt} problem\hspace{0pt} here\hspace{0pt}.\hspace{0pt}Wait\hspace{0pt},\hspace{0pt} in\hspace{0pt}%
 the\hspace{0pt} case\hspace{0pt} when\hspace{0pt} assignments\hspace{0pt} are\hspace{0pt} made\hspace{0pt}%
 for\hspace{0pt} each\hspace{0pt} prime\hspace{0pt},\hspace{0pt} but\hspace{0pt},\hspace{0pt} in\hspace{0pt}%
 the\hspace{0pt} original\hspace{0pt} problem\hspace{0pt},\hspace{0pt} the\hspace{0pt} numerator\hspace{0pt}%
 and\hspace{0pt} denominator\hspace{0pt} are\hspace{0pt} reduced\hspace{0pt} to\hspace{0pt} lowest\hspace{0pt}%
 terms\hspace{0pt}.\hspace{0pt} Which\hspace{0pt} in\hspace{0pt} original\hspace{0pt} fraction\hspace{0pt}%
 it\hspace{0pt}'s\hspace{0pt} not\hspace{0pt} required\hspace{0pt} that\hspace{0pt} the\hspace{0pt} numerator\hspace{0pt}%
 and\hspace{0pt} denominator\hspace{0pt} multiply\hspace{0pt} to\hspace{0pt} \hspace{0pt}2\hspace{0pt}%
0\hspace{0pt}!\hspace{0pt};\hspace{0pt} rather\hspace{0pt},\hspace{0pt} the\hspace{0pt} reduced\hspace{0pt}%
 fraction\hspace{0pt} \textbackslash{}\hspace{0pt}(\textbackslash{}\hspace{0pt}frac\hspace{0pt}\{\hspace{0pt}%
a\hspace{0pt}\}\{\hspace{0pt}b\hspace{0pt}\}\textbackslash{}\hspace{0pt})\hspace{0pt} has\hspace{0pt}%
 the\hspace{0pt} product\hspace{0pt} a\hspace{0pt}*b\hspace{0pt} which\hspace{0pt} is\hspace{0pt} \hspace{0pt}%
2\hspace{0pt}0\hspace{0pt} factorial\hspace{0pt}.\hspace{0pt} Wait\hspace{0pt},\hspace{0pt} that\hspace{0pt}%
 needs\hspace{0pt} to\hspace{0pt} be\hspace{0pt} resolved\hspace{0pt}.\hspace{0pt}Wait\hspace{0pt},\hspace{0pt}%
 perhaps\hspace{0pt} that\hspace{0pt}'s\hspace{0pt} the\hspace{0pt} confusion\hspace{0pt}.\hspace{0pt}%
 The\hspace{0pt} original\hspace{0pt} number\hspace{0pt} is\hspace{0pt} some\hspace{0pt} fraction\hspace{0pt}%
 as\hspace{0pt} part\hspace{0pt} of\hspace{0pt} between\hspace{0pt} \hspace{0pt}0\hspace{0pt} and\hspace{0pt}%
 \hspace{0pt}1\hspace{0pt}.\hspace{0pt} When\hspace{0pt} written\hspace{0pt} in\hspace{0pt} lowest\hspace{0pt}%
 terms\hspace{0pt},\hspace{0pt} numerator\hspace{0pt} a\hspace{0pt} and\hspace{0pt} denominator\hspace{0pt}%
 b\hspace{0pt} must\hspace{0pt} satisfy\hspace{0pt} that\hspace{0pt}.\hspace{0pt} However\hspace{0pt}%
,\hspace{0pt} a\hspace{0pt} different\hspace{0pt} way\hspace{0pt} to\hspace{0pt} phrase\hspace{0pt} this\hspace{0pt}%
 is\hspace{0pt}:\hspace{0pt} a\hspace{0pt} is\hspace{0pt} numerator\hspace{0pt} and\hspace{0pt} b\hspace{0pt}%
 is\hspace{0pt} denominator\hspace{0pt} in\hspace{0pt} reduced\hspace{0pt} form\hspace{0pt} (\hspace{0pt}%
so\hspace{0pt} gcd\hspace{0pt}(a\hspace{0pt},\hspace{0pt} b\hspace{0pt})=\hspace{0pt}1\hspace{0pt}),\hspace{0pt}%
 and\hspace{0pt} when\hspace{0pt} you\hspace{0pt} compute\hspace{0pt} the\hspace{0pt} product\hspace{0pt}%
 a\hspace{0pt}*b\hspace{0pt},\hspace{0pt} this\hspace{0pt} product\hspace{0pt} is\hspace{0pt} equal\hspace{0pt}%
 to\hspace{0pt} \hspace{0pt}2\hspace{0pt}0\hspace{0pt}!.\hspace{0pt}H\hspace{0pt}ence\hspace{0pt},\hspace{0pt}%
 the\hspace{0pt} question\hspace{0pt} is\hspace{0pt} equivalent\hspace{0pt} to\hspace{0pt}:\hspace{0pt}%
 how\hspace{0pt} many\hspace{0pt} pairs\hspace{0pt} (\hspace{0pt}a\hspace{0pt},\hspace{0pt} b\hspace{0pt}%
)\hspace{0pt} are\hspace{0pt} there\hspace{0pt} with\hspace{0pt} gcd\hspace{0pt}(a\hspace{0pt},\hspace{0pt}%
 b\hspace{0pt})=\hspace{0pt}1\hspace{0pt},\hspace{0pt} \hspace{0pt}0\hspace{0pt} \textless{}\hspace{0pt}%
 a\hspace{0pt} \textless{}\hspace{0pt} b\hspace{0pt},\hspace{0pt} and\hspace{0pt} a\hspace{0pt} *\hspace{0pt}%
 b\hspace{0pt} =\hspace{0pt} \hspace{0pt}2\hspace{0pt}0\hspace{0pt}!\hspace{0pt}Which\hspace{0pt} translates\hspace{0pt}%
 as\hspace{0pt},\hspace{0pt} (\hspace{0pt}since\hspace{0pt} \hspace{0pt}2\hspace{0pt}0\hspace{0pt}!\hspace{0pt}%
 is\hspace{0pt} fixed\hspace{0pt} and\hspace{0pt} a\hspace{0pt}*b\hspace{0pt}=\hspace{0pt}2\hspace{0pt}%
0\hspace{0pt}!),\hspace{0pt} how\hspace{0pt} many\hspace{0pt} ways\hspace{0pt} can\hspace{0pt} \hspace{0pt}%
2\hspace{0pt}0\hspace{0pt}!\hspace{0pt} be\hspace{0pt} split\hspace{0pt} into\hspace{0pt} two\hspace{0pt}%
 cop\hspace{0pt}r\hspace{0pt}ime\hspace{0pt} factors\hspace{0pt} a\hspace{0pt} and\hspace{0pt} b\hspace{0pt}%
,\hspace{0pt} both\hspace{0pt} positive\hspace{0pt} integers\hspace{0pt},\hspace{0pt} a\hspace{0pt} \textless{}\hspace{0pt}%
 b\hspace{0pt}.\hspace{0pt}And\hspace{0pt} this\hspace{0pt} number\hspace{0pt} is\hspace{0pt} equal\hspace{0pt}%
 to\hspace{0pt} \hspace{0pt}2\hspace{0pt}\^{}\{\hspace{0pt}k\hspace{0pt} -\hspace{0pt}1\hspace{0pt}\},\hspace{0pt}%
 where\hspace{0pt} k\hspace{0pt} is\hspace{0pt} the\hspace{0pt} number\hspace{0pt} of\hspace{0pt} distinct\hspace{0pt}%
 prime\hspace{0pt} factors\hspace{0pt} of\hspace{0pt} \hspace{0pt}2\hspace{0pt}0\hspace{0pt}!.\hspace{0pt}%
But\hspace{0pt} since\hspace{0pt} the\hspace{0pt} number\hspace{0pt} of\hspace{0pt} distinct\hspace{0pt}%
 prime\hspace{0pt} factors\hspace{0pt} of\hspace{0pt} \hspace{0pt}2\hspace{0pt}0\hspace{0pt}!\hspace{0pt}%
 is\hspace{0pt} \hspace{0pt}8\hspace{0pt} (\hspace{0pt}since\hspace{0pt} primes\hspace{0pt} \hspace{0pt}%
2\hspace{0pt},\hspace{0pt} \hspace{0pt}3\hspace{0pt},\hspace{0pt} \hspace{0pt}5\hspace{0pt},\hspace{0pt}%
 \hspace{0pt}7\hspace{0pt},\hspace{0pt} \hspace{0pt}1\hspace{0pt}1\hspace{0pt},\hspace{0pt}1\hspace{0pt}%
3\hspace{0pt},\hspace{0pt}1\hspace{0pt}7\hspace{0pt},\hspace{0pt}1\hspace{0pt}9\hspace{0pt}).\hspace{0pt}%
 Hence\hspace{0pt},\hspace{0pt} \hspace{0pt}2\hspace{0pt}\^{}\{\hspace{0pt}8\hspace{0pt} -\hspace{0pt}%
1\hspace{0pt}\}=\hspace{0pt} \hspace{0pt}1\hspace{0pt}2\hspace{0pt}8\hspace{0pt}.\hspace{0pt} Hence\hspace{0pt}%
,\hspace{0pt} result\hspace{0pt}.\hspace{0pt} Wait\hspace{0pt},\hspace{0pt} this\hspace{0pt} would\hspace{0pt}%
 be\hspace{0pt} \hspace{0pt}2\hspace{0pt}\^{}\{\hspace{0pt}k\hspace{0pt}-\hspace{0pt}1\hspace{0pt}\}=\hspace{0pt}%
1\hspace{0pt}2\hspace{0pt}8\hspace{0pt}.\hspace{0pt} Alternatively\hspace{0pt},\hspace{0pt} total\hspace{0pt}%
 number\hspace{0pt} of\hspace{0pt} subsets\hspace{0pt} is\hspace{0pt} \hspace{0pt}2\hspace{0pt}\^{}\hspace{0pt}%
8\hspace{0pt}=\hspace{0pt}2\hspace{0pt}5\hspace{0pt}6\hspace{0pt},\hspace{0pt} each\hspace{0pt} ordered\hspace{0pt}%
 pairs\hspace{0pt}.\hspace{0pt} Taking\hspace{0pt} half\hspace{0pt} because\hspace{0pt} a\hspace{0pt}%
 \textless{}\hspace{0pt} b\hspace{0pt} hence\hspace{0pt} \hspace{0pt}2\hspace{0pt}5\hspace{0pt}6\hspace{0pt}%
/\hspace{0pt}2\hspace{0pt}=\hspace{0pt}1\hspace{0pt}2\hspace{0pt}8\hspace{0pt}.\hspace{0pt} So\hspace{0pt}%
 confirms\hspace{0pt}.\hspace{0pt}\heatbox{0.982,0.655,0.655}{But}\hspace{0pt}\heatbox{0.982,0.655,0.655}{ let}\hspace{0pt}%
\heatbox{0.982,0.655,0.655}{ me}\hspace{0pt}\heatbox{0.982,0.655,0.655}{ check}\hspace{0pt}\heatbox{0.982,0.655,0.655}{ with}\hspace{0pt}%
\heatbox{0.982,0.655,0.655}{ the}\hspace{0pt}\heatbox{0.982,0.655,0.655}{ }\hspace{0pt}\heatbox{0.982,0.655,0.655}{6}\hspace{0pt}%
\heatbox{0.982,0.655,0.655}{!}\hspace{0pt}\heatbox{0.982,0.655,0.655}{ example}\hspace{0pt}\heatbox{0.982,0.655,0.655}{.}\hspace{0pt}%
\heatbox{0.982,0.655,0.655}{ For}\hspace{0pt}\heatbox{0.982,0.655,0.655}{ }\hspace{0pt}\heatbox{0.982,0.655,0.655}{6}\hspace{0pt}%
\heatbox{0.982,0.655,0.655}{!,}\hspace{0pt}\heatbox{0.982,0.655,0.655}{ which}\hspace{0pt}\heatbox{0.982,0.655,0.655}{ had}\hspace{0pt}%
\heatbox{0.982,0.655,0.655}{ instead}\hspace{0pt}\heatbox{0.982,0.655,0.655}{ three}\hspace{0pt}\heatbox{0.982,0.655,0.655}{ distinct}\hspace{0pt}%
\heatbox{0.982,0.655,0.655}{ prime}\hspace{0pt}\heatbox{0.982,0.655,0.655}{ factors}\hspace{0pt}\heatbox{0.982,0.655,0.655}{ (}\hspace{0pt}%
\heatbox{0.982,0.655,0.655}{2}\hspace{0pt}\heatbox{0.982,0.655,0.655}{,}\hspace{0pt}\heatbox{0.982,0.655,0.655}{3}\hspace{0pt}%
\heatbox{0.982,0.655,0.655}{,}\hspace{0pt}\heatbox{0.982,0.655,0.655}{5}\hspace{0pt}\heatbox{0.982,0.655,0.655}{),}\hspace{0pt}%
\heatbox{0.982,0.655,0.655}{ number}\hspace{0pt}\heatbox{0.982,0.655,0.655}{ of}\hspace{0pt}\heatbox{0.982,0.655,0.655}{ cop}\hspace{0pt}%
\heatbox{0.982,0.655,0.655}{r}\hspace{0pt}\heatbox{0.982,0.655,0.655}{ime}\hspace{0pt}\heatbox{0.982,0.655,0.655}{ pairs}\hspace{0pt}%
\heatbox{0.982,0.655,0.655}{ where}\hspace{0pt}\heatbox{0.982,0.655,0.655}{ a}\hspace{0pt}\heatbox{0.982,0.655,0.655}{ \textless{}}\hspace{0pt}%
\heatbox{0.982,0.655,0.655}{ b}\hspace{0pt}\heatbox{0.982,0.655,0.655}{ is}\hspace{0pt}\heatbox{0.982,0.655,0.655}{ }\hspace{0pt}%
\heatbox{0.982,0.655,0.655}{4}\hspace{0pt}\heatbox{0.982,0.655,0.655}{ (}\hspace{0pt}\heatbox{0.982,0.655,0.655}{by}\hspace{0pt}%
\heatbox{0.982,0.655,0.655}{ explicit}\hspace{0pt}\heatbox{0.982,0.655,0.655}{ count}\hspace{0pt}\heatbox{0.982,0.655,0.655}{ above}\hspace{0pt}%
\heatbox{0.982,0.655,0.655}{).}\hspace{0pt}\heatbox{0.982,0.655,0.655}{ But}\hspace{0pt}\heatbox{0.982,0.655,0.655}{ }\hspace{0pt}%
\heatbox{0.982,0.655,0.655}{2}\hspace{0pt}\heatbox{0.982,0.655,0.655}{\^{}\{}\hspace{0pt}\heatbox{0.982,0.655,0.655}{3}\hspace{0pt}%
\heatbox{0.982,0.655,0.655}{-}\hspace{0pt}\heatbox{0.982,0.655,0.655}{1}\hspace{0pt}\heatbox{0.982,0.655,0.655}{\}=}\hspace{0pt}%
\heatbox{0.982,0.655,0.655}{ }\hspace{0pt}\heatbox{0.982,0.655,0.655}{4}\hspace{0pt}\heatbox{0.982,0.655,0.655}{ correct}\hspace{0pt}%
\heatbox{0.982,0.655,0.655}{.}\hspace{0pt}\heatbox{0.982,0.655,0.655}{ Hence}\hspace{0pt}\heatbox{0.982,0.655,0.655}{,}\hspace{0pt}%
\heatbox{0.982,0.655,0.655}{ Therefore}\hspace{0pt}\heatbox{0.982,0.655,0.655}{,}\hspace{0pt}\heatbox{0.982,0.655,0.655}{ the}\hspace{0pt}%
\heatbox{0.982,0.655,0.655}{ formula}\hspace{0pt}\heatbox{0.982,0.655,0.655}{ is}\hspace{0pt}\heatbox{0.982,0.655,0.655}{ }\hspace{0pt}%
\heatbox{0.982,0.655,0.655}{2}\hspace{0pt}\heatbox{0.982,0.655,0.655}{\^{}\{}\hspace{0pt}\heatbox{0.982,0.655,0.655}{k}\hspace{0pt}%
\heatbox{0.982,0.655,0.655}{ -}\hspace{0pt}\heatbox{0.982,0.655,0.655}{1}\hspace{0pt}).\hspace{0pt}H\hspace{0pt}%
ence\hspace{0pt},\hspace{0pt} for\hspace{0pt} \hspace{0pt}2\hspace{0pt}0\hspace{0pt}!\hspace{0pt} with\hspace{0pt}%
 \hspace{0pt}8\hspace{0pt} prime\hspace{0pt} factors\hspace{0pt},\hspace{0pt} number\hspace{0pt} of\hspace{0pt}%
 cop\hspace{0pt}r\hspace{0pt}ime\hspace{0pt} pairs\hspace{0pt} with\hspace{0pt} a\hspace{0pt} \textless{}\hspace{0pt}%
 b\hspace{0pt} and\hspace{0pt} a\hspace{0pt}*b\hspace{0pt}=\hspace{0pt}2\hspace{0pt}0\hspace{0pt}!\hspace{0pt}%
 is\hspace{0pt} \hspace{0pt}2\hspace{0pt}\^{}\{\hspace{0pt}8\hspace{0pt}-\hspace{0pt}1\hspace{0pt}\}=\hspace{0pt}%
1\hspace{0pt}2\hspace{0pt}8\hspace{0pt},\hspace{0pt} which\hspace{0pt} seems\hspace{0pt} correct\hspace{0pt}%
.\hspace{0pt}But\hspace{0pt} Wait\hspace{0pt},\hspace{0pt} what\hspace{0pt} about\hspace{0pt} multip\hspace{0pt}%
licity\hspace{0pt}.\hspace{0pt} For\hspace{0pt} example\hspace{0pt},\hspace{0pt} \hspace{0pt}1\hspace{0pt}%
 and\hspace{0pt} \hspace{0pt}7\hspace{0pt}2\hspace{0pt}0\hspace{0pt} were\hspace{0pt} co\hspace{0pt}-pr\hspace{0pt}%
ime\hspace{0pt} (\hspace{0pt}a\hspace{0pt}=\hspace{0pt}1\hspace{0pt},\hspace{0pt} b\hspace{0pt}=\hspace{0pt}%
7\hspace{0pt}2\hspace{0pt}0\hspace{0pt}).\hspace{0pt} Then\hspace{0pt} you\hspace{0pt} had\hspace{0pt}%
 \hspace{0pt}1\hspace{0pt}6\hspace{0pt} and\hspace{0pt} \hspace{0pt}4\hspace{0pt}5\hspace{0pt},\hspace{0pt}%
 and\hspace{0pt} \hspace{0pt}9\hspace{0pt} and\hspace{0pt} \hspace{0pt}8\hspace{0pt}0\hspace{0pt},\hspace{0pt}%
 and\hspace{0pt} \hspace{0pt}5\hspace{0pt} and\hspace{0pt} \hspace{0pt}1\hspace{0pt}4\hspace{0pt}4\hspace{0pt}%
.\hspace{0pt} How\hspace{0pt} is\hspace{0pt} it\hspace{0pt} multip\hspace{0pt}licative\hspace{0pt},\hspace{0pt}%
 related\hspace{0pt} to\hspace{0pt} the\hspace{0pt} number\hspace{0pt} of\hspace{0pt} prime\hspace{0pt}%
 factors\hspace{0pt}?\hspace{0pt}\heatbox{0.981,0.645,0.645}{Yes}\hspace{0pt}\heatbox{0.981,0.645,0.645}{,}\hspace{0pt}%
\heatbox{0.981,0.645,0.645}{ but}\hspace{0pt}\heatbox{0.981,0.645,0.645}{ }\hspace{0pt}\heatbox{0.981,0.645,0.645}{6}\hspace{0pt}%
\heatbox{0.981,0.645,0.645}{!}\hspace{0pt}\heatbox{0.981,0.645,0.645}{ is}\hspace{0pt}\heatbox{0.981,0.645,0.645}{ }\hspace{0pt}%
\heatbox{0.981,0.645,0.645}{7}\hspace{0pt}\heatbox{0.981,0.645,0.645}{2}\hspace{0pt}\heatbox{0.981,0.645,0.645}{0}\hspace{0pt}%
\heatbox{0.981,0.645,0.645}{,}\hspace{0pt}\heatbox{0.981,0.645,0.645}{ and}\hspace{0pt}\heatbox{0.981,0.645,0.645}{ had}\hspace{0pt}%
\heatbox{0.981,0.645,0.645}{ three}\hspace{0pt}\heatbox{0.981,0.645,0.645}{ prime}\hspace{0pt}\heatbox{0.981,0.645,0.645}{ factors}\hspace{0pt}%
\heatbox{0.981,0.645,0.645}{.}\hspace{0pt}\heatbox{0.981,0.645,0.645}{ The}\hspace{0pt}\heatbox{0.981,0.645,0.645}{ number}\hspace{0pt}%
\heatbox{0.981,0.645,0.645}{ of}\hspace{0pt}\heatbox{0.981,0.645,0.645}{ cop}\hspace{0pt}\heatbox{0.981,0.645,0.645}{r}\hspace{0pt}%
\heatbox{0.981,0.645,0.645}{ime}\hspace{0pt}\heatbox{0.981,0.645,0.645}{ pairs}\hspace{0pt}\heatbox{0.981,0.645,0.645}{ a}\hspace{0pt}%
\heatbox{0.981,0.645,0.645}{ \textless{}}\hspace{0pt}\heatbox{0.981,0.645,0.645}{ b}\hspace{0pt}\heatbox{0.981,0.645,0.645}{ where}\hspace{0pt}%
\heatbox{0.981,0.645,0.645}{ a}\hspace{0pt}\heatbox{0.981,0.645,0.645}{*b}\hspace{0pt}\heatbox{0.981,0.645,0.645}{=}\hspace{0pt}%
\heatbox{0.981,0.645,0.645}{7}\hspace{0pt}\heatbox{0.981,0.645,0.645}{2}\hspace{0pt}\heatbox{0.981,0.645,0.645}{0}\hspace{0pt}%
\heatbox{0.981,0.645,0.645}{ is}\hspace{0pt}\heatbox{0.981,0.645,0.645}{ }\hspace{0pt}\heatbox{0.981,0.645,0.645}{4}\hspace{0pt}%
\heatbox{0.981,0.645,0.645}{,}\hspace{0pt}\heatbox{0.981,0.645,0.645}{ which}\hspace{0pt}\heatbox{0.981,0.645,0.645}{ is}\hspace{0pt}%
\heatbox{0.981,0.645,0.645}{ }\hspace{0pt}\heatbox{0.981,0.645,0.645}{2}\hspace{0pt}\heatbox{0.981,0.645,0.645}{\^{}\{}\hspace{0pt}%
\heatbox{0.981,0.645,0.645}{3}\hspace{0pt}\heatbox{0.981,0.645,0.645}{-}\hspace{0pt}\heatbox{0.981,0.645,0.645}{1}\hspace{0pt}%
\heatbox{0.981,0.645,0.645}{\}=}\hspace{0pt}\heatbox{0.981,0.645,0.645}{4}\hspace{0pt}.\hspace{0pt}\heatbox{0.984,0.688,0.688}{Similarly}\hspace{0pt}%
\heatbox{0.984,0.688,0.688}{,}\hspace{0pt}\heatbox{0.984,0.688,0.688}{ if}\hspace{0pt}\heatbox{0.984,0.688,0.688}{ n}\hspace{0pt}%
\heatbox{0.984,0.688,0.688}{ is}\hspace{0pt}\heatbox{0.984,0.688,0.688}{ a}\hspace{0pt}\heatbox{0.984,0.688,0.688}{ square}\hspace{0pt}%
\heatbox{0.984,0.688,0.688}{-free}\hspace{0pt}\heatbox{0.984,0.688,0.688}{ number}\hspace{0pt}\heatbox{0.984,0.688,0.688}{,}\hspace{0pt}%
\heatbox{0.984,0.688,0.688}{ which}\hspace{0pt}\heatbox{0.984,0.688,0.688}{ }\hspace{0pt}\heatbox{0.984,0.688,0.688}{7}\hspace{0pt}%
\heatbox{0.984,0.688,0.688}{2}\hspace{0pt}\heatbox{0.984,0.688,0.688}{0}\hspace{0pt}\heatbox{0.984,0.688,0.688}{ is}\hspace{0pt}%
\heatbox{0.984,0.688,0.688}{ not}\hspace{0pt}\heatbox{0.984,0.688,0.688}{ square}\hspace{0pt}\heatbox{0.984,0.688,0.688}{-free}\hspace{0pt}%
\heatbox{0.984,0.688,0.688}{ (}\hspace{0pt}\heatbox{0.984,0.688,0.688}{as}\hspace{0pt}\heatbox{0.984,0.688,0.688}{ }\hspace{0pt}%
\heatbox{0.984,0.688,0.688}{7}\hspace{0pt}\heatbox{0.984,0.688,0.688}{2}\hspace{0pt}\heatbox{0.984,0.688,0.688}{0}\hspace{0pt}%
\heatbox{0.984,0.688,0.688}{=}\hspace{0pt}\heatbox{0.984,0.688,0.688}{1}\hspace{0pt}\heatbox{0.984,0.688,0.688}{6}\hspace{0pt}%
\heatbox{0.984,0.688,0.688}{*}\hspace{0pt}\heatbox{0.984,0.688,0.688}{4}\hspace{0pt}\heatbox{0.984,0.688,0.688}{5}\hspace{0pt}%
\heatbox{0.984,0.688,0.688}{=}\hspace{0pt}\heatbox{0.984,0.688,0.688}{2}\hspace{0pt}\heatbox{0.984,0.688,0.688}{4}\hspace{0pt}%
\heatbox{0.984,0.688,0.688}{\^{}}\hspace{0pt}\heatbox{0.984,0.688,0.688}{2}\hspace{0pt}\heatbox{0.984,0.688,0.688}{*}\hspace{0pt}%
\heatbox{0.984,0.688,0.688}{5}\hspace{0pt}).\hspace{0pt}But\hspace{0pt} in\hspace{0pt} general\hspace{0pt}%
,\hspace{0pt} n\hspace{0pt}!\hspace{0pt} will\hspace{0pt} have\hspace{0pt} prime\hspace{0pt} factors\hspace{0pt}%
 multiplic\hspace{0pt}ities\hspace{0pt}.\hspace{0pt}But\hspace{0pt} our\hspace{0pt} problem\hspace{0pt}%
 states\hspace{0pt} n\hspace{0pt}=\hspace{0pt}2\hspace{0pt}0\hspace{0pt}.\hspace{0pt} So\hspace{0pt} \hspace{0pt}%
2\hspace{0pt}0\hspace{0pt}!.\hspace{0pt}\heatbox{0.984,0.689,0.689}{Regardless}\hspace{0pt}\heatbox{0.984,0.689,0.689}{ of}\hspace{0pt}%
\heatbox{0.984,0.689,0.689}{ the}\hspace{0pt}\heatbox{0.984,0.689,0.689}{ multiplic}\hspace{0pt}\heatbox{0.984,0.689,0.689}{ities}\hspace{0pt}%
\heatbox{0.984,0.689,0.689}{ (}\hspace{0pt}\heatbox{0.984,0.689,0.689}{since}\hspace{0pt}\heatbox{0.984,0.689,0.689}{ in}\hspace{0pt}%
\heatbox{0.984,0.689,0.689}{ order}\hspace{0pt}\heatbox{0.984,0.689,0.689}{ for}\hspace{0pt}\heatbox{0.984,0.689,0.689}{ a}\hspace{0pt}%
\heatbox{0.984,0.689,0.689}{ and}\hspace{0pt}\heatbox{0.984,0.689,0.689}{ b}\hspace{0pt}\heatbox{0.984,0.689,0.689}{ to}\hspace{0pt}%
\heatbox{0.984,0.689,0.689}{ be}\hspace{0pt}\heatbox{0.984,0.689,0.689}{ cop}\hspace{0pt}\heatbox{0.984,0.689,0.689}{r}\hspace{0pt}%
\heatbox{0.984,0.689,0.689}{ime}\hspace{0pt}\heatbox{0.984,0.689,0.689}{,}\hspace{0pt}\heatbox{0.984,0.689,0.689}{ the}\hspace{0pt}%
\heatbox{0.984,0.689,0.689}{ entire}\hspace{0pt}\heatbox{0.984,0.689,0.689}{ allocation}\hspace{0pt}\heatbox{0.984,0.689,0.689}{ of}\hspace{0pt}%
\heatbox{0.984,0.689,0.689}{ each}\hspace{0pt}\heatbox{0.984,0.689,0.689}{ prime}\hspace{0pt}\heatbox{0.984,0.689,0.689}{ has}\hspace{0pt}%
\heatbox{0.984,0.689,0.689}{ to}\hspace{0pt}\heatbox{0.984,0.689,0.689}{ go}\hspace{0pt}\heatbox{0.984,0.689,0.689}{ to}\hspace{0pt}%
\heatbox{0.984,0.689,0.689}{ one}\hspace{0pt}\heatbox{0.984,0.689,0.689}{ factor}\hspace{0pt}\heatbox{0.984,0.689,0.689}{).}\hspace{0pt}%
\heatbox{0.984,0.689,0.689}{ Therefore}\hspace{0pt}\heatbox{0.984,0.689,0.689}{,}\hspace{0pt}\heatbox{0.984,0.689,0.689}{ how}\hspace{0pt}%
\heatbox{0.984,0.689,0.689}{ primes}\hspace{0pt}\heatbox{0.984,0.689,0.689}{ are}\hspace{0pt}\heatbox{0.984,0.689,0.689}{ assigned}\hspace{0pt}%
\heatbox{0.984,0.689,0.689}{ has}\hspace{0pt}\heatbox{0.984,0.689,0.689}{ a}\hspace{0pt}\heatbox{0.984,0.689,0.689}{ per}\hspace{0pt}%
\heatbox{0.984,0.689,0.689}{-pr}\hspace{0pt}\heatbox{0.984,0.689,0.689}{ime}\hspace{0pt}\heatbox{0.984,0.689,0.689}{ dependence}\hspace{0pt}%
.\hspace{0pt}Wait\hspace{0pt},\hspace{0pt} so\hspace{0pt} since\hspace{0pt} for\hspace{0pt} each\hspace{0pt}%
 prime\hspace{0pt},\hspace{0pt} regardless\hspace{0pt} of\hspace{0pt} exponent\hspace{0pt} (\hspace{0pt}%
even\hspace{0pt} or\hspace{0pt} odd\hspace{0pt}),\hspace{0pt} we\hspace{0pt} must\hspace{0pt} give\hspace{0pt}%
 it\hspace{0pt} entirely\hspace{0pt} to\hspace{0pt} a\hspace{0pt} or\hspace{0pt} b\hspace{0pt},\hspace{0pt}%
 which\hspace{0pt} leads\hspace{0pt} to\hspace{0pt} these\hspace{0pt} cop\hspace{0pt}r\hspace{0pt}ime\hspace{0pt}%
 pairs\hspace{0pt}.\hspace{0pt} So\hspace{0pt} seems\hspace{0pt} that\hspace{0pt} the\hspace{0pt} number\hspace{0pt}%
 is\hspace{0pt},\hspace{0pt} in\hspace{0pt} general\hspace{0pt},\hspace{0pt} \hspace{0pt}2\hspace{0pt}%
\^{}\{\hspace{0pt}k\hspace{0pt}\}\hspace{0pt} ordered\hspace{0pt} cop\hspace{0pt}r\hspace{0pt}ime\hspace{0pt}%
 pairs\hspace{0pt} (\hspace{0pt}a\hspace{0pt},b\hspace{0pt}),\hspace{0pt} and\hspace{0pt} because\hspace{0pt}%
 n\hspace{0pt}!\hspace{0pt} is\hspace{0pt} not\hspace{0pt} a\hspace{0pt} square\hspace{0pt} (\hspace{0pt}%
unless\hspace{0pt} for\hspace{0pt} example\hspace{0pt} primes\hspace{0pt} with\hspace{0pt} even\hspace{0pt}%
 ex\hspace{0pt}ponents\hspace{0pt}),\hspace{0pt} it\hspace{0pt} depends\hspace{0pt}.

...(Omit)...
 but\hspace{0pt} depending\hspace{0pt} on\hspace{0pt} which\hspace{0pt} primes\hspace{0pt}.\hspace{0pt}%
 But\hspace{0pt} here\hspace{0pt} is\hspace{0pt} the\hspace{0pt} cr\hspace{0pt}ux\hspace{0pt}:\hspace{0pt}%
 instead\hspace{0pt} of\hspace{0pt} trying\hspace{0pt} to\hspace{0pt} group\hspace{0pt},\hspace{0pt} this\hspace{0pt}%
 is\hspace{0pt} again\hspace{0pt} difficult\hspace{0pt} to\hspace{0pt} calculate\hspace{0pt} by\hspace{0pt}%
 separate\hspace{0pt} method\hspace{0pt};\hspace{0pt} but\hspace{0pt} via\hspace{0pt} generating\hspace{0pt}%
 the\hspace{0pt} cop\hspace{0pt}rim\hspace{0pt}ality\hspace{0pt} by\hspace{0pt} prime\hspace{0pt} division\hspace{0pt}%
,\hspace{0pt} and\hspace{0pt} noting\hspace{0pt} in\hspace{0pt} N\hspace{0pt}!\hspace{0pt} is\hspace{0pt}%
 fixed\hspace{0pt},\hspace{0pt} so\hspace{0pt} that\hspace{0pt} the\hspace{0pt} only\hspace{0pt} way\hspace{0pt}%
 to\hspace{0pt} a\hspace{0pt}*b\hspace{0pt}=N\hspace{0pt}!\hspace{0pt}:\hspace{0pt}\heatbox{0.981,0.647,0.647}{Since}\hspace{0pt}%
\heatbox{0.981,0.647,0.647}{ a}\hspace{0pt}\heatbox{0.981,0.647,0.647}{ and}\hspace{0pt}\heatbox{0.981,0.647,0.647}{ b}\hspace{0pt}%
\heatbox{0.981,0.647,0.647}{ are}\hspace{0pt}\heatbox{0.981,0.647,0.647}{ cop}\hspace{0pt}\heatbox{0.981,0.647,0.647}{r}\hspace{0pt}%
\heatbox{0.981,0.647,0.647}{ime}\hspace{0pt}.\hspace{0pt}\heatbox{0.983,0.682,0.682}{H}\hspace{0pt}\heatbox{0.983,0.682,0.682}{ence}\hspace{0pt}%
\heatbox{0.983,0.682,0.682}{ equivalence}\hspace{0pt}\heatbox{0.983,0.682,0.682}{ between}\hspace{0pt}%
\heatbox{0.983,0.682,0.682}{ choices}\hspace{0pt}\heatbox{0.983,0.682,0.682}{ over}\hspace{0pt}\heatbox{0.983,0.682,0.682}{ subsets}\hspace{0pt}%
\heatbox{0.983,0.682,0.682}{ of}\hspace{0pt}\heatbox{0.983,0.682,0.682}{ primes}\hspace{0pt}.\hspace{0pt}%
H\hspace{0pt}ence\hspace{0pt},\hspace{0pt} since\hspace{0pt} ordered\hspace{0pt} cop\hspace{0pt}r\hspace{0pt}%
imes\hspace{0pt},\hspace{0pt} \hspace{0pt}2\hspace{0pt}\^{}\hspace{0pt}pairs\hspace{0pt}.\hspace{0pt}%
Each\hspace{0pt} with\hspace{0pt} a\hspace{0pt} Unique\hspace{0pt} Sub\hspace{0pt}sets\hspace{0pt} of\hspace{0pt}%
 primes\hspace{0pt} assigned\hspace{0pt} to\hspace{0pt} a\hspace{0pt}.\hspace{0pt}H\hspace{0pt}ence\hspace{0pt}%
,\hspace{0pt} since\hspace{0pt} ordering\hspace{0pt} for\hspace{0pt} each\hspace{0pt} subset\hspace{0pt}%
,\hspace{0pt} half\hspace{0pt} will\hspace{0pt} be\hspace{0pt} such\hspace{0pt} that\hspace{0pt} the\hspace{0pt}%
 a\hspace{0pt} is\hspace{0pt} less\hspace{0pt} than\hspace{0pt} b\hspace{0pt}.\hspace{0pt}H\hspace{0pt}%
ence\hspace{0pt} if\hspace{0pt} done\hspace{0pt} for\hspace{0pt} k\hspace{0pt} primes\hspace{0pt},\hspace{0pt}%
 because\hspace{0pt} N\hspace{0pt}!\hspace{0pt} is\hspace{0pt} not\hspace{0pt} a\hspace{0pt} square\hspace{0pt}%
,\hspace{0pt} then\hspace{0pt} exactly\hspace{0pt} half\hspace{0pt} of\hspace{0pt} those\hspace{0pt} subsets\hspace{0pt}%
 gives\hspace{0pt} a\hspace{0pt}*a\hspace{0pt} less\hspace{0pt} than\hspace{0pt} N\hspace{0pt}!\hspace{0pt}%
 ,\hspace{0pt} that\hspace{0pt} is\hspace{0pt},\hspace{0pt} a\hspace{0pt} \textless{}\hspace{0pt} sqrt\hspace{0pt}%
(N\hspace{0pt}!).\hspace{0pt}But\hspace{0pt} for\hspace{0pt} all\hspace{0pt} possible\hspace{0pt} assignments\hspace{0pt}%
,\hspace{0pt} the\hspace{0pt} total\hspace{0pt} number\hspace{0pt} of\hspace{0pt} cases\hspace{0pt} where\hspace{0pt}%
 a\hspace{0pt} \textless{}\hspace{0pt} b\hspace{0pt} should\hspace{0pt} be\hspace{0pt} equal\hspace{0pt}%
 to\hspace{0pt} \hspace{0pt}2\hspace{0pt}\^{}\{\hspace{0pt}k\hspace{0pt}-\hspace{0pt}1\hspace{0pt}\}\hspace{0pt}%
 as\hspace{0pt} reasoned\hspace{0pt} before\hspace{0pt}.\hspace{0pt}Therefore\hspace{0pt},\hspace{0pt}%
 the\hspace{0pt} previous\hspace{0pt} answer\hspace{0pt} \hspace{0pt}1\hspace{0pt}2\hspace{0pt}8\hspace{0pt}%
 must\hspace{0pt} be\hspace{0pt} correct\hspace{0pt},\hspace{0pt} hence\hspace{0pt} The\hspace{0pt} answer\hspace{0pt}%
 would\hspace{0pt} be\hspace{0pt} \hspace{0pt}1\hspace{0pt}2\hspace{0pt}8\hspace{0pt}.\hspace{0pt}\heatbox{0.974,0.507,0.507}{But}\hspace{0pt}%
\heatbox{0.974,0.507,0.507}{ before}\hspace{0pt}\heatbox{0.974,0.507,0.507}{ }\hspace{0pt}\heatbox{0.974,0.507,0.507}{ closing}\hspace{0pt}%
\heatbox{0.974,0.507,0.507}{,}\hspace{0pt}\heatbox{0.974,0.507,0.507}{ verifying}\hspace{0pt}\heatbox{0.974,0.507,0.507}{ with}\hspace{0pt}%
\heatbox{0.974,0.507,0.507}{ our}\hspace{0pt}\heatbox{0.974,0.507,0.507}{ }\hspace{0pt}\heatbox{0.974,0.507,0.507}{6}\hspace{0pt}%
\heatbox{0.974,0.507,0.507}{!}\hspace{0pt}\heatbox{0.974,0.507,0.507}{ example}\hspace{0pt}\heatbox{0.974,0.507,0.507}{ was}\hspace{0pt}%
\heatbox{0.974,0.507,0.507}{ key}\hspace{0pt}.\hspace{0pt}\heatbox{0.975,0.530,0.530}{In}\hspace{0pt}%
\heatbox{0.975,0.530,0.530}{ the}\hspace{0pt}\heatbox{0.975,0.530,0.530}{ }\hspace{0pt}\heatbox{0.975,0.530,0.530}{6}\hspace{0pt}%
\heatbox{0.975,0.530,0.530}{!}\hspace{0pt}\heatbox{0.975,0.530,0.530}{ example}\hspace{0pt}\heatbox{0.975,0.530,0.530}{ (}\hspace{0pt}%
\heatbox{0.975,0.530,0.530}{k}\hspace{0pt}\heatbox{0.975,0.530,0.530}{=}\hspace{0pt}\heatbox{0.975,0.530,0.530}{3}\hspace{0pt}%
\heatbox{0.975,0.530,0.530}{ primes}\hspace{0pt}\heatbox{0.975,0.530,0.530}{),}\hspace{0pt}\heatbox{0.975,0.530,0.530}{ number}\hspace{0pt}%
\heatbox{0.975,0.530,0.530}{ of}\hspace{0pt}\heatbox{0.975,0.530,0.530}{ cop}\hspace{0pt}\heatbox{0.975,0.530,0.530}{r}\hspace{0pt}%
\heatbox{0.975,0.530,0.530}{ime}\hspace{0pt}\heatbox{0.975,0.530,0.530}{ pairs}\hspace{0pt}\heatbox{0.975,0.530,0.530}{ with}\hspace{0pt}%
\heatbox{0.975,0.530,0.530}{ a}\hspace{0pt}\heatbox{0.975,0.530,0.530}{ \textless{}}\hspace{0pt}\heatbox{0.975,0.530,0.530}{ b}\hspace{0pt}%
\heatbox{0.975,0.530,0.530}{=}\hspace{0pt}\heatbox{0.975,0.530,0.530}{4}\hspace{0pt}\heatbox{0.975,0.530,0.530}{=}\hspace{0pt}%
\heatbox{0.975,0.530,0.530}{2}\hspace{0pt}\heatbox{0.975,0.530,0.530}{\^{}\{}\hspace{0pt}\heatbox{0.975,0.530,0.530}{3}\hspace{0pt}%
\heatbox{0.975,0.530,0.530}{-}\hspace{0pt}\heatbox{0.975,0.530,0.530}{1}\hspace{0pt}\heatbox{0.975,0.530,0.530}{\}}\hspace{0pt}%
.\hspace{0pt}\heatbox{0.973,0.490,0.490}{And}\hspace{0pt}\heatbox{0.973,0.490,0.490}{ with}\hspace{0pt}%
\heatbox{0.973,0.490,0.490}{ actual}\hspace{0pt}\heatbox{0.973,0.490,0.490}{ calculation}\hspace{0pt}%
\heatbox{0.973,0.490,0.490}{ matching}\hspace{0pt}.\hspace{0pt}\heatbox{0.969,0.413,0.413}{Therefore}\hspace{0pt}%
\heatbox{0.969,0.413,0.413}{,}\hspace{0pt}\heatbox{0.969,0.413,0.413}{ Conclusion}\hspace{0pt}:\hspace{0pt}%
The\hspace{0pt} number\hspace{0pt} of\hspace{0pt} rational\hspace{0pt} numbers\hspace{0pt} between\hspace{0pt}%
 \hspace{0pt}0\hspace{0pt} and\hspace{0pt} \hspace{0pt}1\hspace{0pt} such\hspace{0pt} that\hspace{0pt}%
 when\hspace{0pt} written\hspace{0pt} in\hspace{0pt} lowest\hspace{0pt} terms\hspace{0pt} as\hspace{0pt}%
 a\hspace{0pt}/b\hspace{0pt},\hspace{0pt} the\hspace{0pt} product\hspace{0pt} of\hspace{0pt} numerator\hspace{0pt}%
 and\hspace{0pt} denominator\hspace{0pt} is\hspace{0pt} \hspace{0pt}2\hspace{0pt}0\hspace{0pt}!=\hspace{0pt}%
1\hspace{0pt}ÃĹ\hspace{0pt}2\hspace{0pt}ÃĹ\hspace{0pt}3\hspace{0pt}...\hspace{0pt}ÃĹ\hspace{0pt}2\hspace{0pt}%
0\hspace{0pt} is\hspace{0pt} given\hspace{0pt} by\hspace{0pt} \hspace{0pt}2\hspace{0pt}\^{}\{\hspace{0pt}%
number\hspace{0pt} of\hspace{0pt} prime\hspace{0pt} factors\hspace{0pt} of\hspace{0pt} \hspace{0pt}2\hspace{0pt}%
0\hspace{0pt}!\hspace{0pt} -\hspace{0pt}1\hspace{0pt}\}\hspace{0pt} =\hspace{0pt}2\hspace{0pt}\^{}\{\hspace{0pt}%
8\hspace{0pt}-\hspace{0pt}1\hspace{0pt}\}=\hspace{0pt}1\hspace{0pt}2\hspace{0pt}8\hspace{0pt}.\hspace{0pt}%
\heatbox{0.976,0.538,0.538}{**}\hspace{0pt}\heatbox{0.976,0.538,0.538}{Final}\hspace{0pt}\heatbox{0.976,0.538,0.538}{ Answer}\hspace{0pt}%
**\hspace{0pt}\textbackslash{}\hspace{0pt}boxed\hspace{0pt}\{\hspace{0pt}1\hspace{0pt}2\hspace{0pt}8\hspace{0pt}%
\}\hspace{0pt}\heatbox{0.965,0.335,0.335}{\textless{}|im\_start|\textgreater{}}\hspace{0pt}\heatbox{0.965,0.335,0.335}{answer}\hspace{0pt}%
\hspace{0pt}\heatbox{0.982,0.654,0.654}{Answer}\hspace{0pt}\heatbox{0.982,0.654,0.654}{:}\hspace{0pt}%
\heatbox{0.982,0.654,0.654}{ Given}\hspace{0pt}\heatbox{0.982,0.654,0.654}{ a}\hspace{0pt}\heatbox{0.982,0.654,0.654}{ rational}\hspace{0pt}%
\heatbox{0.982,0.654,0.654}{ number}\hspace{0pt}\heatbox{0.982,0.654,0.654}{ between}\hspace{0pt}\heatbox{0.982,0.654,0.654}{ }\hspace{0pt}%
\heatbox{0.982,0.654,0.654}{0}\hspace{0pt}\heatbox{0.982,0.654,0.654}{ and}\hspace{0pt}\heatbox{0.982,0.654,0.654}{ }\hspace{0pt}%
\heatbox{0.982,0.654,0.654}{1}\hspace{0pt}\heatbox{0.982,0.654,0.654}{,}\hspace{0pt}\heatbox{0.982,0.654,0.654}{ we}\hspace{0pt}%
\heatbox{0.982,0.654,0.654}{ need}\hspace{0pt}\heatbox{0.982,0.654,0.654}{ to}\hspace{0pt}\heatbox{0.982,0.654,0.654}{ write}\hspace{0pt}%
\heatbox{0.982,0.654,0.654}{ it}\hspace{0pt}\heatbox{0.982,0.654,0.654}{ as}\hspace{0pt}\heatbox{0.982,0.654,0.654}{ a}\hspace{0pt}%
\heatbox{0.982,0.654,0.654}{ fraction}\hspace{0pt}\heatbox{0.982,0.654,0.654}{ in}\hspace{0pt}\heatbox{0.982,0.654,0.654}{ lowest}\hspace{0pt}%
\heatbox{0.982,0.654,0.654}{ terms}\hspace{0pt}\heatbox{0.982,0.654,0.654}{ \textbackslash{}}\hspace{0pt}%
\heatbox{0.982,0.654,0.654}{(\textbackslash{}}\hspace{0pt}\heatbox{0.982,0.654,0.654}{frac}\hspace{0pt}%
\heatbox{0.982,0.654,0.654}{\{}\hspace{0pt}\heatbox{0.982,0.654,0.654}{a}\hspace{0pt}\heatbox{0.982,0.654,0.654}{\}\{}\hspace{0pt}%
\heatbox{0.982,0.654,0.654}{b}\hspace{0pt}\heatbox{0.982,0.654,0.654}{\}\textbackslash{}}\hspace{0pt}%
\heatbox{0.982,0.654,0.654}{)}\hspace{0pt}\heatbox{0.982,0.654,0.654}{ and}\hspace{0pt}\heatbox{0.982,0.654,0.654}{ calculate}\hspace{0pt}%
\heatbox{0.982,0.654,0.654}{ the}\hspace{0pt}\heatbox{0.982,0.654,0.654}{ product}\hspace{0pt}\heatbox{0.982,0.654,0.654}{ of}\hspace{0pt}%
\heatbox{0.982,0.654,0.654}{ the}\hspace{0pt}\heatbox{0.982,0.654,0.654}{ resulting}\hspace{0pt}\heatbox{0.982,0.654,0.654}{ numerator}\hspace{0pt}%
\heatbox{0.982,0.654,0.654}{ and}\hspace{0pt}\heatbox{0.982,0.654,0.654}{ denominator}\hspace{0pt}\heatbox{0.982,0.654,0.654}{.}\hspace{0pt}%
\heatbox{0.982,0.654,0.654}{ We}\hspace{0pt}\heatbox{0.982,0.654,0.654}{ need}\hspace{0pt}\heatbox{0.982,0.654,0.654}{ to}\hspace{0pt}%
\heatbox{0.982,0.654,0.654}{ determine}\hspace{0pt}\heatbox{0.982,0.654,0.654}{ how}\hspace{0pt}\heatbox{0.982,0.654,0.654}{ many}\hspace{0pt}%
\heatbox{0.982,0.654,0.654}{ such}\hspace{0pt}\heatbox{0.982,0.654,0.654}{ rational}\hspace{0pt}\heatbox{0.982,0.654,0.654}{ numbers}\hspace{0pt}%
\heatbox{0.982,0.654,0.654}{ exist}\hspace{0pt}\heatbox{0.982,0.654,0.654}{ where}\hspace{0pt}\heatbox{0.982,0.654,0.654}{ the}\hspace{0pt}%
\heatbox{0.982,0.654,0.654}{ product}\hspace{0pt}\heatbox{0.982,0.654,0.654}{ \textbackslash{}(}\hspace{0pt}%
\heatbox{0.982,0.654,0.654}{a}\hspace{0pt}\heatbox{0.982,0.654,0.654}{ \textbackslash{}}\hspace{0pt}\heatbox{0.982,0.654,0.654}{times}\hspace{0pt}%
\heatbox{0.982,0.654,0.654}{ b}\hspace{0pt}\heatbox{0.982,0.654,0.654}{ =}\hspace{0pt}\heatbox{0.982,0.654,0.654}{ }\hspace{0pt}%
\heatbox{0.982,0.654,0.654}{2}\hspace{0pt}\heatbox{0.982,0.654,0.654}{0}\hspace{0pt}\heatbox{0.982,0.654,0.654}{!\textbackslash{}}\hspace{0pt}%
).\hspace{0pt}1\hspace{0pt}.\hspace{0pt} **\hspace{0pt}Prime\hspace{0pt} Factor\hspace{0pt}ization\hspace{0pt}%
 of\hspace{0pt} \hspace{0pt}2\hspace{0pt}0\hspace{0pt}!\hspace{0pt}**\hspace{0pt}:\hspace{0pt} \hspace{0pt}%
 -\hspace{0pt} The\hspace{0pt} prime\hspace{0pt} factors\hspace{0pt} of\hspace{0pt} \hspace{0pt}2\hspace{0pt}%
0\hspace{0pt}!\hspace{0pt} are\hspace{0pt} \hspace{0pt}2\hspace{0pt},\hspace{0pt} \hspace{0pt}3\hspace{0pt}%
,\hspace{0pt} \hspace{0pt}5\hspace{0pt},\hspace{0pt} \hspace{0pt}7\hspace{0pt},\hspace{0pt} \hspace{0pt}%
1\hspace{0pt}1\hspace{0pt},\hspace{0pt} \hspace{0pt}1\hspace{0pt}3\hspace{0pt},\hspace{0pt} \hspace{0pt}%
1\hspace{0pt}7\hspace{0pt},\hspace{0pt} and\hspace{0pt} \hspace{0pt}1\hspace{0pt}9\hspace{0pt}.\hspace{0pt}%
 These\hspace{0pt} are\hspace{0pt} \hspace{0pt}8\hspace{0pt} distinct\hspace{0pt} primes\hspace{0pt}.\hspace{0pt}%
2\hspace{0pt}.\hspace{0pt} **\hspace{0pt}C\hspace{0pt}opr\hspace{0pt}ime\hspace{0pt} P\hspace{0pt}airs\hspace{0pt}%
**\hspace{0pt}:\hspace{0pt} \hspace{0pt} -\hspace{0pt} For\hspace{0pt} \textbackslash{}(\hspace{0pt}a\hspace{0pt}%
 \textbackslash{}\hspace{0pt}times\hspace{0pt} b\hspace{0pt} =\hspace{0pt} \hspace{0pt}2\hspace{0pt}0\hspace{0pt}%
!\textbackslash{}\hspace{0pt})\hspace{0pt} and\hspace{0pt} \textbackslash{}\hspace{0pt}(\textbackslash{}\hspace{0pt}%
gcd\hspace{0pt}(a\hspace{0pt},\hspace{0pt} b\hspace{0pt})\hspace{0pt} =\hspace{0pt} \hspace{0pt}1\hspace{0pt}%
\textbackslash{}\hspace{0pt}),\hspace{0pt} each\hspace{0pt} prime\hspace{0pt} factor\hspace{0pt} of\hspace{0pt}%
 \hspace{0pt}2\hspace{0pt}0\hspace{0pt}!\hspace{0pt} must\hspace{0pt} be\hspace{0pt} assigned\hspace{0pt}%
 entirely\hspace{0pt} to\hspace{0pt} either\hspace{0pt} \textbackslash{}(\hspace{0pt}a\hspace{0pt}\textbackslash{})\hspace{0pt}%
 or\hspace{0pt} \textbackslash{}(\hspace{0pt}b\hspace{0pt}\textbackslash{}\hspace{0pt}).\hspace{0pt} This\hspace{0pt}%
 results\hspace{0pt} in\hspace{0pt} \textbackslash{}(\hspace{0pt}2\hspace{0pt}\^{}\hspace{0pt}8\hspace{0pt}%
 =\hspace{0pt} \hspace{0pt}2\hspace{0pt}5\hspace{0pt}6\hspace{0pt}\textbackslash{})\hspace{0pt} cop\hspace{0pt}%
r\hspace{0pt}ime\hspace{0pt} pairs\hspace{0pt} \textbackslash{}\hspace{0pt}((\hspace{0pt}a\hspace{0pt}%
,\hspace{0pt} b\hspace{0pt})\textbackslash{}\hspace{0pt}).\hspace{0pt}\heatbox{0.983,0.671,0.671}{3}\hspace{0pt}%
\heatbox{0.983,0.671,0.671}{.}\hspace{0pt}\heatbox{0.983,0.671,0.671}{ **}\hspace{0pt}\heatbox{0.983,0.671,0.671}{Count}\hspace{0pt}%
\heatbox{0.983,0.671,0.671}{ing}\hspace{0pt}\heatbox{0.983,0.671,0.671}{ P}\hspace{0pt}\heatbox{0.983,0.671,0.671}{airs}\hspace{0pt}%
\heatbox{0.983,0.671,0.671}{ \textbackslash{}(}\hspace{0pt}\heatbox{0.983,0.671,0.671}{a}\hspace{0pt}%
\heatbox{0.983,0.671,0.671}{ \textless{}}\hspace{0pt}\heatbox{0.983,0.671,0.671}{ b}\hspace{0pt}\heatbox{0.983,0.671,0.671}{\textbackslash{}}\hspace{0pt}%
\heatbox{0.983,0.671,0.671}{)**}\hspace{0pt}:\hspace{0pt} \hspace{0pt} -\hspace{0pt} Since\hspace{0pt}%
 \textbackslash{}(\hspace{0pt}a\hspace{0pt}\textbackslash{})\hspace{0pt} and\hspace{0pt} \textbackslash{}(\hspace{0pt}%
b\hspace{0pt}\textbackslash{})\hspace{0pt} are\hspace{0pt} cop\hspace{0pt}r\hspace{0pt}ime\hspace{0pt}%
 and\hspace{0pt} \textbackslash{}(\hspace{0pt}a\hspace{0pt} \textbackslash{}\hspace{0pt}times\hspace{0pt}%
 b\hspace{0pt} =\hspace{0pt} \hspace{0pt}2\hspace{0pt}0\hspace{0pt}!\textbackslash{}\hspace{0pt}),\hspace{0pt}%
 each\hspace{0pt} pair\hspace{0pt} \textbackslash{}\hspace{0pt}((\hspace{0pt}a\hspace{0pt},\hspace{0pt}%
 b\hspace{0pt})\textbackslash{}\hspace{0pt})\hspace{0pt} and\hspace{0pt} \textbackslash{}\hspace{0pt}%
((\hspace{0pt}b\hspace{0pt},\hspace{0pt} a\hspace{0pt})\textbackslash{}\hspace{0pt})\hspace{0pt} are\hspace{0pt}%
 distinct\hspace{0pt}.\hspace{0pt} To\hspace{0pt} find\hspace{0pt} pairs\hspace{0pt} where\hspace{0pt}%
 \textbackslash{}(\hspace{0pt}a\hspace{0pt} \textless{}\hspace{0pt} b\hspace{0pt}\textbackslash{}\hspace{0pt}%
),\hspace{0pt} we\hspace{0pt} divide\hspace{0pt} the\hspace{0pt} total\hspace{0pt} number\hspace{0pt}%
 of\hspace{0pt} pairs\hspace{0pt} by\hspace{0pt} \hspace{0pt}2\hspace{0pt},\hspace{0pt} excluding\hspace{0pt}%
 the\hspace{0pt} case\hspace{0pt} \textbackslash{}(\hspace{0pt}a\hspace{0pt} =\hspace{0pt} b\hspace{0pt}%
\textbackslash{})\hspace{0pt} (\hspace{0pt}which\hspace{0pt} is\hspace{0pt} impossible\hspace{0pt} since\hspace{0pt}%
 \hspace{0pt}2\hspace{0pt}0\hspace{0pt}!\hspace{0pt} is\hspace{0pt} not\hspace{0pt} a\hspace{0pt} square\hspace{0pt}%
).\hspace{0pt}4\hspace{0pt}.\hspace{0pt} **\hspace{0pt}Result\hspace{0pt}**\hspace{0pt}:\hspace{0pt} \hspace{0pt}%
 -\hspace{0pt} The\hspace{0pt} number\hspace{0pt} of\hspace{0pt} cop\hspace{0pt}r\hspace{0pt}ime\hspace{0pt}%
 pairs\hspace{0pt} \textbackslash{}\hspace{0pt}((\hspace{0pt}a\hspace{0pt},\hspace{0pt} b\hspace{0pt}%
)\textbackslash{}\hspace{0pt})\hspace{0pt} with\hspace{0pt} \textbackslash{}(\hspace{0pt}a\hspace{0pt}%
 \textless{}\hspace{0pt} b\hspace{0pt}\textbackslash{})\hspace{0pt} is\hspace{0pt} \textbackslash{}\hspace{0pt}%
(\textbackslash{}\hspace{0pt}frac\hspace{0pt}\{\hspace{0pt}2\hspace{0pt}5\hspace{0pt}6\hspace{0pt}\}\{\hspace{0pt}%
2\hspace{0pt}\}\hspace{0pt} =\hspace{0pt} \hspace{0pt}1\hspace{0pt}2\hspace{0pt}8\hspace{0pt}\textbackslash{}\hspace{0pt}%
).\hspace{0pt}Thus\hspace{0pt},\hspace{0pt} the\hspace{0pt} number\hspace{0pt} of\hspace{0pt} rational\hspace{0pt}%
 numbers\hspace{0pt} between\hspace{0pt} \hspace{0pt}0\hspace{0pt} and\hspace{0pt} \hspace{0pt}1\hspace{0pt}%
 for\hspace{0pt} which\hspace{0pt} the\hspace{0pt} product\hspace{0pt} of\hspace{0pt} the\hspace{0pt}%
 numerator\hspace{0pt} and\hspace{0pt} denominator\hspace{0pt} in\hspace{0pt} lowest\hspace{0pt} terms\hspace{0pt}%
 is\hspace{0pt} \textbackslash{}(\hspace{0pt}2\hspace{0pt}0\hspace{0pt}!\textbackslash{}\hspace{0pt})\hspace{0pt}%
 is\hspace{0pt} \textbackslash{}\hspace{0pt}(\textbackslash{}\hspace{0pt}boxed\hspace{0pt}\{\hspace{0pt}%
1\hspace{0pt}2\hspace{0pt}8\hspace{0pt}\}\textbackslash{}\hspace{0pt}).\hspace{0pt}\textless{}|im\_end|\textgreater{}\hspace{0pt}%

\par
\endgroup